\documentclass[twoside,11pt]{article}

\usepackage{jmlr2e}

\usepackage{amsmath}
\usepackage{mathtools}
\usepackage[usenames]{color}
\usepackage{mathabx}
\usepackage{bm}
\usepackage{multirow}

\usepackage{chngcntr}
\counterwithout{equation}{section}
\allowdisplaybreaks[1]

\usepackage{subfigure}

\newtheorem{assumption}[theorem]{Assumption}

\newcommand{\LD}{\langle}
\newcommand{\RD}{\rangle}

\def\wy{\widecheck{y}}
\def\wZk{\widecheck{z_k}}

\def\wZik{\widecheck{Z_{ik}}}
\def\wZij{\widecheck{Z_{ij}}}
\def\wZi{\widecheck{\bZ_{i}}}
\def\wS(X){\widecheck{S(\bx)}}
\def\wyi{\widecheck{y_i}}
\def\wSXi{\widecheck{S(\bX_i)}}

\def\wT_j(X){\widecheck{T}_j(X)}

\newcommand{\bTheta}{\bm{\Theta}}
\newcommand{\btheta}{\bm{\theta}}
\newcommand{\bomega}{\bm{\omega}}
\newcommand{\bSigma}{\bm{\Sigma}}
\newcommand{\bOmega}{\bm{\Omega}}
\newcommand{\be}{\bm{e}}
\newcommand{\bLambda}{\bm{\Lambda}}
\newcommand{\bT}{\bm{T}}
\newcommand{\bQ}{\bm{Q}}
\newcommand{\bl}{\bm{l}}
\newcommand{\bI}{\bm{I}}
\newcommand{\bV}{\bm{V}}
\newcommand{\bv}{\bm{v}}
\newcommand{\bU}{\bm{U}}

\newcommand{\bA}{\bm{A}}
\newcommand{\ba}{\bm{a}}
\newcommand{\diag}{{\rm diag}}
\newcommand{\bsigma}{\bm{\sigma}}
\newcommand{\vb}{\mathbf{v}}
\newcommand{\xb}{\mathbf{x}}
\newcommand{\bX}{\bm{X}}
\newcommand{\bZ}{\bm{Z}}
\newcommand{\bignorm}[1]{\bigg|\bigg|#1\bigg|\bigg|}

\newcommand{\bbeta}{\bm{\beta}}
\newcommand{\bB}{\bm{B}}
\newcommand{\ttbeta}{\tilde{\bbeta}}
\newcommand{\hbeta}{\hat{\bbeta}}
\newcommand{\tbeta}{\bbeta^\star}
\newcommand{\tOmega}{\bOmega^\star}
\newcommand{\tomega}{\bomega^\star}
\newcommand{\homega}{\hat{\bomega}}
\newcommand{\hl}{\hat{\bl}}
\newcommand{\tSigma}{\bSigma^\star}
\newcommand{\hOmega}{\hat{\bOmega}}
\newcommand{\hSigma}{\hat{\bSigma}}
\newcommand{\hL}{\hat{L}}
\newcommand{\barL}{\bar{L}}
\newcommand{\tB}{\bB^\star}
\newcommand{\ttB}{\tilde{\bB}}
\newcommand{\hB}{\hat{\bB}}
\newcommand{\VAR}{\text{Var}}

\newcommand{\mR}{\mathbb{R}}
\newcommand{\mE}{\mathbb{E}}
\newcommand{\mF}{\mathcal{F}}
\newcommand{\hSigmajk}{\hat{\Sigma}_{jk}}
\newcommand{\tSigmajk}{\Sigma^\star_{jk}}

\def\bx{{\boldsymbol{x}}}
\def\bz{{\boldsymbol{z}}}
\def\0{{\boldsymbol{0}}}
\def\S{\mathcal{S}}
\def\H{{\mathcal H}}
\def\I{\mathcal{I}}
\def\T{\mathcal{T}}
\def\P{\mathcal{P}}
\def\tH{\tilde{\mathcal H}}
\def\tP{\tilde{\mathcal P}}

\usepackage{xargs}
\usepackage[colorinlistoftodos,prependcaption,textsize=tiny]{todonotes}
\newcommandx{\unsure}[2][1=]{\todo[linecolor=red,backgroundcolor=red!25,bordercolor=red,#1]{#2}}
\newcommandx{\change}[2][1=]{\todo[linecolor=blue,backgroundcolor=blue!25,bordercolor=blue,#1]{#2}}
\newcommandx{\info}[2][1=]{\todo[linecolor=OliveGreen,backgroundcolor=OliveGreen!25,bordercolor=OliveGreen,#1]{#2}}
\newcommandx{\improvement}[2][1=]{\todo[linecolor=red,backgroundcolor=red!25,bordercolor=red,#1]{#2}}

\usepackage{lastpage}
\jmlrheading{20}{2019}{1--\pageref{LastPage}}{10/18; Revised 5/19}{10/19}{18-705}{Sen Na, Zhuoran Yang, Zhaoran Wang, and Mladen Kolar}

\ShortHeadings{High-dimensional Varying Index Coefficient Models via Stein's Identity}{Na, Yang, Wang, and Kolar}
\firstpageno{1}

\begin{document}

\title{High-dimensional Varying Index Coefficient Models via Stein's Identity}
\author{\name Sen Na \email senna@uchicago.edu \\
\addr Department of Statistics\\
University of Chicago\\
Chicago, IL 60637, USA
\AND
\name Zhuoran Yang \email  zy6@princeton.edu\\
\addr Department of Operations Research and Financial Engineering\\
Princeton University\\
Princeton, NJ 08544, USA
\AND
\name Zhaoran Wang \email  zhaoran.wang@northwestern.edu\\
\addr Department of Industrial Engineering and Management Sciences\\
Northwestern University\\
Evanston, IL 60208, USA
\AND
\name Mladen Kolar \email  mkolar@chicagobooth.edu\\
\addr The University of Chicago Booth School of Business\\
Chicago, IL 60637, USA }

\editor{Kenji Fukumizu}

\maketitle

\begin{abstract}%
We study the parameter estimation problem for a varying index coefficient model in high dimensions. Unlike the most existing works that iteratively estimate the parameters and link functions, based on the generalized Stein's identity, we propose computationally efficient estimators for the high-dimensional parameters without estimating the link functions. We consider two different setups where we either estimate each sparse parameter vector individually or estimate the parameters simultaneously as a sparse or low-rank matrix. For all these cases, our estimators are shown to achieve optimal statistical rates of convergence (up to logarithmic terms in the low-rank setting). Moreover, throughout our analysis, we only require the covariate to satisfy certain moment conditions, which is significantly weaker than the Gaussian or elliptically symmetric assumptions that are commonly made in the existing literature. Finally, we conduct extensive numerical experiments to corroborate the theoretical results.
\end{abstract}


\section{Introduction}\label{sec:1}

We consider the problem of estimating parameters in a high-dimensional varying index coefficient model with following form
\begin{align}\label{mod:1}
y =  \sum_{j=1}^{d_2}z_j \cdot f_j(\LD \bx, \tbeta_j\RD) + \epsilon,
\end{align}
where $y$ is response variable, $\bx = (x_1, \ldots, x_{d_1})^\top \in\mR^{d_1}$ and $\bz = (z_1, \ldots, z_{d_2})^\top \in \mR^{d_2}$ are given covariates, and $\epsilon$ is random noise with $\mE[\epsilon \mid \bx,\bz] = 0$.
For $j\in[d_2]$\footnote{For any integer $d$, we	denote $[d] = \{1,2,...,d\}$.}, $\tbeta_j = (\beta_{j1}^\star, \ldots, \beta_{jd_1}^\star)^\top$ are the coefficient vectors, that is parameters, which vary with different covariates $z_j$, and $f_j(\cdot)$ are unknown nonparametric link functions. For identification purposes, we can always permute $\tbeta_j$ and multiply by a scalar such that
\begin{equation}
\label{eq:identification}
\tbeta_j \in \{ \bbeta \in \mR^{d_1} : \| \bbeta \|_2 = 1 \text{ and } \beta_{1} > 0\}, \quad j = 1,\ldots, d_2.
\end{equation}
All further restrictions on parameters will only be considered under
(\ref{eq:identification}).

Model \eqref{mod:1} has been introduced by \cite{ma2015varying} as a flexible generalization of a number of well studied semiparametric
statistical models \citep[see also][]{xue2012empirical}. When $z_j = 1$
for all $j \in [d_2]$, the model reduces to the additive single-index
model \citep{chen1991estimation, carroll1997generalized}, which can
also be viewed as a two-layer neural network with $d_2$ hidden
nodes. When $d_1 = 1$ and $\tbeta_j = 1$ for $j=1,\ldots, d_2$, the
model \eqref{mod:1} reduces to the varying coefficient model proposed
in \cite{Cleveland91local} and \cite{Hastie1993Varying}, with wide
applications in scientific areas such as economics and medical science
\citep{fan2008statistical}. Varying coefficient models allow the
coefficients of $\bz$ to be smooth functions of $\bx$, thus incorporating nonlinear interactions between $\bx$ and $\bz$.  Model
(\ref{mod:1}) is also easily interpreted in real applications because
it inherits features from both single-index model and varying coefficient model, while being able to capture complex multivariate nonlinear structure.

Our focus is on the case when the dimension of $\bx$ is high, which makes estimation of the coefficients difficult. Existing procedures estimate the unknown functions and coefficients iteratively. First,
with the signal parameters $\{ \tbeta_j \}_{j \in [d_2]}$ fixed, one estimates the functions $\{ f_j(\cdot) \}_{j \in [d_2]}$ using a nonparametric method, such as local polynomial estimator. Next, using
the estimated link functions, one re-estimates the coefficients. While
the global minimizer has desirable properties (see \cite{xue2012empirical} and \cite{ma2015varying} and the references
therein), the loss function is usually nonconvex and it is computationally intractable to obtain the global optima. For high-dimensional single-index models ($d_2 = 1$ and $\bz = 1$), when the distribution of $\bx$ is known, the signal parameter can be estimated directly by fitting Lasso \citep{tibshirani96regression}. Such an estimator is shown to achieve minimax-optimal statistical rate of convergence
\citep{plan2016generalized, plan2017high}. Thus, the following
question naturally arises:

\emph{Is it possible to estimate signal parameters $\{ \tbeta_j \}_{j \in [d_2]}$ in \eqref{mod:1} with both statistical accuracy and computational efficiency?}

In this work, we provide a positive answer to above question. Specifically, we focus on the problem of estimating the parameter matrix $\tB = (\tbeta_1, \ldots, \tbeta_{d_2})\in\mR^{d_1\times d_2}$ in the high-dimensional setting where the sample size is much smaller than $d_1\times d_2$ and $\tB$ is either sparse or low-rank.  We utilize
the score functions and the generalized Stein's identity \citep{Stein1972bound, Stein2004Use} to estimate the unknown
coefficients through a regularized least-square regression problem,
without learning the unknown functions $\{ f_j(\cdot) \}_{j \in [d_2]}$.  We prove that the estimators achieve (near) optimal statistical rates of convergence under weak moment conditions, which make our procedure suitable for heavy-tailed data, using a careful truncation argument. Finally, our estimator can be computed as a solution to a convex optimization problem.

{\noindent \bf Main Contributions.} Our contributions are three-fold. First, we propose a computationally efficient estimation procedure for the varying index coefficient model in high dimensions. Different from existing work, our approach does not need to estimate the unknown functions $\{ f_j \}_{j \in [d_2]}$. Second, when $\tB$ is sparse, we prove that the proposed estimator achieves the optimal statistical rate of convergence, while when $\tB$ is low-rank, our estimator is shown to be near-optimal (up to a logarithmic factor). Finally, we provide thorough numerical experiments with both synthetic and real data to back up our theories.

{\noindent \bf Related Work.}  There is a plethora of literature on
the varying coefficient model, first proposed in \cite{Cleveland91local} and \cite{Hastie1993Varying}, where the coefficients are modeled as nonparametric functions of $\bx$.  See \cite{fan2008statistical} for a detailed review. \cite{Xia1999single}, \cite{fan2003adaptive}, and \cite{xue2012empirical} considered model in~\eqref{mod:1} with $\tbeta_j = \tbeta$ for all $j \in [d_2]$ and estimated it with standard nonparametric techniques. \cite{ma2015varying} proposed model \eqref{mod:1} and developed a profile least-square approach to estimate the coefficients. Unfortunately, the estimator is defined as a solution to a constrained optimization problem with nonconvex objective function that can be hard to globally optimize in practice. This should be contrasted to estimators that are based on solving convex optimization problems.

Another related line of research is on the high-dimensional single-index model (SIM) with sparse coefficient vector, which is a special case of model \eqref{mod:1}. Most of the existing results require either knowing the distribution of $\bx$ or strong assumptions on the link functions.  Specifically, \cite{thrampoulidis2015lasso}, \cite{neykov2016l1}, \cite{plan2016generalized}, and \cite{plan2017high} all showed that
when $\bx$ is a standard Gaussian and the link function satisfies certain conditions, Lasso estimators could also work for SIM with the
same theoretical guarantee as if the link function is not present. To
relax the Gaussian assumption, \cite{goldstein2016structured} proposed a modified Lasso-type estimator when $\bx$ has an elliptically symmetric distribution. Moreover, using the generalized Stein's identity, \cite{yang2017high} proposed a soft-thresholding estimator for SIM when the distribution of $\bx$ is known, and \cite{Na2018High} proposed estimators for single-index volatility model. Our work can be viewed as the extension of this work from two aspects. First, as mentioned before, our model in \eqref{mod:1} includes the single-index model considered in \cite{yang2017high} as a special case, which can be recovered by letting $d_2 = 1$ and $\bz = 1$. In comparison, our model contains $d_2$ unknown signal vectors $\{\tbeta_j \}_{j\in [d_2]}$, where $d_2$ itself can be large compared to the sample size.  As we will show later, when the parameter matrix $\tB$ has a certain low-dimensional structure, such as being sparse or low rank, we construct efficient estimators that enjoy sharp statistical rates of convergence even when $d_2$ is large and $\{z_j \}_{j \in [d_2]}$ are dependent, heavy-tailed. Second, from the methodology perspective, our estimators are based on combining the first-order Stein's identity and the dependence structure of the covariate $\bz$, which is a novel observation in index models and generalizes the method in \cite{yang2017high} for SIM.  In particular, we extract the information of the parameter matrix $\tB$ via multiplying the response $y$ by the score function of $\bx$ and the covariate $\bz$ adjusted by its precision matrix (see (\ref{equ:7})). When $\bz$ has independent components, our procedure for estimating each single sparse vector $\tbeta_j$ is equivalent to fitting $d_2$ SIMs using the procedure proposed in \cite{yang2017high}. However, even having independent $\{z_j\}_{j\in[d_2]}$, their method fails when estimating $\tB$ as a whole with either low-rank or sparse structure, since they only have one coefficient and cannot deal with varying coefficient models without accounting for $\bz$ in Stein’s identity. Furthermore, to handle dependence structure of $\bz$ in general, we construct suitable
precision matrix estimators for heavy-tailed variable in either low or
high dimensions based on the CLIME procedure \citep{Cai2011Constrained} and the soft truncation technique, which is
of independent interests. Besides aforementioned related literature, a sequence of work \citep{zhu2006sliced, jiang2014variable, zhang2018high, lin2017optimality, Lin2018consistency} applied the sliced inverse regression (SIR) technique on high-dimensional SIM, which is generalized from \cite{li1991sliced}. But all these works require the distribution of $\bx$ to be Gaussian or elliptical. To resolve this
limitation, \cite{Babichev2018Slicea} incorporated SIR with both first-order and second-order score function when fitting a low-dimensional index model, while the high-dimensional analysis is
not included.

Furthermore, our work is also related to the study of additive index
model, which is more challenging than \eqref{mod:1}, and there is very
much work in this direction. Most existing work focuses on estimating
the signal parameters and the link functions together in the low-dimensional setting. See, for example, \cite{yuan2011identifiability}, \cite{wang2015estimation}, and
\cite{chen2016generalized}. When the covariate is Gaussian and the
link functions are known, \cite{sedghi2016provable} proposed to estimate the signal parameters via tensor decomposition. These works
are not comparable with ours as we consider a different model and our
goal is to efficiently estimate the high-dimensional parameters.

Lastly, our estimation methodology utilizes the generalized Stein's identity \citep{Stein2004Use}, which extends the well-known Stein's
identity for Gaussian distribution \citep{Stein1972bound} to general
distributions whose density satisfies certain regularity conditions. This identity is widely applied in probability, statistics, and machine learning. We point the reader to \cite{Chen2011Normal}, \cite{chwialkowski2016kernel}, \cite{liu2016kernelized}, \cite{liu2016stein}, and \cite{Liu2018Action} for recent applications.

\noindent{\bf Notations:} Throughout the paper, we use boldface letters, $\bv, \bV$, to denote vector or matrix and their elements will be denoted as $v_i$, $V_{ij}$. For any vector $\bv$ and $p\geq 1$, $\|\bv\|_p$ is the vector $\ell_p$-norm, with the usual extension $\|\bv\|_0 = |\text{supp}(\bv)| = |\{i: v_i\neq 0\}|$. Given a matrix $\bV\in\mR^{m\times n}$, we let $\|\bV\|_p$ be the induced $p$-norm. We denote the nuclear norm and Frobenius norm as $\|\bV\|_*$ and $\|\bV\|_F$, respectively. We also define $\|\bV\|_{p,q} = \big(\sum_{j=1}^n(\sum_{i=1}^m|V_{ij}|^p)^{q/p}\big)^{1/q}$, which is basically computing the vector $\ell_p$-norm for each column and then computing the $\ell_q$ norm for those $n$ numbers. We also define
$\|\bV\|_{\max} = \|\bV\|_{\infty, \infty}$ and $\text{supp}(\bV) = \{(i,j): V_{ij}\neq 0\}$. For two matrices $\bV, \bU$ with the same dimension, we let $\LD \bV, \bU\RD = \text{trace}(\bV^T\bU) =
\sum_{i,j=1}^{n}V_{ij}U_{ij}$. When presenting the result, we use $a\lesssim b$ ($\gtrsim$) to denote $a\leq c\cdot b$ ($\geq$) for some
constant $c$ that we are less interested in. Also, we have $a\asymp b\Leftrightarrow a\lesssim b$ and $a\gtrsim b$. Given a threshold $\lambda$, we define the soft-thresholding function $\T_{\lambda}(\cdot)$ as follows: (i) when $\ba\in\mR^d$, we let
$\T_{\lambda}(\ba)\in\mR^d$ with $[\T_{\lambda}(\ba)]_i=(1-\lambda/|a_i|)_+a_i$; (ii) when
$\bA\in\mR^{d_1\times d_2}$, suppose its singular value decomposition
can be written as $\bA = \bU\diag(\bsigma)\bV^T$, then we let
$\T_{\lambda}(\bA) = \bU\text{diag}(\hat{\bsigma})\bV^T$ where
$\hat{\sigma}_i = (\sigma_i-\lambda)_+$.

\section{Estimation via the Generalized Stein's Identity}\label{sec:2}

In this section, we present the main idea for estimating coefficients
in model \eqref{mod:1}. Our estimator relies on the generalized
Stein's identity \citep{Stein2004Use}, which we state next.

\begin{theorem}[Generalized Stein's identity, \cite{Stein2004Use}]\label{thm:1}

Suppose $\bv \in \mR^d$ is a random vector with differentiable positive density $p_{\bv}:\mR^d\rightarrow\mR$. We define its score function as $S_{\bv}: \mR^d\rightarrow \mR^d$, $S_{\bv}(\vb) = -\nabla \log p_{\bv}(\vb)$. If a differentiable function $f:\mR^{d}\rightarrow\mR$ together with $\bv$ satisfies \textit{regularity condition}: $|p_{\bv}(\vb)|\rightarrow 0$ as $\|\vb\|\rightarrow \infty$ and $\mE[|f(\bv)S_{\bv}(\bv)|] \vee \mE[|\nabla f(\bv)|]<\infty$, then we have
\begin{align}\label{equ:1}
\mE[f(\bv)S_{\bv}(\bv)] = \mE[\nabla f(\bv)].
\end{align}
In particular, when $\bv\sim N(\bm{0},\bI_d)$, we have $$\mE[g(\bv)\bv]=\mE[\nabla g(\bv)].$$
\end{theorem}

We drop off the subscript of density and score function to make
notation concise. In order to use Theorem~\ref{thm:1} for estimation
of coefficients in model \eqref{mod:1}, we require the following regularity condition.

\begin{assumption}[Regularity]\label{ass:1}

We assume that $\bx$, $\bz$ in \eqref{mod:1} are independent and the density function $p(\cdot)$ of $\bx$ is positive and differentiable. For any $j\in[d_2]$, we assume function $\tilde{f}_j: \mR^{d_1}\rightarrow\mR$, defined to be $\tilde{f}_j(\xb) = f_j(\LD \xb, \tbeta_j\RD)$, together with variable $\bx$ satisfies the \textit{regularity condition} of Theorem~\ref{thm:1}. Further, let $\mu_j \coloneqq \mE[f_j'(\LD \bx, \tbeta_j\RD)]$ and we assume $\mu_j\neq0$. In addition, we assume covariate $\bz$ are standardized with $\mE[z_j] = 0$ and $\mE[z_j^2] = 1$, $\forall j\in[d_2]$.

\end{assumption}

Note that the standardization of $\bz$ is made only to simplify our presentation. It is easy to extend to a general $\bz$ using the fact
that $\mE[\bz(\bz - \mE[\bz])^T] = \text{Var}(\bz)$, where diagonal
entries of $\text{Var}(\bz)$ can be assumed to be one without loss of
generality, as the variance of each $z_j$ can be absorbed into
$f_j(\cdot)$. Since $\mE[\bz]$ is easy to estimate at a fast enough
rate, we can replace $\bz$ by $\bz-\mE[\bz]$ whenever necessary in
analysis for the general $\bz$. See equation (\ref{equ:7}) for
example. Under Assumption \ref{ass:1}, Stein's identity will allow us
to extract the unknown coefficient parameter, which is proportional to
the derivative of the corresponding unknown function in the index
model. To clarify, note that
\begin{align}\label{equ:2}
\mE[f_j(\LD \bx,\tbeta_j\RD)S(\bx)] = \mE[\tilde{f}_j(\bx)S(\bx)] \stackrel{(\ref{equ:1})}{=} \mE[\nabla\tilde{f}_j(\bx)] = \mu_j\tbeta_j\coloneqq \ttbeta_j.
\end{align}
The condition $\mu_j\neq 0$ ensures that the above expectation will
not vanish and further $\tbeta_j$ can be fully identifiable from
$\ttbeta_j$ due to (\ref{eq:identification}).

With this setup, we start with a warm-up example and illustrate how to
estimate coefficients $\{\tbeta_j \}_{j\in[d_2]}$ when
$\bx \sim N(\bm{0},\bI_{d_1})$, $\bz \sim N(\bm{0}, \bI_{d_2})$ and
$\bx$ and $\bz$ are independent.  The extension to heavy-tailed
distributions is presented in the next section. Similar to
(\ref{equ:2}), for any $k \in [d_2]$, Stein's identity gives us
\begin{align}\label{equ:3}
\mE[y\cdot z_k\cdot \bx]&=\sum_{j=1}^{d_2}\mE[z_j z_k f_j(\langle\tbeta_j,\bx \rangle)\bx]=\mE[f_k(\langle\tbeta_k,\bx\rangle)\bx]\stackrel{(\ref{equ:2})} {=}\mu_k\tbeta_k=\ttbeta_k.
\end{align}
Under Assumption \ref{ass:1} and identifiability condition
\eqref{eq:identification}, the above equation allows us to form an
estimator for $\tbeta_k$ by minimizing the following population loss
\begin{align}\label{eq:pop_loss}
\ttbeta_k = \arg\min_{\bbeta_k} L_k(\bbeta_k) = \arg\min_{\bbeta_k}\bigg\{ \|\bbeta_k\|_2^2-2\mE [y\cdot z_k\cdot\LD\bbeta_k, \bx\RD]\bigg\}.
\end{align}

Given $n$ i.i.d.~copies of $(y, \bx, \bz)$, $\{y_i, \bX_i, \bZ_i \}_{i=1}^{n}$, we obtain an estimator of $\ttbeta_k$ by replacing the expectation in \eqref{eq:pop_loss} with a sample mean
\begin{align}\label{eq:emp_loss:warmup}
\hbeta_k = \arg\min_{\bbeta_k} \hL_k(\bbeta_k) + R_k(\bbeta_k) = \arg\min_{\bbeta_k}\bigg\{\|\bbeta_k\|^2 - \frac{2}{n}\sum_{i=1}^{n}y_iZ_{ik}\langle \bbeta_k, \bX_i\rangle+\lambda_k \|\bbeta_k\|_1\bigg\},
\end{align}
where $R_k(\bbeta_k)$ is a penalty function that imposes desired
structural assumptions on the estimate. In a high-dimensional setting,
it is common to assume that $\ttbeta_k$ is sparse, so here we use the
$\ell_1$-norm penalty with $R_k(\bbeta_k) = \lambda_k\|\bbeta_k\|_1$.
Note that the loss function in \eqref{eq:pop_loss} can also be written
as
\begin{align*}
L(\bbeta_k) = \mE[(y-z_k\langle \bx, \bbeta_k\rangle)^2],
\end{align*}
which leads to an alternative form for the estimator with a design matrix
\begin{align*}
\arg\min_{\bbeta_k}\bigg\{\frac{1}{n}\sum_{i=1}^{n}(y_i-Z_{ik}\bX_i^T\bbeta_k)^2+\lambda_k\|\bbeta_k\|_1\bigg\}.
\end{align*}
Finally, we note that the estimator in \eqref{eq:emp_loss:warmup} can
be obtained in a closed form
\begin{align*}
\hbeta_k = \T_{\lambda_k/2}\left(\frac{1}{n}\sum_{i=1}^ny_iZ_{ik}\bX_i\right),
\end{align*}
where $\T(\cdot)$ is the soft-thresholding operator defined at the end
of Section \ref{sec:1}.

Our first result establishes the convergence rate for the estimator in
\eqref{eq:emp_loss:warmup}. We present the result for a slightly more
general setting where $\bz$ has independent sub-Gaussian components with $\|z_j\|_{\psi_2}=\Upsilon_{z_j}$, $\forall j\in[d_2]$.\footnote{For centered random variable $x$, we define $\|x\|_{\psi_1} = \sup_{p\geq 1} p^{-1}(\mE|x|^p)^{1/p}$, $\|x\|_{\psi_2} = \sup_{p\geq 1} p^{-1/2}(\mE|x|^p)^{1/p}$. We call $x$ a sub-exponential random variable if $\|x\|_{\psi_1}<\infty$. We call $x$ a sub-Gaussian random variable with proxy variance $\|x\|_{\psi_2}^2$ if $\|x\|_{\psi_2}<\infty$. See \cite{Vershynin2012Introduction} for detailed properties.}

\begin{theorem}[Warm-up]\label{thm:conv_warmup}

Consider model \eqref{mod:1} with $\|{\tbeta_k}\|_0 \leq s$ for $k \in [d_2]$, $\bx\sim N(\0, \bI_{d_1})$, components of $\bz$ are independent with $\|z_k\|_{\psi_2} = \Upsilon_{z_k} \leq \Upsilon_{\bz}$ for $k\in[d_2]$ and independent of $\bx$, and $y$ is sub-exponential with $\|y\|_{\psi_1} \leq \Upsilon_y$. Furthermore assume that Assumption \ref{ass:1} holds. The estimator in \eqref{eq:emp_loss:warmup} with $\lambda_k = 4\Upsilon \sqrt{\log n/n}$, for a constant $\Upsilon$ that depends on $\Upsilon_y$ and $\Upsilon_{\bz}$ only, satisfies
\begin{align*}
\|\hbeta_k-\ttbeta_k\|_2\leq\frac{3}{2}\sqrt{s}\lambda_k \text{\ \ and\ \ } \|\hbeta_k-\ttbeta_k\|_1\leq 6s\lambda_k, \text{\ \ \ } \forall k\in[d_2]
\end{align*}
with probability at least $1-d_2d_1/n^2$.

\end{theorem}

Theorem~\ref{thm:conv_warmup} established rate of convergence for the estimator of $\ttbeta_k$. In particular, it suggests that with high
probability we have $\forall k\in[d_2]$,
\begin{align*}
\|\hbeta_k-\ttbeta_k\|_2\lesssim\sqrt{\frac{s\log n}{n}}\text{\ \ and\ \ } \|\hbeta_k-\ttbeta_k\|_1\lesssim s\sqrt{\frac{\log n}{n}},
\end{align*}
which matches the optimal rate of convergence for sparse vectors
recovery under the setting when $n\ll d_1 \ll n^2$
\citep{lin2017optimality}. It is useful to note that sub-exponential
assumption on $y$ is mild. For example, it is satisfied when $\{f_j\}_{j\in[d_2]}$ can be dominated by a linear function and $\epsilon$ is sub-exponential. The result follows from a bound on
$\|\nabla \hat{L}_k(\tilde{\bbeta}_k)\|_{\infty}$, which is presented
in the following lemma.

\begin{lemma}\label{lem:warmup:bound_score}

Under the conditions of Theorem~\ref{thm:conv_warmup}, we have $\forall k\in[d_2]$,
\begin{align*}
P\left(\|\nabla \hat{L}_k(\tilde{\bbeta}_k)\|_{\infty}>2\Upsilon\sqrt{\frac{\log n}{n}}\right)<\frac{d_1}{n^2}.
\end{align*}

\end{lemma}

Under the identifiability condition \eqref{eq:identification}, we have
the following corollary for the normalized estimator.

\begin{corollary}\label{cor:1}

Suppose the conditions of Theorem \ref{thm:conv_warmup} are satisfied, then for sufficiently large $n$ (threshold depends on $s$ and $\min_{j\in[d_2]}(|\mu_j|\wedge \beta_{j1}^\star)$), we have
\begin{align*}
\bignorm{\text{sign}(\hat{\beta}_{k1})\frac{\hbeta_k}{\|\hbeta_k\|_2}-\tbeta_k}_2\lesssim\sqrt{\frac{s\log n}{n}}\text{\ \ and\ \ } 	\bignorm{\text{sign}(\hat{\beta}_{k1})\frac{\hbeta_k}{\|\hbeta_k\|_2}-\tbeta_k}_1\lesssim s\sqrt{\frac{\log n}{n}}, \text{\ \ \ } \forall k\in[d_2]
\end{align*}
with probability $1-d_2d_1/n^2$. Further we can get
$\|\hB-\tB\|_F\lesssim\sqrt{\frac{d_2s\log n}{n}}$ where $\hB$ saves
normalized estimators by column.

\end{corollary}

Note that the order of $\|\hB - \tB\|_F$ in Corollary \ref{cor:1} is proven under the column-wise sparsity. Details under the assumption that $\tB$ is fully sparse will be discussed later.

In the next few sections, we will build on the illustrative example studied in this section and generalize our results to a heavy-tailed setting, which will improve the applicability of the estimator. Furthermore, we will consider estimation of all coefficients $\{\tbeta_j\}_{j\in[d_2]}$ simultaneously by imposing structural assumptions on the coefficient matrix $\tB$. Let $\ttB = (\ttbeta_1,...,\ttbeta_{d_2})$, we usually focus on establishing statistical guarantee for estimating $\ttB$ as it keeps the same structure as $\tB$, and conversely, $\tB$ is fully identifiable from $\ttB$ under~\eqref{eq:identification}.

\section{Overview of Results}\label{sec:3}

In this section, we introduce weak moment assumption and provide an overview of the proposed estimators and their statistical convergence rates. Our theoretical analysis is separated into two cases: (i) estimating a single sparse coefficient vector $\tbeta_k$; (ii) estimating the coefficient matrix $\tB$. In the former case, we assume covariate $\bz$ has independent entries so that we can extract one specific parameter, while in the latter case, we impose either low-rank or sparse
structure on $\tB$ and relax the requirement for independence of $\bz$ by modifying the estimation procedure to include the estimator of the precision matrix. We build our theoretical results on following weak moment condition.

\begin{assumption}[Finite $p$-th moment]\label{ass:3}

We say finite $p$-th moment holds if there exists a constant $M_p>0$ such that
\begin{align*}
\mE[y^p] \vee \mE[S(\bx)_j^p] \vee \mE[z_k^p]\leq M_p, \ \ \forall j\in[d_1], k\in[d_2].
\end{align*}

\end{assumption}

This condition is used throughout all of our theoretical analysis. In sparse vector recovery, we require finite $6$-th moment, while in low-rank matrix recovery, we only require finite $4$-th moment. Note
that though we cannot assume $S(\bx)$ is sub-Gaussian, it turns out
assuming $S(\bx)$ to have finite moment is still reasonable in the
sense that even for some heavy-tailed distributions, such as
$t$-distribution and Gamma distribution, their score variables still
satisfy the finite moment assumption for some $p$. On the other hand,
assumptions for applying Stein's identity always boil down to finite moment conditions. For example, \cite{yang2017high} required finite $4$-th moment when estimating SIM. In order to estimate varying indices, we require additional two moments to be finite.

\begin{table}[!tb]
	\centering
	\small \addtolength{\tabcolsep}{-4.5pt}
	\begin{tabular}{|c|c| c| c|  }
		\hline
		\multicolumn{4}{|c|}{Single sparse vector recovery}\\
		\hline
		Type & Moment condition & Dimension &  Rate\\
		\hline
		 \multirow{2}{*}{Warm-up } & \multirow{2}{7.5em}{$\bx, \bz\sim N(0,\bI)$, indep; $y\sim\text{subE}$} & \multirow{2}{*}{$s\ll n\ll d_1\ll n^2$} & \multirow{2}{*}{$\sqrt{\frac{s\log n}{n}}$}\\
		& & & \\
		\cline{1-4}
		\multirow{2}{*}{General} & \multirow{2}{*}{$p=6$} & \multirow{2}{*}{$s\ll n\ll d_1$} &\multirow{2}{*}{$\sqrt{\frac{s\log d_1d_2}{n}}$}\\
		& & & \\
		\hline
		\multicolumn{4}{|c|}{Low-rank matrix recovery}\\
		\hline
		\multirow{2}{*}{Sparse precision} & \multirow{2}{*}{$p=4$} & \multirow{2}{*}{{$(r,w)\ll (d_1, d_2)\ll n$}} & \multirow{2}{*}{{$\sqrt{\frac{r(d_1+d_2)\log(d_1+d_2)}{n}}\vee w\sqrt{\frac{r\log d_2}{n}}$}}\\
		& & & \\
		\cline{1-4}
		\multirow{2}{*}{General precision}& \multirow{2}{*}{$p=4$} & \multirow{2}{*}{{$r\ll (d_1,d_2)\ll n$}} & \multirow{2}{8em}{{$\sqrt{\frac{r(d_1+d_2)\log(d_1+d_2)}{n}}$}}\\
		& & & \\
		\cline{1-4}
		\multirow{2}{*}{Indep. entries in $\bz$} & \multirow{2}{*}{$p=4$} & \multirow{2}{*}{{$r\ll (d_1,d_2)\ll n$}} & \multirow{2}{*}{{$\sqrt{\frac{r(d_1+d_2)\log(d_1+d_2)}{n}}$}}\\
		& & & \\
		\hline
		\multicolumn{4}{|c|}{Sparse matrix recovery}\\
		\hline
		 \multirow{2}{8em}{Column sparse \& sparse precision} & \multirow{2}{*}{$p=6$} & \multirow{2}{*}{{$(s, d_2)\ll n\ll d_1$}} & \multirow{2}{*}{$\sqrt{\frac{sd_2\log d_1d_2}{n}}$}\\
		& & &\\
		\cline{1-4}
		\multirow{2}{8em}{Column sparse \& general precision}  & \multirow{2}{*}{$p=6$} & \multirow{2}{*}{{$(s, d_2)\ll n\ll d_1$}}
		& \multirow{2}{*}{$\sqrt{\frac{sd_2\log d_1d_2}{n}} \vee \frac{d_2\sqrt{s\log d_2}}{\sqrt{n}}$}\\
		& & & \\
		\cline{1-4}
		\multirow{2}{8em}{Fully sparse \& sparse precision}  & \multirow{2}{*}{$p=6$} & \multirow{2}{*}{{$(s, w)\ll n\ll (d_1, d_2)$}}
		& \multirow{2}{*}{$\sqrt{\frac{s\log d_1d_2}{n}}$}\\
		& & & \\
		\cline{1-4}
		\multirow{2}{8em}{Fully sparse \& indep. entries in $\bz$}  & \multirow{2}{*}{$p=6$} & \multirow{2}{*}{{$s\ll n\ll (d_1, d_2)$}}
		& \multirow{2}{*}{$\sqrt{\frac{s\log d_1d_2}{n}}$}\\
		& & &\\
		\hline

	\end{tabular}
	\caption{Convergence Rate for $\|\cdot\|_2$ or $\|\cdot\|_F$.}
	\label{tab:1}
\end{table}

Our results are summarized in Table~\ref{tab:1}. In summary, we achieve $\sqrt{s\log d_1/n}$ rate for estimating a single sparse vector, while $\sqrt{s\log d_1d_2/n}$ for estimating a fully sparse parameter matrix. Both of them attain the minimax rate considering the case where all unknown link functions $f_j(\cdot)$ are identity functions. For low-rank estimation, we achieve $\sqrt{r(d_1+d_2)\log(d_1+d_2)/n}$ rate, which is also comparable with results in \cite{plan2016generalized, goldstein2016structured}, though it only attains near-optimal rate up to the logarithmic factor. Note that estimating precision matrix of $\bz$ can be conducted independently from our main procedure and any advanced, suitable estimators can be plugged into our approach. To make the paper compact but self-contained, we only consider estimating a general low-dimensional precision matrix for heavy-tailed $\bz$ as an illustration, and discuss the high-dimensional sparse precision matrix estimation in appendix. Basically, if the precision matrix of $\bz$ is sparse, we can estimate it by CLIME procedure \citep{Cai2011Constrained} with a careful truncation of the sample covariance to attain optimal rate of convergence, even though $\bz$ only has finite certain moment. The proposed precision matrix estimator for heavy-tailed variable is used as a plug-in estimator whenever $\bz$ has non-diagonal covariance matrix. Detailed estimation procedures and corresponding error rates are showed in Section~\ref{sec:5} and Appendix~\ref{supple3}, respectively.

\section{Sparse Vector Recovery}\label{sec:4}

In this section, we present an extension of the estimator discussed in
Section~\ref{sec:2} to heavy-tailed data. Applying Theorem~\ref{thm:1}
after replacing $\bx$ by $S(\bx)$ in \eqref{equ:3} leads to
\begin{align*}
\mE[y\cdot z_k\cdot S(\bx)]&=\sum_{j=1}^{d_2}\mE[z_jz_kf_j(\LD\tbeta_j,\bx\RD)S(\bx)]=\mE[f_k(\LD\tbeta_k,\bx\RD)S(\bx)]\stackrel{(\ref{equ:2})}{=}\mu_k\tbeta_k = \ttbeta_k,
\end{align*}
under the independence condition that $\mE[z_jz_k] = 0$ for $j\neq k$,
which we maintain throughout the section but relax it in Section \ref{sec:5} and \ref{sec:6}. The above identity allows us to estimate the direction of $\tbeta_k$ by estimating the left hand side even in the setting with heavy-tailed data. However, in order to get a fast rate of convergence we will require the covariates and the response to be appropriately truncated.

Given a threshold $\tau>0$, we define the truncation of a vector
$\bv\in\mR^d$ as $\widecheck{\bv}\in\mR^d$ whose coordinates are
defined by $[\widecheck{\bv}]_i = v_i $ if $|v_i|\leq\tau$ and $0$
otherwise. Our estimator for $\ttbeta_k$ is given as
\begin{align}\label{equ:6}
\hbeta_k&=\arg\min_{\bbeta_k}\barL_k(\bbeta_k) + R_k(\bbeta_k) = \arg\min_{\bbeta_k}\bigg\{\|\bbeta_k\|^2-\frac{2}{n}\sum_{i=1}^{n}\wy_i\wZik\langle\bbeta_k,\wSXi\rangle+\lambda_k\|\bbeta_k\|_1 \bigg\},
\end{align}
which can be obtained in a closed form as
\begin{align*}
\hbeta_k = \T_{\lambda_k/2}\bigg(n^{-1}\sum_{i=1}^n\wyi\wZik\wSXi\bigg).
\end{align*}
Compared to the estimator in \eqref{eq:emp_loss:warmup}, we have
replaced $\bX_i$ by $S(\bX_i)$ and have carefully truncated the data
to obtain the following result.

\begin{theorem}[Single sparse vector recovery]\label{thm:4}

Consider the model \eqref{mod:1} with $\|\tbeta_k\|_0 \leq s$, $\forall k \in [d_2]$.  Suppose Assumption \ref{ass:1}, Assumption \ref{ass:3} $(p=6)$ hold and $\mE[z_jz_k] = 0$ for $j\neq k$, then the estimator defined in \eqref{equ:6} with $\lambda_k = 76\sqrt{M_6\log d_1d_2/n}$ and $\tau = (M_6n/\log d_1d_2)^{1/6}/2$ satisfies
\begin{align*}
\|\hat{\bbeta}_k-\tilde{\bbeta}_k\|_2\leq \frac{3}{2}\sqrt{s}\lambda_k \text{\ \ and\ \ }\|\hat{\bbeta}_k-\tilde{\bbeta}_k\|_1\leq 6s\lambda_k, \text{\ \ \ } \forall k\in[d_2],
\end{align*}
with probability at least $1-2/d_1^2d_2^2$.

\end{theorem}

The theorem establishes that
\begin{align*}
\|\hat{\bbeta}_k-\tilde{\bbeta}_k\|_2\lesssim \sqrt{\frac{s \log d_1d_2}{n}} \text{\ \ and\ \ }
\|\hat{\bbeta}_k-\tilde{\bbeta}_k\|_1\lesssim s\sqrt{\frac{\log d_1d_2}{n}}
\end{align*}
with high probability. When $d_2 = 1$ the rate matches the minimax
rate established in \cite{lin2017optimality}. Our proof technique requires finite $6$-th moment, which ensures that the truncated
variables do not lose too much information. This assumption can be
compared to boundedness of the $4$-th moment in estimation of a single-index model \citep{yang2017high}. We require a stronger assumption due to estimation in a more general model. Theorem
\ref{thm:4} follows from a bound on $\|\nabla \bar{L}_k(\tilde{\bbeta}_k)\|_{\infty}$ given in the following lemma.

\begin{lemma}\label{lem:6}

Under the conditions of Theorem \ref{thm:4},
\begin{align*}
P\bigg(\|\nabla \bar{L}_k(\tilde{\bbeta}_k)\|_{\infty}\leq 38\sqrt{\frac{M_6\log d_1d_2}{n}}, \text{\ \ } \forall k\in[d_2]\bigg)\geq 1 - \frac{2}{d_1^2d_2^2}.
\end{align*}

\end{lemma}

From the standard analysis of the $\ell_1$-penalized methods in
\cite{Buehlmann2011book}, we know that the penalty parameter
$\lambda_k$ should be set as $c\|\nabla \hat{L}_k(\tilde{\bbeta}_k)\|_{\infty}$ for some $c >
0$. Moreover, we see threshold $\tau$ has the order $\tau \asymp 1/\lambda_k^{2/p}$, where $p$ is the number of moments variables have. Thus, the more moments the variables have, the smaller the threshold level $\tau$ would be, which is also consistent with our intuition.

We note that the estimator in \eqref{equ:6} crucially depends on the
independence among coordinates of $\bz$. Without this assumption the
estimator is not valid. In what follows, we study estimators of the
matrix $\tB$ as a whole, by imposing either low-rank or sparse
structure, instead of estimating the matrix column by column.

\section{Low-rank Matrix Recovery}\label{sec:5}

In this section, we propose an estimator for $\tB$ in model \eqref{mod:1}, which has near optimal rate of convergence under an
assumption that $\tB$ is low-rank. We relax the condition that
$\mE[z_jz_k] = 0$ as assumed earlier, by estimating the inverse of the
covariance of $\bz$, also called the precision matrix. Let $\tSigma = \mE[\bz\bz^\top]\in\mR^{d_2\times d_2}$ and $\tOmega = (\tSigma)^{-1}$. We consider two cases: (i) when $d_2$ is in low dimensions, we have no structural assumptions on the precision matrix $\tOmega$; (ii) when $d_2$ is in high dimensions, we assume that the precision matrix is in the set $\mF_w^K$ for some $w$ and $K$, where
\begin{align*}
\mF_w^K = \bigg\{\bOmega\succeq \0:  \|\bOmega\|_{0,\infty}\leq w, \|\bOmega\|_2\leq K, \|\bOmega^{-1}\|_2\leq K\bigg\}.
\end{align*}
The set above is borrowed from \cite{Cai2011Constrained} and it
controls upper and lower bounds on eigenvalues of $\tOmega$, as well
as the maximal sparsity over columns. Since estimating the precision
matrix itself is a well studied topic and can be conducted independently from estimating model~(\ref{mod:1}), we only take the
former case as an example. For the latter case, the sparsity structure
on precision matrix can allow us to study the model with both
$d_1, d_2$ in high dimensions. We will discuss how to make use of the
CLIME procedure \citep{Cai2011Constrained} with truncated sample
covariance to estimate the precision matrix for heavy-tailed variable
in Appendix \ref{supple3}.

We start by writing down the identifiability relationship. Under
Assumption \ref{ass:1}, we have
\begin{align}\label{equ:7}
\mE[y\cdot S(\bx)\bz^T]\tOmega&=\sum_{j=1}^{d_2}\mE[f_j(\langle\tbeta_j,\bx\rangle)S(\bx)]\mE[z_j\cdot \bz^T]\tOmega \notag \\
& =\sum_{j=1}^{d_2}\ttbeta_j\be_j^T\tSigma\tOmega = (\ttbeta_1,...,\ttbeta_{d_2}) = \ttB,
\end{align}
where $\be_j \in \mR^{d_2}$ is the $j$-th canonical basis vector. This
relationship allows us to estimate the $\ttB$ as a minimizer of the
population loss,
\begin{align}\label{equ:8}
\ttB = \arg\min_{\bB}\bigg\{\|\bB\|_{F}^2-2\mE[y\cdot \langle S(\bx)\bz^T\tOmega, \bB\rangle]\bigg\}.
\end{align}
In order to use the above relationship, we will separately estimate
$\mE[y\cdot S(\bx)\bz^T]$ and $\tOmega$.

Let
\begin{equation}\label{equ:9}
\begin{aligned}
\phi(x)=\begin{cases}
-\log (1-x+x^2/2)\text{\ \ \ if\ } x\leq0, \\
\log(1+x+x^2/2)\text{\ \ \ \ \ \ if\ } x> 0
\end{cases}
\end{aligned}
\end{equation}
be the soft truncation function. When $|x|$ is small $\phi(x)\approx x$; when $|x|$ is relatively large $\phi(x)$ dramatically shrinks $x$, while still being monotonically increasing. This function has been widely used in robust mean estimation, especially for variables that only have finite certain moments. For the univariate case, suppose $v_1,\ldots, v_n$ to be a sequence of i.i.d.~samples distributed as $v_i\sim v$ that only has finite $2$nd moment. \cite{Catoni2012Challenging} proposed a simple mean estimator defined as $(n\kappa)^{-1}\sum_{i=1}^{n}\phi(\kappa v_i)$ with a properly chosen $\kappa$ and showed sub-Gaussian tail
bound. \cite{Minsker2018Sub} generalized to the multivariate case
using estimator $(n\kappa)^{-1} \sum_{i=1}^{n}\phi(\kappa \bV_i)$, where $\bV_i$ are independent random Hermitian matrices with finite
$2$nd moments and $\phi(\cdot)$ is applied to the eigenvalues only. Similarly, when $\bV_i$ are not Hermitian, we can define a dimension-free matrix soft truncation function $\Phi(\cdot)$ using Hermitian dilation as follows: for a matrix $\bV$, let
$\begin{pmatrix}
  \0 & \bV\\
  \bV^T & \0
\end{pmatrix} = \bQ\bLambda \bQ^T$ be the eigenvalue decomposition of the Hermitian dilation of $\bV$. Let $\tilde{\bU} = \bQ\phi(\bLambda)\bQ^T$, where $\phi(\bLambda)$ is applied entrywise. Then $\Phi(\bV)$ is defined to be the upper right corner matrix of $\tilde{\bU}$ with the same dimension as $\bV$. Using the truncation function $\Phi(\cdot)$, our estimator for $\mE[yS(\bx)\bz^T]$ is further defined as
\begin{align}\label{equ:10}
\frac{1}{n\kappa_1}\sum_{i=1}^n\Phi(\kappa_1 y_i\cdot S(\bX_i)\bZ_i^T),
\end{align}
where $\kappa_1 > 0$ is a user-specified parameter. In the later
section, we will also apply the function $\Phi(\cdot)$ when estimating
precision matrix for heavy-tailed variables in low dimensions.  With
the matrix in (\ref{equ:10}), our estimator of $\ttB$ is given as
\begin{align}\label{equ:12}
\hB=\arg\min_{\bB}\bigg\{\|\bB\|_{F}^2-\frac{2}{n\kappa_1}\sum_{i=1}^{n}\langle \Phi(\kappa_1 y_i\cdot S(\bX_i)\bZ_i^T)\hOmega, \bB\rangle+\lambda\|\bB\|_*\bigg\},
\end{align}
where $\hOmega$ is an estimator of $\tOmega$. The penalty function
$\lambda\|\bB\|_*$ biases the estimated matrix $\hB$ to be low-rank.
Note that the estimator $\hB$ can be obtained in a closed form as
\begin{align*}
\hB =\T_{\lambda/2}\bigg(\frac{1}{n\kappa_1}\sum_{i=1}^{n}\Phi(\kappa_1 y_i\cdot S(\bX_i)\bZ_i^T)\hOmega\bigg).
\end{align*}

We characterize convergence rate for the estimator \eqref{equ:12}
in the next theorem and discuss estimation of a general low-dimensional $\tOmega$ for heavy-tailed covariate later.

\begin{theorem}[Low-rank matrix recovery]\label{thm:5}

Consider the model (\ref{mod:1}) with $\text{rank}(\tB)\leq r$. Suppose Assumption \ref{ass:1}, Assumption \ref{ass:3} $(p=4)$ hold and furthermore suppose a precision matrix estimator $\hOmega$ satisfies
\begin{align*}
P\big(\|\hOmega - \tOmega\|_2\leq \H(n,d_2)\big)\geq 1 - \P(n,d_2).
\end{align*}
Denote $K = \|\tSigma\|_2 \vee \|\tOmega\|_2$. If we set $\kappa_1 = \sqrt{\frac{2\log(d_1+d_2)}{n(d_1+d_2)M_4^{3/2}}}$ and
\begin{align*}
\lambda \geq 16KM_4^{3/4}\sqrt{\frac{(d_1+d_2)\log(d_1+d_2)}{n}} + 4K\max_{j\in[d_2]} |\mu_j|\cdot\|\tB\|_2\cdot\H(n,d_2),
\end{align*}
then the estimator \eqref{equ:12} satisfies
\begin{align*}
\|\hB - \ttB\|_F\leq &3\sqrt{r}\lambda \text{\ \ and\ \ }	\|\hB - \ttB\|_*\leq 24 r\lambda.
\end{align*}
with probability at least $1-2/(d_1+d_2)^2 - \P(n,d_2)$.

\end{theorem}

The theorem follows from the following concentration result.

\begin{lemma}\label{lem:7}

Under the conditions in Theorem \ref{thm:5}, we have
\begin{align*}
\bignorm{ \frac{1}{n\kappa_1}\sum_{i=1}^{ n}\Phi(\kappa_1 y_i\cdot S(\bX_i)\bZ_i^T)-\mE[y\cdot S(\bx)\bz^T]}_2\leq 4 M_4^{3/4}\sqrt{\frac{(d_1+d_2)\log(d_1+d_2)}{n}},
\end{align*}
with probability at least $1 - \frac{2}{(d_1+d_2)^2}$.

\end{lemma}

Different from Theorem \ref{thm:4}, the penalty parameter $\lambda$
depends also on a term that comes from estimating $\tOmega$. Specifically, if it holds that $\mE[z_jz_k] = 0$, we can get the following corollary immediately.

\begin{corollary}\label{thm:7}

Suppose the conditions of Theorem~\ref{thm:5} are satisfied. In addition, suppose that $\mE[\bz\bz^\top] = \bI_{d_2}$. If we set $\kappa_1 = \sqrt{\frac{2\log(d_1+d_2)}{n(d_1+d_2)M_4^{3/2}}}$ and $\lambda = 16M_4^{3/4} \sqrt{\frac{(d_1+d_2)\log(d_1+d_2)}{n}}$, then
\begin{align*}
\|\hB - \ttB\|_F\leq &3\sqrt{r}\lambda \text{\ \ and\ \ }	\|\hB - \ttB\|_*\leq 24 r\lambda,
\end{align*}
with probability at least $1-2/(d_1+d_2)^2$.

\end{corollary}

Next, we briefly discuss how to estimate the precision matrix $\tOmega$, noting that any suitable estimator for heavy-tailed data
can be used. In a general case, when no additional structural assumptions are available, we can invert the soft truncated empirical
covariance matrix as
\begin{align}\label{equ:17}
\hOmega = \hSigma^{-1} \quad\text{where}\quad \hSigma = \frac{1}{n\kappa_2}\sum_{i=1}^{n}\Phi(\kappa_2 \bZ_i\bZ_i^T).
\end{align}
We will show that $\hSigma$ is invertible for sufficiently large
$n$. In particular, we prove $\|\hSigma- \tSigma\|_2\lesssim \sqrt{d_2\log d_2/n}$. Thus, $\hSigma$ is invertible when $\sqrt{d_2\log d_2/n} < \lambda_{\min}(\tSigma)$, where $\lambda_{\min}(\tSigma)$ denotes the minimum eigenvalue of $\tSigma$. The following lemma characterizes the rate of convergence.

\begin{lemma}\label{lem:8a}

Set $\kappa_2 = \sqrt{\frac{2\log d_2}{nd_2M_4^{1/2}}}$. If $n\geq 64\sqrt{M_4}K^2d_2\log d_2$, the estimator \eqref{equ:17} satisfies
\begin{align*}
P\bigg(\|\hOmega - \tOmega\|_2\leq 8K^2M_4^{1/4}\sqrt{\frac{d_2\log d_2}{n}}\bigg)\geq 1-\frac{2}{d_2^2}.
\end{align*}

\end{lemma}

In fact, we only need finite $2$nd moment for $\bz$ to make $\hOmega$ in (\ref{equ:17}) consistent. Combining the rate obtained in Lemma \ref{lem:8a} with that of Theorem \ref{thm:5}, we observe that
\begin{align*}
\|\hB - \ttB\|_F \lesssim \sqrt{\frac{r(d_1+d_2)\log(d_1+d_2)}{n}}.
\end{align*}
with high probability. In particular, the rate of convergence is governed by the rate obtained in Lemma~\ref{lem:7} and the estimation of the precision matrix contributes to the higher order terms. Furthermore, we note that the rate is optimal up to logarithmic terms \citep{rohde2011estimation}. Similar rate is shown in estimating the single-index model \citep{plan2016generalized, goldstein2016structured, yang2017high}.  Estimation of a high-dimensional sparse precision matrix is presented in Appendix \ref{supple3}.

\section{Sparse Matrix  Recovery}\label{sec:6}

In this section, we consider the setting as in Section~\ref{sec:5}, but with the parameter matrix $\tB$ being sparse rather than low-rank.
Different from (\ref{equ:10}), here we can simply estimate $\mE[y\cdot S(\bx)\bz^T]$ by
\begin{align}\label{equ:18}
\frac{1}{n}\sum_{i=1}^{n}\wy_i\cdot\wSXi\wZi^T
\end{align}
for some truncation threshold $\tau>0$. We apply hard truncation instead of soft truncation (based on $\phi(\cdot)$) in (\ref{equ:18}), since we are going to bound the max norm of the error, instead of the operator norm as we did in Section \ref{sec:5}. The hard truncation can give a fast rate when bounding the max norm, while a slow rate when bounding the operator norm. The soft truncation works in the opposite way. Finally, our estimator is defined as
\begin{align}\label{equ:13}
\hB = \arg\min_{\bB}\bigg\{\|\bB\|_F^2-\frac{2}{n}\sum_{i=1}^{n}\LD \wy_i\cdot\wSXi \wZi^T\hOmega, \bB\RD + \lambda\|\bB\|_{1,1}\bigg\}.
\end{align}

We obtain the following rate of convergence for $\hB$.

\begin{theorem}[Sparse matrix recovery (column-wise sparse)]\label{thm:8}

Consider the model (\ref{mod:1}) with $\|\tbeta_k\|_0\leq s$ for all $k\in[d_2]$. Suppose Assumption \ref{ass:1} and Assumption \ref{ass:3} $(p=6)$ hold and furthermore suppose that the precision matrix estimator $\hOmega$ satisfies
\begin{align*}
P(\|\hOmega - \tOmega\|_{\max}\leq \tH(n,d_2))\geq 1 - \tP(n,d_2).
\end{align*}
If $\tau = (M_6n/\log d_1 d_2)^{1/6}/2$ in \eqref{equ:18} and
\begin{align*}
\lambda \geq 76\|\tOmega\|_1\sqrt{\frac{M_6\log d_1d_2}{n}} + 4\max_{j\in[d_2]}|\mu_j|\cdot \|\tB\tSigma\|_{\infty}\tH(n,d_2),
\end{align*}
then
\begin{align*}
\|\hB - \ttB\|_F\leq 2\sqrt{sd_2}\lambda \text{\ \ and\ \ } \|\hB - \ttB\|_{1,1}\leq 8sd_2\lambda,
\end{align*}
with probability at least $1-2/d_1^2d_2^2 - \tP(n,d_2)$.

\end{theorem}

Different from Theorem \ref{thm:5}, we require the bound $\|\hOmega - \tOmega\|_{\max}$ with high probability here because $\|\cdot\|_{\max}$ is the dual norm of $\|\cdot\|_{1,1}$. Note that $\|\hOmega - \tOmega\|_{\max}\leq \|\hOmega - \tOmega\|_{2}$, so we
can simply have $\tH(n,d_2) = \H(n, d_2)$ and $\tP(n,d_2) = \P(n, d_2)$ for estimation in low dimensions where $\H(n,d_2)$ and $\P(n,d_2)$ come from Lemma \ref{lem:8a}. However, this bound is not sharp for CLIME procedure in high dimensions. Above theorem follows from the following lemma.

\begin{lemma}\label{lem:10}

Under the conditions in Theorem \ref{thm:8},
\begin{align*}
\bignorm{\mE[y\cdot S(\bx)\bz^T] -  \frac{1}{n}\sum_{i=1}^{n}\wy_i\cdot\wSXi\wZi^T}_{\max}\leq 19 \sqrt{\frac{M_6\log d_1d_2}{n}}
\end{align*}
with probability at least $1 - 2/d_1^2d_2^2$.

\end{lemma}

Note that the rate obtained in Theorem~\ref{thm:8} is the same as the one obtained in Theorem~\ref{thm:4}, which required the assumption
that $\mE[z_jz_k] = 0$. Furthermore, we observe that the same proof provided for Theorem~\ref{thm:8} can be used under the setting that
$n\ll d_1\wedge d_2$ and $\tB$ is fully sparse with $\|\tB\|_{0,1}\leq s$. Estimation of a high-dimensional precision matrix is discussed in Appendix \ref{supple3}, but we note that the error in estimating the precision matrix in any case only contributes higher order terms and our final rate is
\begin{align*}
\|\hB - \ttB\|_F\lesssim \sqrt{\frac{s\log d_1d_2}{n}}
\text{\ \ and\ \ } \|\hB - \ttB\|_{1,1}\lesssim s \sqrt{\frac{\log d_1d_2}{n}}
\end{align*}
with probability at least $1-2/d_1d_2 - 2/d_2^2$. It follows by
combining Theorem \ref{thm:8} with Lemma \ref{lem:8b}. Finally, similar to Corollary \ref{thm:7}, when $\tSigma = \bI_{d_2}$, we can set $\tH(n, d_2) = \tP(n,d_2) = 0$ in Theorem \ref{thm:8} to derive the same optimal rate.

Until now, we have shown a comprehensive theoretical analysis for the
model (\ref{mod:1}). When estimating a single sparse vector, we assume
that $\bz$ has independent entries. When estimating a parameter matrix $\tB$, we relax this assumption by incorporating an estimate of the precision matrix. Based on our analysis, we see the error occurred in
estimating $\mE[y\cdot S(\bx)\bz^T]$ will always be the dominant term,
while estimation of the precision matrix only contributes higher order terms.

\section{Numerical Experiment}\label{sec:7}

\begin{figure}[!b]
	\centering     
	\subfigure[$f^{(1)}_k$ function]{\label{Link1}\includegraphics[width=49.9mm]{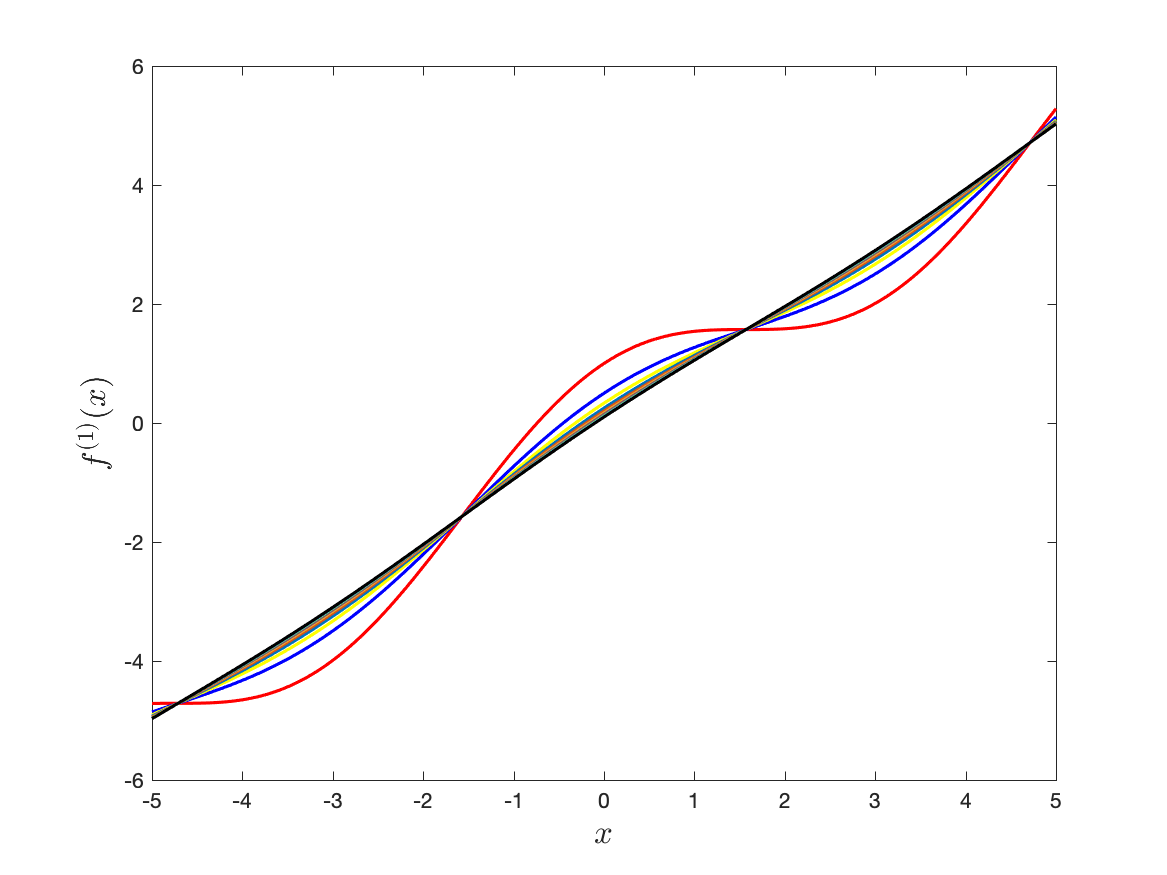}}
	\subfigure[$f^{(2)}_k$ function]{\label{Link2}\includegraphics[width=49.9mm]{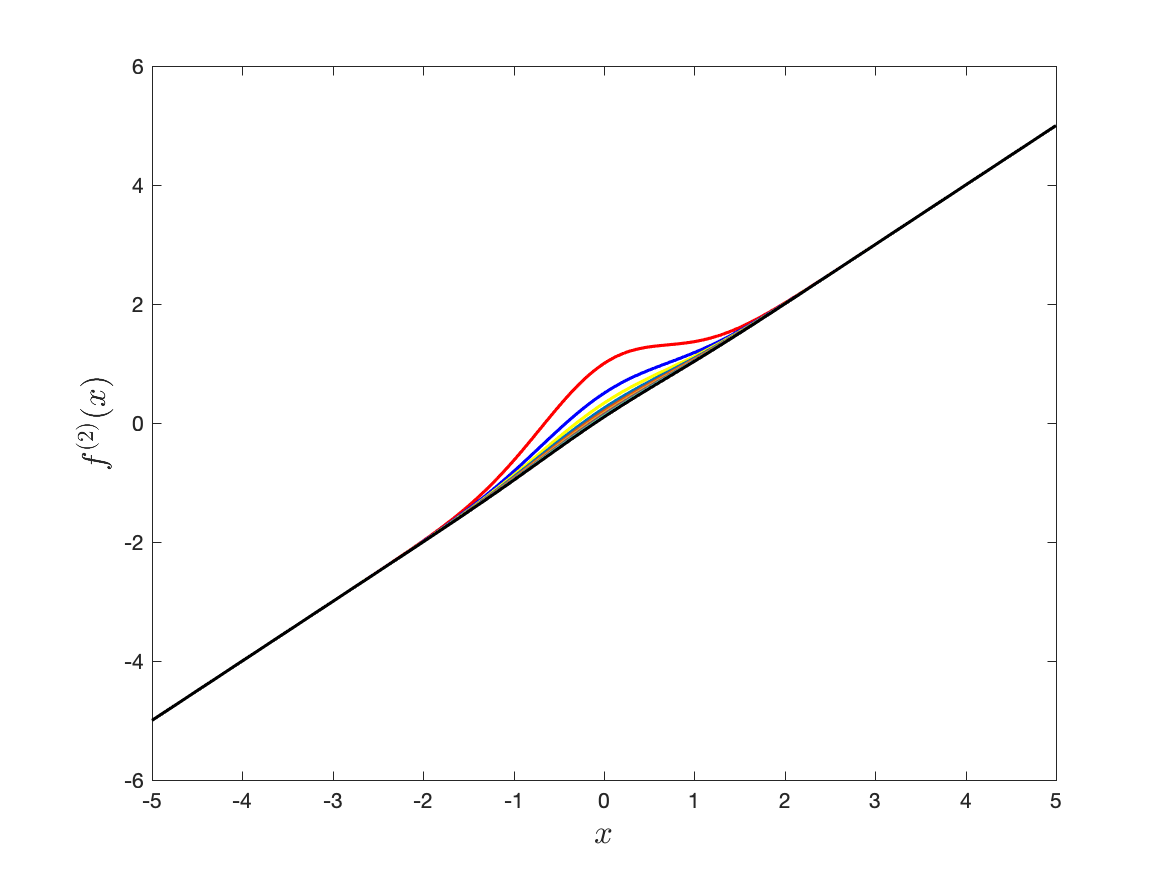}}
	\subfigure[$f^{(3)}_k$ function]{\label{Link3}\includegraphics[width=49.9mm]{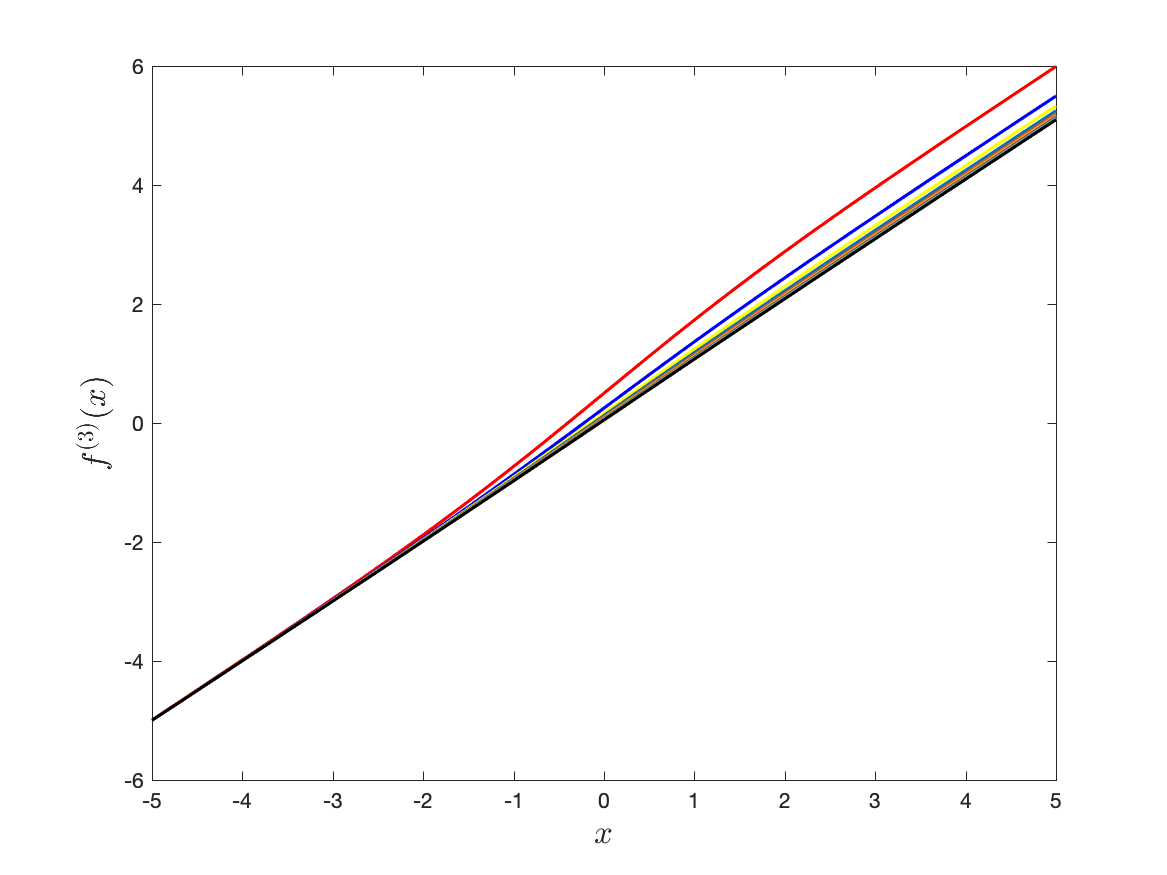}}
	\subfigure[$f^{(4)}_k$ function]{\label{Link4}\includegraphics[width=49.9mm]{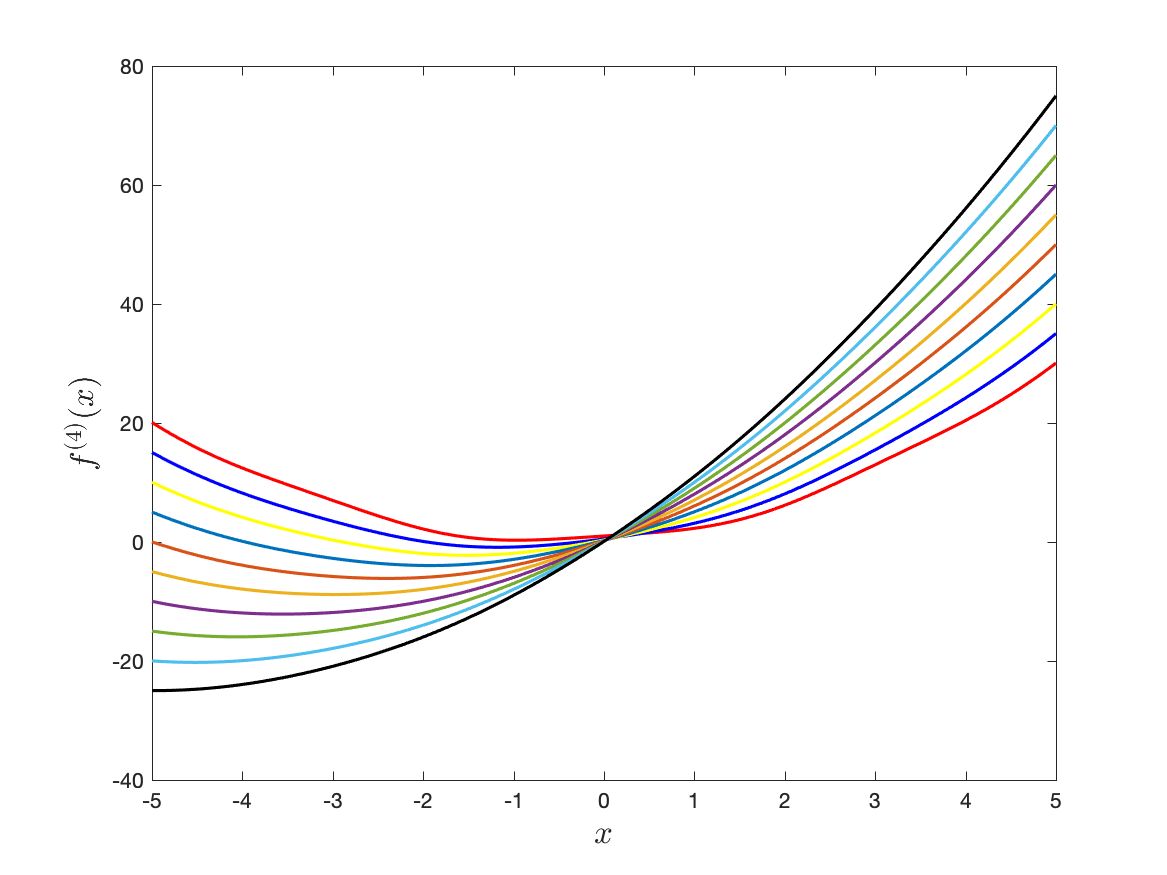}}
	\subfigure[$f^{(5)}_k$ function]{\label{Link5}\includegraphics[width=49.9mm]{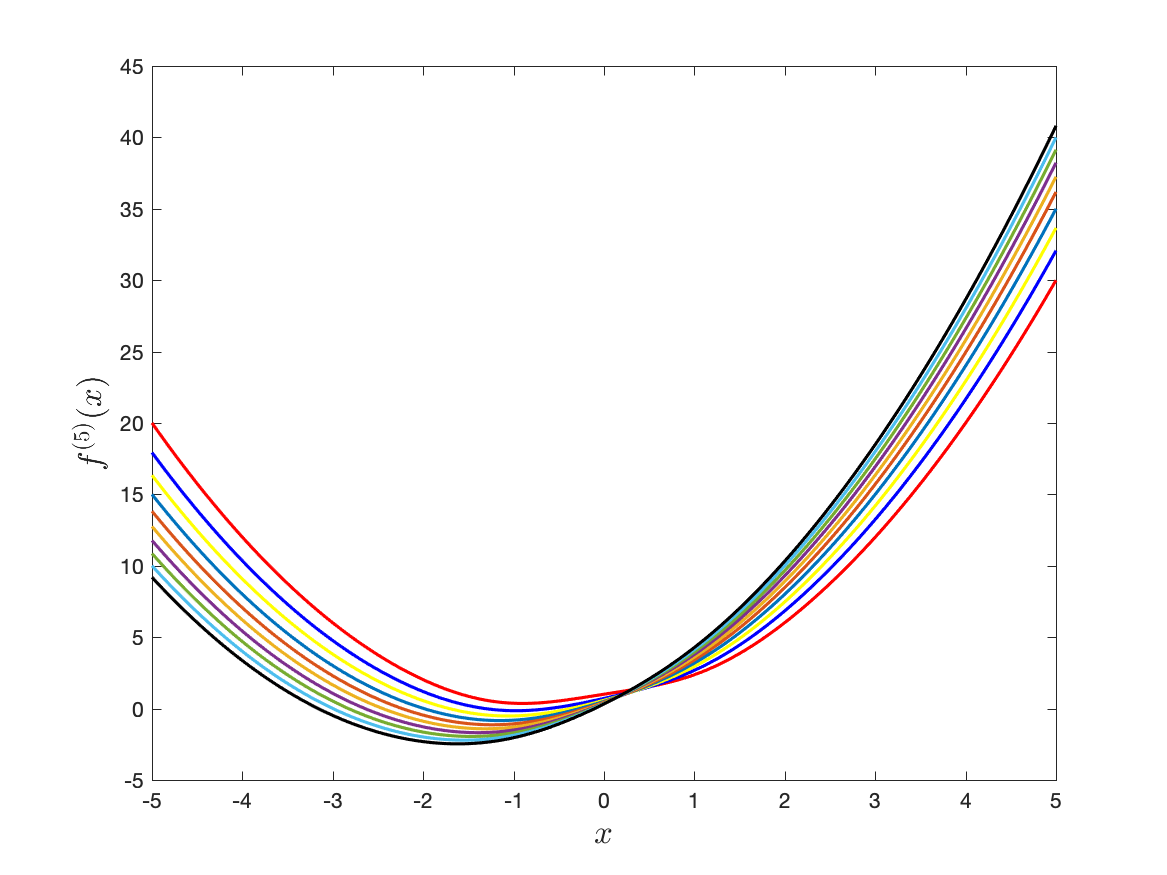}}
	\subfigure[$f^{(6)}_k$ function]{\label{Link6}\includegraphics[width=49.9mm]{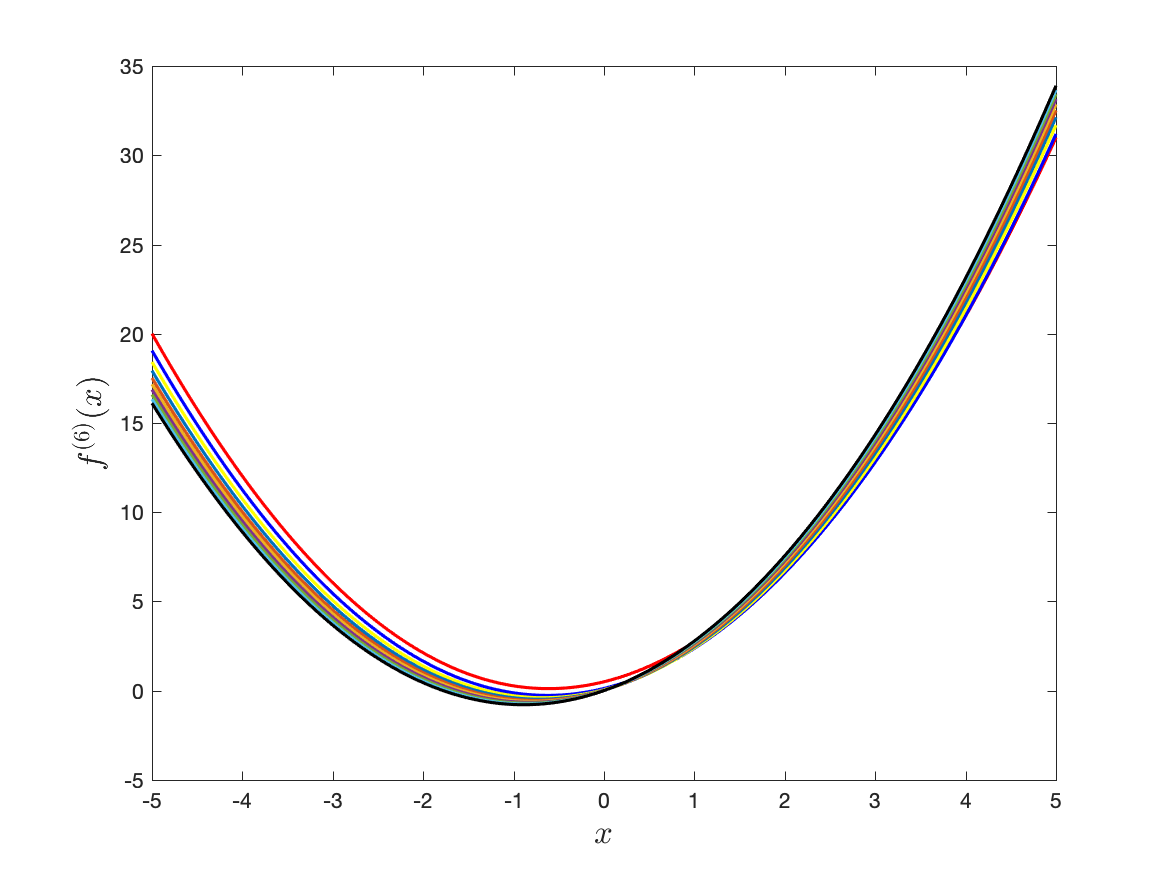}}
	\caption{{The link functions used in simulations. When $k$ varies from $1$ to $10$, the line moves from red to black. The link functions in the first row are essentially linear functions combined with different fluctuations, while in the second row they are quadratic. As $k$ increases, the fluctuation is more and more moderate.}}\label{fig:1}
\end{figure}

In this section, we illustrate the performance of our proposed estimators in different simulation settings and carry out a real data
application in the end. In simulation studies, the link function is
set to one of the following six forms:
\begin{align*}
f^{(1)}_k(x) =& x + \frac{1}{k}\cos(x);\text{\ \ \ \ \ \ \ \ \ \ \ }f^{(2)}_k(x) = x + \frac{1}{k}\exp(-x^2);\text{\ \ \ \ \ }  f^{(3)}_k(x)  = x+\frac{1}{k}\frac{\exp(x)}{1+\exp(x)};\\
f^{(4)}_k(x) =& x^2 + kx + \frac{1}{k}\cos^2(x); \text{\ \ } f^{(5)}_k(x) = x^2 + \sqrt{k}x + \frac{1}{\sqrt{k}}\exp(-x^2); \\
f^{(6)}_k(x) =& x^2 + k^{\frac{1}{4}}x + \frac{1}{k^2}\frac{\exp(x)}{1 + \exp(x)}. \text{\ \ \ } &
\end{align*}
Their plots are shown in Figure \ref{fig:1}. For all simulations we let $\epsilon\sim N(0,0.01)$. To measure the estimation accuracy we
use the cosine distance defined by $\cos(\hat{\bbeta}, \tbeta) = 1 -
|\hat{\bbeta}^T\tbeta|/\|\hat{\bbeta}\|_2$. Note that we do not normalize $\hbeta$ and change its direction according to the sign of
its first entry because cosine distance is more suitable for verifying
our matrix results and it allows us to generate $\tbeta$ without restricting the first entry to be positive. We can easily see $\cos(\hat{\bbeta}_k,\tbeta_k)\asymp\|\hat{\bbeta}_k-\ttbeta_k\|^2_2$,
$\forall k\in[d_2]$. For a matrix estimator, we will sum up cosine
distance over all columns, which is a rough surrogate of $\|\hB - \ttB\|_F^2$. Our results are averaged of 40 independent runs.

\subsection{Single Sparse Vector} \label{sec:single-sparse-vector}

We set $d_1 = 100$, $d_2=20$, $s = 10$, and vary $n$ only. For $i\in[n]$ and $k\in[d_2]$, we let $\bX_i$ be independent and identically generated from one of distributions listed in Table \ref{tab:2}, and $Z_{ik} \in \{-1, 1\}$ with equal probability and independent from other coordinates. To generate $d_1$-dimensional
parameter $\tbeta_k$, we first generate the support of nonzero coefficients $S_k$ uniformly at random and then let $\{[\tbeta_{k}]_l\}_{l\in S_k}\stackrel{iid}{\sim}\frac{1}{\sqrt{s}}\cdot{\rm Unif}(\{-1,1\})$. According to Theorem~\ref{thm:conv_warmup} and \ref{thm:4}, we set $\lambda_k = 30\sqrt{\log d_1d_2/n}$ and
$\tau = 2(n/\log d_1d_2)^{1/6}$.

\begin{table}[!t]
	\centering
	\begin{tabular}{ l|c|ccr }
		\hline
		Distribution & parameter & score function \\
          \hline
          \hline
		Gaussian & $\mu = 0$; $\sigma = 1$ & $s(x) = x$\\
		\hline
		Beta & $\alpha = 8$; $\beta = 8$ & $s(x) = \frac{14x-7}{(1-x)x}$\\
		\hline
		Gamma & $k = 8$; $\theta = 0.1$ & $s(x) =10-\frac{7}{x}$\\
		\hline
		Student's t & $\nu = 13$ & $s(x) = \frac{14x}{13+x^2}$\\
		\hline
		Rayleigh &  $\sigma = 1$ & $s(x)=x-\frac{1}{x}$\\
		\hline
		Weibull & $k = 7$; $\lambda = 1$ & $s(x) = 7x^6 - \frac{6}{x}$\\
          \hline
          \hline
	\end{tabular}
	\caption{\textit{Distribution of $\bx$}}
	\label{tab:2}
\end{table}

\begin{figure}[t]
\centering     
\subfigure[Gaussian for $\tbeta_1$]{\label{w1}\includegraphics[width=49.9mm]{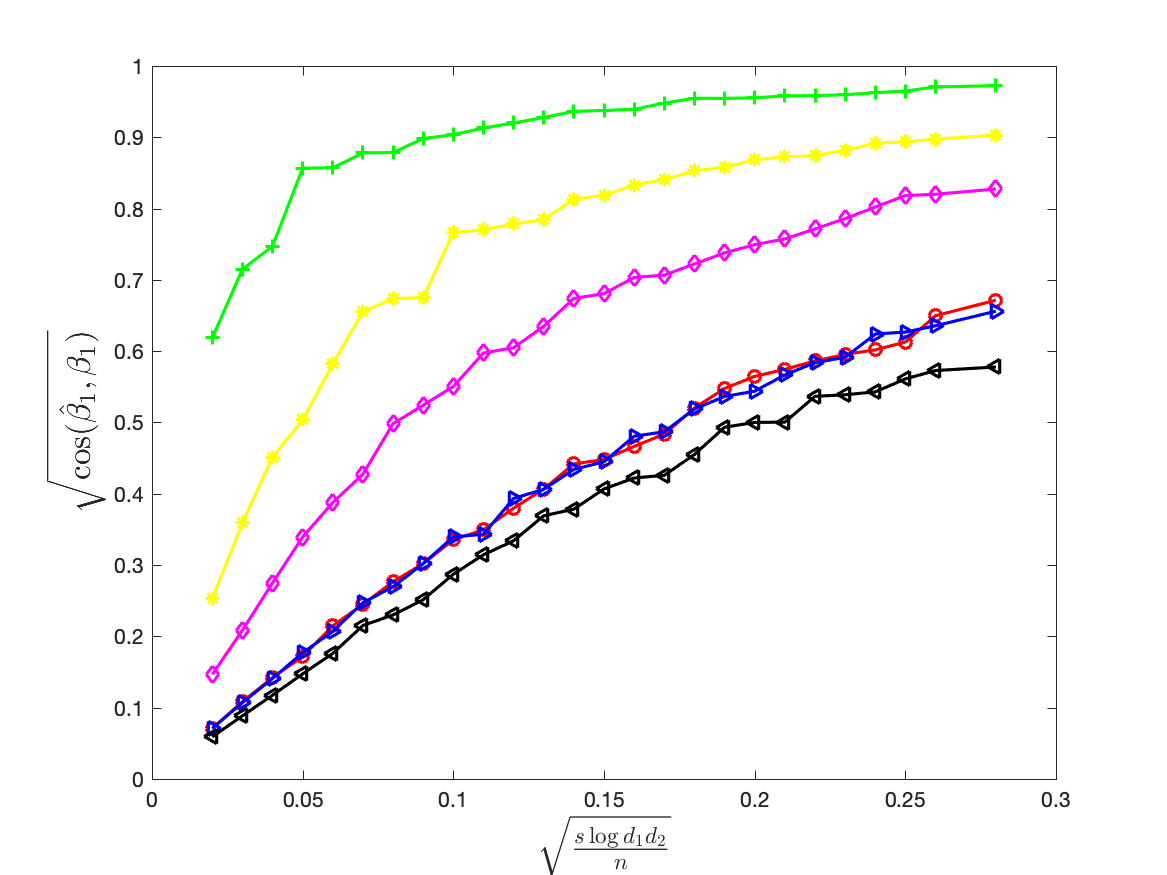}}
\subfigure[Gaussian for $\tbeta_{d_2/2}$]{\label{w2}\includegraphics[width=49.9mm]{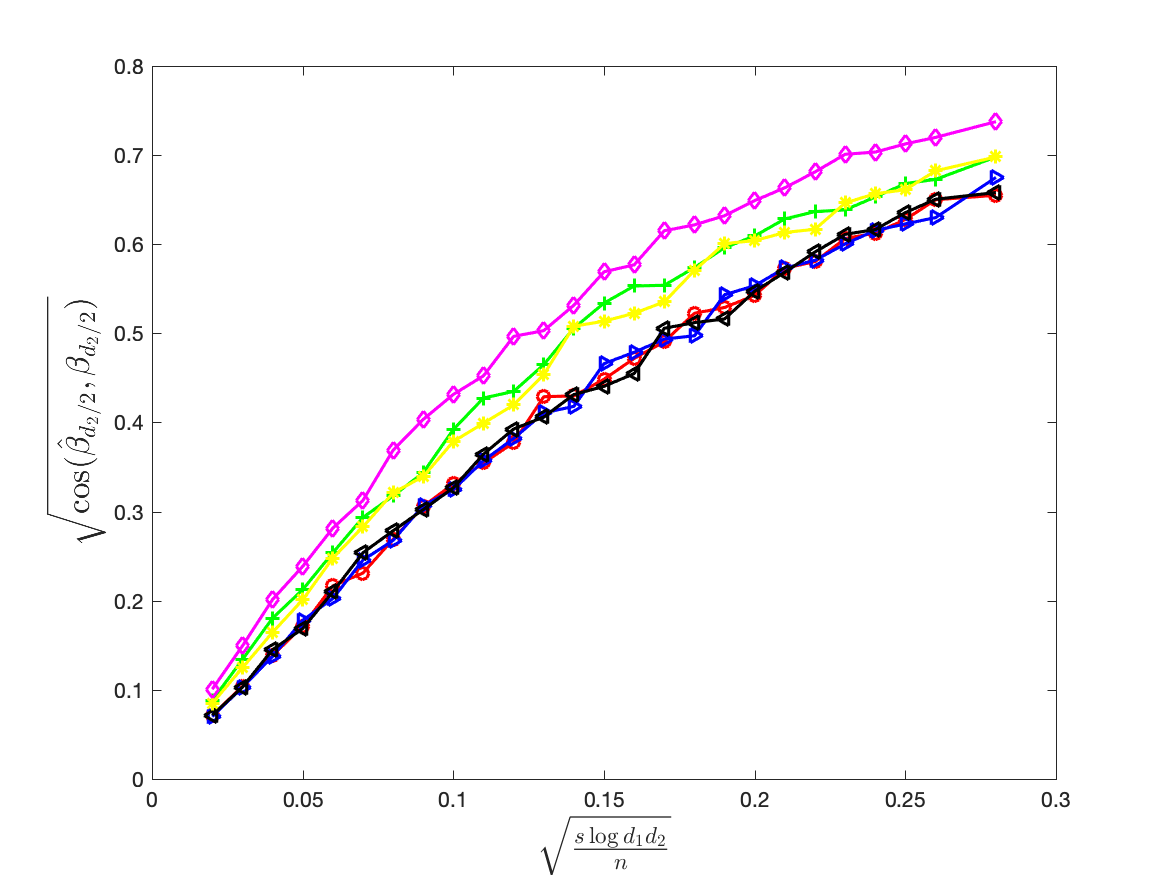}}
\subfigure[Gaussian for $\tbeta_{d_2}$]{\label{w3}\includegraphics[width=49.9mm]{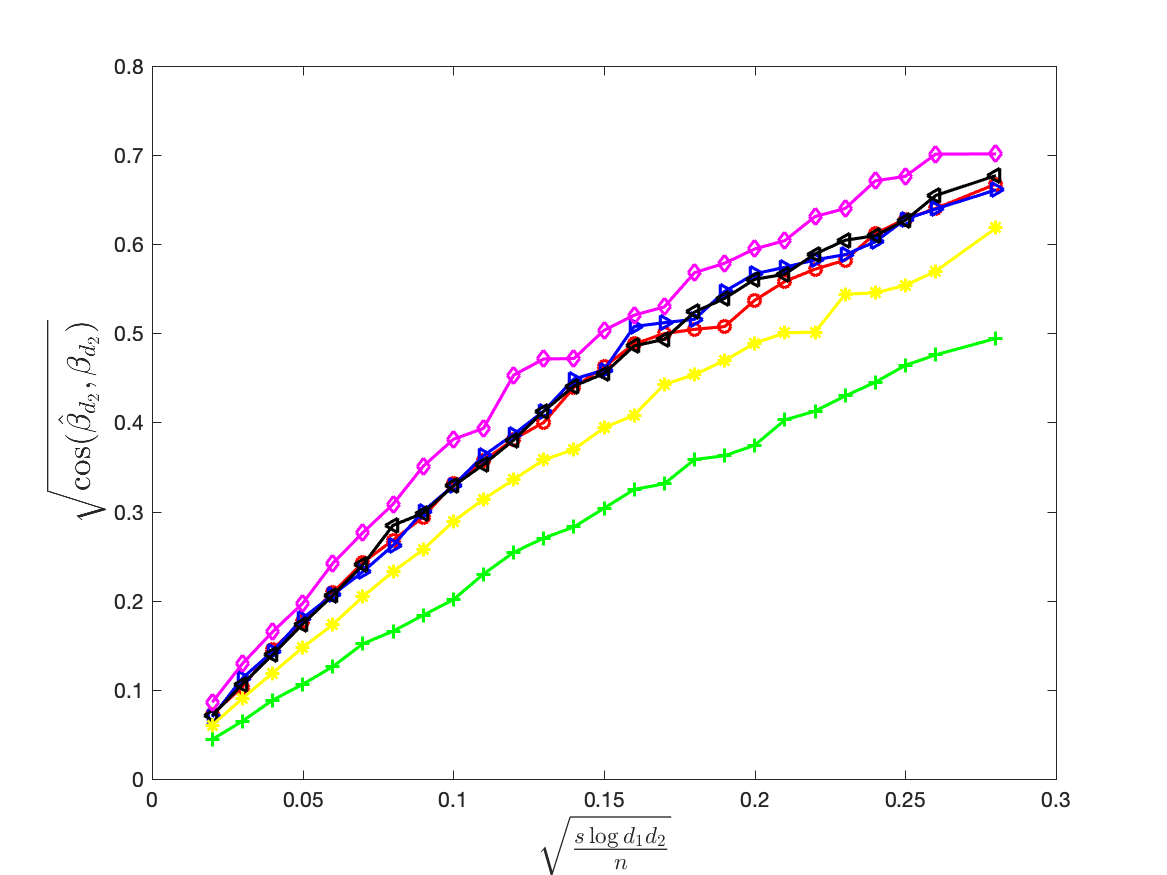}}
\subfigure[Beta for $\tbeta_1$]{\label{beta1}\includegraphics[width=49.9mm]{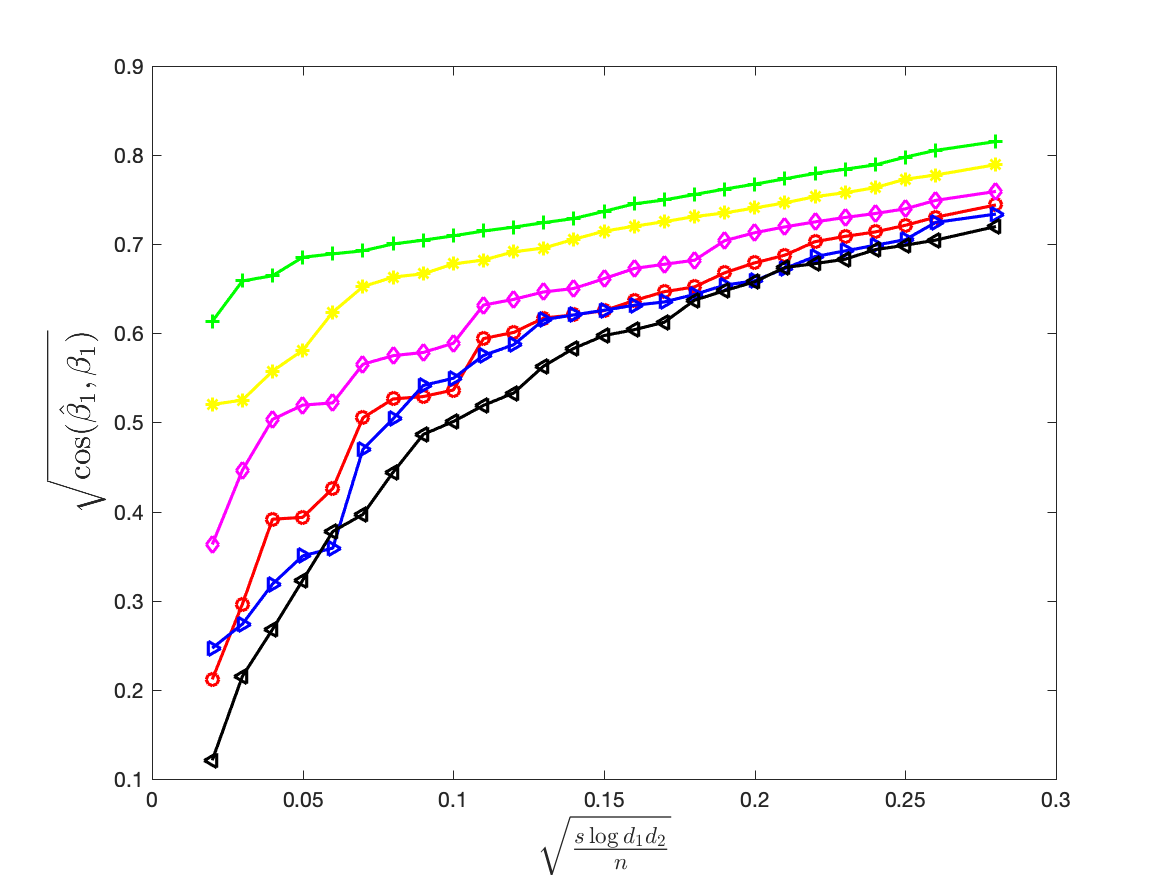}}
\subfigure[Beta for $\tbeta_{d_2/2}$]{\label{beta2}\includegraphics[width=49.9mm]{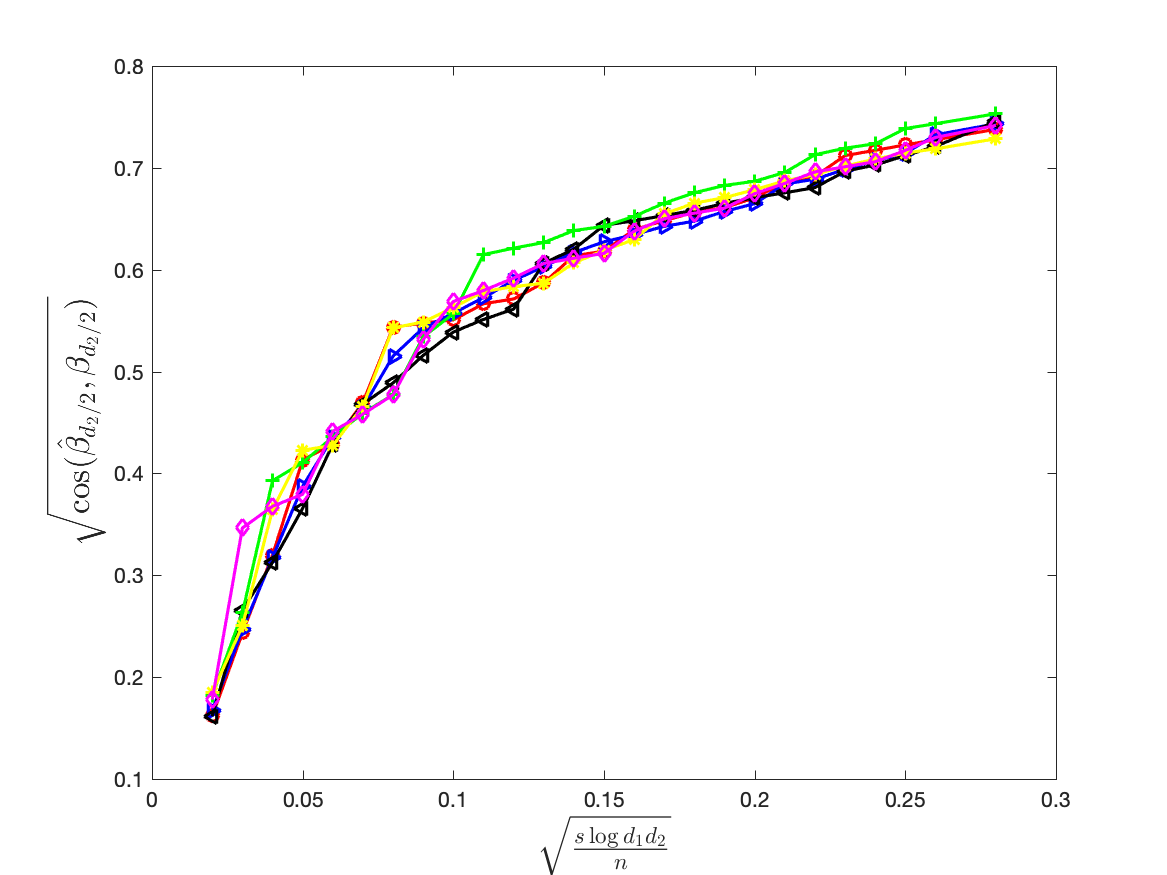}}
\subfigure[Beta for $\tbeta_{d_2}$]{\label{beta3}\includegraphics[width=49.9mm]{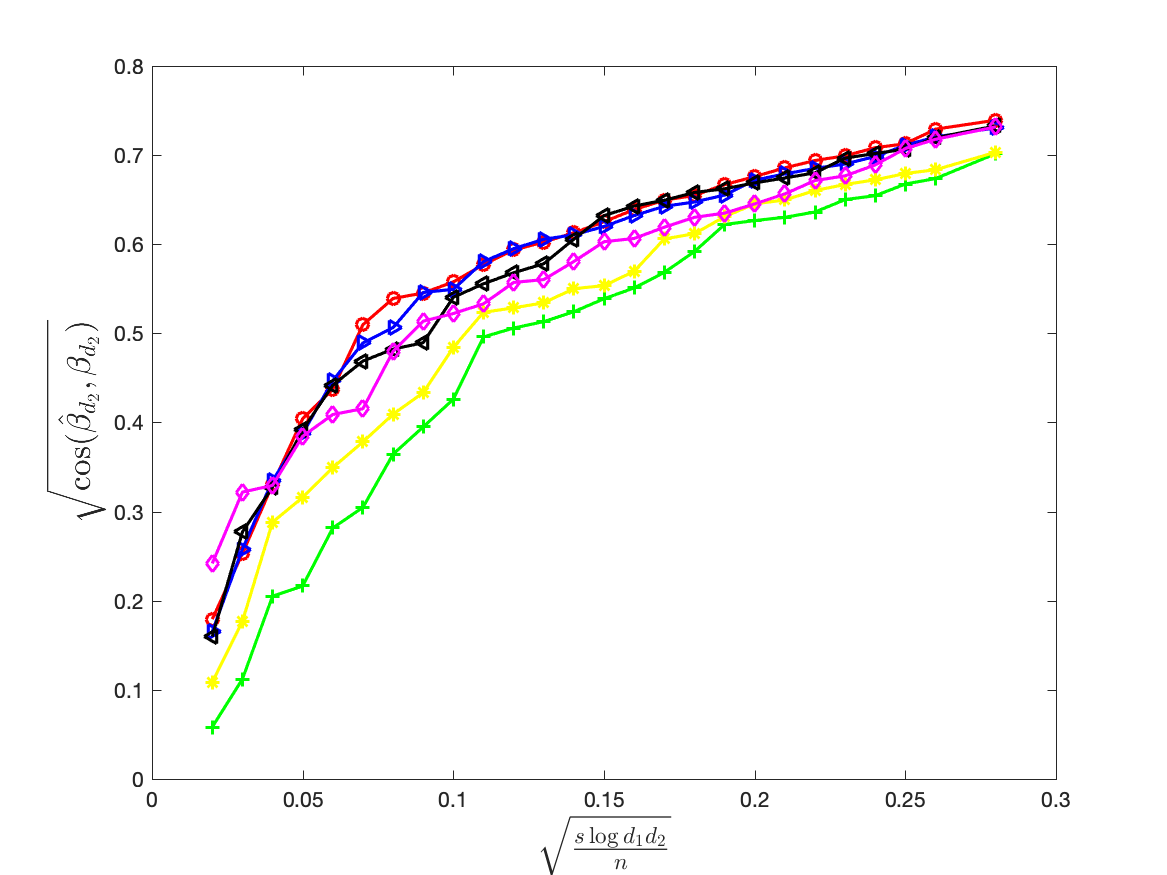}}
\subfigure[Gamma for $\tbeta_1$]{\label{gamma1}\includegraphics[width=49.9mm]{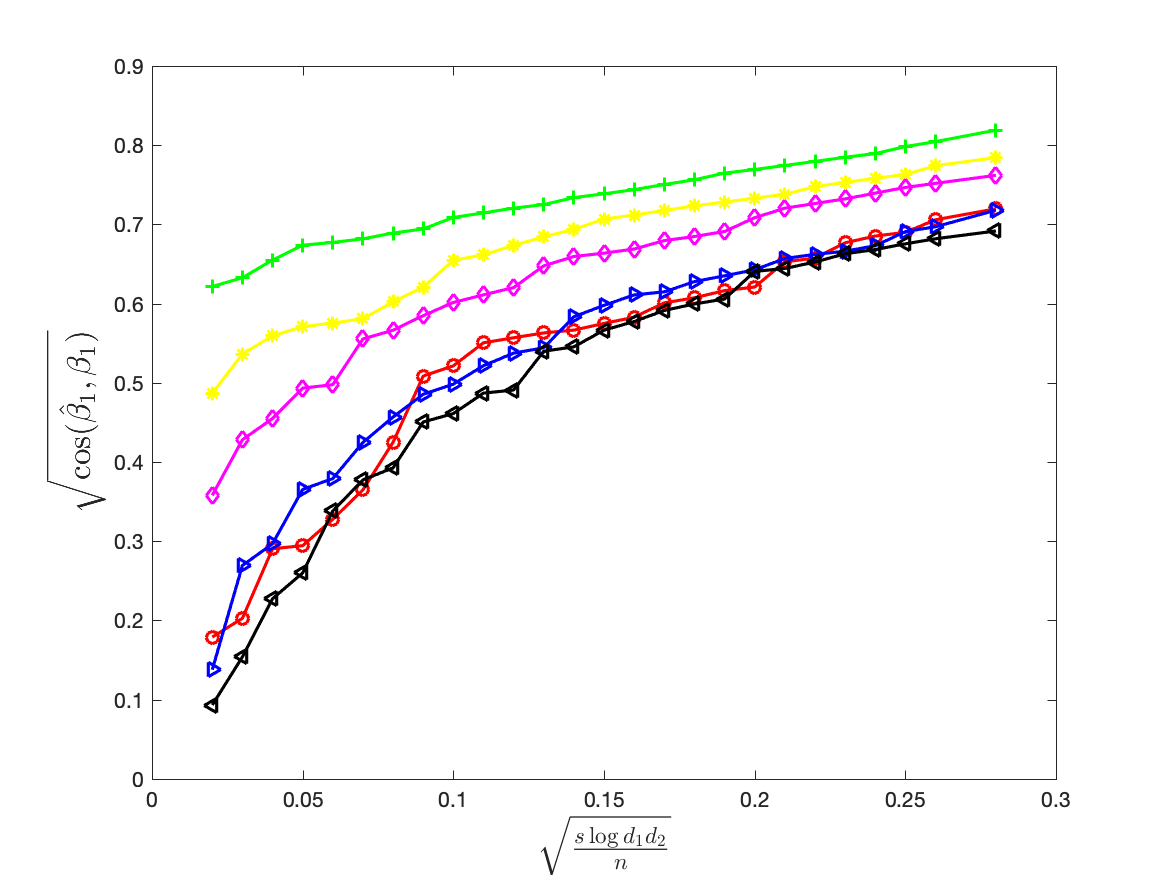}}
\subfigure[Gamma for $\tbeta_{d_2/2}$]{\label{gamma2}\includegraphics[width=49.9mm]{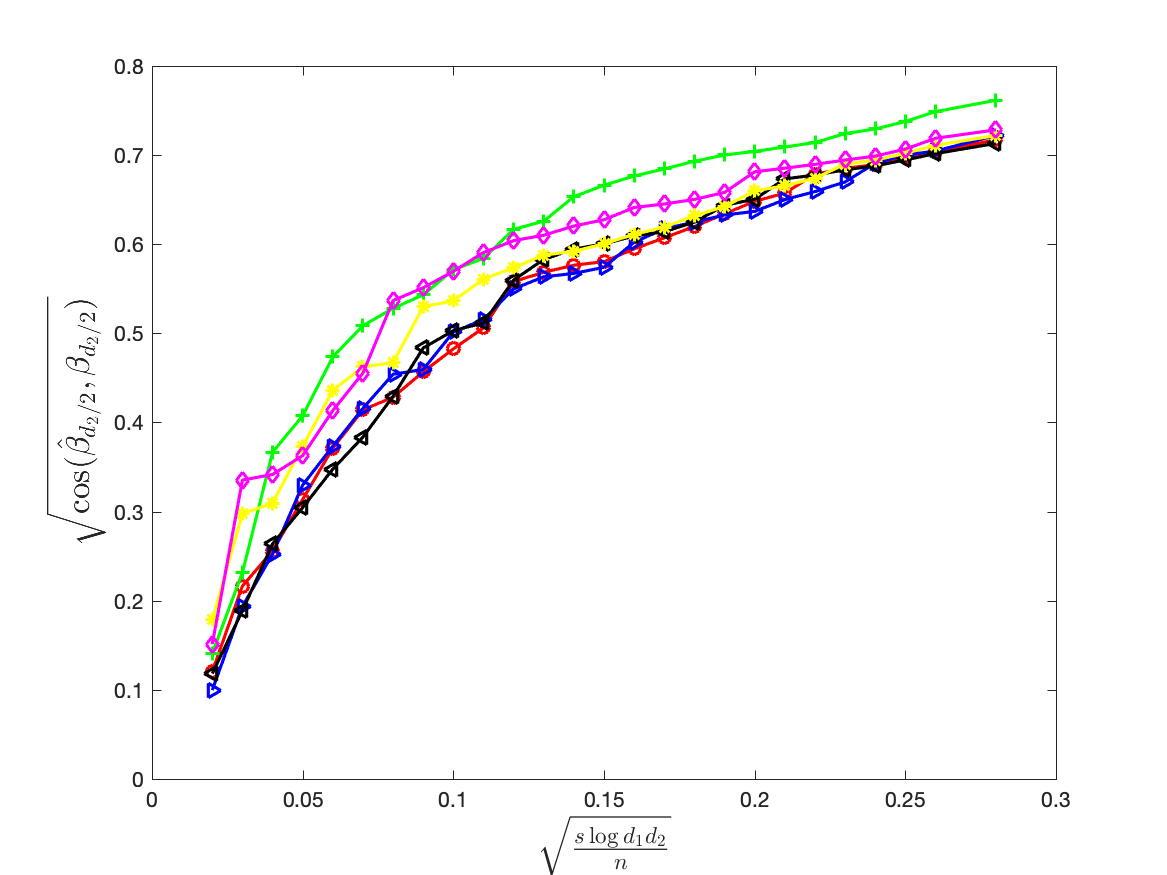}}
\subfigure[Gamma for $\tbeta_{d_2}$]{\label{gamma3}\includegraphics[width=49.9mm]{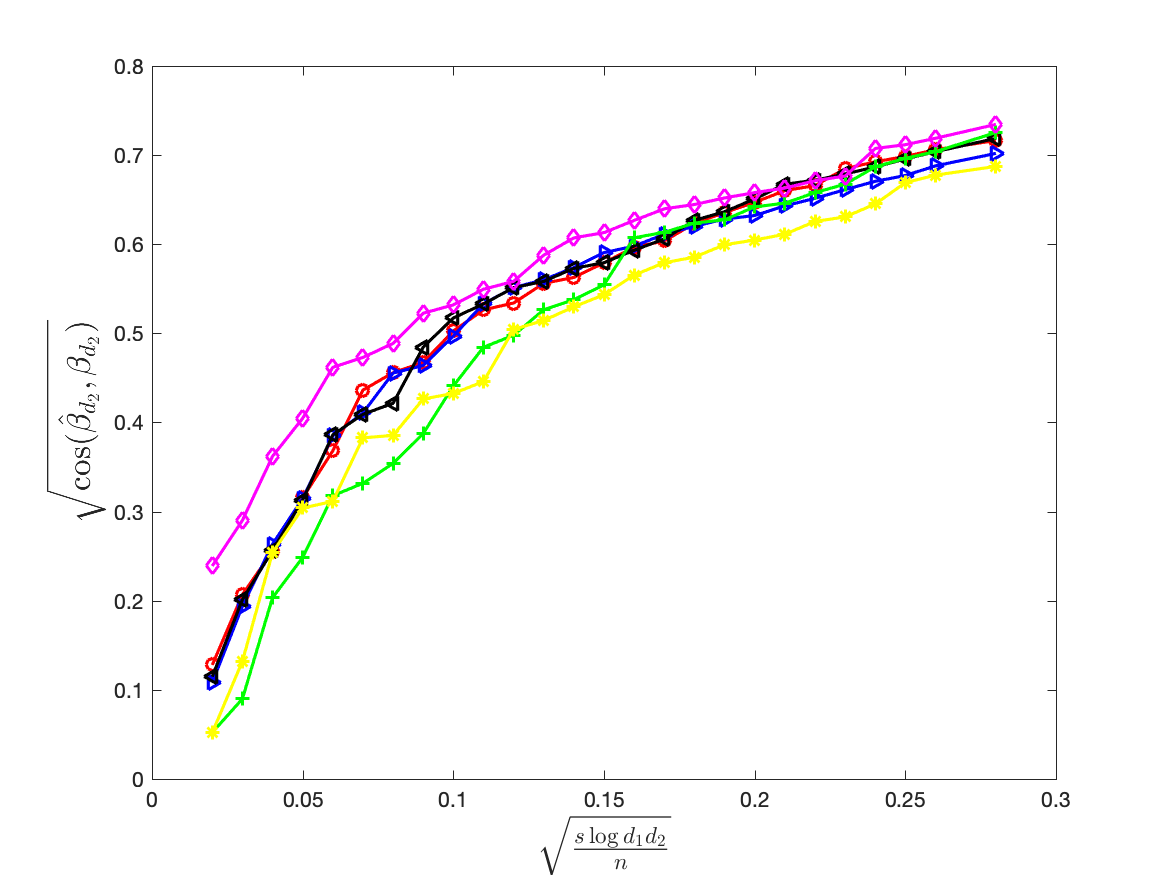}}
\includegraphics[width=7.5cm,height=0.36cm]{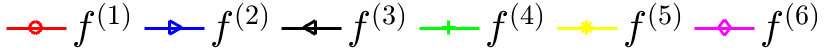}
\caption{{Sparse vector estimation plot (I). This figure shows cosine distance trend for error of estimating single sparse parameter in model (\ref{mod:1}). Six lines indicates six different types of link functions. Each row represents one type of distribution for design. }}\label{fig:2}
\end{figure}

Figure \ref{fig:2} illustrates the error trend for estimating $\tbeta_1$, $\tbeta_{d_2/2}$ and $\tbeta_{d_2}$ under different link functions and $\bx$ following one of the following distributions: Gaussian, Beta or Gamma. Additional results are shown in Figure~\ref{fig:3} in Appendix \ref{supple4} for $\bx$ following $t_{13}$, Rayleigh or Weibull distributions. From the plots, we see the error for using linear link functions ($f^{(1)}$ to $f^{(3)}$) is generally smaller than the error for using quadratic link functions ($f^{(4)}$ to $f^{(6)}$), no matter which distribution we use for design. But in all settings the scaled error plots have a linear trend when $n\gg s\log d_1d_2$, which is consistent with Theorem \ref{thm:4}. Another interesting observation is that, as $k$ increases, the estimation will be more and more precise. This can also
be seen from our theoretical results. Note that quadratic function $f(x) = x^2$ brings the singularity to the estimation problem even for the Gaussian design since $\mE[f'(\LD \bx, \tbeta\RD)] = 0$. Thus, when $k$ is small, the scalar $\mu_k$, appeared in (\ref{equ:3}), will be close to zero, which implies that the estimation problem is almost singular. Fortunately, when $k$ is large, the linear term in quadratic functions will have more contributions and $\mu_k\approx k$. From Corollary \ref{cor:1} and identity (\ref{equ:3}), we see that $\mu_k$ amplifies the magnitude of the signal $\tbeta_k$ that characterizes the difficulty of estimation problem. Therefore a larger $\mu_k$ implies an easier problem. Figure \ref{fig:4}---\ref{fig:6} in Appendix \ref{supple4} provide additional simulation results where we plot the error trend for all $d_2$
parameters, which make this observation more clear.

\subsection{Low-rank Matrix}

We study the estimation of $\tB$ under the low-rank assumption. We let $d_1=d_2=25$, $r = 5$. The distribution of $\bx$ is set as described in Table \ref{tab:2}. For generating $\tB$, we first generate two random orthogonal matrices $\bU\in\mR^{d_1\times d_1}$ and $\bV\in\mR^{d_2\times d_2}$, then generate $d_1\times d_2$ diagonal matrix $\bLambda$ by first sampling $r$ locations from $[\min\{d_1, d_2\}]$ and further setting each entry to be $\{-\frac{1}{\sqrt{r}}, \frac{1}{\sqrt{r}}\}$ with equal probability. Finally we let $\tB = \bU\bLambda\bV^T$. We consider both independent and dependent $\bz$. For the independent case, $\bz$ is generated as in the previous section. For the dependent case, we generate $\bz$ as follows. Using the Gaussian copula with the correlation matrix $\bSigma = 0.2\cdot\pmb{1}_{d_2}\pmb{1}_{d_2}^T + 0.8\pmb{I}_{d_2}$, we generate $d_2$-dimensional random vectors with support in $[0, 1]^{d_2}$. Then we apply inverse transformation on each coordinate to make it marginally distributed as $t_7$. Note that under our setup $\bz$ has dependent coordinates with each coordinate being $t_7$ distribution. However, the true covariance matrix of $\bz$ is not $\bSigma$ anymore and is even unknown. According to Theorem \ref{thm:5}, we set $\kappa_1 = 2\sqrt{\log(d_1+d_2)/n(d_1+d_2)}$ and $\lambda = 12\sqrt{(d_1+d_2)\log(d_1+d_2)/n}$. The precision matrix estimator we use is defined in (\ref{equ:17}) with $\kappa_2 = 2\sqrt{\log d_2/nd_2}$, suggested by Lemma~\ref{lem:8a}.

\begin{figure}[t]
\centering     
\subfigure[Gaussian]{\label{LMgau}\includegraphics[width=49.3mm]{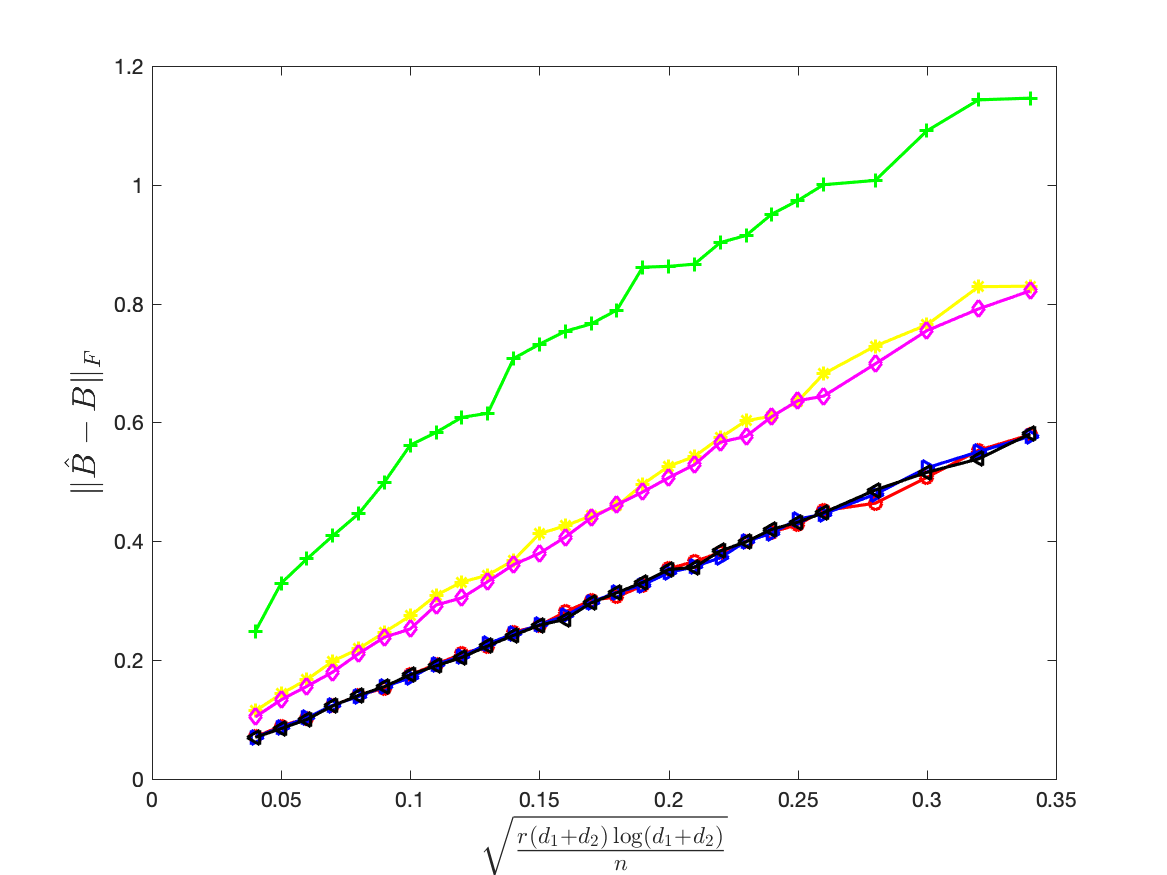}}
\subfigure[Beta]{\label{LMbeta}\includegraphics[width=49.3mm]{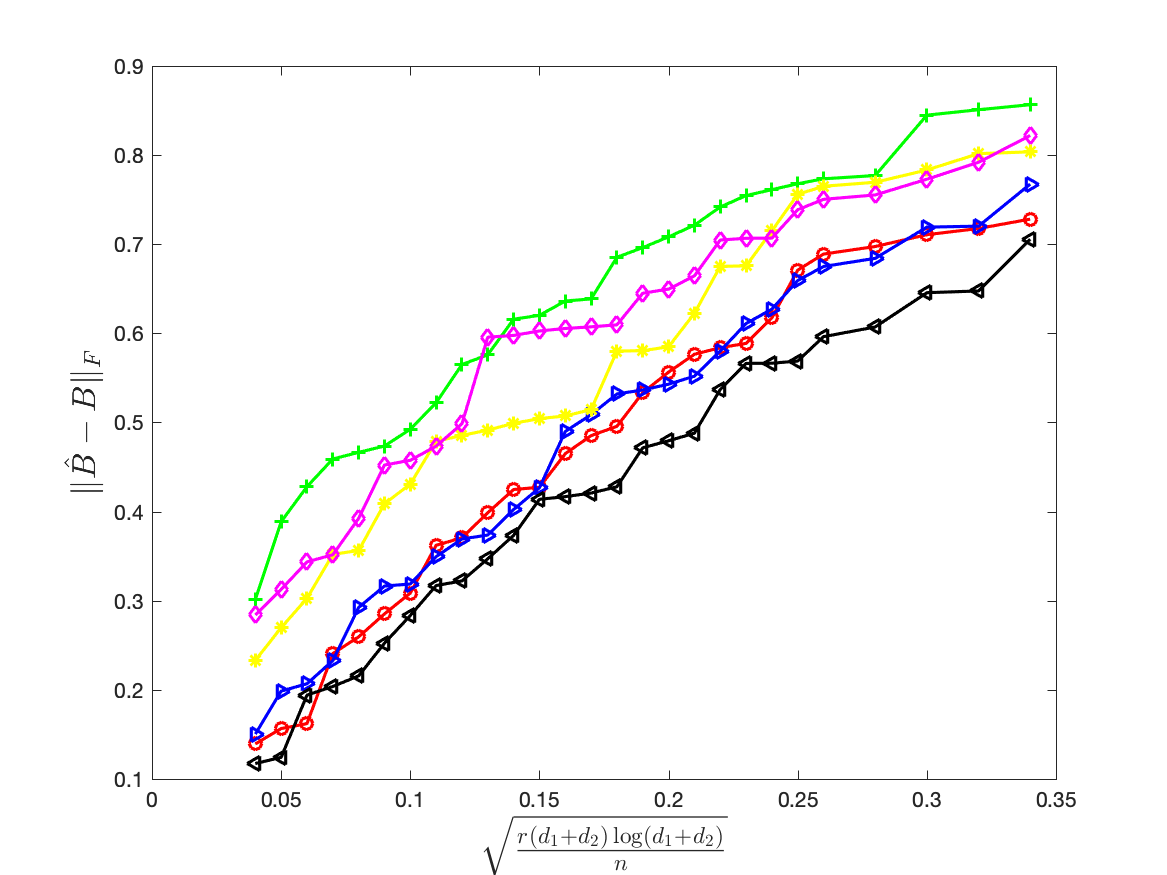}}
\subfigure[Gamma]{\label{LMgam}\includegraphics[width=49.3mm]{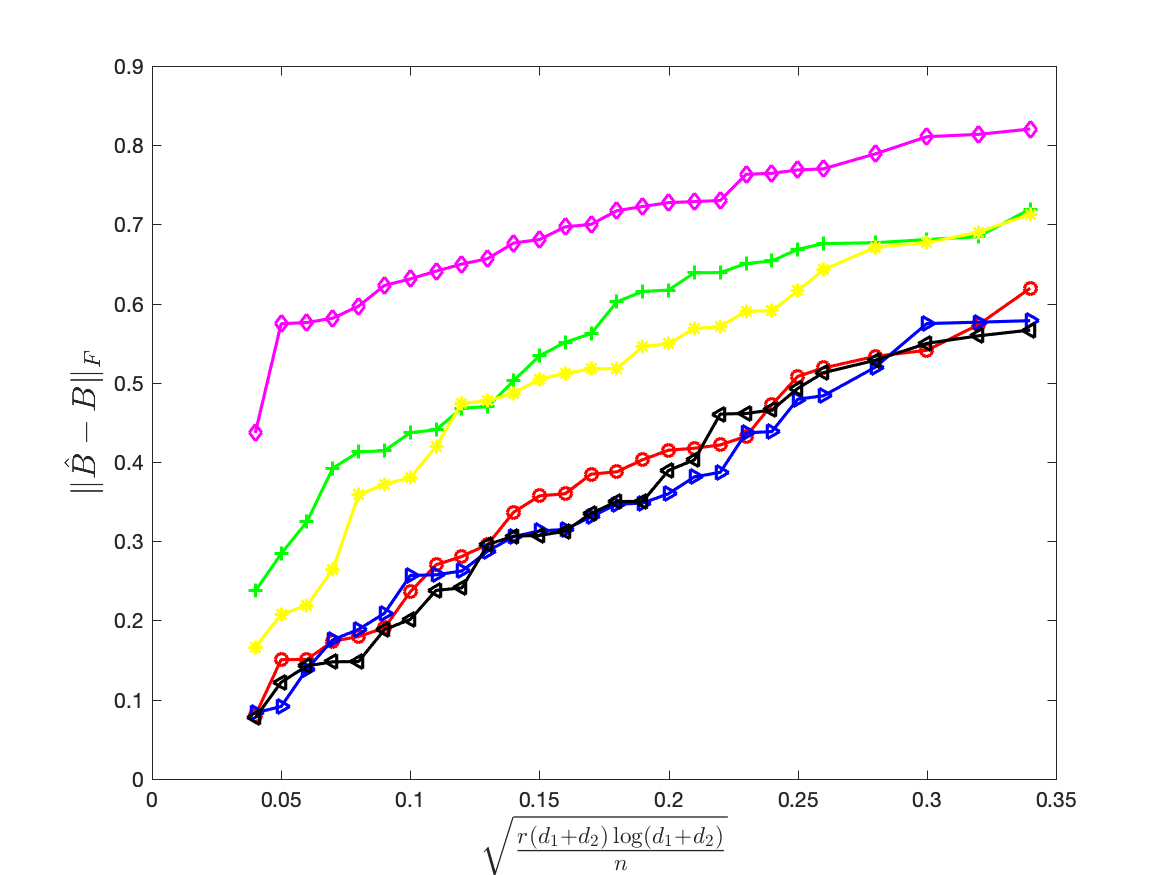}}
\subfigure[$t_{13}$]{\label{LMt13}\includegraphics[width=49.3mm]{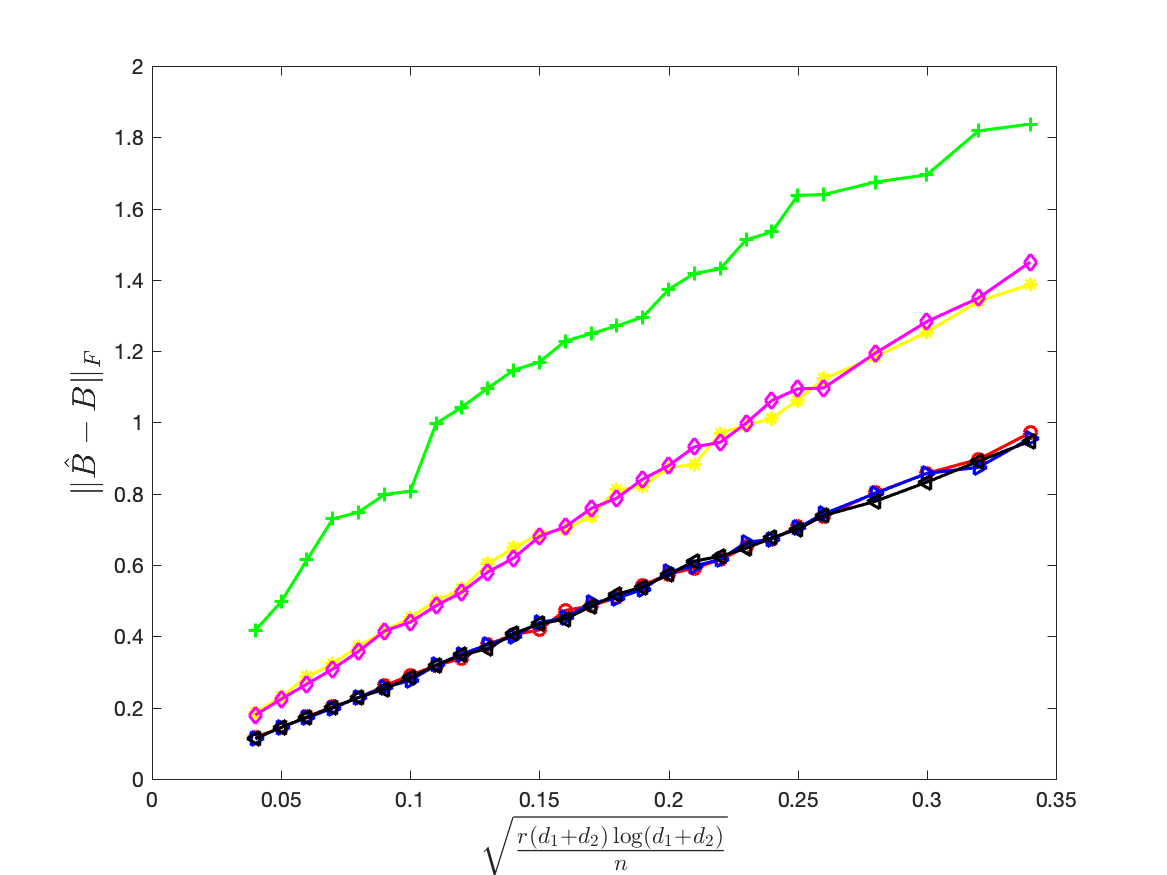}}
\subfigure[Rayleigh]{\label{LMray}\includegraphics[width=49.3mm]{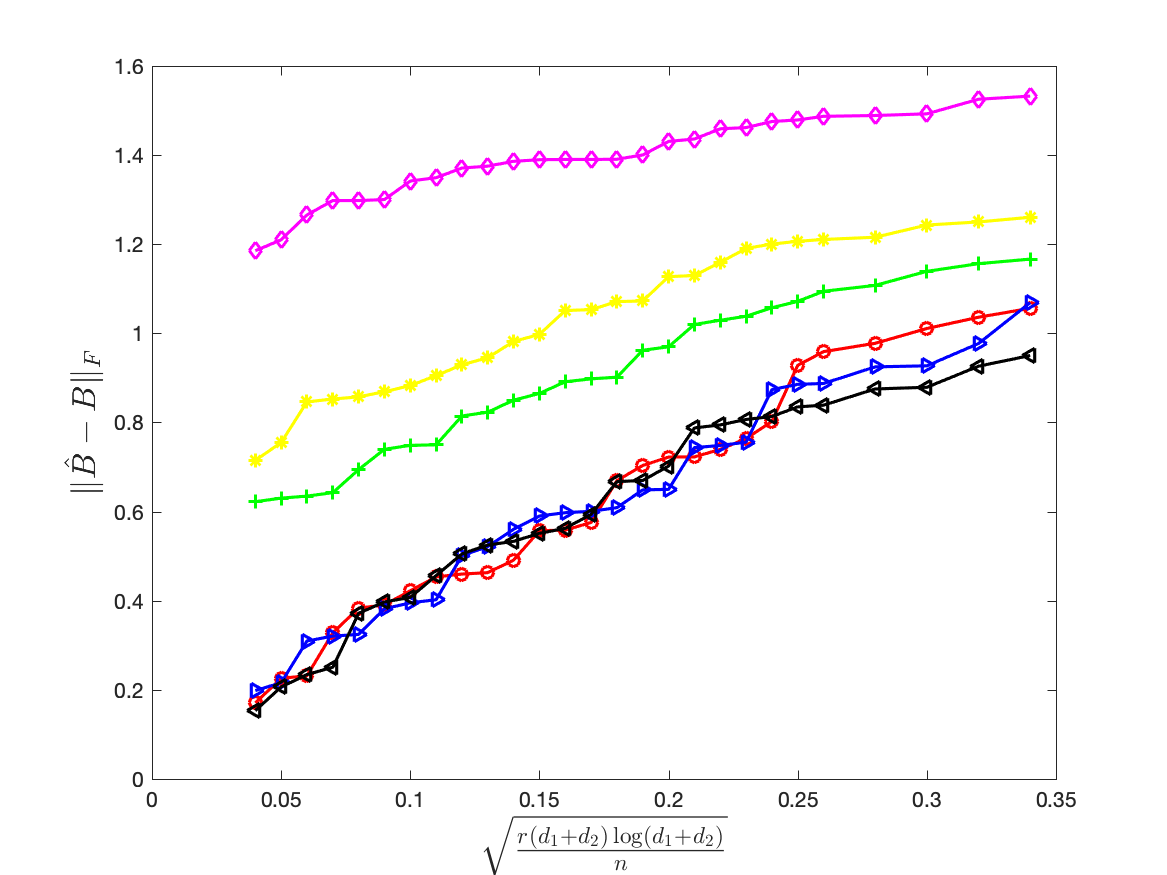}}
\subfigure[Weibull]{\label{LMwei}\includegraphics[width=49.3mm]{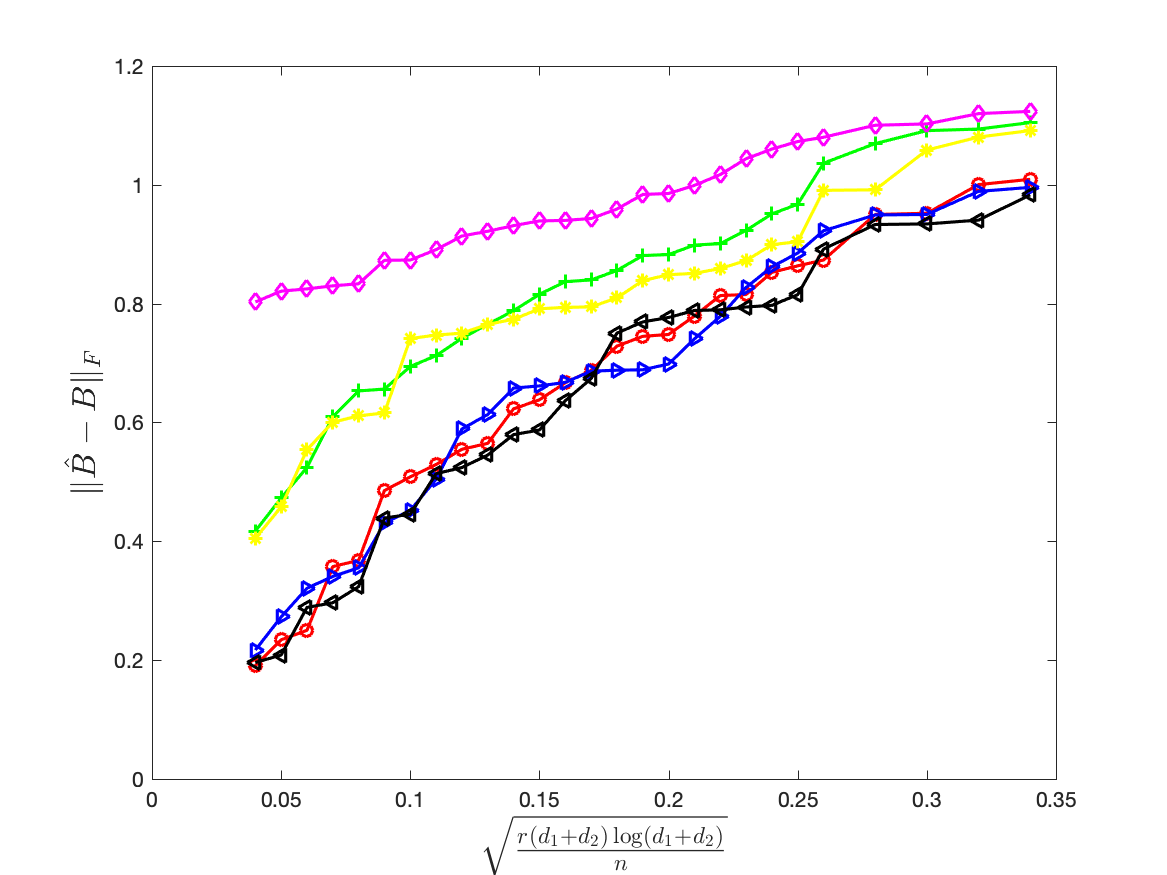}}
\includegraphics[width=7.5cm,height=0.36cm]{Figure/funcleg.png}
\caption{{Low-rank matrix estimation plot. This figure shows $\|\hB - \ttB\|_F$ error of estimating low-rank parameter matrix in model (\ref{mod:1}). Six lines indicates six different types of link functions. Components of $\bz$ are independent.}}\label{fig:7a}
\end{figure}

\begin{figure}[t]
\centering     
\subfigure[Gaussian]{\label{LMgau1}\includegraphics[width=49.9mm]{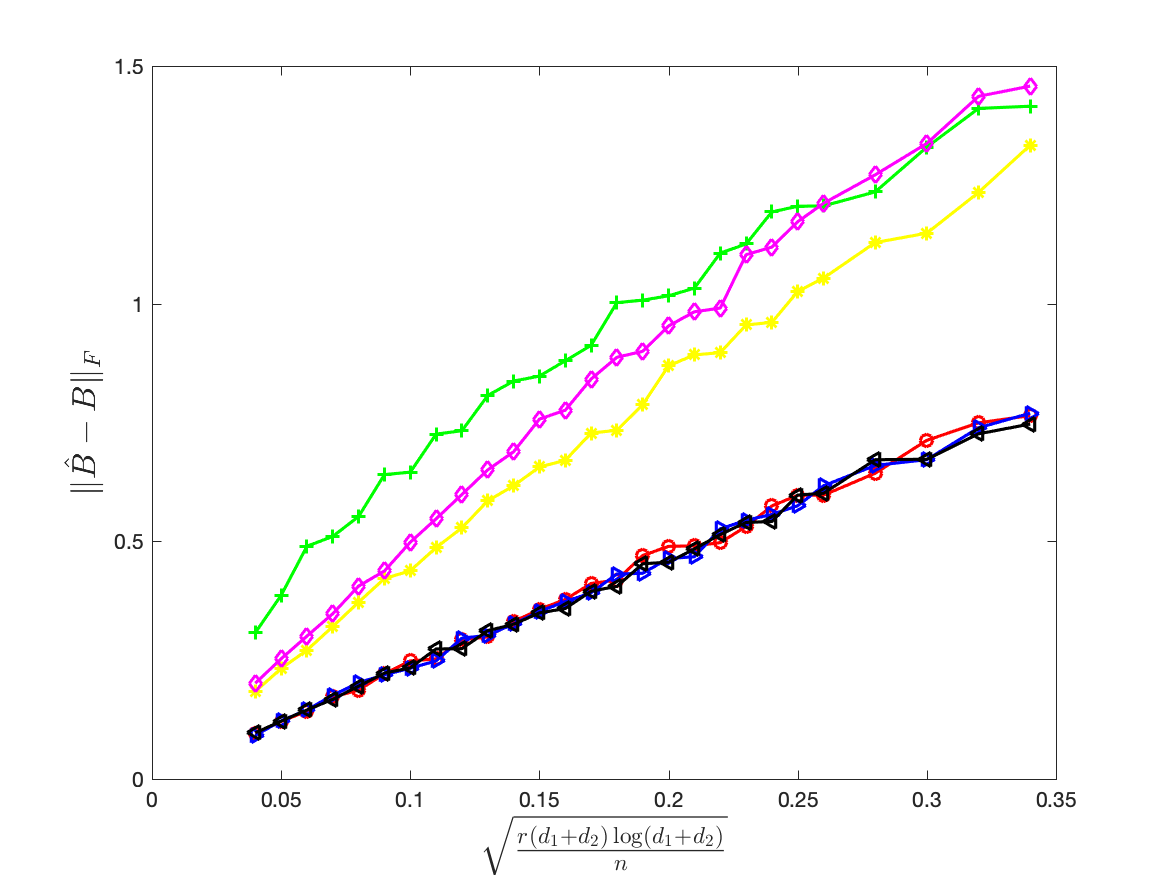}}
\subfigure[Beta]{\label{LMbeta1}\includegraphics[width=49.9mm]{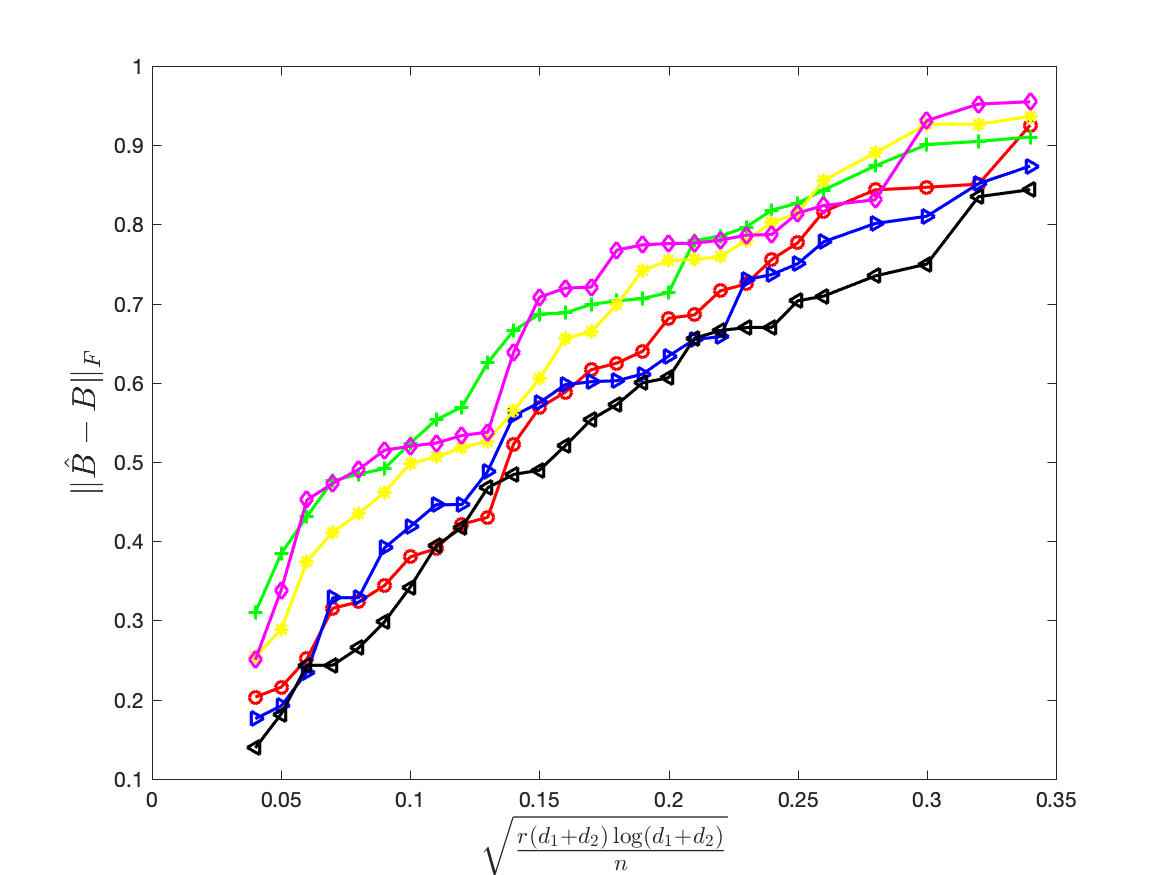}}
\subfigure[Gamma]{\label{LMgam1}\includegraphics[width=49.9mm]{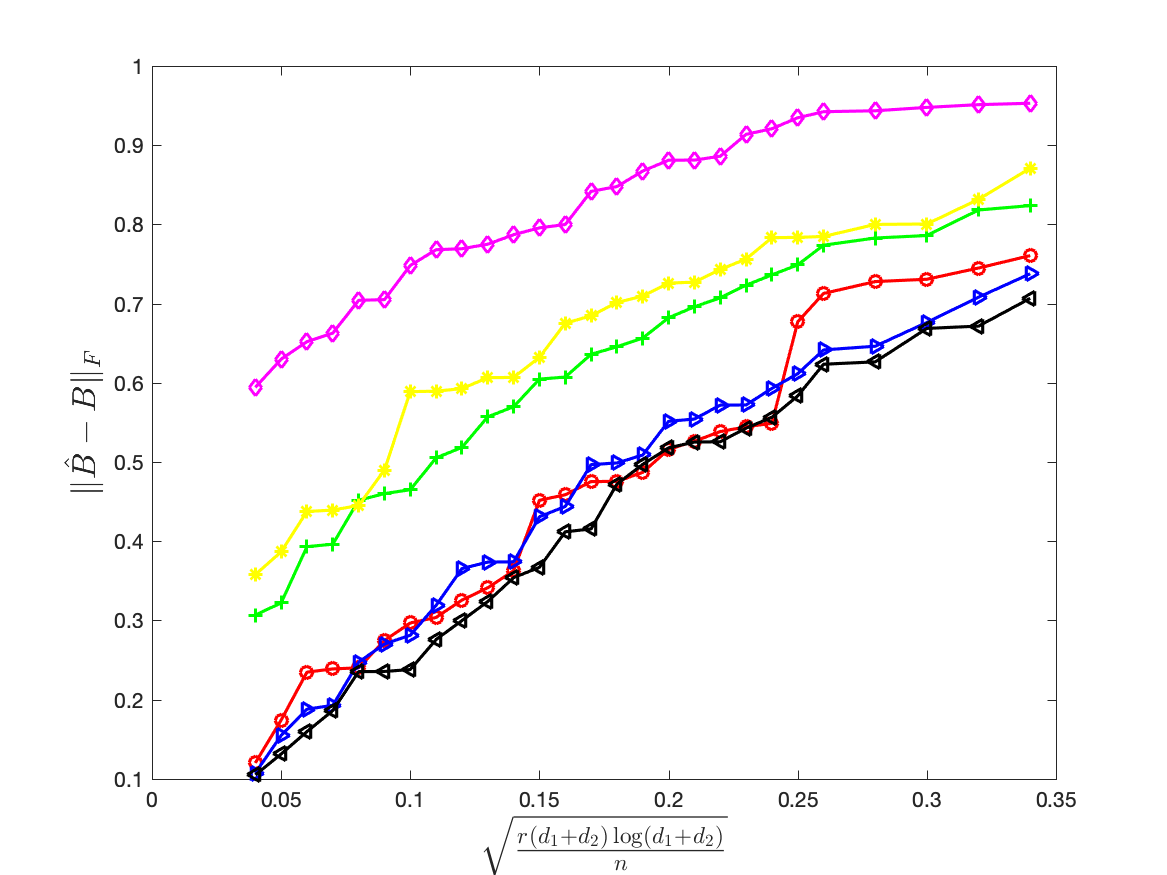}}
\subfigure[$t_{13}$]{\label{LMt131}\includegraphics[width=49.9mm]{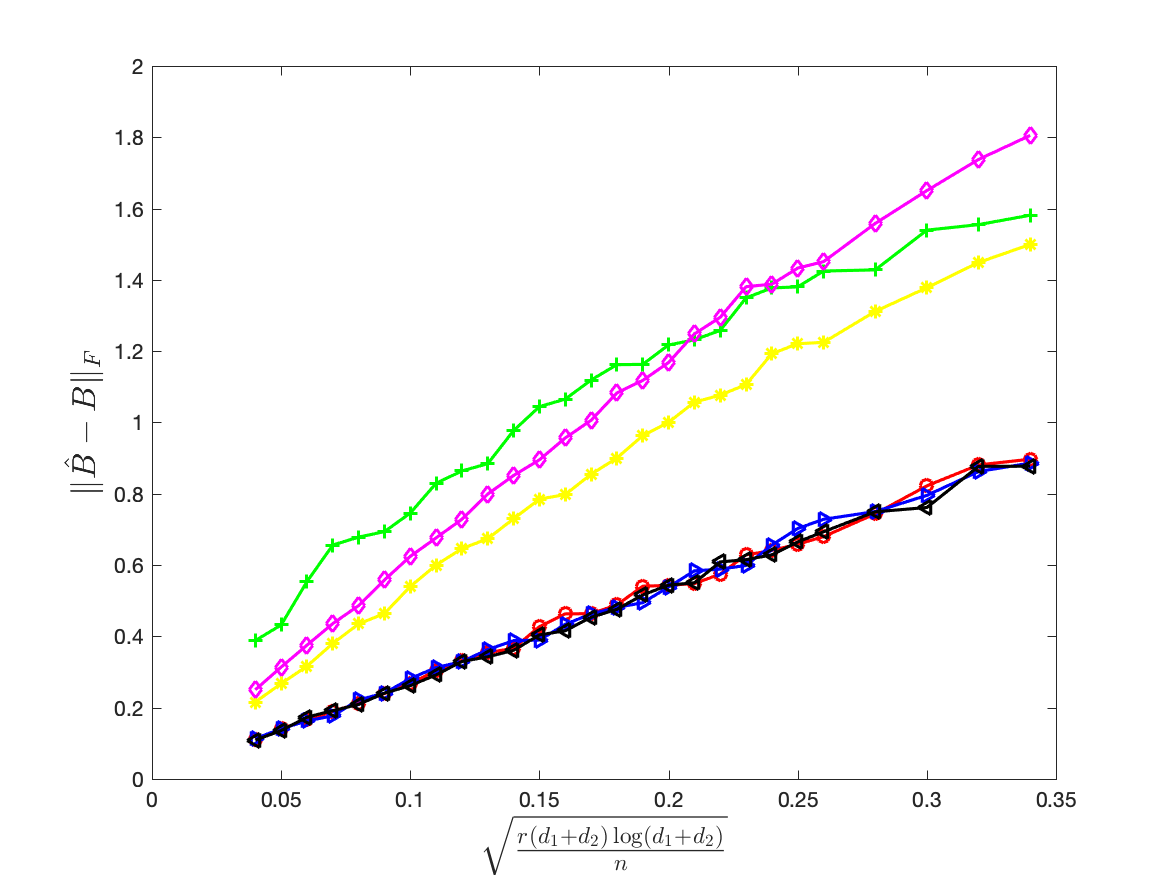}}
\subfigure[Rayleigh]{\label{LMray1}\includegraphics[width=49.9mm]{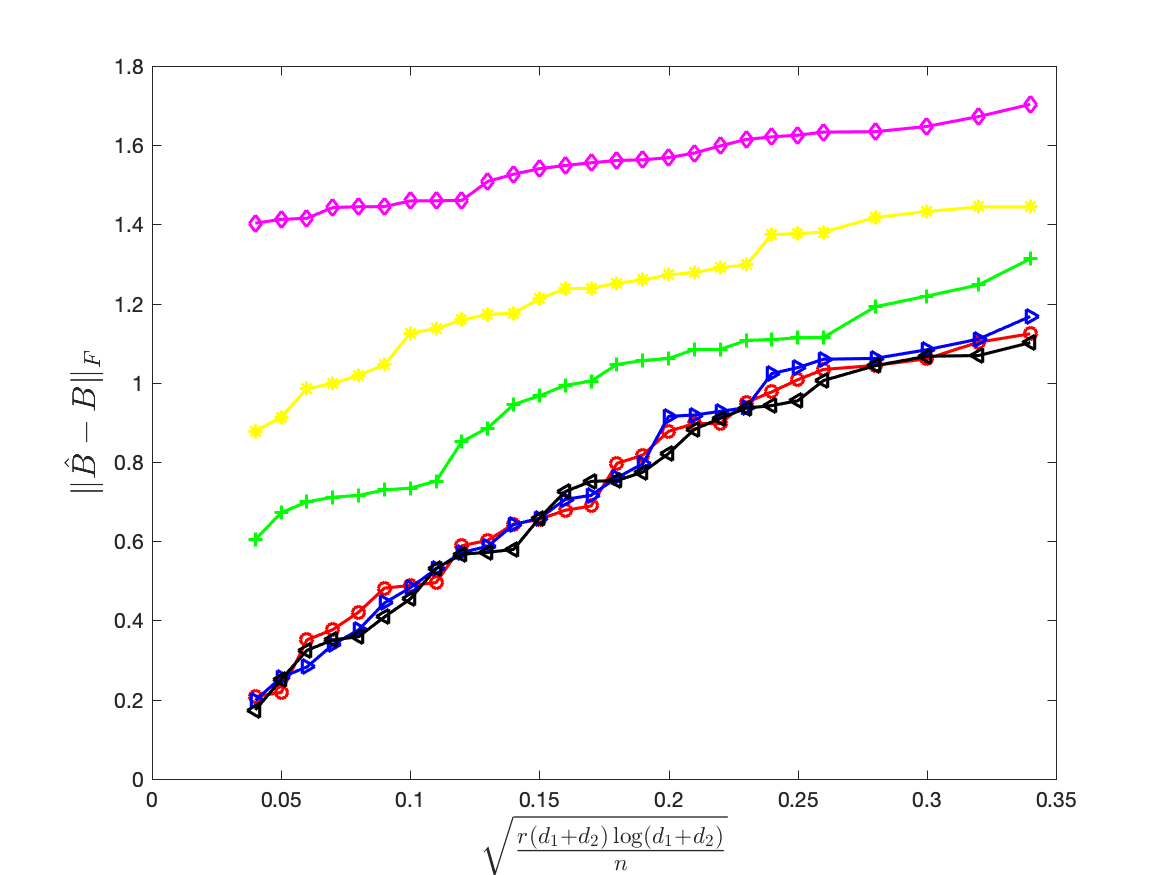}}
\subfigure[Weibull]{\label{LMwei1}\includegraphics[width=49.9mm]{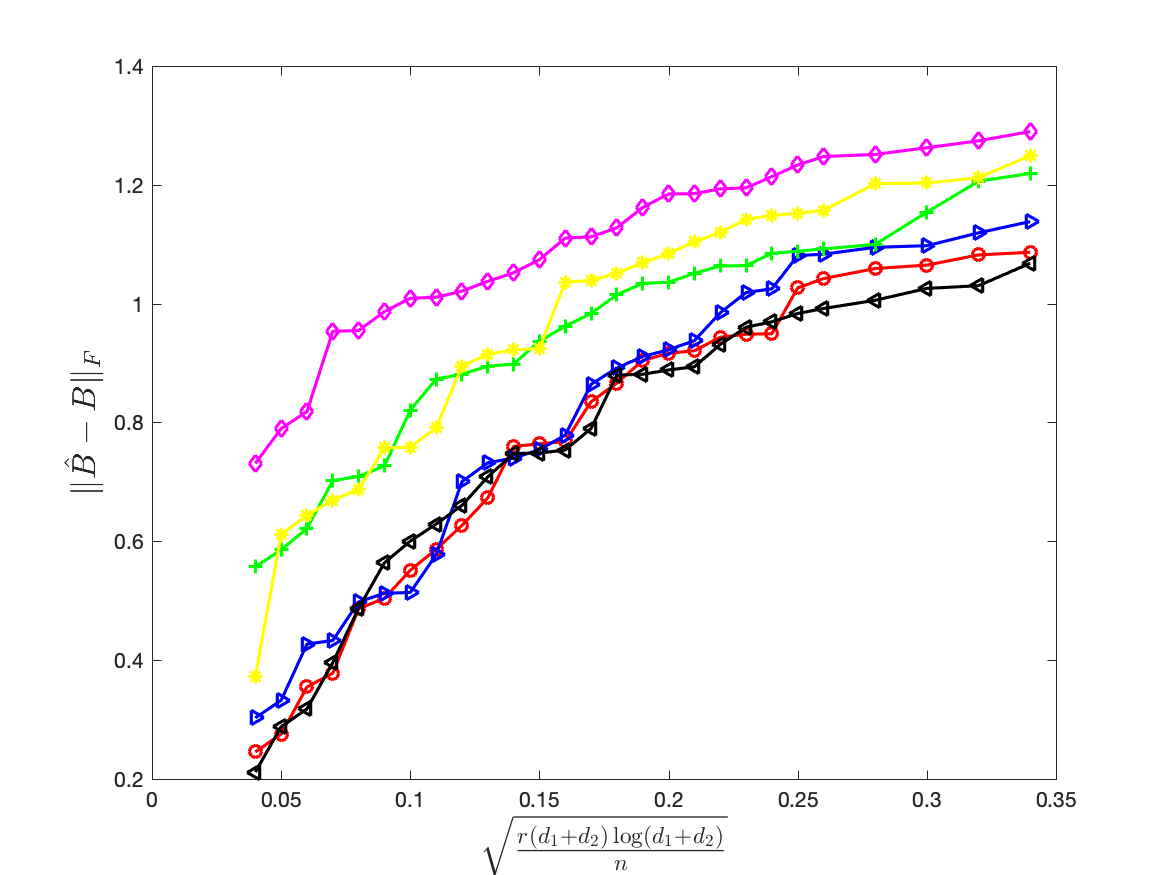}}
\includegraphics[width=7.5cm,height=0.36cm]{Figure/funcleg.png}
\caption{{Low-rank matrix estimation plot. This figure shows $\|\hB - \ttB\|_F$ error of estimating low-rank parameter matrix in model (\ref{mod:1}). Six lines indicates six different types of link functions. Components of $\bz$ are dependent.}}\label{fig:7b}
\end{figure}

Figure \ref{fig:7a} and \ref{fig:7b} summarize our results. We observe a linear trend for all link functions, though the quadratic link functions result in a larger error overall. Also, for each distribution, dependent or independent $\bz$ have very similar error trends among all six types of link functions. This observation verifies our theoretical results in Theorem \ref{thm:5} and Lemma \ref{lem:8a} that estimating precision matrix only contributes higher order error.

\subsection{Sparse Matrix}

\begin{figure}[!t]
\centering     
\subfigure[Gaussian]{\label{SMgau1}\includegraphics[width=49.9mm]{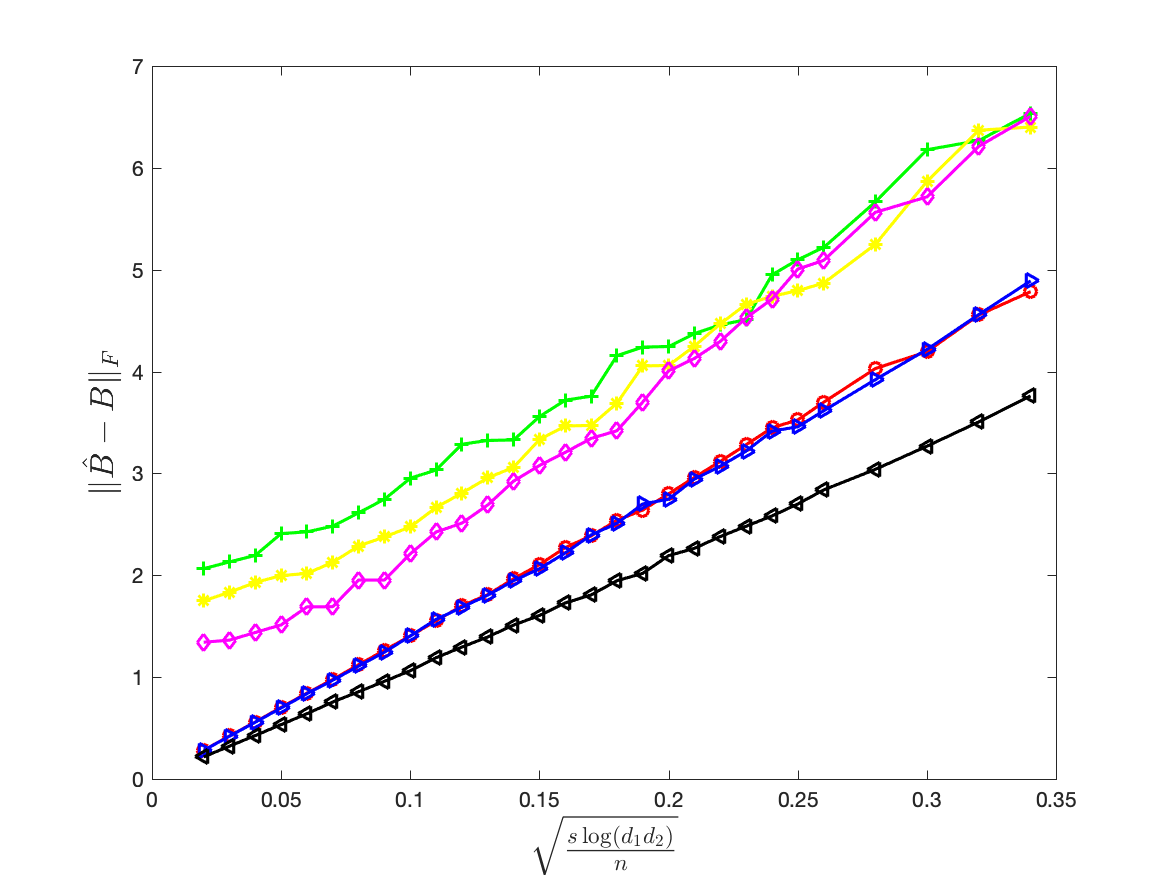}}
\subfigure[Beta]{\label{SMbeta1}\includegraphics[width=49.9mm]{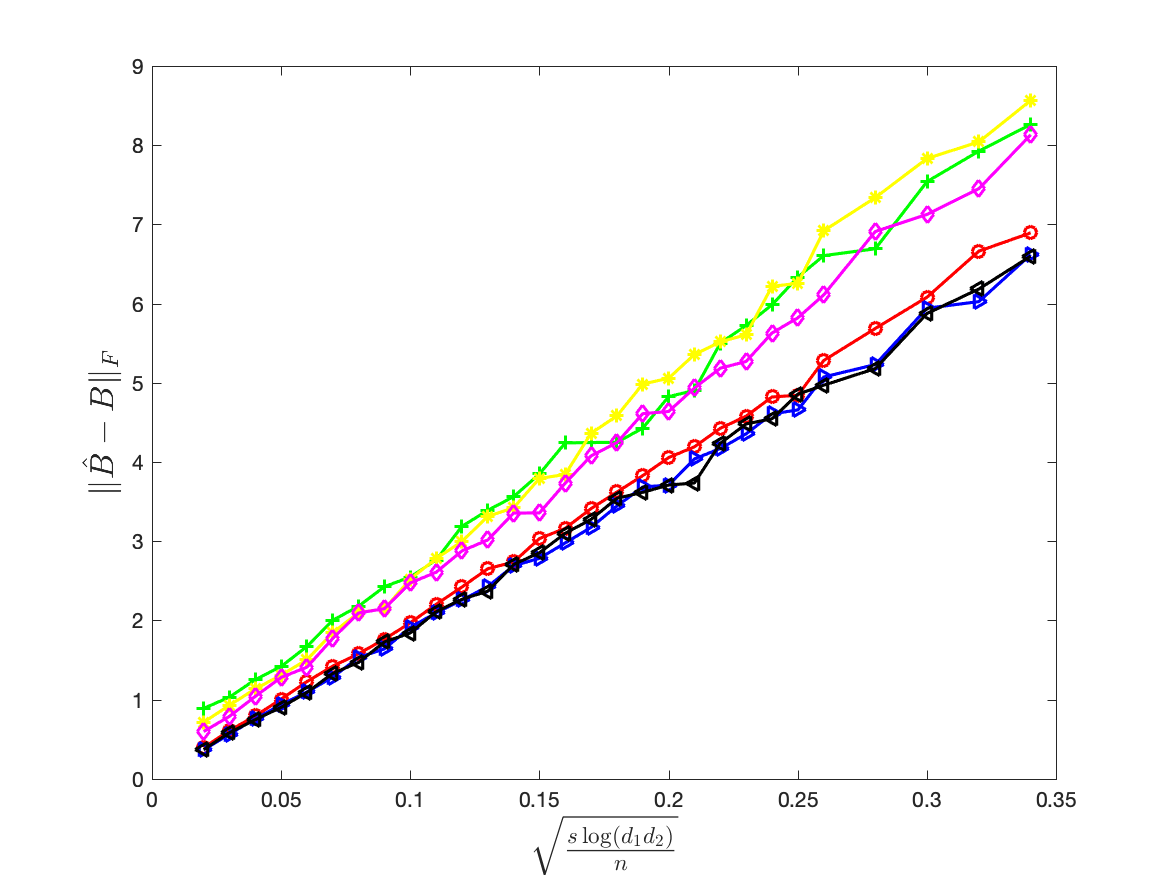}}
\subfigure[Gamma]{\label{SMgam1}\includegraphics[width=49.9mm]{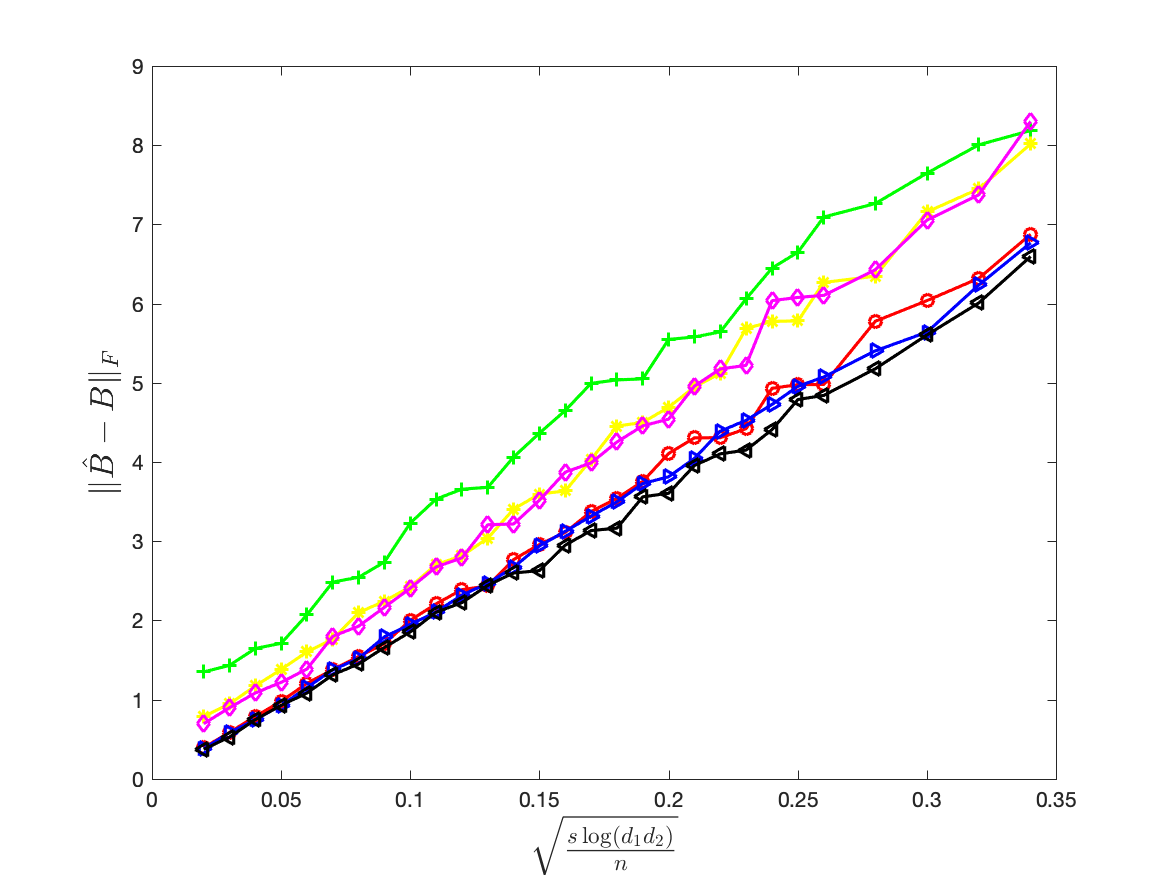}}
\subfigure[$t_{13}$]{\label{SMt131}\includegraphics[width=49.9mm]{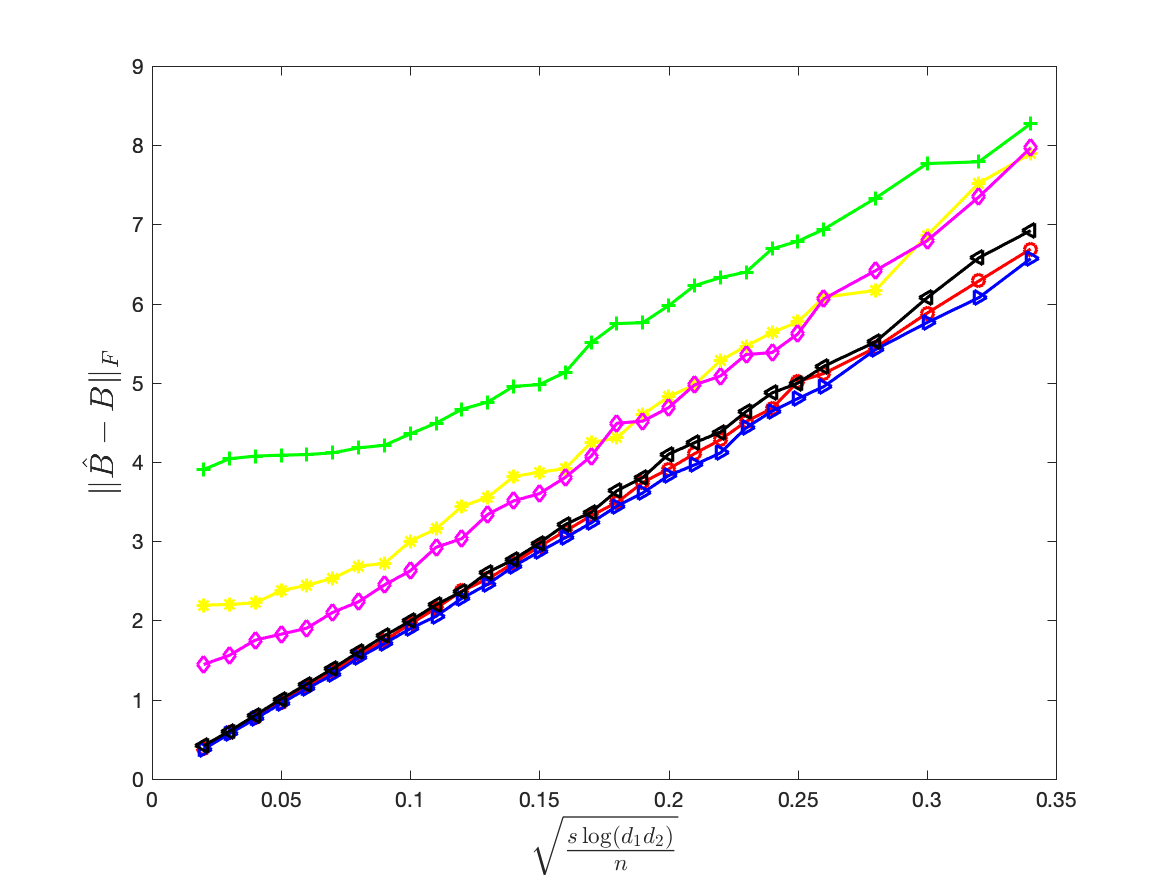}}
\subfigure[Rayleigh]{\label{SMray1}\includegraphics[width=49.9mm]{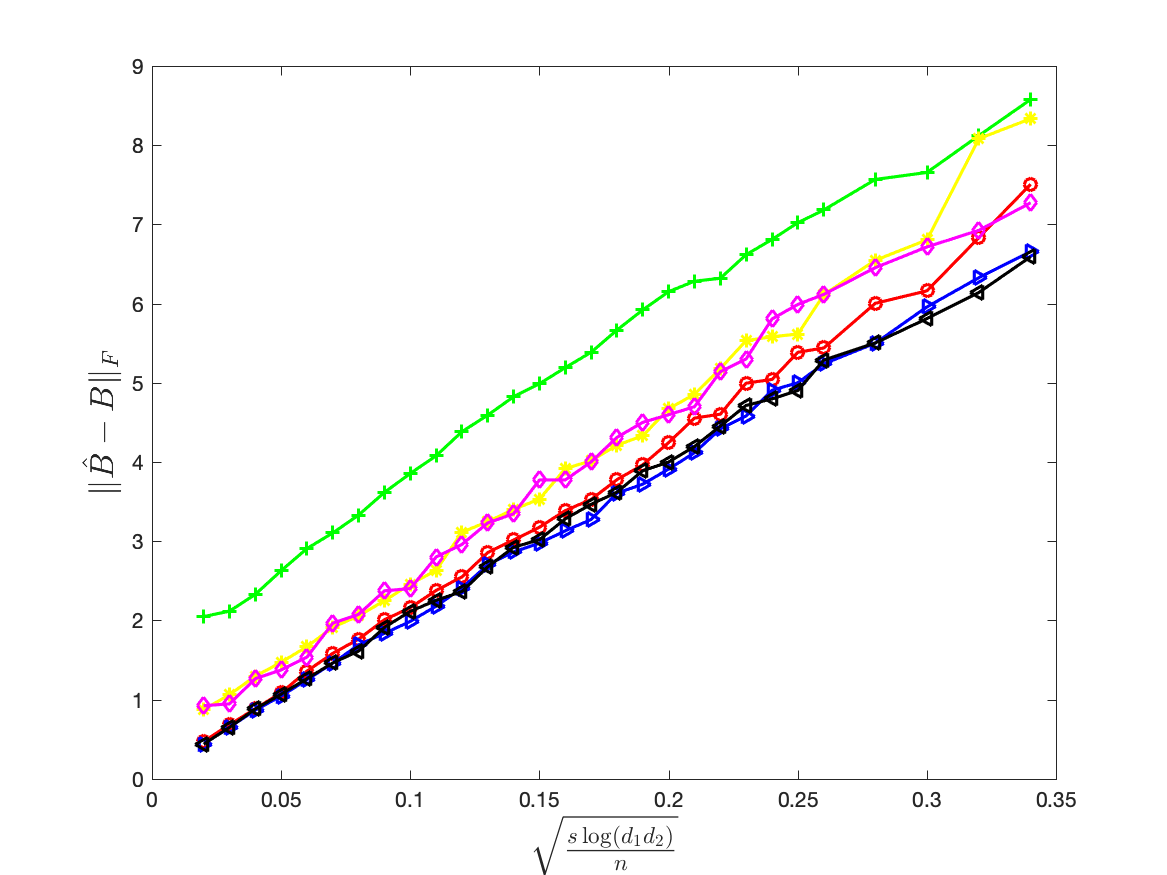}}
\subfigure[Weibull]{\label{SMwei1}\includegraphics[width=49.9mm]{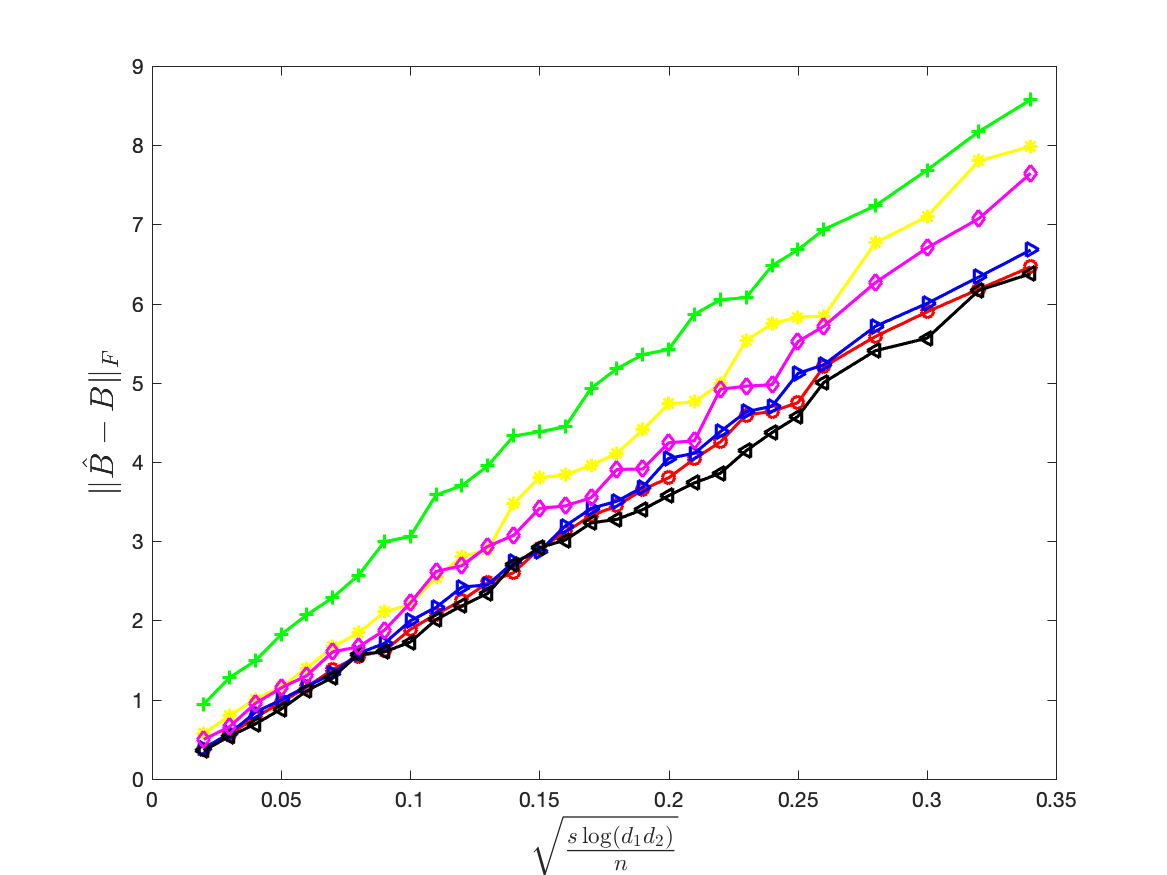}}
\includegraphics[width=8cm,height=0.36cm]{Figure/funcleg.png}
\caption{{Sparse matrix estimation plot. This figure shows $\|\hB - \ttB\|_F$ error of estimating sparse parameter matrix in model (\ref{mod:1}) with dependent $\bz$. Six lines indicates six different types of link functions. All above simulation results are consistent with Theorem \ref{thm:8}.}}\label{fig:8}
\end{figure}

For the sparse matrix recovery, we only consider estimating fully sparse $\tB$ with dependent covariate $\bz$. We let $d_1 = 100$, $d_2 = 50$, $s = 10$. The considered distributions for design $\bx$ are in Table \ref{tab:2}. $\tB$ is generated by first generating the support of nonzero entries and then setting each entry on the support to be $\{-\frac{1}{\sqrt{s}}, \frac{1}{\sqrt{s}}\}$ with equal probability. For generating dependent $\bz$, we define the sparse precision matrix $\Theta = (\theta_{ij})_{i,j=1}^{d_2}$ as follows
\begin{align*}
\theta_{ij} = \begin{cases}
1 & \text{if\ } i=j,\\
0.2 & \text{if\ } |i-j|=1,\\
0 & \text{otherwise}.
\end{cases}
\end{align*}
Then we calculate the corresponding covariance matrix $\Theta^{-1}$
and convert it to correlation matrix. Following the same steps as in
the previous section, we use Gaussian copula to derive dependent
covariates $\bz$, whose marginal distribution for each coordinate is
$t_7$. The precision matrix of $\bz$ has the same sparsity structure
as $\Theta$, even though it is not equal to $\Theta$. According to
Theorem \ref{thm:8}, we set $\tau = 2(n/\log d_1d_2)^{1/6}$ and
$\lambda = 10\sqrt{\log d_1d_2/n}$. The sparse precision matrix
estimator is defined in (\ref{equ:11}) and (\ref{equ:16}) with truncation threshold $2(n/\log d_2)^{1/4}$ and $\gamma = 10\sqrt{\log d_2/n}$. We use default settings in CVX package \citep{Grant2008Graph, Grant2012cvx} to solve (\ref{equ:11}) efficiently. The error plot is shown in Figure \ref{fig:8} and linear trends for errors appear again, as explained by Theorem \ref{thm:8}.

\subsection{Real Data Application}\label{realdata}

In this subsection, we will further illustrate the proposed method by
analyzing a \textit{Coffea canephora} genetic data set.\footnote{The data set was obtained from Dryad at \href{https://datadryad.org/resource/doi:10.5061/dryad.1139fm7}{https://datadryad.org/resource/doi:10.5061/dryad.1139fm7.}} The data set collects three important traits (phenotypes): production of coffee beans $(y_1)$, leaf rust incidence $(y_2)$ and yield of green
beans $(y_3)$, from two recurrent selection populations $(x_1)$ of
\textit{Coffea canephora} and each one of traits is evaluated at two
locations $(x_2)$. For each individual sample, the single nucleotide
polymorphisms (SNPs)---genotype $\{z_j\}$---are identified by Genotyping-by-Sequencing. In particular, the population group $(x_1)$
and the evaluation location $(x_2)$ are two confounders which may
modify the effect of each SNP $(z_j)$ on traits $(y_1, y_2, y_3)$. To study this higher-level effect from confounders, we consider the
varying index coefficient model in (\ref{mod:1}), which in this case
is written as
\begin{align}\label{mod:2}
y_i = \sum_{j=1}^{d_2}z_j\cdot f_j\big((\beta_{j}^{1, i})^\star x_1 + (\beta_{j}^{2, i})^\star x_2\big) + \epsilon, \text{\ \ \ } \forall i = 1, 2, 3,
\end{align}
and focus on estimating parameter matrices $\tB_i = \big((\bbeta^{1, i})^\star, (\bbeta^{2, i})^\star\big)^T\in\mR^{2\times d_2}$.

\begin{figure}[t]
\centering     
\subfigure[Production]{\label{hist:prod}\includegraphics[width=50mm]{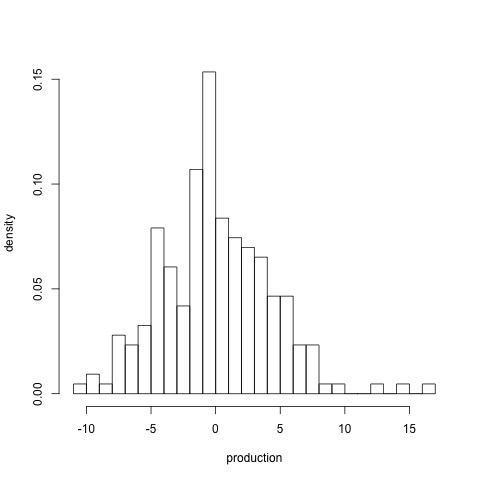}}
\subfigure[Leaf rust]{\label{hist:rust}\includegraphics[width=50mm]{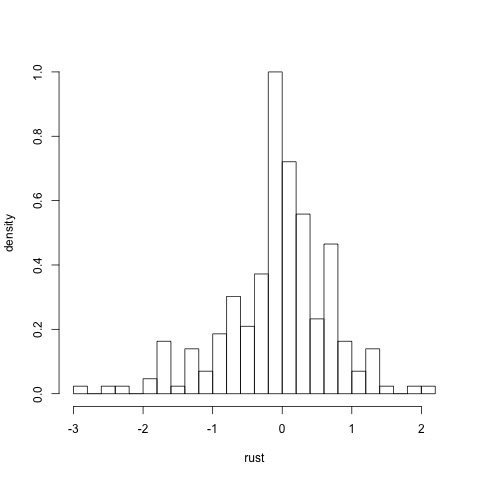}}
\subfigure[Green beans]{\label{hist:green}\includegraphics[width=50mm]{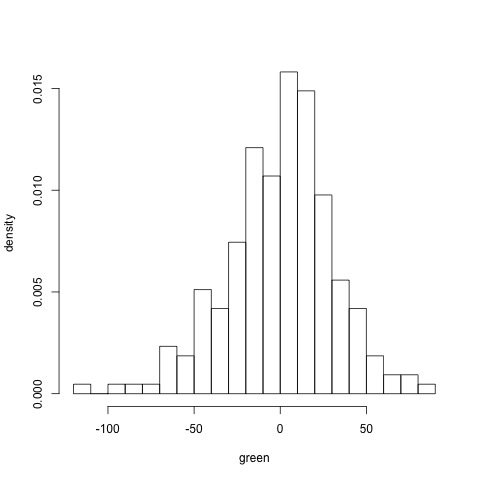}}
\caption{{Histograms for three traits: production of coffee beans, leaf rust incidence and yield of green beans.}}\label{fig:9}
\end{figure}

We follow the preprocessing steps described in \cite{Ferrao2018Accurate}. The data set has $n = 215$ samples evaluated at two locations in total, where $n_1 = 119$ of them are collected from the first population group with $45748$ SNPs measured, while $n_2 = 96$ of them are collected from the second population group with $59332$ SNPs measured. There are $38106$ SNPs in common and we select $d_2 = 250$ from them uniformly at random. To make individuals independent from each other, in each group we only use the data evaluated at the first location for the first half of individuals and the data evaluated at the second location for the second half of individuals. Continuous distributions are used for the confounders $x_1$ and $x_2$. In particular, we let $x_1\sim N(0, 1)$, if the observation is from the first population group and $x_1\sim N(50, 1)$, if it is from the second population group. Analogously, we have $x_2\sim t_{13}$, if the observation is evaluated at the first location and $x_2\sim 50 + t_{13}$, if it is evaluated at the second location. Here $t_{13}$ refers to the $t$ distribution with degree of freedom $13$. Under our setup, the confounders come from the following mixture distribution:
\begin{align}\label{conf:dist}
x_1\sim \frac{n_1}{n} N(0, 1) + \frac{n_2}{n}N(50, 1), \text{\ \ \ } x_2\sim \frac{1}{2}t_{13} + \frac{1}{2}(50 + t_{13}).
\end{align}
The corresponding score functions are computed based on the above
distributions. We note that the data set is a high-dimensional one,
since $n < d_2$, and $\tB_i$ cannot be estimated using many of the
related works, where typically $d_2<5$ \citep{Huang2013Profile,
  Guo2016Generalized, Zhao2017Quantile}. The histograms for three
traits are provided in Figure \ref{fig:9}. Note that they have been
normalized as in \cite{Ferrao2018Accurate}. We also refer to \cite{Ferrao2018Accurate} for other basic statistical analysis and
detailed preprocessing steps on this data set.

\begin{figure}[t]
	\centering     
	\subfigure[Production]{\label{tra1:prod}\includegraphics[width=49.9mm]{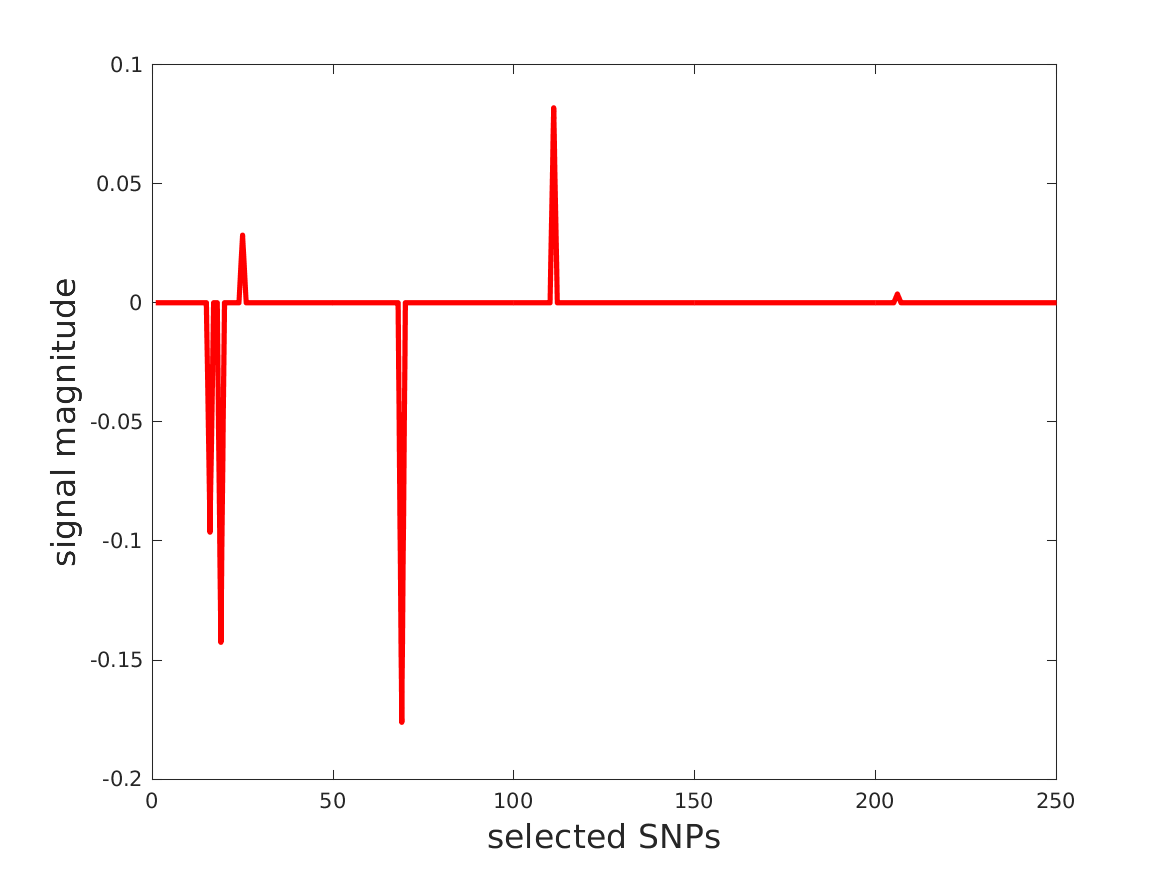}}
	\subfigure[Leaf rust]{\label{tra1:rust}\includegraphics[width=49.9mm]{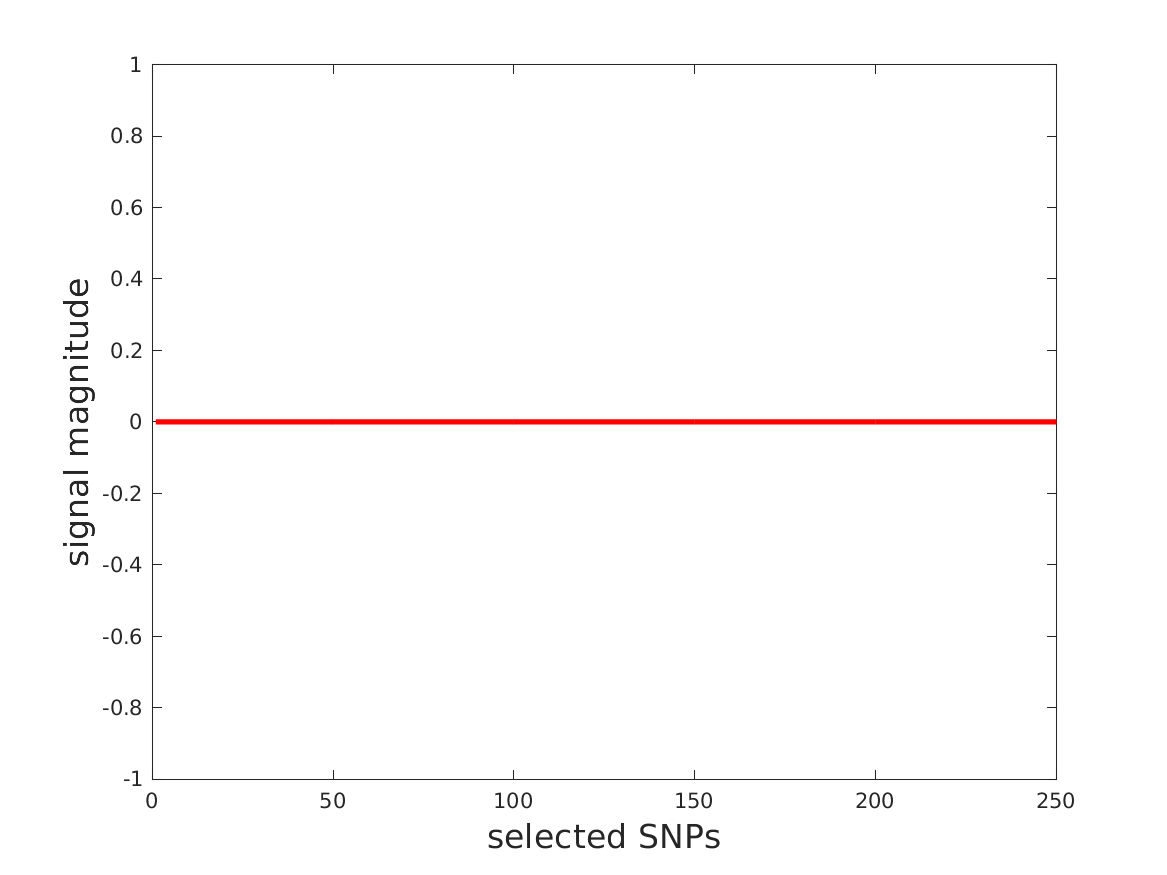}}
	\subfigure[Green beans]{\label{tra1:green}\includegraphics[width=49.9mm]{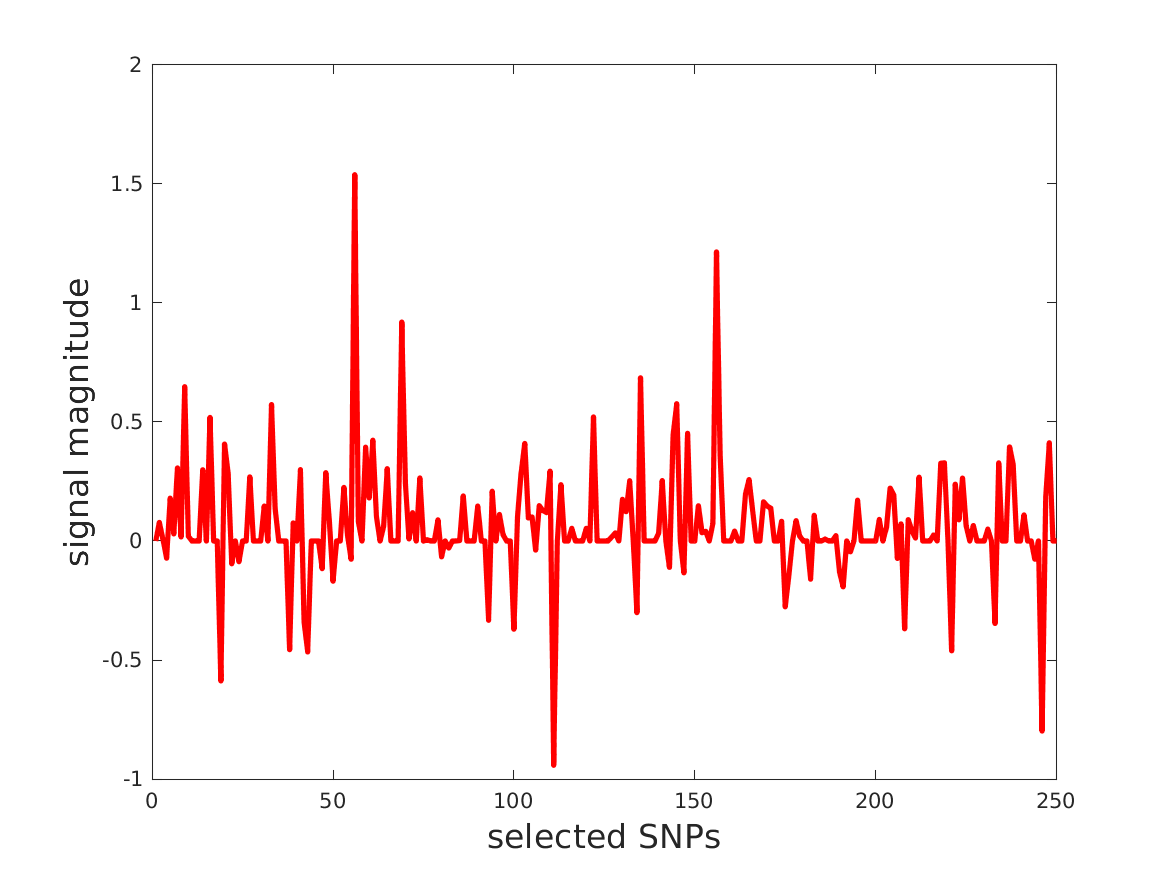}}
	\subfigure[Production]{\label{tra2:prod}\includegraphics[width=49.9mm]{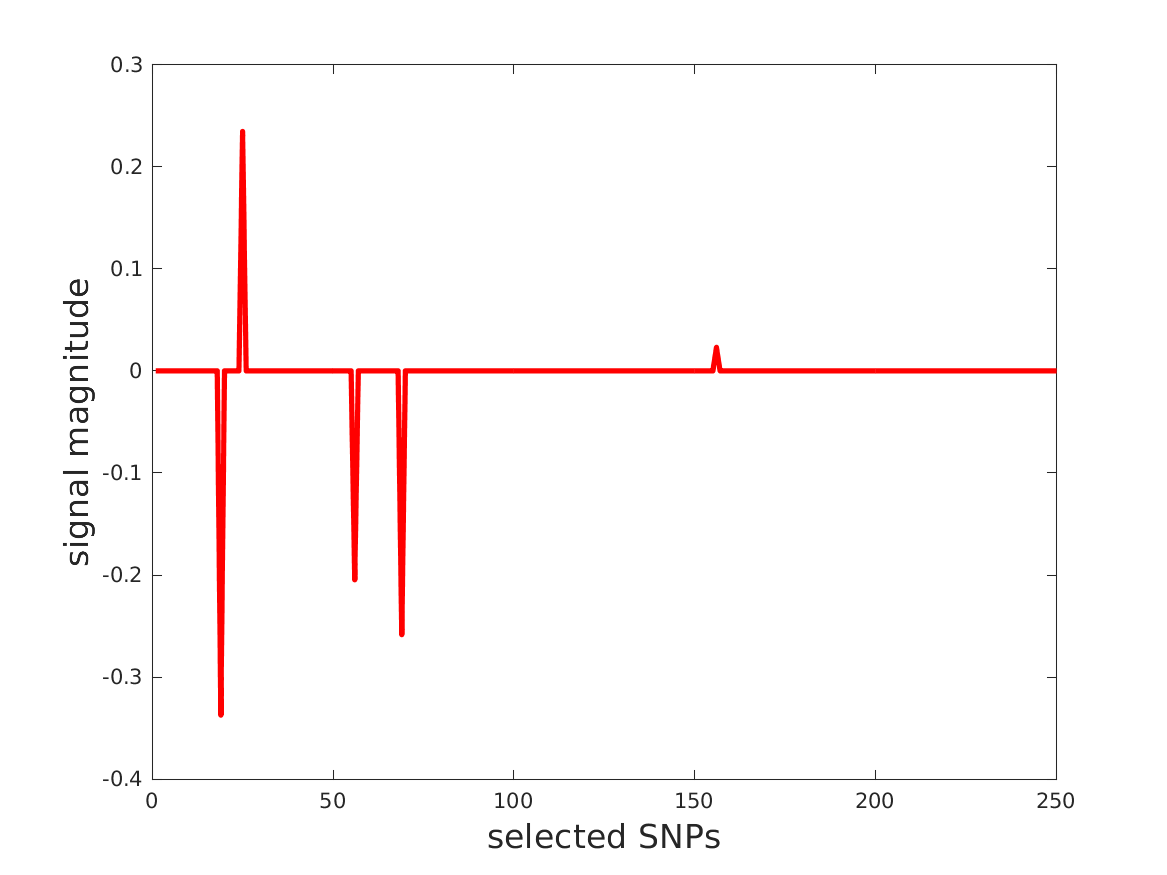}}
	\subfigure[Leaf rust]{\label{tra2:rust}\includegraphics[width=49.9mm]{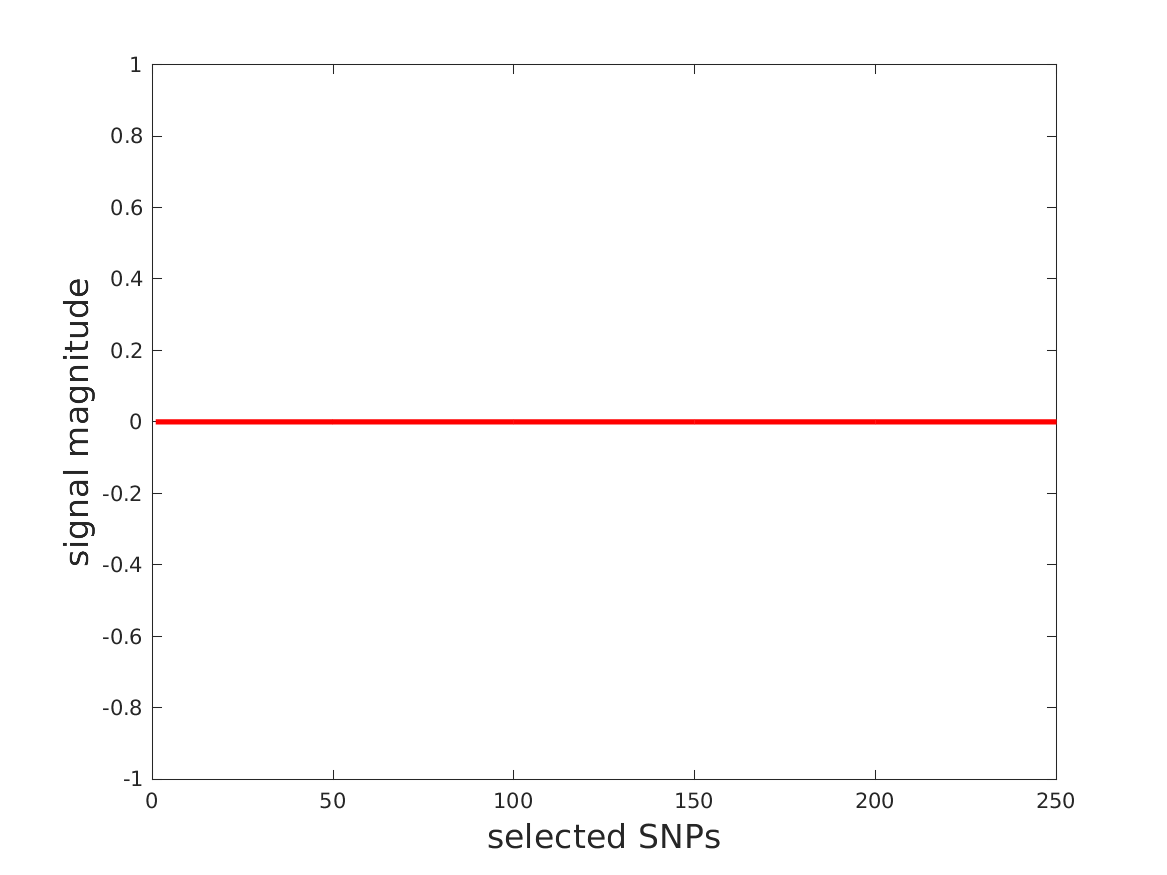}}
	\subfigure[Green beans]{\label{tra2:green}\includegraphics[width=49.9mm]{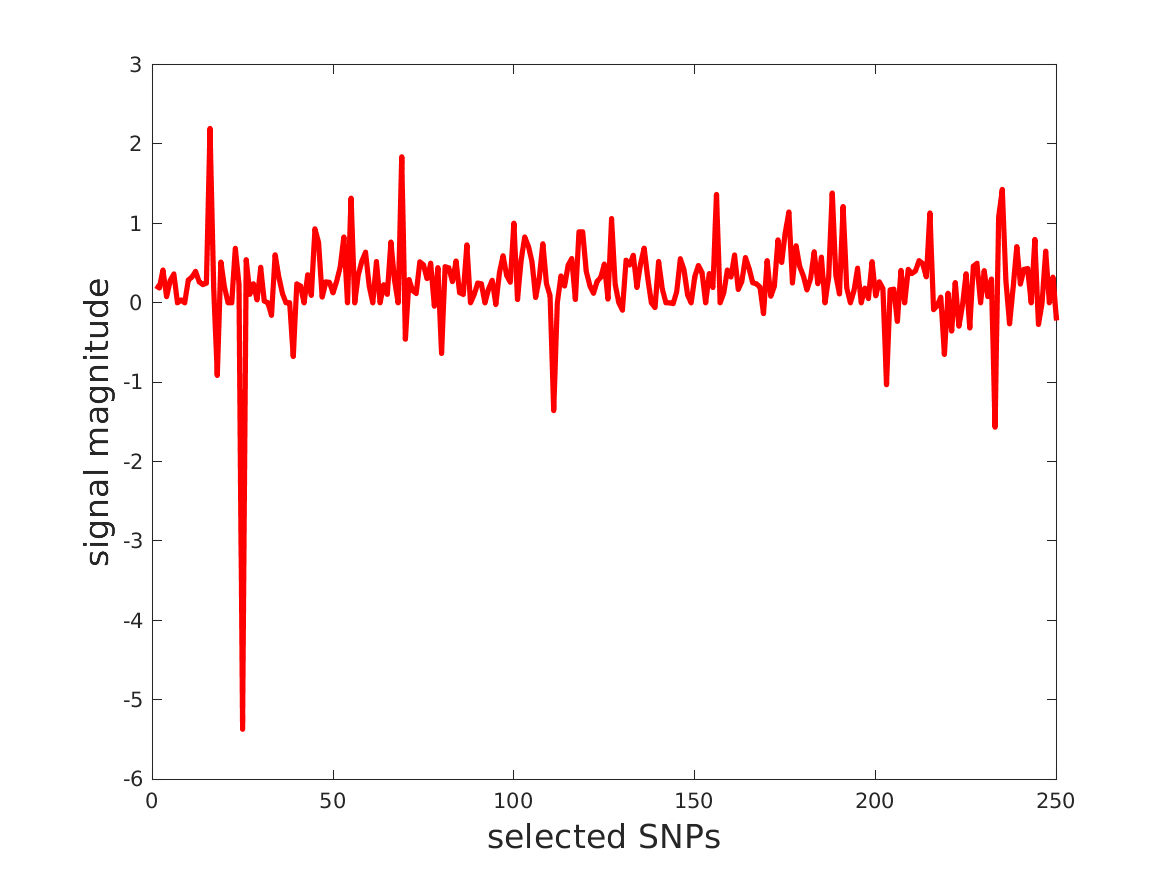}}
	\caption{{Signal trajectories. Each plot corresponds to a pair of confounders and responses, i.e. $(l, i)$ plot shows the trajectory of $(\bbeta^{l, i})^\star$ where $l = 1, 2$ indexes the confounder and $i = 1, 2, 3$ indexes the response.}}\label{fig:10}
\end{figure}

\begin{figure}[t]
	\centering     
	\subfigure[Production]{\label{tra1:prod:1}\includegraphics[width=49.9mm]{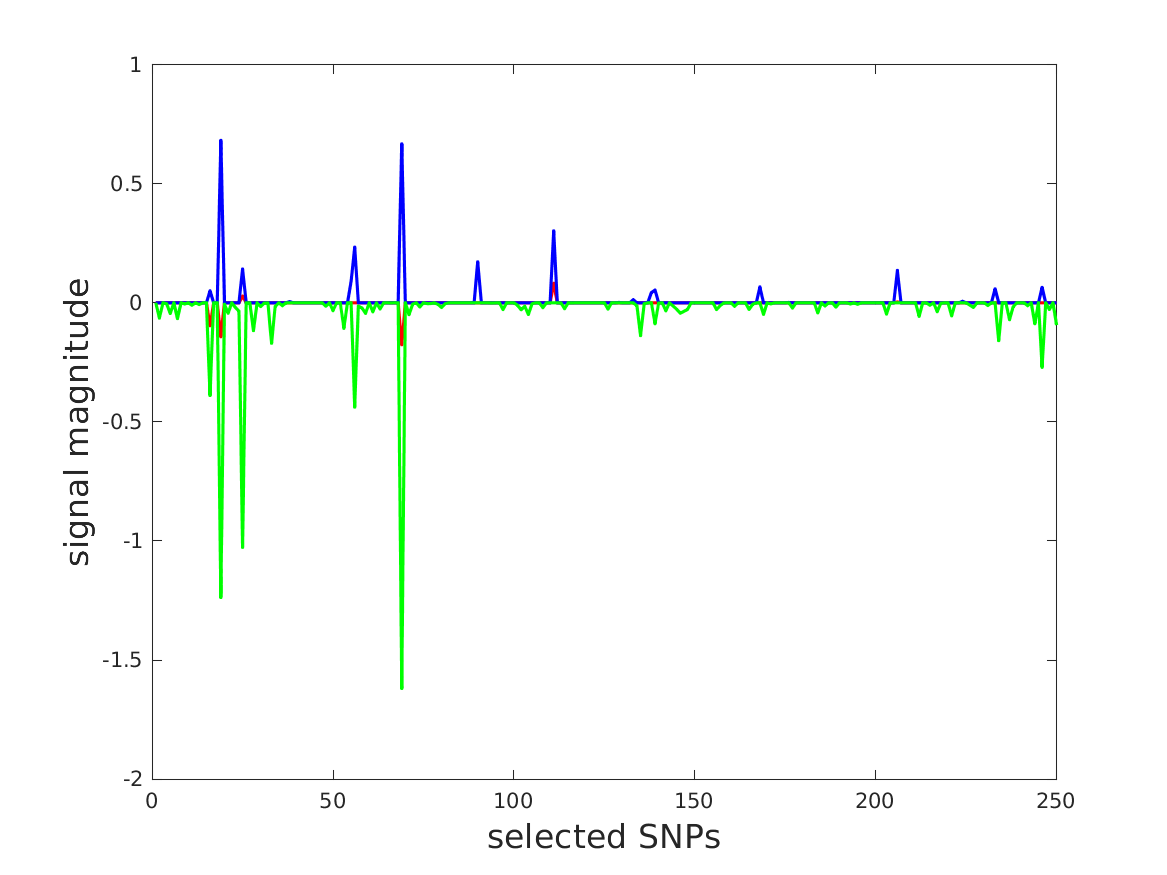}}
	\subfigure[Leaf rust]{\label{tra1:rust:1}\includegraphics[width=49.9mm]{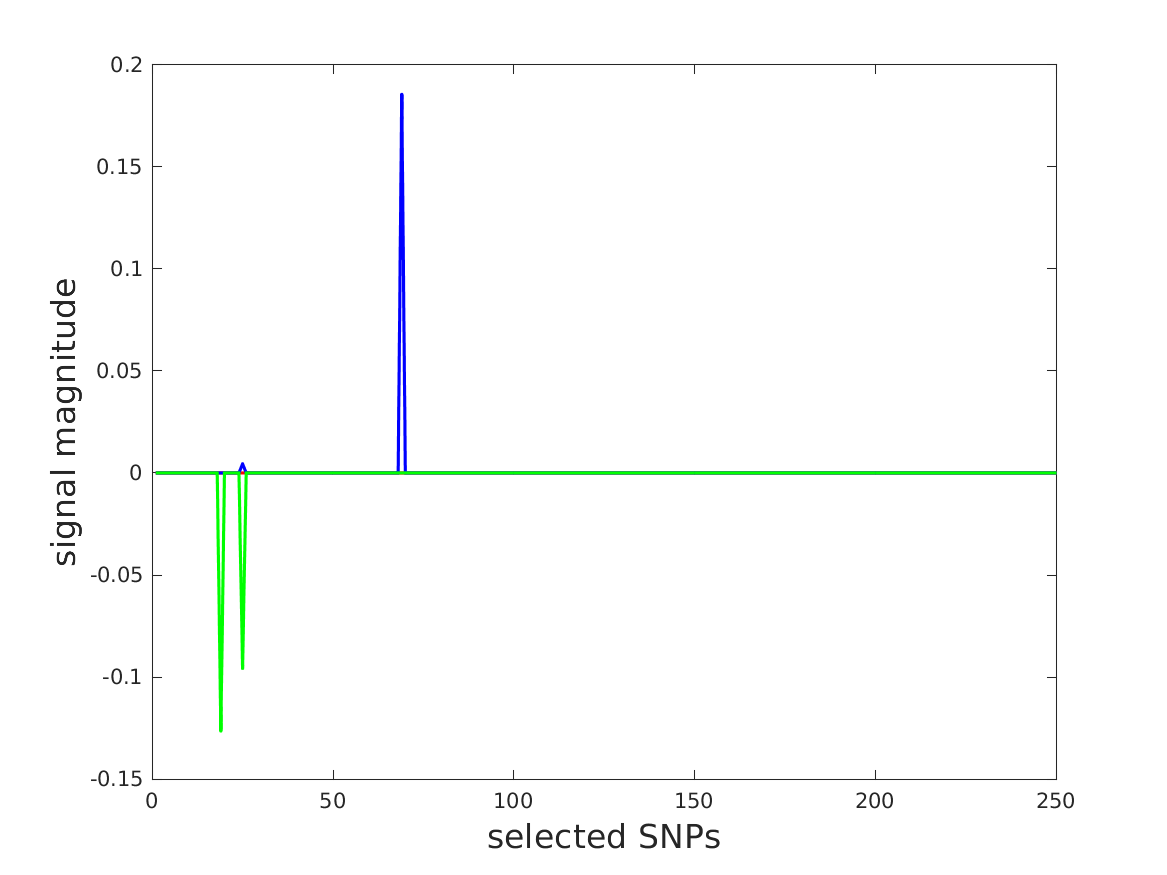}}
	\subfigure[Green beans]{\label{tra1:green:1}\includegraphics[width=49.9mm]{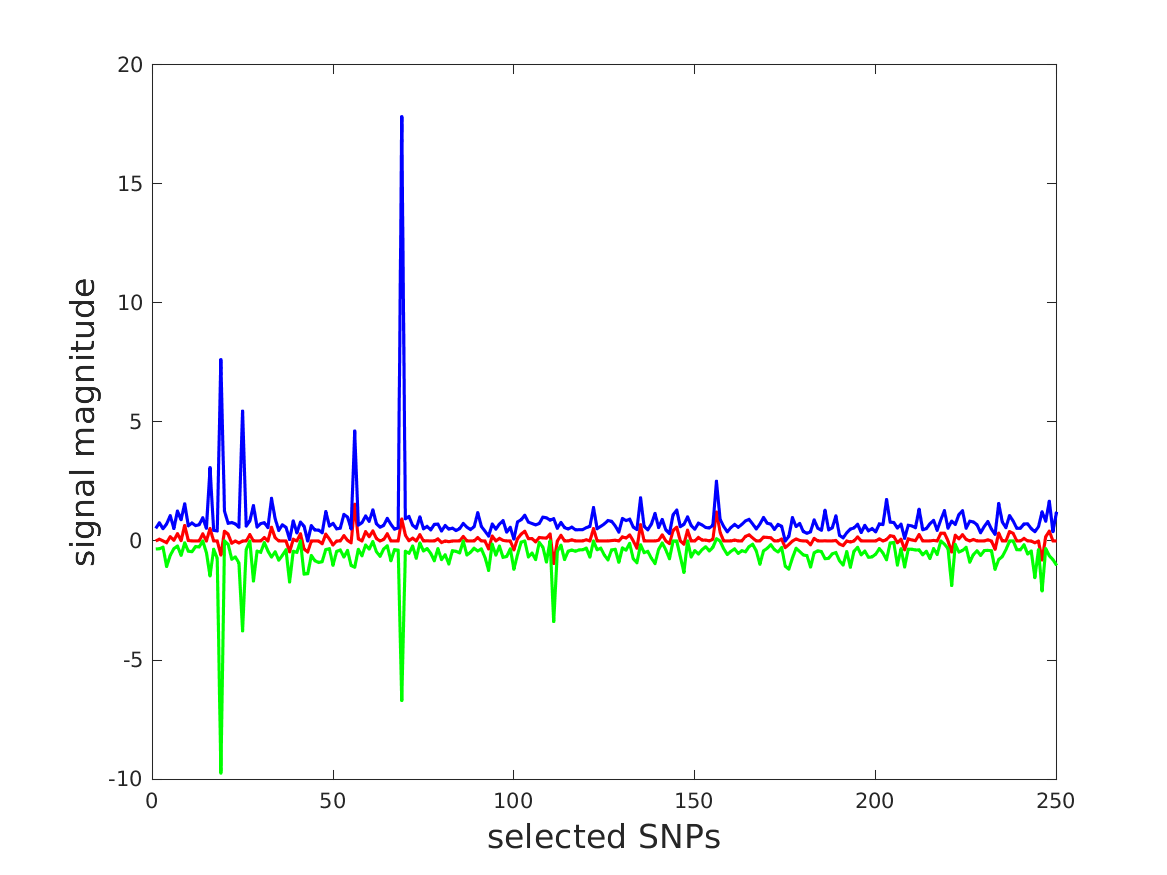}}
	\subfigure[Production]{\label{tra2:prod:1}\includegraphics[width=49.9mm]{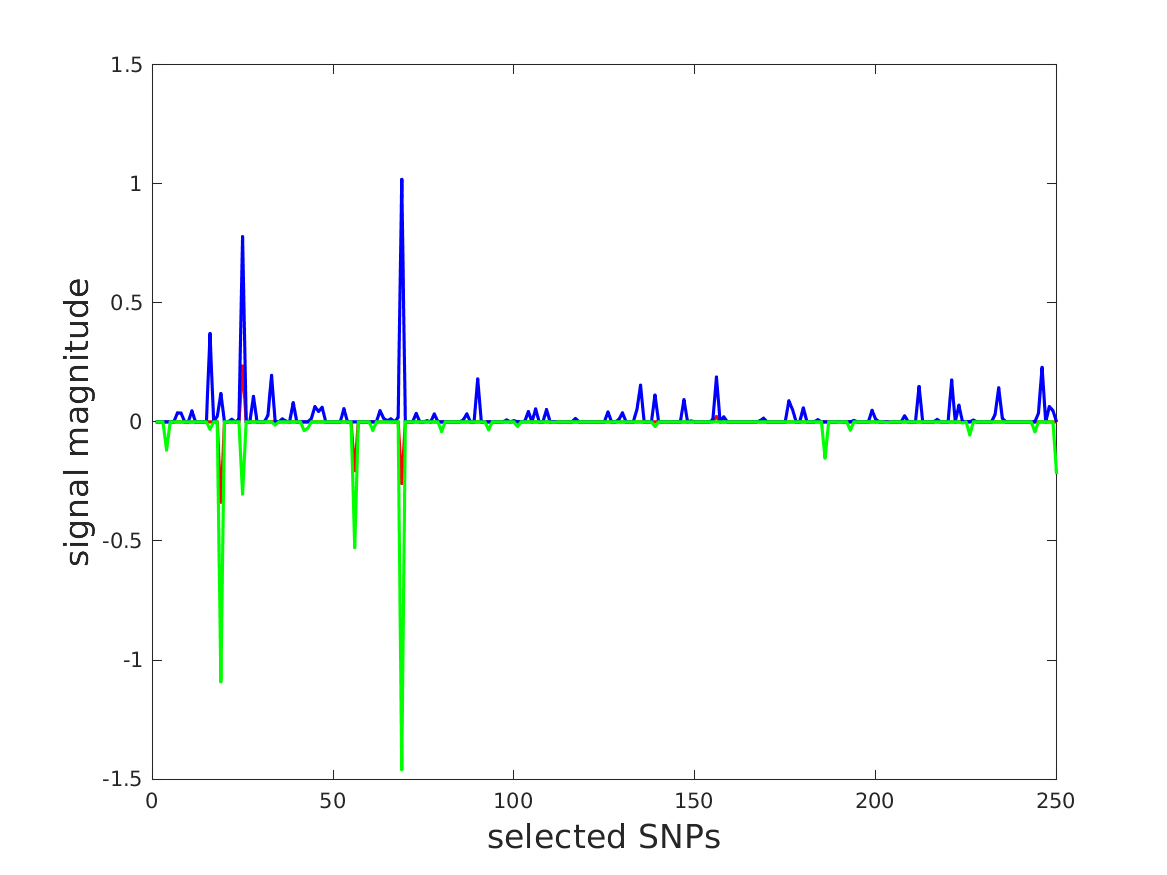}}
	\subfigure[Leaf rust]{\label{tra2:rust:1}\includegraphics[width=49.9mm]{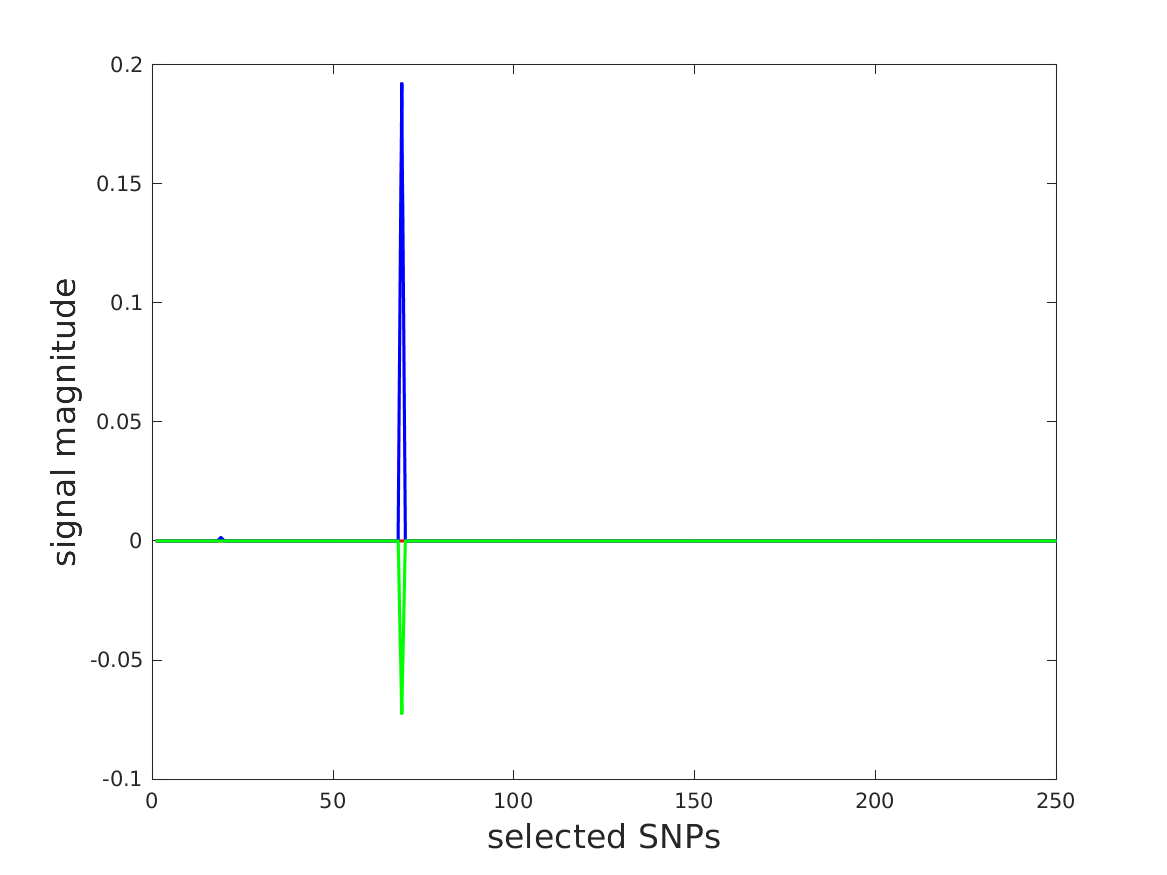}}
	\subfigure[Green beans]{\label{tra2:green:1}\includegraphics[width=49.9mm]{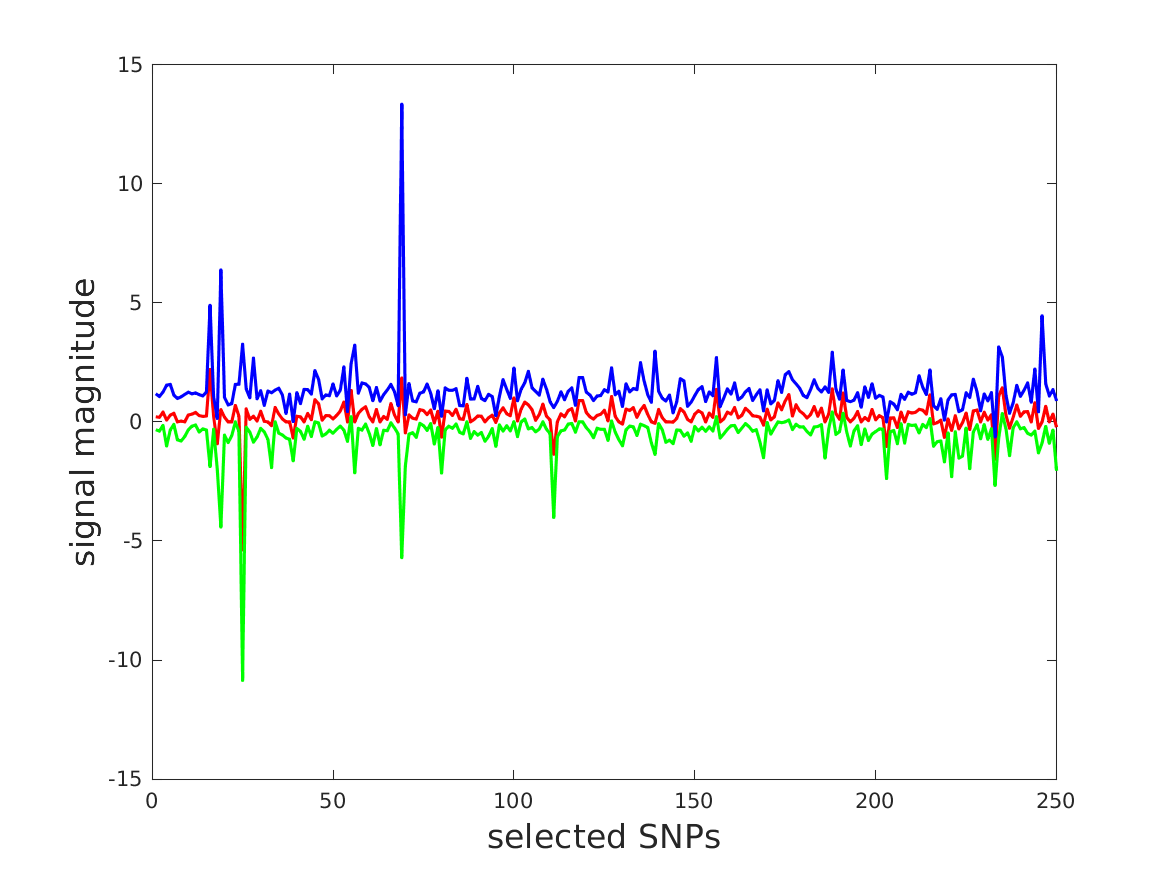}}
	\caption{{Signal trajectories for $95\%$ confidence interval (nonparametric bootstrapping). The $(l, i)$ plot shows the confidence interval trajectory of $(\bbeta^{l, i})^\star$ where $l = 1, 2$ indexes the confounder and $i = 1, 2, 3$ indexes the response. The blue line indicates the upper bound, the red line indicates the estimator, and the green line indicates the lower bound.  }}\label{fig:11}
\end{figure}

For each $i = 1, 2, 3$, we assume that $\tB_i$ is fully sparse. We compute the sparse matrix estimator via $(\ref{equ:13})$ with $\tau = \big(n/\log d_1d_2\big)^6$ and $\lambda = \sqrt{\log d_1d_2/n}$. The sparse precision matrix is estimated by conducting CLIME procedure with $\gamma = 5\sqrt{\log d_2/n}$. Throughout the experiment, we apply the hard truncation on confounders only since other variables are light-tailed already (for example, SNPs take value in $\{-1, 0, 1\}$). The estimated signal trajectories are drawn in Figure \ref{fig:10}. From the plots, we observe that when fitting $y_1$ under the model (\ref{mod:2}), both confounders may cause different effect sizes for $5$ out of $250$ selected SNPs, and $3$ of them affect common SNPs. However, confounders have no effects on selected SNPs when fitting $y_2$, while have dense effects when fitting $y_3$. In the latter case, we should mention the signals with large magnitude are still sparse.

In addition, we provide the $95\%$ confidence interval for each trajectory using a nonparametric bootstrap: we sample the data set with
replacement for $100$ times and for each data set we estimate $\tB_i$
using the same method with the same parameters, then compute $0.025$ and $0.975$ quantiles to construct the confidence interval. The results are shown in Figure \ref{fig:11} where the width of confidence interval indicates the variance of our estimators. We see some of identified signals have large variance but all unidentified signals have small variance. As a complement of analysis, we also conduct parametric bootstrapping: we generate new confounders $x_1$, $x_2$ from distribution in (\ref{conf:dist}) for $10000$ times, and for each new data set we estimate $\tB_i$ with the same setup, then use quantiles to construct confidence interval. The plots for parametric bootstrapping are provided in Figure \ref{fig:12} in Appendix
\ref{supple4} and similar observations appear again. Our code is available for download at:
\url{https://github.com/senna1128/Varying-Index-Coefficient-Models}.

\section{Conclusion}\label{sec:8}

In this paper, we proposed new estimators based on Stein's identity
for varying index coefficient models. By utilizing score function, we can either estimate a single sparse vector or a low-rank/sparse parameter matrix. The estimator can handle dependent covariates $\bz$ through estimation of the precision matrix and can achieve optimal convergence rate in sparse estimation and near optimal rate in low-rank estimation. In all cases, the estimators we proposed have closed form and are easy to implement. Instead of assuming that covariates $\bx$ follow an elliptical distribution, we only require certain finite moment assumption on response $y$, covariates $\bz$, and score variable $S(\bx)$. We also conduct extensive numerical
experiments to illustrate our result.

There are still many open problems worth exploring. One of future work
is about finite moment assumption. Under the general sparsity assumption, we believe that finite $6$-th moment condition is mild enough, however whether it is possible to relax it further is not clear. Furthermore, we note that all first-order Stein's estimators suffer from $\mu_k=\mE[f_k'(\langle \bx, \tbeta_k\rangle)] =0$. Therefore developing a second-order Stein's estimator is of practical interest and will be explored elsewhere.

\acks{This work was completed in part with resources provided by the
  University of Chicago Research Computing Center. We thank Kenji
  Fukumizu and two anonymous reviewers for helpful comments. }

\newpage

\appendix
\numberwithin{equation}{section}

\section{Estimate of Sparse Precision Matrix}\label{supple3}

We propose an approach to estimate a high-dimensional sparse precision matrix for heavy-tailed variable. Suppose $\bz$ has finite $4$th moment and $\tOmega = (\tSigma)^{-1}$, with $\tSigma = \mE[\bz\bz^T]$, is column sparse. In particular, we assume that $\tOmega\in \mF_w^K$\footnote{See definition in Section \ref{sec:5}.} for some $w$ and $K$. In this setting, we estimate the precision matrix using the CLIME procedure \citep{Cai2011Constrained}
\begin{equation}\label{equ:11}
\begin{aligned}
\min\ & \|\bOmega\|_{1,1},\\
\text{s.t.}\ \ & \|\hSigma\bOmega - \bI_{d_2}\|_{\max}\leq \gamma
\end{aligned}
\end{equation}
but with
\begin{align}\label{equ:16}
\hSigma = \frac{1}{n}\sum_{i=1}^{n}\wZi\wZi^T
\end{align}
being a thresholded estimator of the covariance matrix for some threshold $\tau>0$, and $\gamma$ is a tuning parameter. The linear program in \eqref{equ:11} is the same as in \cite{Cai2011Constrained},
with the difference that we use an estimator of $\tSigma$ that is more suitable for heavy-tailed data.

\begin{lemma}\label{lem:8b}
	
	If 	$\tau = (M_4n/\log d_2)^{1/4}/2$ and $\gamma = 12\|\tOmega\|_1\sqrt{M_4\log d_2/n}$, the estimator \eqref{equ:11} satisfies
	\begin{align*}
	P\bigg(\|\hOmega - \tOmega\|_2\leq 96\|\tOmega\|_1^2w\sqrt{M_4\log d_2/n}\bigg)\geq 1-\frac{2}{d_2^2},
	\end{align*}
	and
	\begin{align*}
	P\bigg(\|\hOmega - \tOmega\|_{\max}\leq 48\|\tOmega\|_1^2\sqrt{M_4\log d_2/n}\bigg)\geq 1-\frac{2}{d_2^2}.
	\end{align*}
	
\end{lemma}

From above lemma, we see the setting for $\gamma$ in (\ref{equ:11}) is oracle in the sense that $\|\tOmega\|_1$ is unknown. \cite{Cai2011Constrained} showed a detailed discussion on this aspect and this dependence could be removed by using a self-calibrated estimator, similar to scaled lasso \citep{Sun2012Sparse}. We should also mention that (\ref{equ:11}) achieves the optimal rate \citep{Cai2012Estimating}.

\section{Proofs of Lemmas}\label{supple1}

Throughout the proof, we frequently utilize the Bernstein's inequality presented in Corollary 2.11 in \cite{boucheron2013concentration}. To simplify subsequent presentation, we define a function to denote the common upper bound:
\begin{align*}
\varphi(t,a,b)=\exp(-\frac{t^2/2}{a+b\cdot t/3}).
\end{align*}
As shown in Bernstein's inequality, usually $a$ measures the total variance and $b$ is bound for a single variable. We also use $M$ as the substitute of $M_p$ ($p$ is certain moment) for simplicity. We summarize all structures we used in the paper for future reference.

\begin{assumption}[Column-wise sparse]\label{ass:4}
	We assume $\|\tbeta_k\|_0\leq s$, $\forall k\in[d_2]$.
\end{assumption}

\begin{assumption}[Fully sparse]\label{ass:7}
	We assume $\tB$ is $s$-sparse: $\|\tB\|_{0,1} = |\text{supp}(\tB)|\leq s$.
\end{assumption}

\begin{assumption}[Low-rank]\label{ass:6}
	We assume $\tB$ satisfies $\text{rank}(\tB)\leq r$.
\end{assumption}

\begin{assumption}[Independence]\label{ass:2}
	We assume $\bz$ satisfies $\mE[z_iz_j] = 0$, $\forall i\neq j\in[d_2]$.
\end{assumption}

\begin{assumption}[Precision matrix restriction]\label{ass:5}
	Define $\tSigma = \mE[\bz\bz^T]$ and let $\tOmega = (\tSigma)^{-1}$, we assume
	\begin{align*}
	\tOmega\in \mF_w^K = \bigg\{\bOmega\in\mR^{d_2\times d_2}: \|\bOmega\|_{0,\infty}\leq w, \|\bOmega\|_2\leq K, \|\bOmega^{-1}\|_2\leq K\bigg\}
	\end{align*}
	for some $w$ and $K$.
\end{assumption}

\subsection{Proof of Lemma \ref{lem:warmup:bound_score}}

Under Assumption \ref{ass:1}, we can get from (\ref{eq:emp_loss:warmup}) that
\begin{align*}
\nabla \hat{L}_k(\ttbeta_k)=2\ttbeta_k-\frac{2}{n}\sum_{i=1}^{n}y_iZ_{ik}\bX_i \stackrel{(\ref{equ:3})}{=} 2(\mE[yz_k\cdot\bx]-\frac{1}{n}\sum_{i=1}^{n}y_iZ_{ik}\bX_i).
\end{align*}
So, for any fixed $j\in[d_1]$, we have
\begin{align}\label{pequ:1}
[\nabla \hat{L}_k(\tilde{\bbeta}_k)]_j=2(\mE[yz_k\cdot x_j]-\frac{1}{n}\sum_{i=1}^{n}y_iZ_{ik}X_{ij}).
\end{align}
Note that $z_kx_j$ is a sub-exponential random variable with
\begin{align}\label{pequ:2}
\|z_kx_j\|_{\psi_1}\leq\|z_k\|_{\psi_2}\|x_j\|_{\psi_2}\leq\Upsilon_{\bz}\Upsilon_{\bx},
\end{align}
where $\Upsilon_{\bx}$ is $\psi_2$-norm of a standard Gaussian variable. So we see $\{y_i, Z_{ik}X_{ij}\}_{i\in[n]}$ are $n$ independent copies of $y$ and $z_kx_j$. Based on Lemma C.4 in \cite{yang2017misspecified} and (\ref{pequ:2}), let $\gamma=\max(\Upsilon_y,\Upsilon_{\bx}\Upsilon_{\bz})$ and we get
\begin{align*}
P(|\frac{1}{n}\sum_{i=1}^{n}y_iZ_{ik}X_{ij}-\mE[yz_k\cdot x_j]|>\Upsilon_{\gamma}\sqrt{\frac{\log n}{n}})<\frac{1}{n^2}
\end{align*}
where $\Upsilon_{\gamma}>0$ only depends on $\gamma$. According to equation (\ref{pequ:1}), we can take union bound and further have
\begin{align*}
P(\|\nabla \hat{L}_k(\tilde{\bbeta}_k)\|_{\infty}>2\Upsilon_{\gamma}\sqrt{\frac{\log n}{n}})<\frac{d_1}{n^2},
\end{align*}
which concludes the proof.

\subsection{Proof of Lemma \ref{lem:6}}

Based on equation (\ref{equ:6}), we know
\begin{align*}
\nabla \bar{L}_k(\tilde{\bbeta}_k)=2\tilde{\bbeta}_k-\frac{2}{n}\sum_{i=1}^{n}\wyi\wZik\wSXi.
\end{align*}
Under Assumption \ref{ass:1} and \ref{ass:2}, we know $\ttbeta_k = \mE[y z_k\cdot S(\bx)]$. So we can separate above gradient into two parts
\begin{align}\label{pequ:11}
\|&\nabla \bar{L}_k(\tilde{\bbeta}_k)\|_{\infty}= 2\|\tilde{\bbeta}_k-\frac{1}{n}\sum_{i=1}^{n}\wy_i\wZik\wSXi\|_{\infty} \nonumber\\
\leq& 2\|\underbrace{\mE[yz_k\cdot S(\bx)]-\frac{1}{n}\sum_{i=1}^{n}\mE[\wy_i\wZik\wSXi]}_{\I_1}\|_{\infty}+2\|\underbrace{\frac{1}{n}\sum_{i=1}^{n}\mE[\wy_i\wZik\wSXi]-\frac{1}{n}\sum_{i=1}^{n}\wy_i\wZik\wSXi}_{\I_2}\|_{\infty}.
\end{align}
We will provide a deterministic bound for $\I_1$ and probabilistic bound for $\I_2$. Let's deal with $\I_1$ first. For any $j\in[d_1]$, we know
\begin{align}\label{pequ:12}
\I_{1j} =& \mE[yz_k\cdot S(\bx)_j]- \mE[\wy \wZk\cdot \wS(X)_j] \nonumber \\
=& \mE\big[yz_k\cdot S(\bx)_j\cdot\pmb{1}_{|y|>\tau\text{\ or\ } |z_k|>\tau\text{\ or\ }|S(\bx)_j|>\tau}\big] \nonumber\\
\leq & \sqrt{\mE[y^2\bz_k^2S(\bx)_j^2]\cdot \big(P(|y|>\tau)+P(|\bz_k|>\tau)+P(|S(\bx)_j|>\tau)\big)} \nonumber\\
\leq & \sqrt[4]{\mE[y^4]\mE[z_k^4]\mE[S(\bx)_j^4]}\frac{\sqrt{3}M^{1/2}}{\tau^3} \nonumber\\
\leq & \frac{2M}{\tau^3}.
\end{align}
Here, the third inequality is from Cauchy-Schwarz inequality; the fourth inequality is Chebyshev inequality; the last inequality is due to Assumption \ref{ass:3} ($p=6$). So from (\ref{pequ:12}), we know
\begin{align}\label{pequ:13}
\|\I_1\|_{\infty}\leq 2M/\tau^3.
\end{align}
For the $\I_2$ term in equation (\ref{pequ:11}), we apply Bernstein's inequality. We have $\forall j\in[d_1]$,
\begin{equation}\label{pequ:14}
\begin{aligned}
&-\tau^3\leq \wy_i\wZik\wSXi_j\leq\tau^3\Longrightarrow C = 2\tau^3,\\
&V_n=\sum_{i=1}^{n}\VAR(\wy_i\wZik\wSXi_j)\leq\sum_{i=1}^{n}\mE[\wy_i^2\wZik^2\wSXi_j^2]\leq nM.
\end{aligned}
\end{equation}
Thus, based on (\ref{pequ:14}), we have $\forall t>0$,
\begin{align}\label{pequ:15}
P(\big|\mE[\wy\wZk\cdot\wS(X)_j]-\frac{1}{n}\sum_{i=1}^{n}\wy_i\wZik\wSXi_j\big|>t)\leq 2\varphi(nt,nM,2\tau^3).
\end{align}
Then we take union bound for (\ref{pequ:15}) and get
\begin{align}\label{pequ:16}
P(\|\I_2\|_{\infty} >t)\leq 2d_1\varphi(nt,nM,2\tau^3)
\end{align}
Combine (\ref{pequ:13}) and (\ref{pequ:16}) together, and take union bound over $k$, we have $\forall t, \tau>0$,
\begin{align}\label{pequ:18}
P(\|\nabla \bar{L}_k(\tilde{\bbeta}_k)\|_{\infty}\leq \frac{4M}{\tau^3}+2t, \text{\ \ } \forall k\in[d_2])\geq 1 - 2d_1d_2\exp(-\frac{nt^2}{2M+2\tau^3 t}).
\end{align}
Suppose, for some positive constant $c_1, c_2$, we let
\begin{align}\label{pequ:19}
t = c_1\sqrt{\log d_1d_2/n}  \text{\ \ and\ \ } \tau = c_2^{1/3}(n/\log d_1d_2)^{1/6}.
\end{align}
Then plug (\ref{pequ:19}) into the right hand side of (\ref{pequ:18}) and get
\begin{align}\label{pequ:20}
2d_1d_2\exp(-\frac{nt^2}{2M+2\tau^3 t}) = 2d_1d_2\exp(-\frac{c_1^2\log d_1d_2}{2M + 2c_1c_2})\leq 2/d_1^2d_2^2,
\end{align}
if
\begin{align}\label{pequ:21}
\frac{c_1^2}{2M+2c_1c_2}\geq 3\Longrightarrow c_1^2 - 6c_1c_2-6M\geq 0.
\end{align}
We can set $c_1 = 3\sqrt{M}$ and $c_2 = \sqrt{M}/8$, which satisfies condition in (\ref{pequ:21}) naturally. Further we know (\ref{pequ:20}) holds. Plug this setting in (\ref{pequ:19}) and (\ref{pequ:18}) and we get
\begin{align}\label{pequ:22}
\|\nabla \bar{L}_k(\tilde{\bbeta}_k)\|_{\infty}\leq 38\sqrt{M\log d_1d_2/n}, \text{\ \ }\forall k\in[d_2]
\end{align}
with probability at least $1-2/d_1^2d_2^2$. This finishes the proof.

\subsection{Proof of Lemma \ref{lem:7}}

Define $\I_3 = \frac{1}{n\kappa_1}\sum_{i=1}^{n}\Phi(\kappa_1 y_i\cdot S(\bX_i)\bZ_i^T)-\mE[y\cdot S(\bx)\bz^T] $, we will apply Corollary 3.1 in \cite{Minsker2018Sub}. Let's first bound the variance. Under Assumption \ref{ass:1}, we know $\mE[S(\bx)_j] = 0, \forall j\in[d_1]$. So for any unit vector $\bv\in\mR^{d_1}$, we have
\begin{align}\label{pequ:25}
\mE[y^2&\cdot \bv^TS(\bx)\bz^T\bz S(\bx)^T\bv]=\mE[y^2\cdot\bz^T\bz\cdot(S(\bx)^T\bv)^2]\leq\sqrt{\mE[y^4]\mE[(\bz^T\bz)^2]\mE[(S(\bx)^T\bv)^4]} \nonumber\\
\leq&M^{1/2}\sqrt{\mE[d_2(\bz_1^4+...+\bz_{d_2}^4)]}\sqrt{\mE[\sum_{i_1=1}^{d_1}\sum_{i_2=1}^{d_1}S(\bx)^2_{i_1}S(\bx)^2_{i_2}\bv^2_{i_1}\bv^2_{i_2}]} \nonumber\\
\leq& d_2M\sqrt{\sum_{i_1=1}^{d_1}\sum_{i_2=1}^{d_1}\mE[S(\bx)^2_{i_1}S(\bx)^2_{i_2}]\bv^2_{i_1}\bv^2_{i_2}} \leq d_2M\sqrt{\sum_{i_1=1}^{d_1}\sum_{i_2=1}^{d_1}\sqrt{\mE[S(\bx)^4_{i_1}]}\sqrt{\mE[S(\bx)^4_{i_2}]}\bv^2_{i_1}\bv^2_{i_2}} \nonumber\\
\leq& d_2M^{3/2}.
\end{align}
The second inequality uses Cauchy-Schwarz inequality; the third inequality uses Assumption~\ref{ass:1}. From (\ref{pequ:25}) we can get
\begin{align}\label{pequ:26}
\|\mE[y^2\cdot S(\bx)\bz^T\bz S(\bx)^T]\|_2\leq d_2M^{3/2}.
\end{align}
Follow the exactly same derivation in (\ref{pequ:25}) we can also have
\begin{align}\label{pequ:27}
\|\mE[y^2\cdot \bz S(\bx)^TS(\bx)\bz ^T]\|_2\leq d_1M^{3/2}.
\end{align}
Thus, combine (\ref{pequ:26}) and (\ref{pequ:27}) together, we have $\forall t>0$
\begin{align}\label{pequ:28}
P(\|\I_3\|_2\geq t)\leq 2(d_1+d_2)\exp(-n\kappa_1 t + \frac{n(d_1+d_2)M^{3/2}\kappa_1^2}{2}).
\end{align}
In above tail bound (\ref{pequ:28}), we let $t = 2M^{3/4}\sqrt{\frac{2(d_1+d_2)\log(d_1+d_2)}{n}}$ and $\kappa_1 = \sqrt{\frac{2\log(d_1+d_2)}{n(d_1+d_2)M^{3/2}}}$ and have
\begin{align}\label{pequ:41}
P\bigg(\|\I_3\|\leq 2M^{3/4}\sqrt{\frac{2(d_1+d_2)\log(d_1+d_2)}{n}}\bigg)\geq 1 - \frac{2}{(d_1+d_2)^2}.
\end{align}
This is consistent with argument of lemma.

\subsection{Proof of Lemma \ref{lem:8a}}

Let's first get concentration rate for $\|\hSigma - \tSigma\|_2=\|\frac{1}{n\kappa_2}\Phi(\kappa_2 Z_iZ_i^T) - \mE[\bz\bz^T]\|_2$. We have $\forall \bv\in\mR^{d_2}$ such that $\|\bv\|_2=1$,
\begin{align*}
\mE[\bv^T\bz\bz^T\bv] = \mE[(\bv^T\bz)^2]\leq \mE[\|\bz\|^2_2]\leq d_2\sqrt{M}.
\end{align*}
Based on Corollary 3.1 in \cite{Minsker2018Sub}, we know $\forall t>0$,
\begin{align}\label{pequ:53}
P(\|\hSigma - \tSigma\|_2\geq t)\leq 2d_2\exp(-n\kappa_2 t + \frac{nd_2\sqrt{M}\kappa_2^2}{2}).
\end{align}
In above (\ref{pequ:53}), we let $t = 2M^{1/4}\sqrt{\frac{2d_2\log d_2}{n}}$ and $\kappa_2 = \sqrt{\frac{2\log d_2}{nd_2M^{1/2}}}$, and have
\begin{align}\label{pequ:54}
P\bigg(\|\hSigma - \tSigma\|_2\leq 2M^{1/4}\sqrt{\frac{2d_2\log d_2}{n}}\bigg)\geq 1-\frac{2}{d_2^2}.
\end{align}
We use matrix perturbation analysis to give bound for $\hOmega$. As shown in Chapter III Theorem 2.5 in \cite{stewart90matrix}, when
\begin{align*}
\|\tOmega(\hSigma - \tSigma)\|_2\leq \|\tOmega\|_2\|\hSigma - \tSigma\|_2\leq \|\tOmega\|_24M^{1/4}\sqrt{d_2\log d_2/n}\leq 1/2,
\end{align*}
we know $\hOmega$ is perforce invertible and satisfies
\begin{align}\label{pequ:55}
\|\hOmega - \tOmega\|_2\leq 2\|\tOmega\|_2^2\|\hSigma -\tSigma\|_2\leq 8\|\tOmega\|_2^2M^{1/4}\sqrt{d_2\log d_2/n},
\end{align}
with probability at least $1-2/d_2^2$. Therefore we finish the proof.

\subsection{Proof of Lemma \ref{lem:10}}

We define $\I_7 = \mE[y\cdot S(\bx)\bz^T] -  \frac{1}{ n}\sum_{i=1}^{n}\wy_i\cdot\wSXi\wZi^T$. For any $j\in[d_1], k\in[d_2]$, we know
\begin{align*}
|(\I_7)_{jk}|\leq |\frac{1}{n}\sum_{i=1}^{n}\wy_i \wZik \wSXi_j - \mE[\wy_i \wZik \wSXi_j ]| + |\frac{1}{n}\sum_{i=1}^{n}\mE[\wy_i \wZik \wSXi_j] - \mE[y\cdot z_k\cdot S(\bx)_j]|.
\end{align*}
Use the same technique as in (\ref{pequ:12})-(\ref{pequ:16}), we can get that $\forall t,\tau>0$
\begin{align*}
P(\|\I_7\|_{\max} > t + \frac{2M}{\tau^3})\leq 2d_1d_2\exp(-\frac{nt^2}{2M + 2\tau^3 t}).
\end{align*}
We let $t = 3\sqrt{M\log d_1d_2/n}$, $\tau = (Mn/\log d_1 d_2)^{1/6}/2$ and have
\begin{align}\label{pequ:57}
P\bigg(\|\I_7\|_{\max}\leq 19\sqrt{\frac{M\log d_1d_2}{n}}\bigg)\geq 1 - 2/d_1^2d_2^2.
\end{align}
So we finish the proof of lemma.

\subsection{Proof of Lemma \ref{lem:8b}}

We first prove a concentration bound for truncated empirical covariance $\hSigma$. $\forall j,k\in[d_2]$,
\begin{multline}\label{pequ:29}
|\hSigmajk - \tSigmajk| = |\frac{1}{n}\sum_{i=1}^{n}\wZij\wZik - \mE[z_jz_k]|\\
\leq \underbrace{|\frac{1}{n}\sum_{i=1}^{n}(\wZij\wZik -\mE[\wZij\wZik])| }_{\I_5}+ \underbrace{|\frac{1}{n}\sum_{i=1}^{n}\mE[\wZij\wZik]- \mE[z_jz_k]|}_{\I_6}.
\end{multline}
Use Bernstein's inequality for $\I_5$, we have
\begin{align*}
&-\tau^2\leq \wZij\wZik\leq \tau^2,\\
&V_n = \sum_{i=1}^{n}\VAR(\wZij\wZik)\leq \sum_{i=1}^{n}\mE[\wZij^2\wZik^2]\leq nM.
\end{align*}
Note that the above last inequality holds no matter whether $j=k$ or not. Thus, $\forall t>0$, we have
\begin{align}\label{pequ:30}
P(|\I_5|>t) \leq 2\varphi (nt,nM,2\tau^2).
\end{align}
For the term $\I_6$, we know
\begin{align}\label{pequ:31}
|\I_6| =& \mE[z_jz_k\cdot\pmb{1}_{\{|z_j|>\tau\text{\ or\ } |z_k|>\tau\}}] \leq \sqrt{\mE[z_j^2z_k^2]\cdot (P(|z_j|>\tau) + P(|z_k|>\tau))} \leq \frac{2M}{\tau^2}.
\end{align}
Combine (\ref{pequ:29}), (\ref{pequ:30}) and (\ref{pequ:31}) together, we get
\begin{align}\label{pequ:32}
|\hSigmajk - \tSigmajk|\leq t + \frac{2M}{\tau^2},
\end{align}
with probability at least $1 - 2\exp(-\frac{nt^2}{2M + 2\tau^2 t})$. Take union bound for (\ref{pequ:32}) we have
\begin{align*}
P(\|\hSigma - \tSigma\|_{\max}\leq t + \frac{2M}{\tau^2})\geq 1 - 2d_2^2\varphi (nt,nM,2\tau^2).
\end{align*}
Let $t = 4\sqrt{M\log d_2/n}$, $\tau = (Mn/\log d_2)^{1/4}/2$, we have
\begin{align}\label{pequ:33}
P\bigg(\|\hSigma - \tSigma\|_{\max}\leq 12 \sqrt{\frac{M\log d_2}{n}}\bigg)\geq 1-\frac{2}{d_2^2}.
\end{align}
Based on this bound, we deal with convex problem (\ref{equ:11}). Suppose $\hOmega = (\homega_1,...,\homega_{d_2})$, we will show each $\homega_j$ is also a solution to following problem:
\begin{equation}\label{pequ:34}
\begin{aligned}
\min_{\bl_j}\ &\|\bl_j\|_1,\\
\text{s.t.}\ \ & \|\hSigma\bl_j - \be_j\|_{\infty}\leq \gamma.
\end{aligned}
\end{equation}
In fact, it's easy to see $\homega_j$ is a feasible point for problem (\ref{pequ:34}), so $\|\hl_j\|_1\leq \|\homega_j\|_1$. Further we know $\|(\hl_1,...,\hl_{d_2})\|_{1,1}\leq \|\hOmega\|_{1,1}$. On the other hand, $\|\homega_j\|_1\leq \|\hl_j\|_1$ for sure. Otherwise $(\homega_1,...,\hl_j,...,\homega_{d_2})$ satisfies condition of (\ref{equ:11}) but with smaller objective value. In this case, we know each $\homega_j$ can also be solved from (\ref{pequ:34}). Note for $\tOmega = (\tomega_1,...,\tomega_{d_2})\in\mR^{d_2\times d_2}$, we have
\begin{align*}
\|\hSigma\tOmega - \bI_{d_2}\|_{\max} = \|(\hSigma - \tSigma + \tSigma)\tOmega - \bI_{d_2}\|_{\max} = \|(\hSigma - \tSigma)\tOmega\|_{\max}\leq \|\hSigma - \tSigma\|_{\max}\|\tOmega\|_1.
\end{align*}
So when $\|\hSigma - \tSigma\|_{\max}\|\tOmega\|_1\leq \gamma$, we know $\tOmega$ is feasible for problem (\ref{equ:11}) and $\tomega_j$ is feasible for problem (\ref{pequ:34}). So we know
\begin{align}\label{pequ:35}
\|\hOmega\|_{1,1}\leq \|\tOmega\|_{1,1}\text{\ \ and\ \ } \|\homega_j\|_1\leq \|\tomega_j\|_1.
\end{align}
Based on (\ref{pequ:35}), we know $\|\hOmega\|_1\leq \|\tOmega\|_1$. Further, we have
\begin{align}\label{pequ:36}
\|\tSigma(\hOmega - \tOmega)\|_{\max}\leq& \|(\tSigma-\hSigma)(\hOmega - \tOmega)\|_{\max} + \|\hSigma(\hOmega - \tOmega)\|_{\max} \nonumber\\
\leq & \|\tSigma - \hSigma\|_{\max}\|\hOmega - \tOmega\|_1 + \|\hSigma\hOmega - \bI_{d_2}\|_{\max}+ \|\hSigma\tOmega - \bI_{d_2}\|_{\max} \nonumber\\
\leq & \|\tSigma - \hSigma\|_{\max}(\|\hOmega\|_1 + \|\tOmega\|_1) + 2\gamma \nonumber\\
\leq & 2 \|\tSigma - \hSigma\|_{\max}\|\tOmega\|_1+2\gamma \nonumber\\
\leq & 4\gamma.
\end{align}
Based on the bound (\ref{pequ:36}), we have
\begin{align}\label{pequ:37}
\|\hOmega - \tOmega\|_{\max} = \|\tOmega\tSigma(\hOmega - \tOmega)\|_{\max}\leq \|\tOmega\|_{\infty}\|\tSigma(\hOmega - \tOmega)\|_{\max}\leq 4\gamma\|\tOmega\|_1.
\end{align}
For the last inequality in (\ref{pequ:37}), we use $\|\tOmega\|_1 = \|\tOmega\|_{\infty}$ because $\tOmega$ is symmetric matrix. To proceed, let's derive the cone condition. Define $\Delta_j = \homega_j - \tomega_j$ and $s_j = \text{supp}(\tomega_j)$. From equation (\ref{pequ:35}) we know
\begin{align}\label{pequ:38}
\|\tomega_j\|_1\geq \|\homega_j\|_1 = \|\Delta_j + \tomega_j\|_1 = \|(\Delta_j + \tomega_j)_{s_j}\|_1 + \|(\Delta_j)_{s_j^c}\|_1\Longrightarrow \|(\Delta_j)_{s_j}\|_1\geq \|(\Delta_j)_{s_j^c}\|_1.
\end{align}
Combine the cone condition in (\ref{pequ:38}) with (\ref{pequ:37}) and Assumption \ref{ass:5}, we know
\begin{align}\label{pequ:61}
\|\Delta_j\|_1\leq 2\|(\Delta_j)_{s_j}\|_1\leq 2 w\|\Delta_j\|_{\infty} = 2w\|\hOmega - \tOmega\|_{\max}\leq 8\|\tOmega\|_1w\gamma.
\end{align}
Therefore, if $\|\hSigma - \tSigma\|_{\max}\|\tOmega\|_1\leq \gamma$, then
\begin{align}\label{pequ:39}
\|\hOmega - \tOmega\|_2\leq \sqrt{\|\hOmega - \tOmega\|_1\|\hOmega - \tOmega\|_{\infty}} = \|\hOmega - \tOmega\|_1\leq 8\|\tOmega\|_1w\gamma.
\end{align}
Based on (\ref{pequ:33}) and (\ref{pequ:39}), we can choose $\gamma = 12\|\tOmega\|_1\sqrt{M\log d_2/n}$, then with probability at least $1- 2/d_2^2$, we have
\begin{align}\label{pequ:40}
\|\hOmega - \tOmega\|_2\leq 96\|\tOmega\|_1^2w\sqrt{M\log d_2/n}.
\end{align}
Further, from (\ref{pequ:37}), we know with probability at least $1-2/d_2^2$
\begin{align}\label{pequ:58}
\|\hOmega - \tOmega\|_{\max}\leq 4\|\tOmega\|_1\gamma\leq 48\|\tOmega\|_1^2\sqrt{M\log d_2/n}.
\end{align}
This concludes the proof.

\section{Proofs of Theorems and Corollaries}\label{supple2}

\subsection{Proof of Theorem \ref{thm:conv_warmup}}

Let's fix $k\in[d_2]$ first. Based on the definition of $\hbeta_k$ in (\ref{eq:emp_loss:warmup}), we have following basic inequality
\begin{align}\label{pequ:3}
\hL_k(\hbeta_k)+\lambda_k\|\hbeta_k\|_1\leq \hL_k(\ttbeta_k)+\lambda_k\|\ttbeta_k\|_1.
\end{align}
We define $\btheta_k=\hbeta_k-\ttbeta_k$ and have
\begin{align}\label{pequ:4}
\hat{L}_k(\hbeta_k)-\hat{L}_k(\ttbeta_k)&=\|\btheta_k\|_2^2+2\langle \ttbeta_k, \btheta_k\rangle-\frac{2}{n}\sum_{i=1}^{n}y_iZ_{ik}\langle \bX_i,\btheta_k\rangle \nonumber\\
&=\|\btheta_k\|_2^2+\langle \nabla \hat{L}_k(\ttbeta_k),\btheta_k\rangle.
\end{align}
Given a vector $\bv\in\mR^{d}$ and an index set $\I\subset[d]$, we define $\bv_{\I}\in\mR^d$ to be $\bv$ restricted on $\I$ as $[\bv_{\I}]_i=\bv_i$ if $i\in\I$ and 0 otherwise. Suppose $S_k$ is the support of $\ttbeta_k$, which is the same as $\tbeta_k$, combine (\ref{pequ:3}) and (\ref{pequ:4}) together and we get
\begin{align}\label{pequ:5}
\|\btheta_k\|_2^2&\leq -\langle \nabla \hat{L}_k(\ttbeta_k),\btheta_k\rangle+\lambda_k\|\ttbeta_k\|_1-\lambda_k\|\hbeta_k\|_1 \nonumber\\
&= -\langle \nabla \hat{L}_k(\ttbeta_k),\btheta_k\rangle+\lambda_k\|(\ttbeta_k)_{S_k}\|_1-\lambda_k\|(\hbeta_k)_{S_k}\|_1-\lambda_k\|(\hbeta_k)_{S_k^C}\|_1 \nonumber\\
&\leq -\langle \nabla \hat{L}_k(\ttbeta_k),\btheta_k\rangle+\lambda_k\|(\btheta_k)_{S_k}\|_1-\lambda_k\|(\btheta_k)_{S_k^C}\|_1 \nonumber\\
&\leq \|\nabla \hat{L}_k(\ttbeta_k)\|_{\infty}\|\btheta_k\|_1+\lambda_k\|(\btheta_k)_{S_k}\|_1-\lambda_k\|(\btheta_k)_{S_k^C}\|_1,
\end{align}
where the third inequality is from triangle inequality and the last is based on H\"older's inequality. If we set $\lambda_k=4\Upsilon_{\gamma}\sqrt{\log n/n}$, based on Lemma \ref{lem:warmup:bound_score}, we have
\begin{align}\label{pequ:6}
\|\nabla \hat{L}(\tilde{\bbeta}_k)\|_{\infty}\leq\frac{\lambda_k}{2}
\end{align}
with probability at least $1-d_1/n^2$. Combine (\ref{pequ:5}) and (\ref{pequ:6}), we know with probability at least  $1-d_1/n^2$,
\begin{align}\label{pequ:7}
\|\btheta_k\|_2^2&\leq\frac{3\lambda_k}{2}\|(\btheta_k)_{S_k}\|_1-\frac{\lambda_k}{2}\|(\btheta_k)_{S_k^C}\|_1.
\end{align}
From (\ref{pequ:7}) we further get the following cone condition:
\begin{align}\label{pequ:8}
\|(\btheta_k)_{S_k^C}\|_1\leq 3\|(\btheta_k)_{S_k}\|_1.
\end{align}
Also from equation (\ref{pequ:7}) and sparsity condition we know
\begin{align*}
\|\btheta_k\|_2^2\leq \frac{3}{2}\lambda_k\|(\btheta_k)_{S_k}\|_1\leq\frac{3}{2}\lambda_k\sqrt{s}\|(\btheta_k)_{S_k}\|_2\leq\frac{3}{2}\lambda_k\sqrt{s}\|\btheta_k\|_2.
\end{align*}
So we have with probability at least $1-d_1/n^2$,
\begin{align*}
\|\btheta_k\|_2\leq\frac{3}{2}\sqrt{s}\lambda_k.
\end{align*}
Further by cone condition in (\ref{pequ:8}) we can get $l_1$-norm convergence rate as
\begin{align*}
\|\btheta_k\|_1=\|(\btheta_k)_{S_k}\|_1+\|(\btheta_k)_{S_k^C}\|_1\leq4\|(\btheta_k)_{S_k}\|_1\leq4\sqrt{s}\|(\btheta_k)_{S_k}\|_2\leq6s\lambda_k.
\end{align*}
By taking the union bound, it's easy to have
\begin{align*}
P(\|\btheta_k\|_2\leq \frac{3}{2}\sqrt{s}\lambda_k \text{\ and\ } \|\btheta_k\|_2\leq 6s\lambda_k, \forall k)\geq 1-\frac{d_2d_1}{n^2}.
\end{align*}

\subsection{Proof of Corollary \ref{cor:1}}

We still fix $k\in[d_2]$ first and then take union bound. Denote $\mu = \min_{j\in[d_2]}|\mu_j|$, from Theorem \ref{thm:conv_warmup}, we know there exists $N(s,\mu)$ such that whenever $n\geq N$, we have
\begin{equation}\label{pequ:9}
\|\hbeta_k\|_2\geq \mu-\|\hbeta_k - \ttbeta_k\|_2\geq \mu - 6\Upsilon\sqrt{s \log n/n}\geq \mu/2.
\end{equation}
with probability at least $1-d_1/n^2$. For either $l_2$-norm or $l_1$-norm, use (\ref{pequ:9}) and we can get
\begin{align}\label{pequ:10}
\|\frac{\hat{\bbeta}_k}{\|\hat{\bbeta}_k\|_2}-\frac{\ttbeta_k}{|\mu_k|}\| =&\frac{\|\hat{\bbeta}_k-\|\hat{\bbeta}_k\|_2/|\mu_k|\cdot\ttbeta_k\|}{\|\hat{\bbeta}_k\|_2}\leq \frac{\|\hat{\bbeta}_k-\tilde{\bbeta}_k\| + \big||\mu_k| - \|\hbeta_k\|_2\big|\|\tbeta_k\|}{\|\hat{\bbeta}_k\|_2} \nonumber\\
\leq &\frac{2}{\mu}\|\hat{\bbeta}_k-\tilde{\bbeta}_k\| + \frac{2}{\mu}\|\tbeta_k\|\cdot\big|\|\tilde{\bbeta}_k\|_2 - \|\hbeta_k\|_2\big| \nonumber\\
\leq & \frac{2}{\mu}\|\hat{\bbeta}_k-\tilde{\bbeta}_k\| + \frac{2}{\mu}\|\tbeta_k\|\cdot\|\hbeta_k-\ttbeta_k \|_2.
\end{align}
Thus, combine (\ref{pequ:10}) with Theorem \ref{thm:conv_warmup}, we know with probability $1-d_1/n^2$
\begin{align*}
\|\frac{\hat{\bbeta}_k}{\|\hat{\bbeta}_k\|_2}-\frac{\ttbeta_k}{|\mu_k|}\|_2\leq& \frac{4}{\mu}\|\hat{\bbeta}_k-\tilde{\bbeta}_k\|_2\lesssim \sqrt{s\log n/n}/\mu,\\
\|\frac{\hat{\bbeta}_k}{\|\hat{\bbeta}_k\|_2}-\frac{\ttbeta_k}{|\mu_k|}\|_1\leq& \frac{2}{\mu}\|\hat{\bbeta}_k-\tilde{\bbeta}_k\|_1 + \frac{2\sqrt{s}}{\mu}\|\hat{\bbeta}_k - \ttbeta_k\|_2\lesssim s\sqrt{\log n/n}/\mu.
\end{align*}
Note that under identifiability condition (\ref{eq:identification}) we have $\tbeta_k = \text{sign}(\tilde{\beta}_{k1})\cdot\frac{\ttbeta_k}{|\mu_k|}$, hence there exists $M(s, N, \min_{j\in[d_2]}\beta_{j1}^\star)$, such that $n\geq M$, $\text{sign}(\hat{\beta}_{k1}) = \text{sign}(\tilde{\beta}_{k1})$. So we can get for either $l_2$-norm or $l_1$-norm
\begin{align*}
\|\frac{\hat{\bbeta}_k}{\|\hat{\bbeta}_k\|_2}-\frac{\ttbeta_k}{|\mu_k|}\| =
\|\text{sign}(\hat{\beta}_{k1})\frac{\hat{\bbeta}_k}{\|\hat{\bbeta}_k\|_2}-\text{sign}(\hat{\beta}_{k1})\frac{\ttbeta_k}{|\mu_k|}\| = \|\text{sign}(\hat{\beta}_{k1})\frac{\hat{\bbeta}_k}{\|\hat{\bbeta}_k\|_2}-\tbeta_k\|.
\end{align*}
By taking the union bound, we can get the conclusion. Particularly, in the worst case, we have
\begin{align*}
\|\hB-\tB\|_F^2 = \sum_{j=1}^{d_2}\|\text{sign}(\hat{\beta}_{k1})\frac{\hat{\bbeta}_k}{\|\hat{\bbeta}_k\|_2}-\tbeta_k\|_2^2\lesssim d_2s\log n/n\mu^2.
\end{align*}
So, we know $\|\hB-\tB\|_F\lesssim \frac{1}{\mu}\sqrt{\frac{d_2s\log n}{n}}$. This concludes the proof.

\subsection{Proof of Theorem \ref{thm:4}}

We still fix $k\in[d_2]$ first. Start from the definition of $\hbeta_k$ in (\ref{equ:6}) and basic inequality, then follow the same steps as in (\ref{pequ:3}), (\ref{pequ:4}), and (\ref{pequ:5}), we finally get
\begin{align}\label{pequ:17}
\|\btheta_k\|_2^2\leq \|\nabla \bar{L}_k(\tilde{\bbeta}_k)\|_{\infty}\|\btheta_k\|_1+\lambda_k\|(\btheta_k)_{\S_k}\|_1-\lambda_k\|(\btheta_k)_{\S_k^C}\|_1.
\end{align}
Based on Lemma \ref{lem:6}, we can let $\lambda_k = 76\sqrt{M\log d_1d_2/n}$ and have $\|\nabla \bar{L}_k(\tilde{\bbeta}_k)\|_{\infty}\leq \lambda_k/2$. Plug into (\ref{pequ:17}) and follow the derivation in equation (\ref{pequ:7}) and (\ref{pequ:8}), we can finally get
\begin{align*}
\|\hat{\bbeta}_k-\tilde{\bbeta}_k\|_2\leq \frac{3}{2}\sqrt{s}\lambda_k \text{\ \ and\ \ }\|\hat{\bbeta}_k-\tilde{\bbeta}_k\|_1\leq 6s\lambda_k.
\end{align*}
Note that above error bound holds uniformly over $k\in[d_2]$ and we finish the proof.

\subsection{Proof of Theorem \ref{thm:5}}

To make notation consistent, let's denote the loss function without penalty defined in (\ref{equ:12}) by $\hL(\bB)$ and define $\I_4 = \hOmega - \tOmega$, then we have
\begin{align}\label{pequ:23}
\nabla\hat{L}(\tilde{\bB})=&2\tilde{\bB}-\frac{2}{ n\kappa_1}\sum_{i=1}^{n}\Phi(\kappa_1 y_i\cdot S(X_i)Z_i^T)\hOmega\nonumber\\
\stackrel{(\ref{equ:7})}{=} &2\bigg(\mE[y\cdot S(\bx)\bz^T]\tOmega-\frac{1}{n\kappa_1}\sum_{i=1}^{ n}\Phi(\kappa_1 y_i\cdot S(X_i)Z_i^T)\hOmega\bigg).
\end{align}
Based on equation (\ref{pequ:23}), we use triangle inequality and get
\begin{align}\label{pequ:24}
&\|\nabla\hat{L}(\tilde{\bB})\|_2 \\
& \leq 2\|\underbrace{\mE[y\cdot S(\bx)\bz^T] - \frac{1}{n\kappa_1}\sum_{i=1}^{ n}\Phi(\kappa_1 y_i\cdot S(X_i)Z_i^T)}_{\I_3}\|_2\|\hOmega\|_2 + 2\|\mE[y\cdot S(\bx)\bz^T]\|_2\|\underbrace{\hOmega - \tOmega}_{\I_4}\|_2 \nonumber\\
& \leq 2\|\I_3\|_2\|\I_4\|_2 + \underbrace{2\|\tOmega\|_2\|\I_3\|_2 + 2\|\mE[y\cdot S(\bx)\bz^T]\|_2\|\I_4\|_2}_{\text{dominant term}}.
\end{align}
Note that
\begin{align}\label{pequ:42}
\|\mE[y\cdot S(\bx)\bz^T]\|_2 =  \|\ttB\tSigma\|_2\leq \max_{j\in[d_2]} |\mu_j|\cdot\|\tB\|_2\|\tSigma\|_2.
\end{align}
So combine (\ref{pequ:41}), (\ref{pequ:24}), (\ref{pequ:42}) together and drop off smaller order term, we can get
\begin{align}\label{equ:new}
P\bigg(\|\nabla\hL(\ttB)\|_2\leq & 8KM^{3/4}\sqrt{\frac{(d_1+d_2)\log(d_1+d_2)}{n}} + 2K\max_{j\in[d_2]} |\mu_j|\cdot\|\tB\|_2\H(n,d_2)\bigg) \nonumber\\
&\geq 1 - \frac{2}{(d_1+d_2)^2} - \P(n,d_2).
\end{align}
So we know under the setup of $\lambda$ as in theorem, we have $\|\nabla\hat{L}(\tilde{\bB})\|_2\leq \lambda/2$ with probability at least $1 - 2/(d_1+d_2)^2 - \P(n,d_2)$. On the other side, start from the definition of $\hB$ in (\ref{equ:12}), we have following basic inequality
\begin{align}\label{pequ:44}
\hat{L}(\hat{\bB})+\lambda\|\hat{\bB}\|_*\leq\hat{L}(\tilde{\bB})+\lambda\|\tilde{\bB}\|_*.
\end{align}
Define $\bTheta=\hat{\bB}-\tilde{\bB}$, we have
\begin{align}\label{pequ:45}
\hat{L}(\hat{\bB})-\hat{L}(\tilde{\bB})&=\|\hat{\bB}\|_F^2-\|\tilde{\bB}\|_F^2-\frac{2}{n\kappa_1}\sum_{i=1}^{n}\langle \Phi(\kappa_1 y_i\cdot S(\bX_i)\bZ_i^T)\hOmega, \hat{\bB}-\tilde{\bB}\rangle \nonumber\\
&=\|\bTheta\|_F^2+2\langle\tilde{\bB},\bTheta\rangle-\frac{2}{n\kappa_1}\sum_{i=1}^{n}\langle\Phi(\kappa_1 y_i\cdot S(\bX_i)\bZ_i^T)\hOmega, \bTheta\rangle \nonumber\\
&=\langle\nabla\hat{L}(\tilde{\bB}),\bTheta\rangle+\|\bTheta\|_F^2.
\end{align}
Combine (\ref{pequ:44}) and (\ref{pequ:45}) together, we have
\begin{align}\label{pequ:46}
\|\bTheta\|_F^2=&-\langle\nabla\hat{L}(\tilde{\bB}),\bTheta\rangle+\hat{L}(\hat{\bB})-\hat{L}(\tilde{\bB}) \nonumber\\
\leq&-\langle\nabla\hat{L}(\tilde{\bB}),\bTheta\rangle+\lambda\|\tilde{\bB}\|_*-\lambda\|\hat{\bB}\|_* \nonumber\\
\leq&\|\nabla\hat{L}(\tilde{\bB})\|_2\|\bTheta\|_*+\lambda\|\tilde{\bB}\|_*-\lambda\|\hat{\bB}\|_*.
\end{align}
Under Assumption \ref{ass:1} and \ref{ass:6}, we know $r = \text{rank}(\tB) = \text{rank}(\tilde{\bB})$. We let $\tilde{\bB}=\bU\bLambda \bV^T$ be its singular value decomposition where diagonal matrix $\bLambda\in\mR^{d_1\times d_2}$ can be expressed as $\begin{pmatrix}
{\bLambda}_{11} & 0\\
0 & 0
\end{pmatrix}$ for ${\bLambda}_{11}\in\mR^{r\times r}$. We define
\begin{align*}
\bT = \bU^T\bTheta \bV = \bT^{(1)}+\bT^{(2)}
\end{align*}
where $\bT^{(1)}=\begin{pmatrix}
0 & 0\\
0 & \bT_{22}
\end{pmatrix}$ and $\bT^{(2)}=\begin{pmatrix}
\bT_{11} & \bT_{12}\\
\bT_{21} & 0
\end{pmatrix}$ have the same corresponding block size as ${\bLambda}$. Then we get
\begin{align}\label{pequ:47}
\|\hat{\bB}\|_*=&\|\tilde{\bB}+\bTheta\|_*=\|\bU({\bLambda}+\bT)\bV^T\|_*=\|\bLambda+\bT\|_* \nonumber\\
\geq&\|{\bLambda}+\bT^{(1)}\|_*-\|\bT^{(2)}\|_*=\|\tilde{\bB}\|_*+\|\bT^{(1)}\|_*-\|\bT^{(2)}\|_*.
\end{align}
The last equality is because of the block diagonal structure of ${\bLambda}$ and $\bT^{(1)}$ and $\|\tilde{\bB}\|_*=\|{\bLambda}\|_*$. Combine (\ref{pequ:47}) and (\ref{pequ:46}), we have
\begin{align}\label{pequ:48}
\|\bTheta\|_F^2\leq\frac{3\lambda}{2}\|\bT^{(2)}\|_*-\frac{\lambda}{2}\|\bT^{(1)}\|_*.
\end{align}
According to (\ref{pequ:48}), we have following cone condition
\begin{align}\label{pequ:49}
\|\bT^{(1)}\|_*\leq 3\|\bT^{(2)}\|_*.
\end{align}
Also form (\ref{pequ:48}) and using Assumption \ref{ass:6}, we get
\begin{align*}
\|\bTheta\|_F^2\leq\frac{3\lambda}{2}\|\bT^{(2)}\|_*\leq \frac{3\lambda}{2}\sqrt{\text{rank}(\bT^{(2)})}\|\bT^{(2)}\|_F\leq3\lambda\sqrt{r}\|\bT^{(2)}\|_F\leq3\lambda\sqrt{r}\|\bTheta\|_F.
\end{align*}
Combining with (\ref{pequ:49}), we know with probability at least $1 - 2/(d_1+d_2)^2 - \P(n,d_2)$,
\begin{align*}
\|\bTheta\|_F\leq &3\sqrt{r}\lambda,\\
\|\bTheta\|_*\leq &\|\bT^{(1)}\|_*+\|\bT^{(2)}\|_*\leq 4\|\bT^{(2)}\|_*\leq 24 r\lambda.
\end{align*}
This concludes the theorem.

\subsection{Proof of Theorem \ref{thm:8}}

Let's denote the loss function without penalty defined in (\ref{equ:13}) by $\hL(\bB)$ and let $\I_4 = \hOmega - \tOmega$. We know
\begin{align}\label{pequ:60}
\nabla\hat{L}(\tilde{\bB})=&2\tilde{\bB}-\frac{2}{ n}\sum_{i=1}^{n}\wy_i\cdot\wSXi\wZi^T \hOmega \stackrel{(\ref{equ:7})}{=}2\mE[y\cdot S(\bx)\bz^T]\tOmega - \frac{2}{ n}\sum_{i=1}^{n}\wy_i\cdot\wSXi\wZi^T \hOmega \nonumber\\
= &2\bigg(\mE[y\cdot S(\bx)\bz^T](\tOmega - \hOmega) + (\mE[y\cdot S(\bx)\bz^T] -  \frac{2}{ n}\sum_{i=1}^{n}\wy_i\cdot\wSXi\wZi^T )\hOmega\bigg).
\end{align}
From (\ref{pequ:60}), we have
\begin{align*}
\|\nabla\hat{L}(\tilde{\bB})\|_{\max}\leq 2\|\I_7 \|_{\max}\|\I_4\|_1 + \underbrace{2\|\I_7\|_{\max}\|\tOmega\|_1 + 2\|\I_4\|_{\max}\|\mE[y\cdot S(\bx)\bz^T]\|_{\infty}}_{\text{dominant term}}.
\end{align*}
Note that
\begin{align}\label{pequ:64}
\|\mE[y\cdot S(\bx)\bz^T]\|_{\infty} = \|\ttB\tSigma\|_{\infty}\leq \max_{j\in[d_2]}|\mu_j|\cdot \|\tB\tSigma\|_{\infty}.
\end{align}
Combine (\ref{pequ:64}) with (\ref{pequ:57}) and drop off the intersection term (smaller order), we know
\begin{multline*}
P\bigg(\|\nabla\hL(\ttB)\|_{\max}> 38\|\tOmega\|_1\sqrt{\frac{M\log d_1d_2}{n}} + 2\max_{j\in[d_2]}|\mu_j|\cdot \|\tB\tSigma\|_{\infty}\tH(n,d_2)\bigg) \\
\leq \frac{2}{d_1^2d_2^2}+\tP(n,d_2).
\end{multline*}
So under the setup in theorem we have $\|\nabla\hat{L}(\tilde{\bB})\|_{\max}\leq \lambda/2$. On the other side, based on definition of $\hB$ in (\ref{equ:13}), we have following basic inequality
\begin{align}\label{pequ:51}
\hat{L}(\hat{\bB})+\lambda\|\hat{\bB}\|_{1,1}\leq\hat{L}(\tilde{\bB})+\lambda\|\tilde{\bB}\|_{1,1}.
\end{align}
Define $\bTheta=\hat{\bB}-\tilde{\bB}$, same with (\ref{pequ:45}) we have
\begin{align}\label{pequ:52}
\hat{L}(\hat{\bB})-\hat{L}(\tilde{\bB})=\langle\nabla\hat{L}(\tilde{\bB}),\bTheta\rangle+\|\bTheta\|_F^2.
\end{align}
Combine (\ref{pequ:51}) and (\ref{pequ:52}), and define $S = \text{supp}(\ttB)$, we have
\begin{align}\label{pequ:67}
\|\bTheta\|_F^2\leq& -\LD \nabla\hat{L}(\tilde{\bB}),\bTheta\RD + \lambda\|\tilde{\bB}\|_{1,1} - \lambda\|\hat{\bB}\|_{1,1} \nonumber\\
\leq &\|\nabla\hat{L}(\tilde{\bB})\|_{\max}\|\bTheta\|_{1,1} + \lambda\|\tilde{\bB}_S\|_{1,1} - \lambda\|\hat{\bB}_S\|_{1,1}-\lambda\|\hat{\bB}_{S^C}\|_{1,1} \nonumber\\
\leq& \|\nabla\hat{L}(\tilde{\bB})\|_{\max}\|\bTheta\|_{1,1} + \lambda\|\bTheta_S\|_{1,1} -\lambda\|\bTheta_{S^C}\|_{1,1}.
\end{align}
So, based on (\ref{pequ:67}), we know, with probability at least $1-2/d_1d_2 - \tP(n,d_2)$,
\begin{align}\label{pequ:68}
\|\bTheta\|_F^2\leq \frac{3\lambda}{2}\|\bTheta_S\|_{1,1}\leq \frac{3\lambda}{2}\sqrt{sd_2}\|\bTheta_S\|_F\Longrightarrow \|\bTheta\|_F\leq 2\sqrt{sd_2}\lambda.
\end{align}
Similarly, we have
\begin{align*}
\|\bTheta\|_{1,1}\leq 4\|\bTheta_S\|_{1,1}\leq 8sd_2\lambda.
\end{align*}
This concludes the proof.

\section{Additional simulation results and plots}\label{supple4}





\begin{figure}[p]
	\centering     
	\subfigure[$t_{13}$ for $\tbeta_1$]{\label{t131}\includegraphics[width=49.9mm]{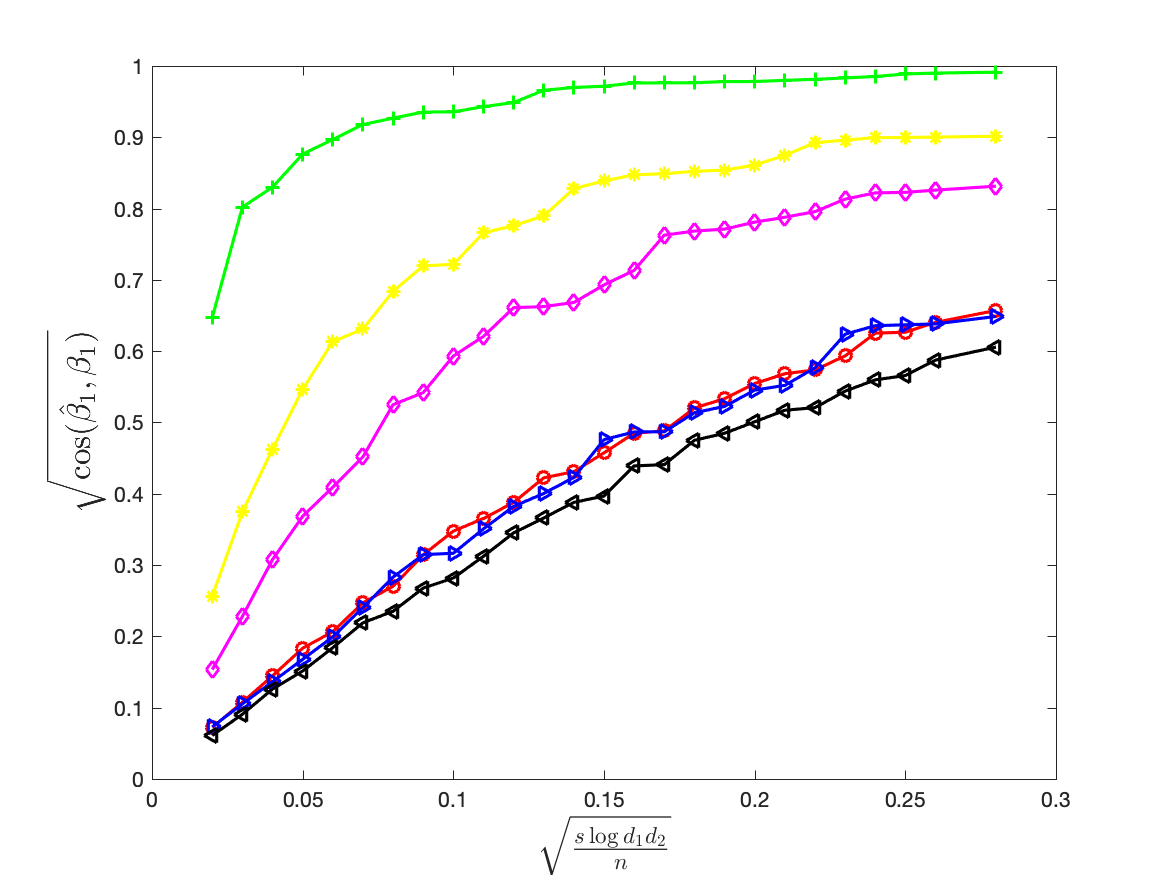}}
	\subfigure[$t_{13}$ for $\tbeta_{d_2/2}$]{\label{t132}\includegraphics[width=49.9mm]{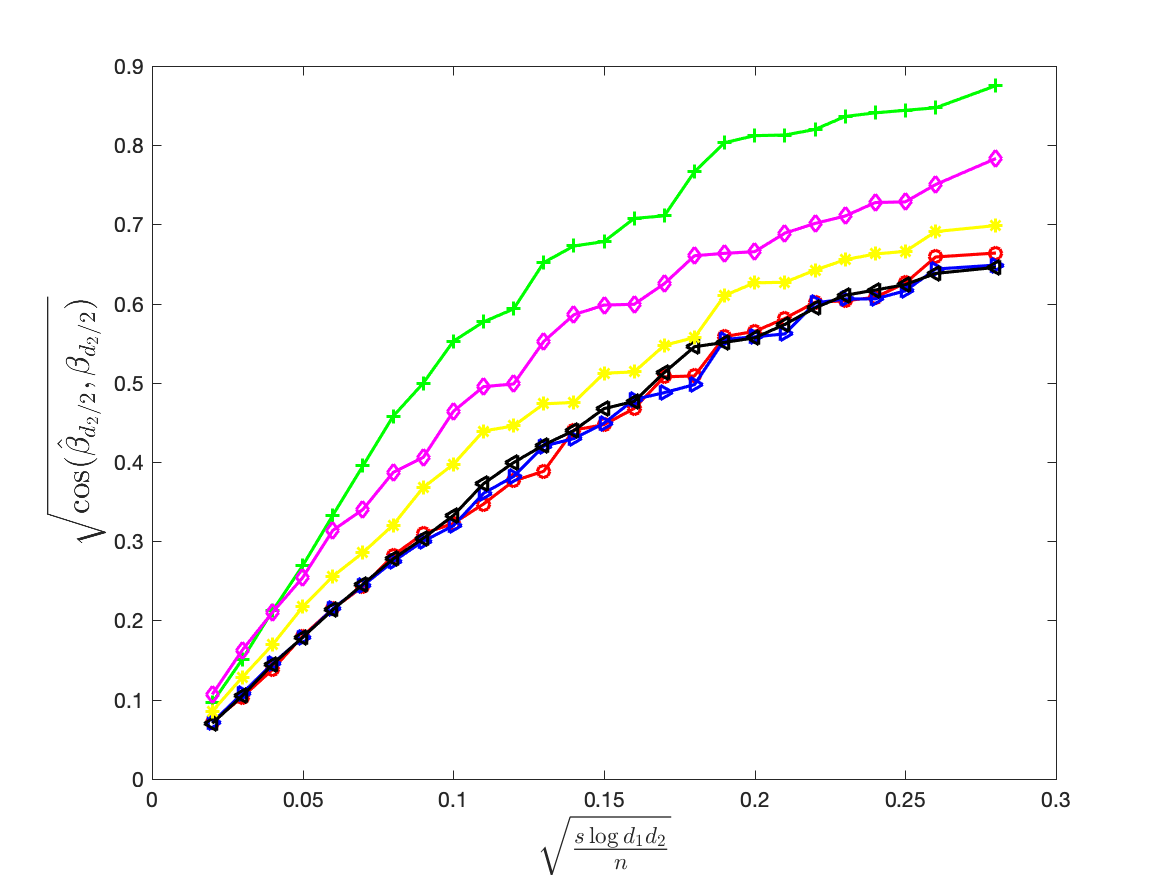}}
	\subfigure[$t_{13}$ for $\tbeta_{d_2}$]{\label{t133}\includegraphics[width=49.9mm]{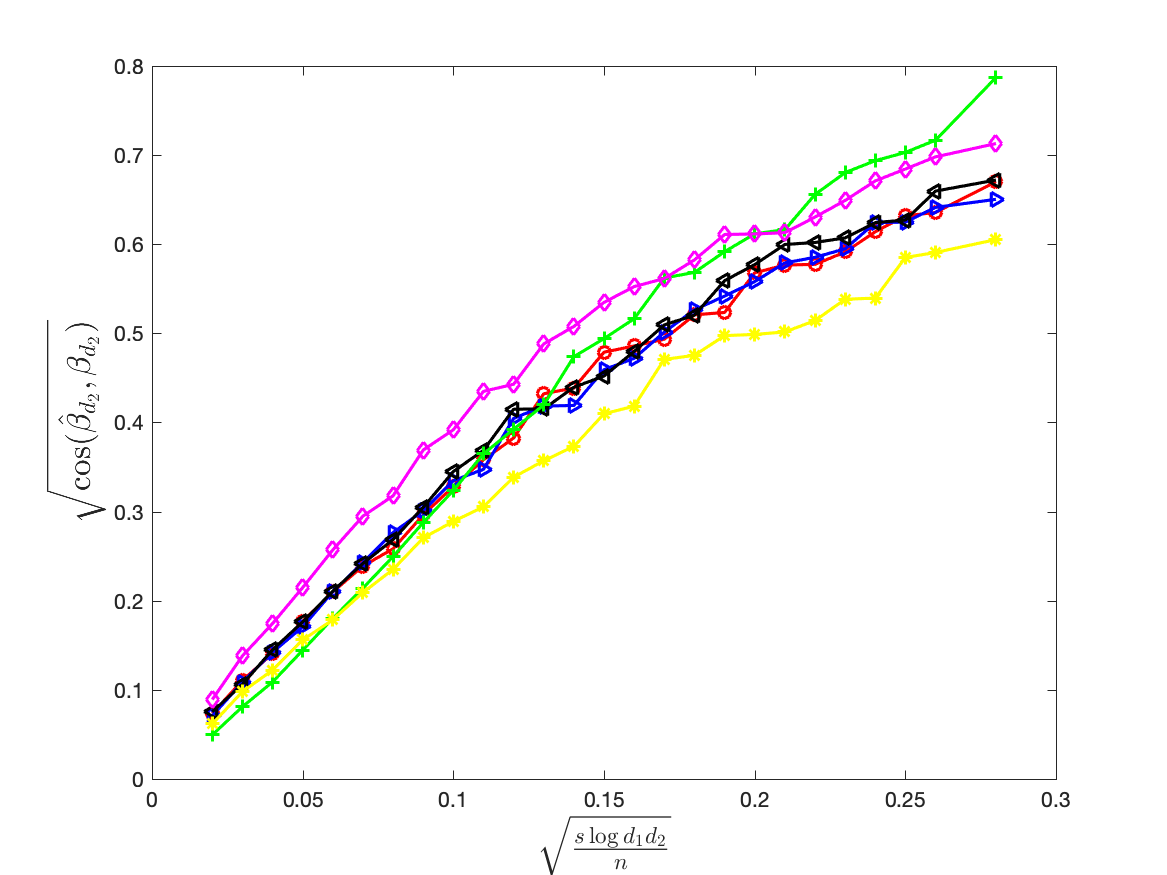}}
	\subfigure[Rayleigh for $\tbeta_1$]{\label{ray1}\includegraphics[width=49.9mm]{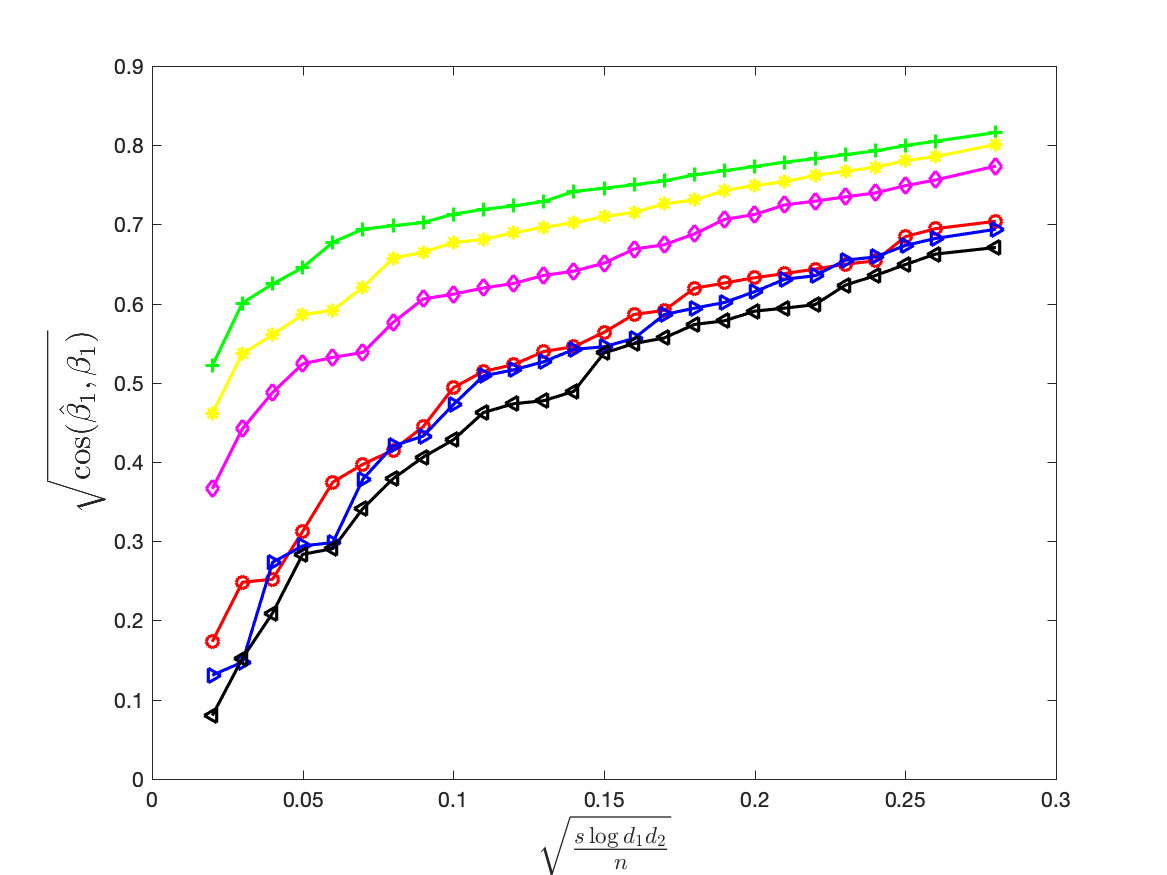}}
	\subfigure[Rayleigh for $\tbeta_{d_2/2}$]{\label{ray2}\includegraphics[width=49.9mm]{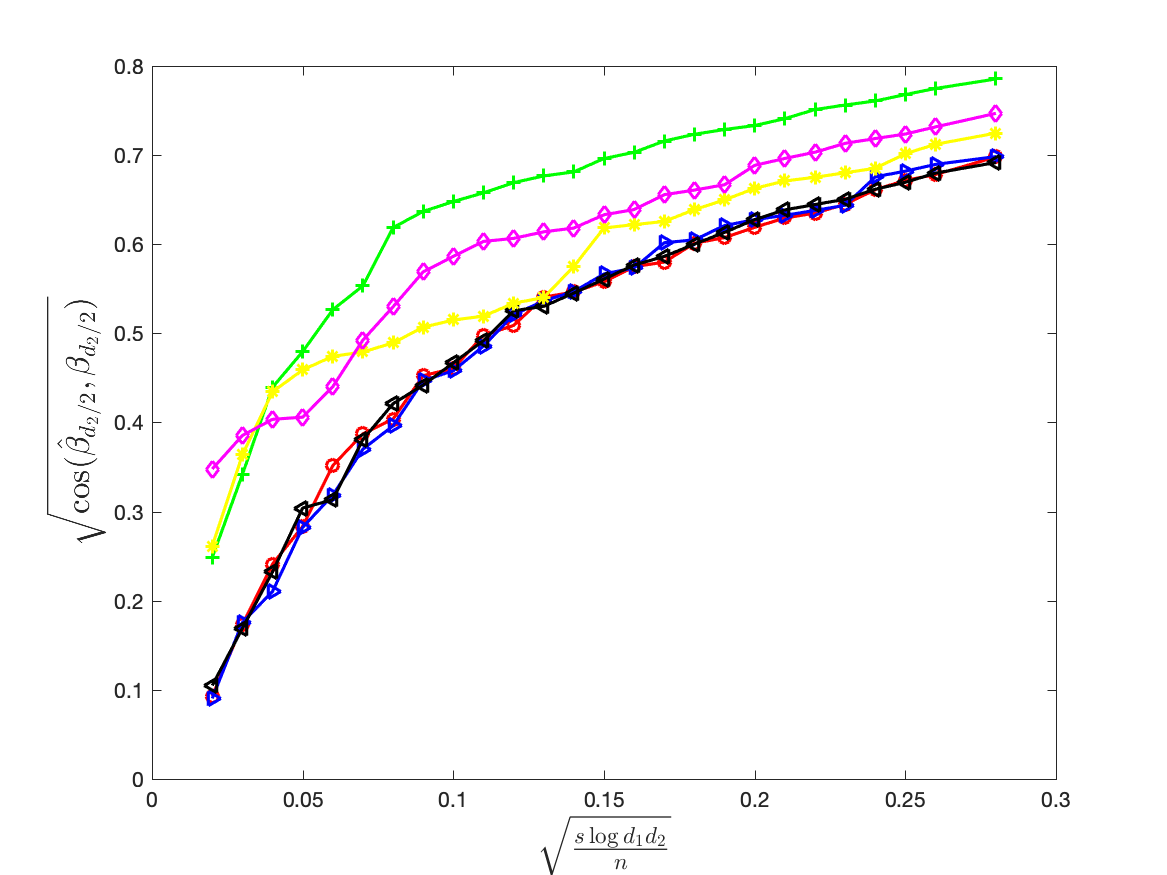}}
	\subfigure[Rayleigh for $\tbeta_{d_2}$]{\label{ray3}\includegraphics[width=49.9mm]{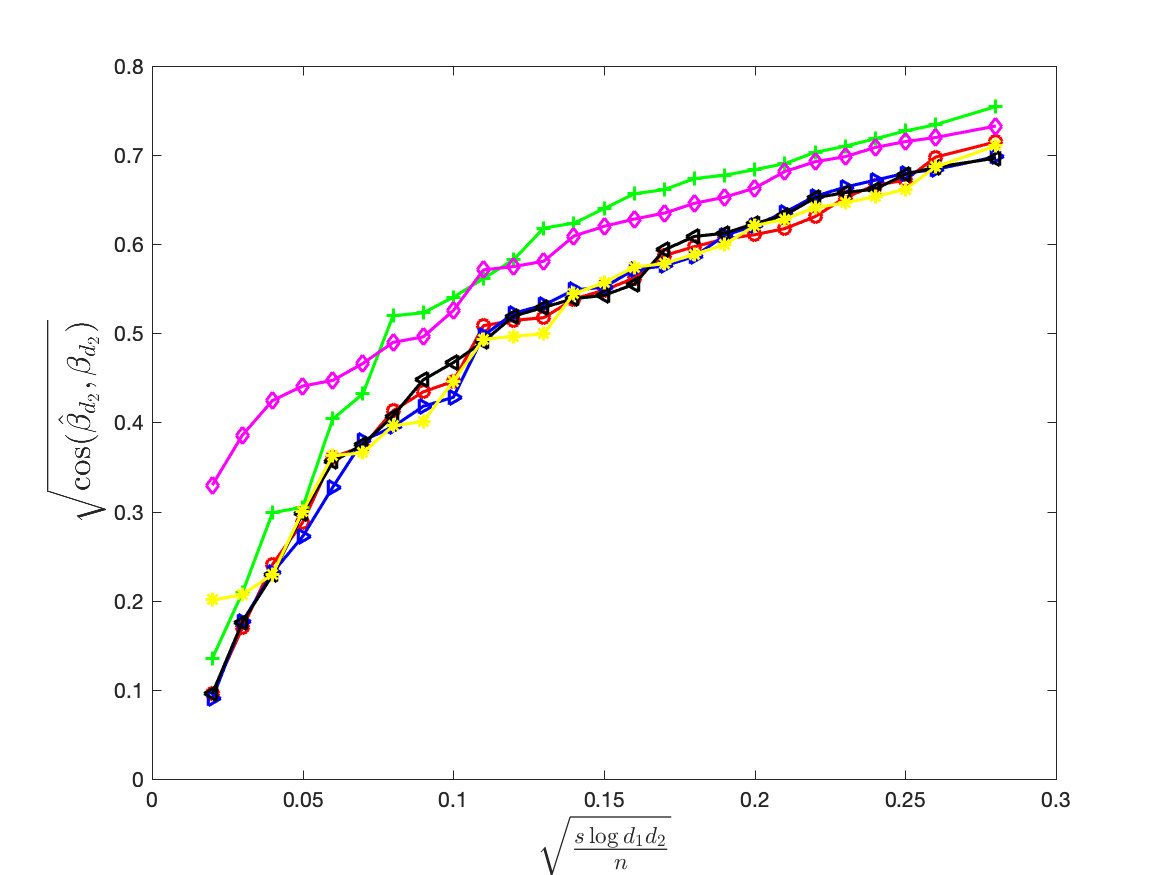}}
	\subfigure[Weibull for $\tbeta_1$]{\label{wei1}\includegraphics[width=49.9mm]{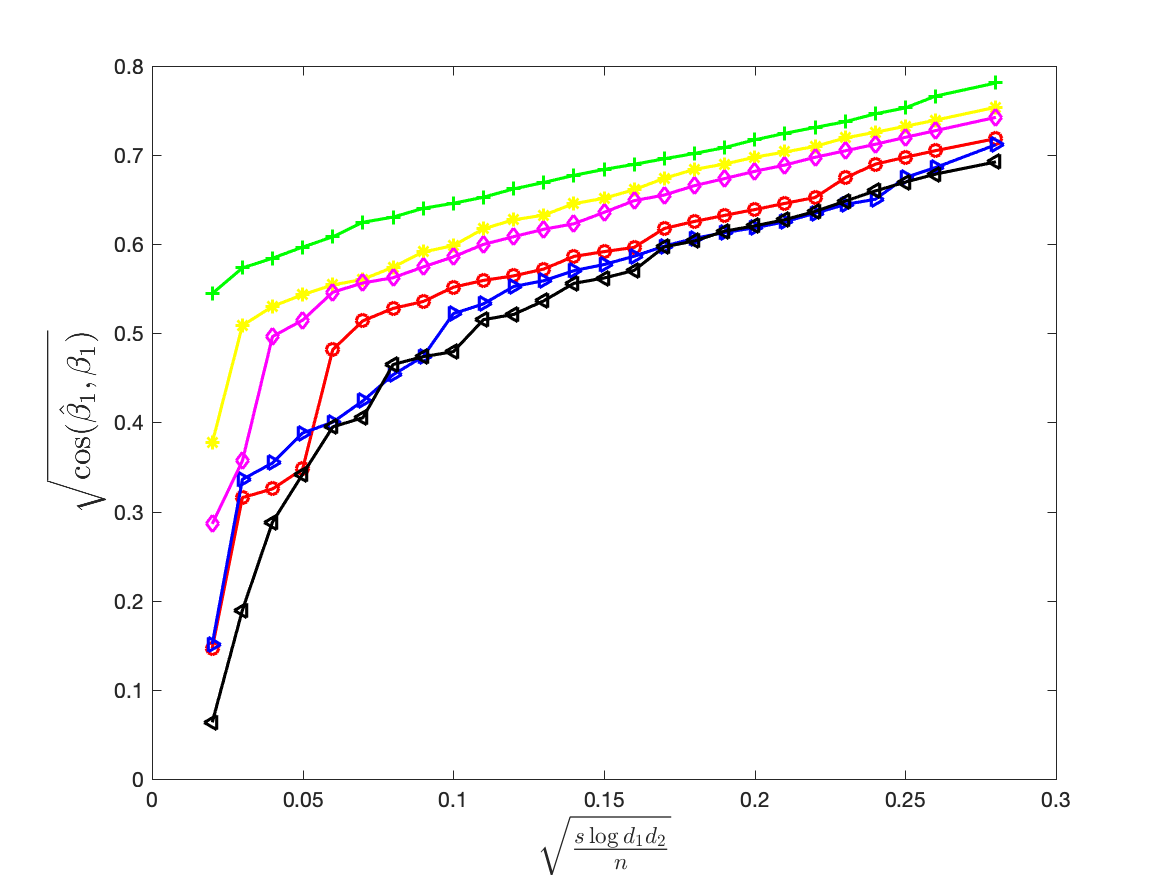}}
	\subfigure[Weibull for $\tbeta_{d_2/2}$]{\label{wei2}\includegraphics[width=49.9mm]{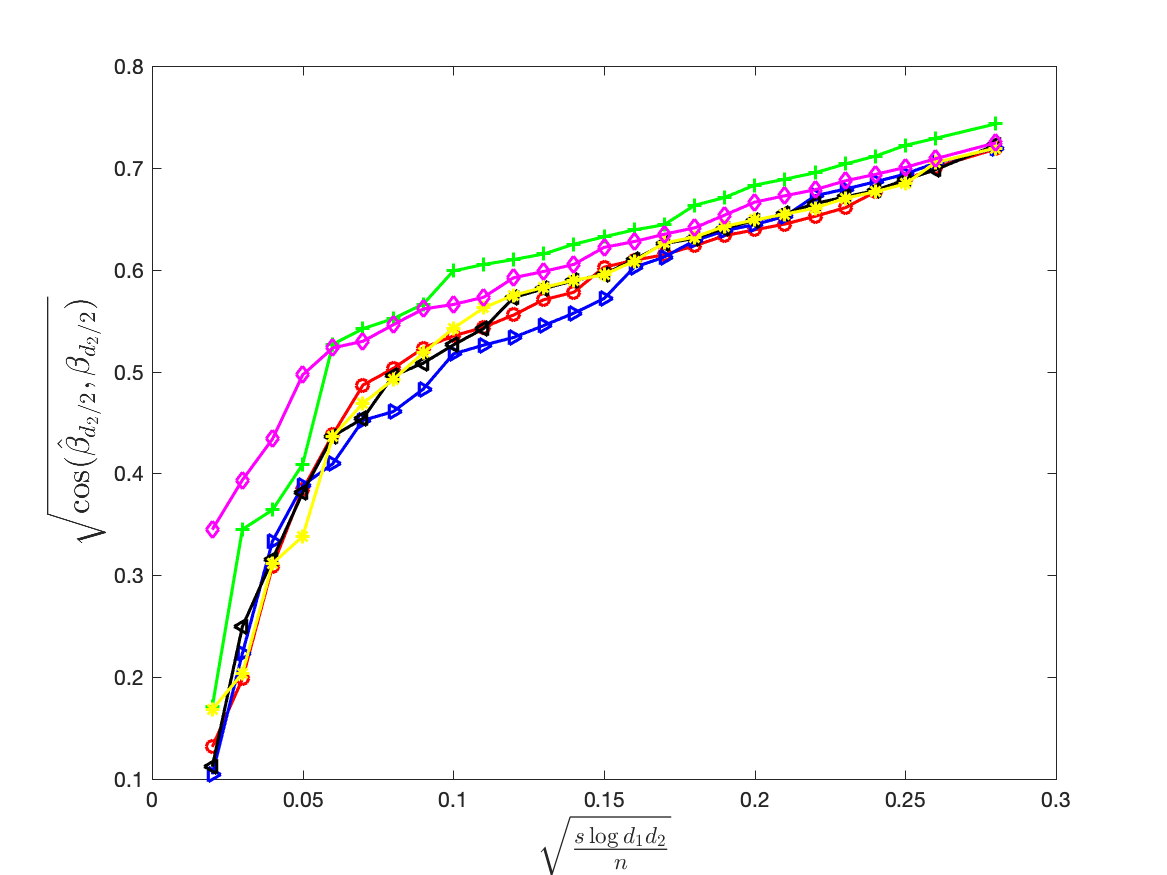}}
	\subfigure[Weibull for $\tbeta_{d_2}$]{\label{wei3}\includegraphics[width=49.9mm]{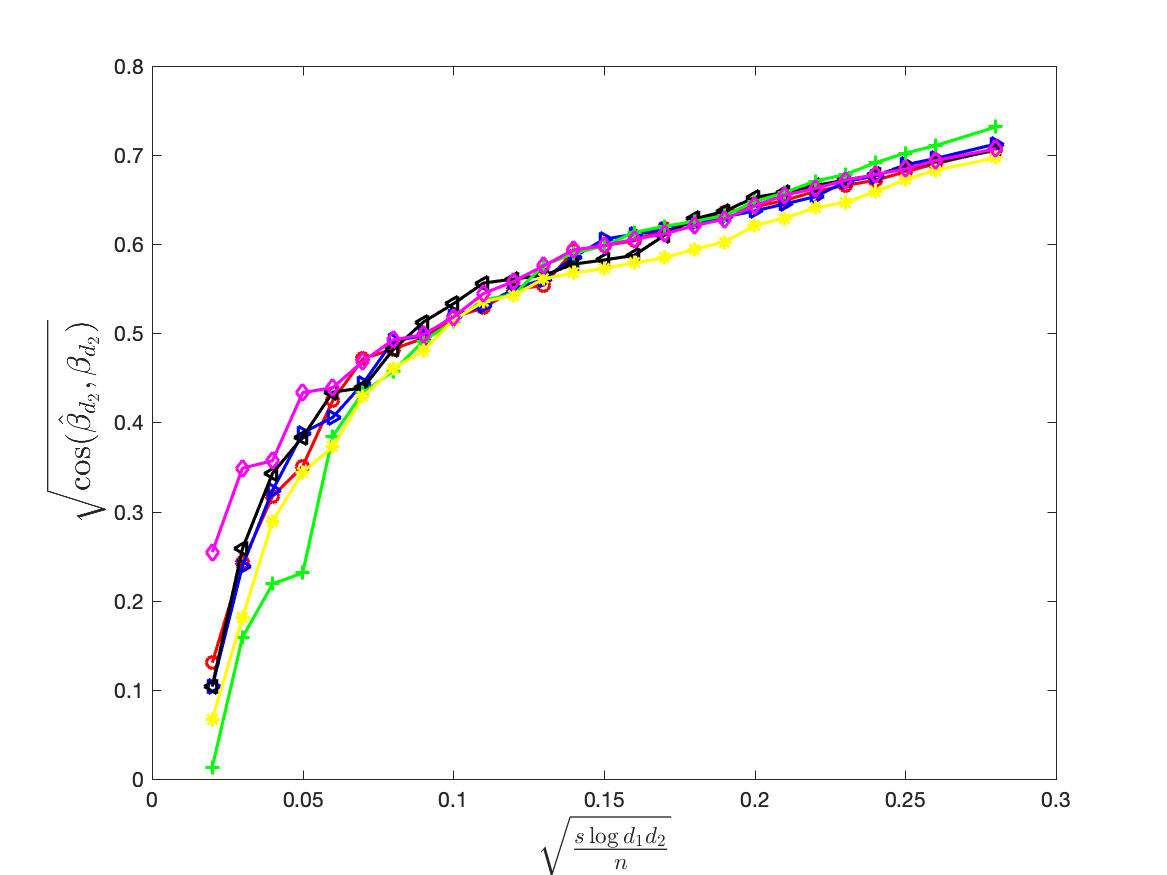}}
	\includegraphics[width=7.5cm,height=0.36cm]{Figure/funcleg.png}
	\caption{{Sparse vector estimation plot (II). This figure shows cosine distance trend for error of estimating single sparse parameter in model (\ref{mod:1}). Six lines indicates six different types of link functions. Each row represents one type of distribution for design. We only plot the error trend for the first, the middle, the last parameter. All above simulation results are consistent with Theorem \ref{thm:4}.}}\label{fig:3}
\end{figure}

\begin{figure}[p]
	\centering     
	\subfigure[Gaussian for $f^{(1)}$]{\label{fw1}\includegraphics[width=49.9mm]{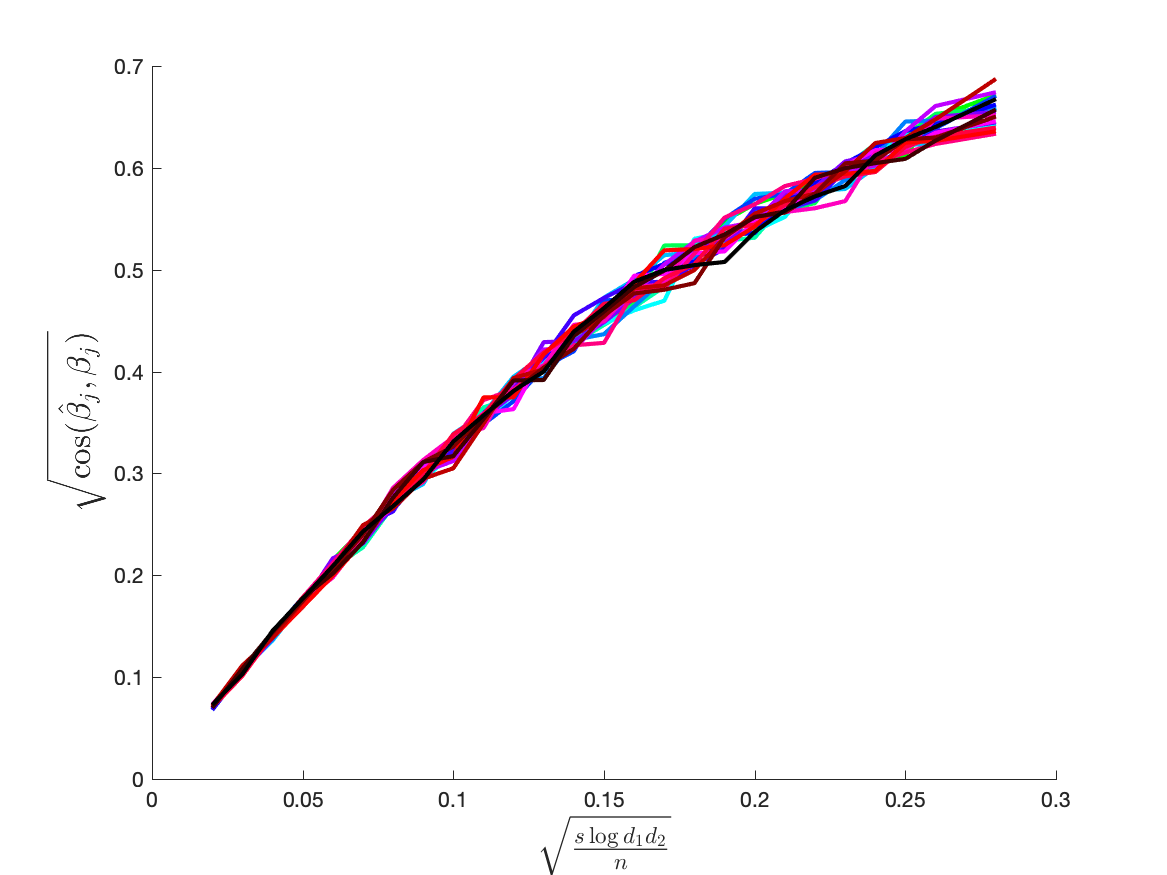}}
	\subfigure[Gaussian for $f^{(2)}$]{\label{fw2}\includegraphics[width=49.9mm]{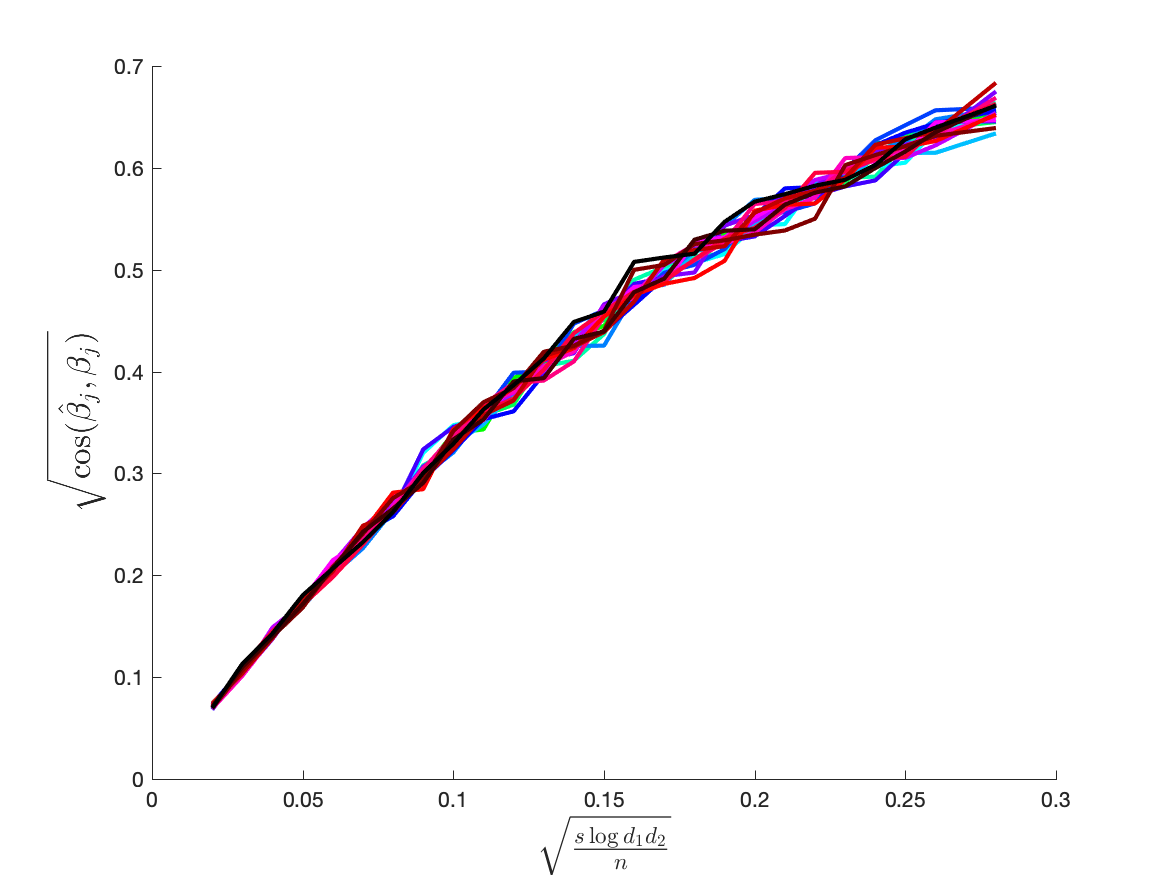}}
	\subfigure[Gaussian for $f^{(3)}$]{\label{fw3}\includegraphics[width=49.9mm]{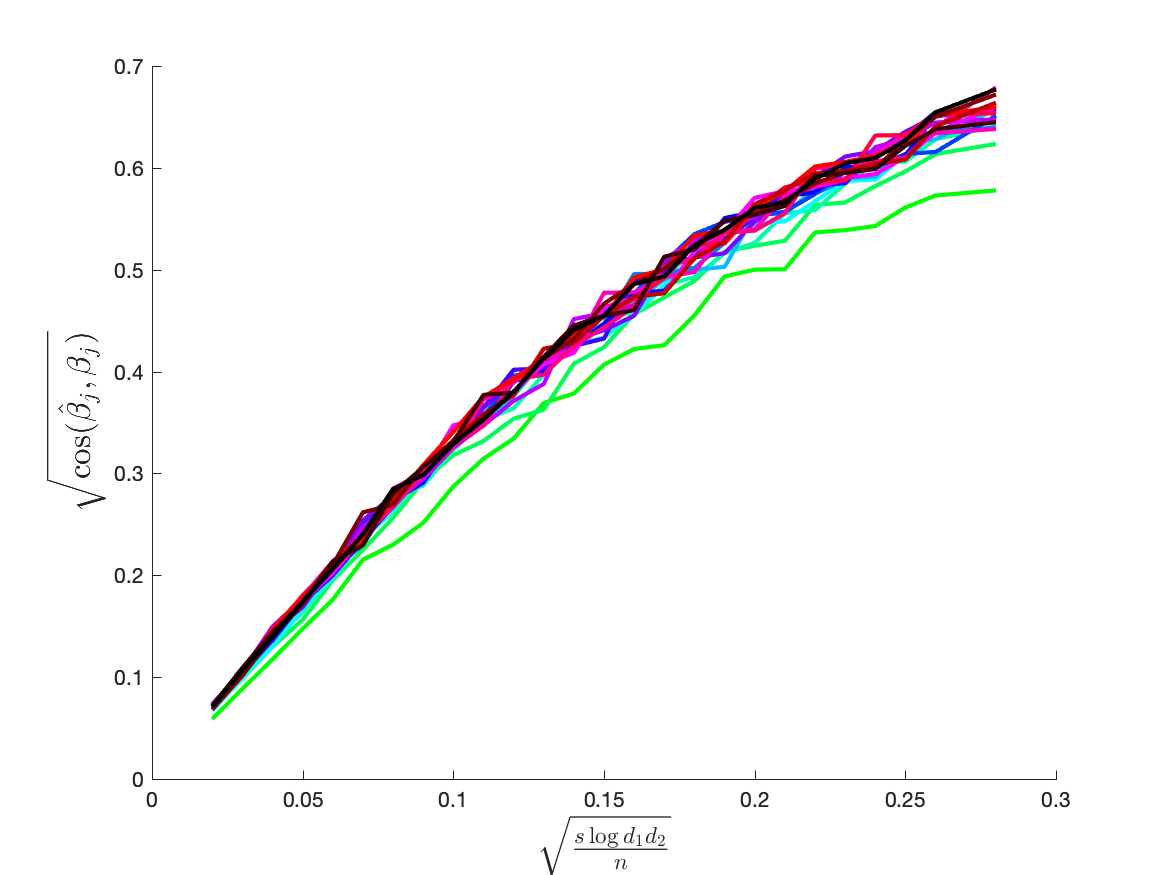}}
	\subfigure[Gaussian for $f^{(4)}$]{\label{fw4}\includegraphics[width=49.9mm]{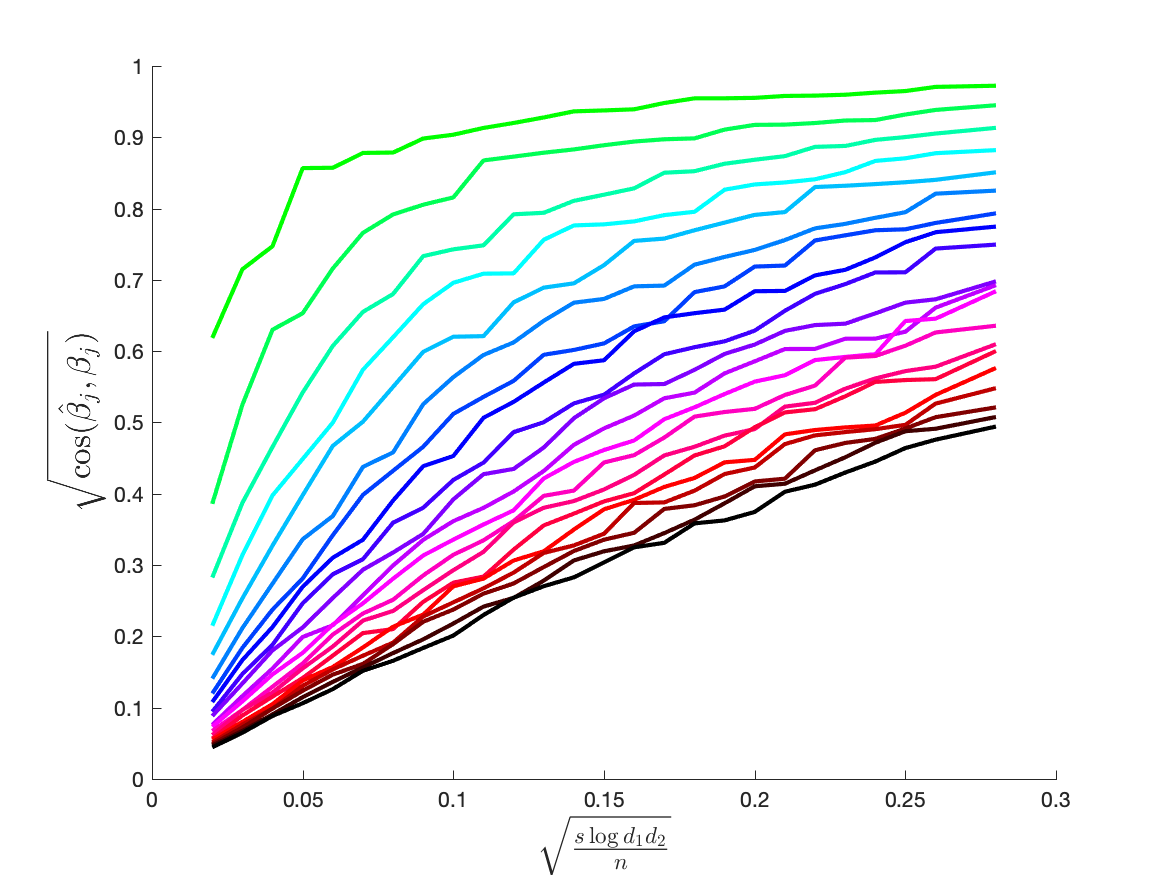}}
	\subfigure[Gaussian for $f^{(5)}$]{\label{fw5}\includegraphics[width=49.9mm]{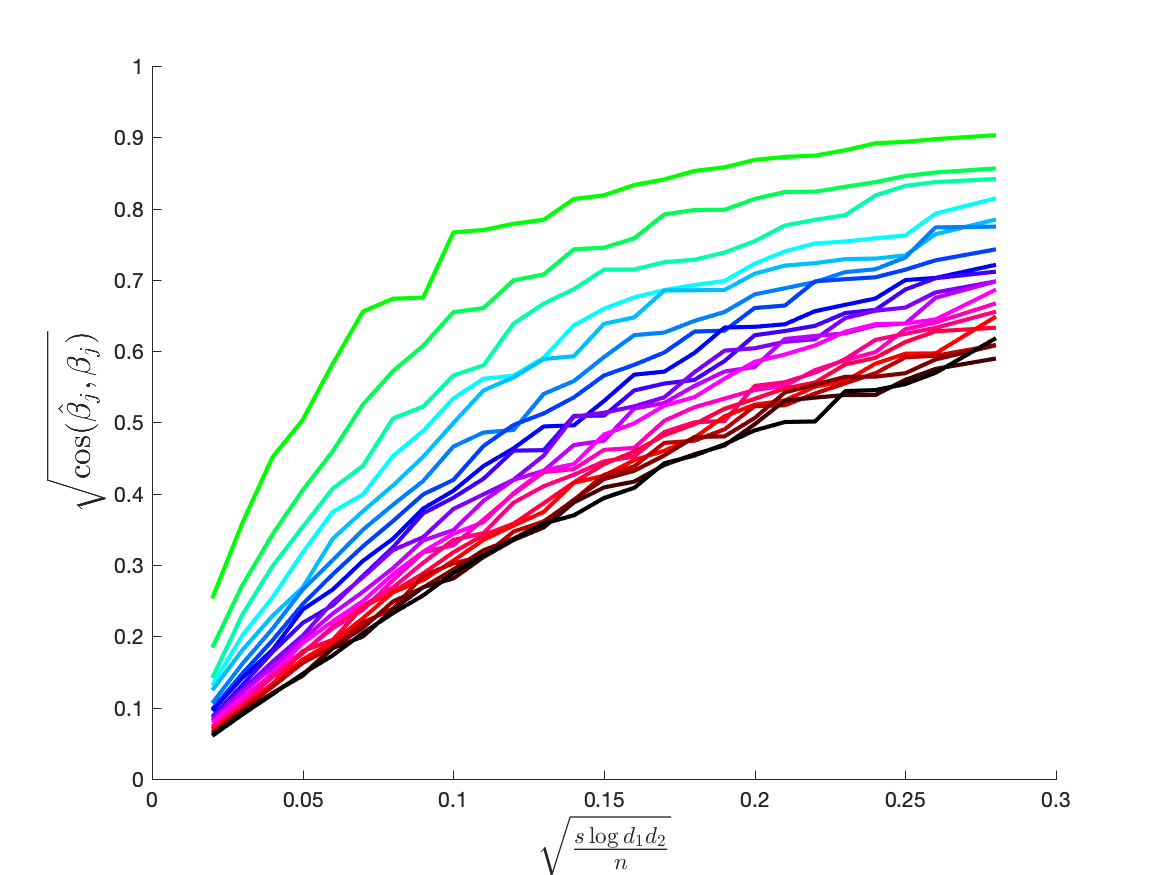}}
	\subfigure[Gaussian for $f^{(6)}$]{\label{fw6}\includegraphics[width=49.9mm]{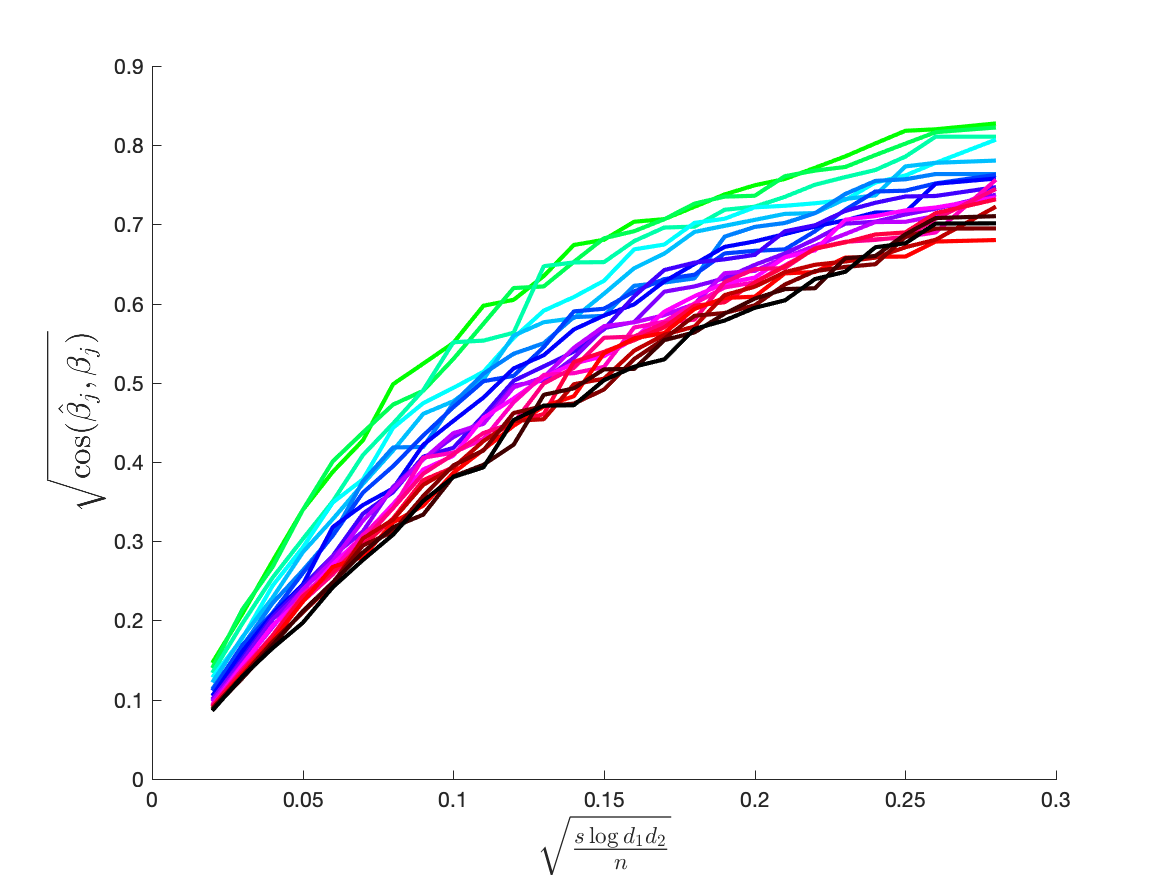}}
	\subfigure[Beta for $f^{(1)}$]{\label{fbeta1}\includegraphics[width=49.9mm]{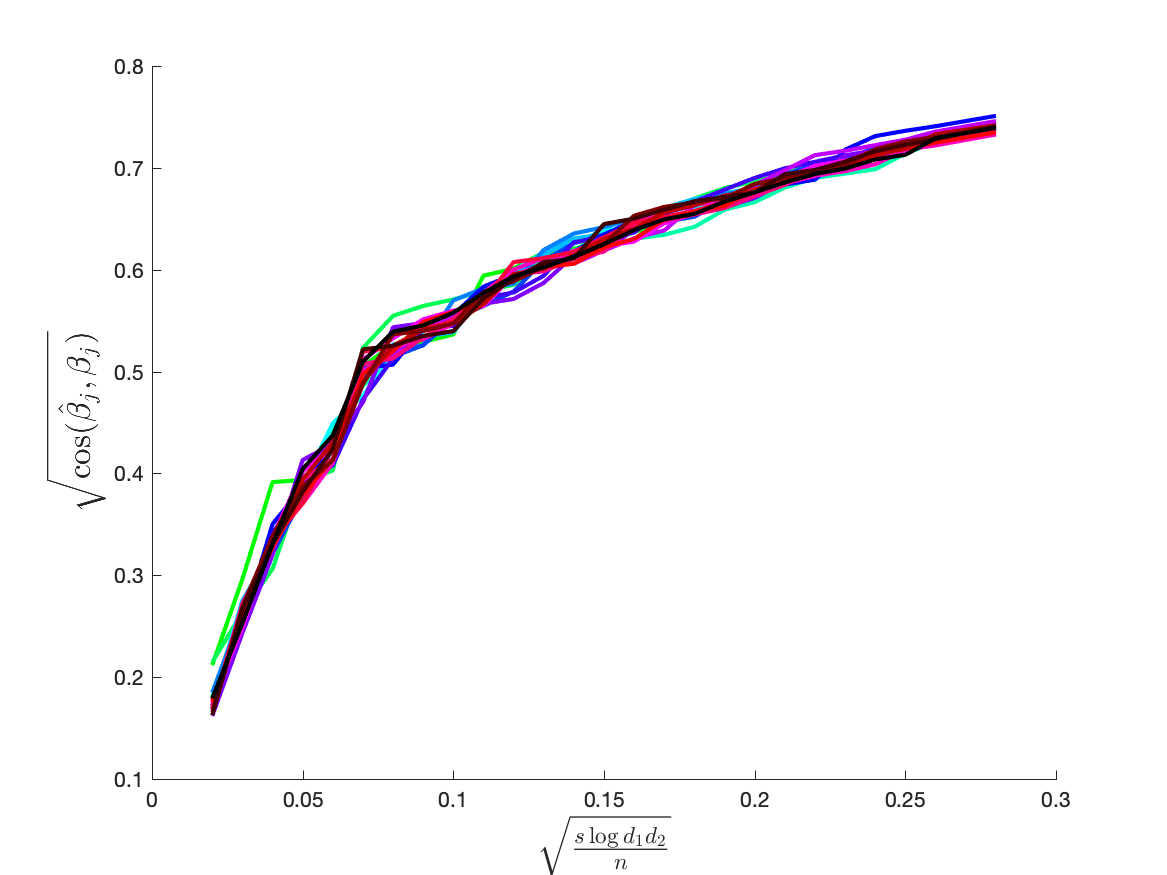}}
	\subfigure[Beta for $f^{(2)}$]{\label{fbeta2}\includegraphics[width=49.9mm]{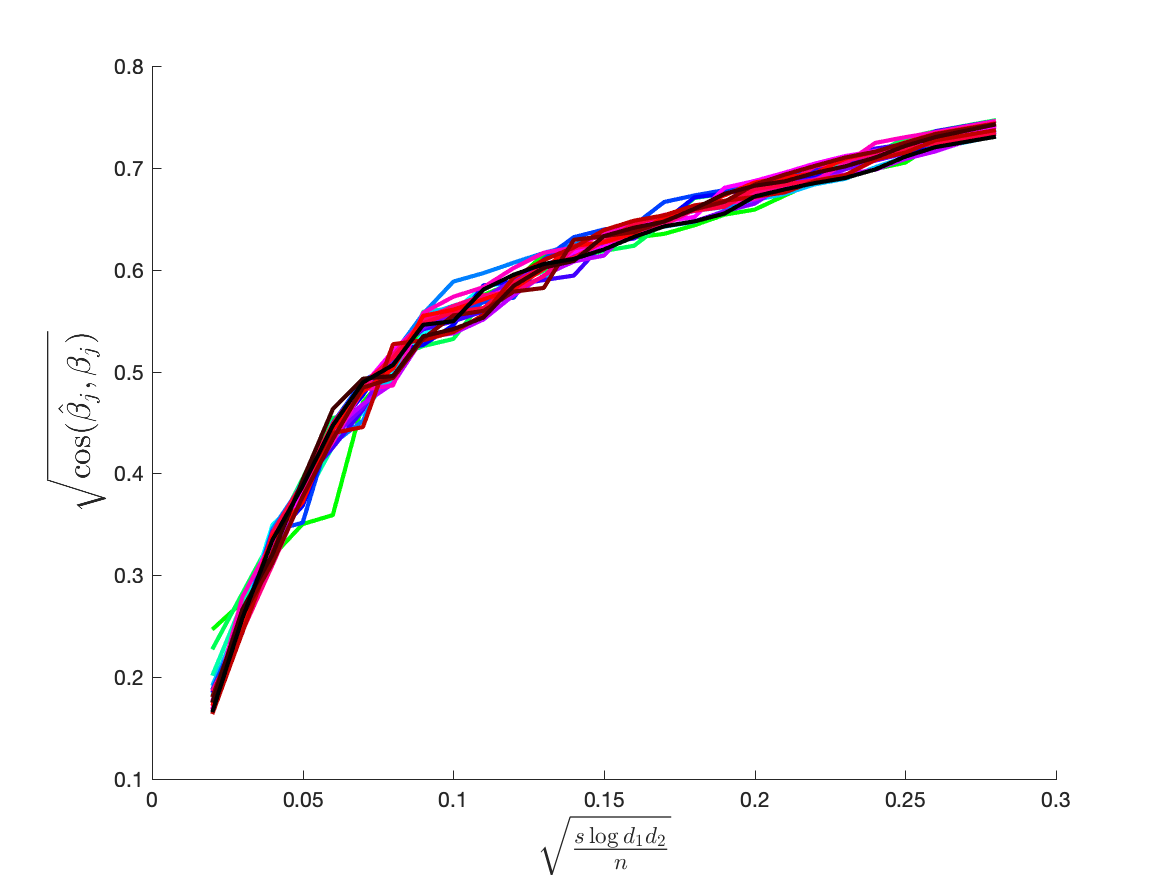}}
	\subfigure[Beta for $f^{(3)}$]{\label{fbeta3}\includegraphics[width=49.9mm]{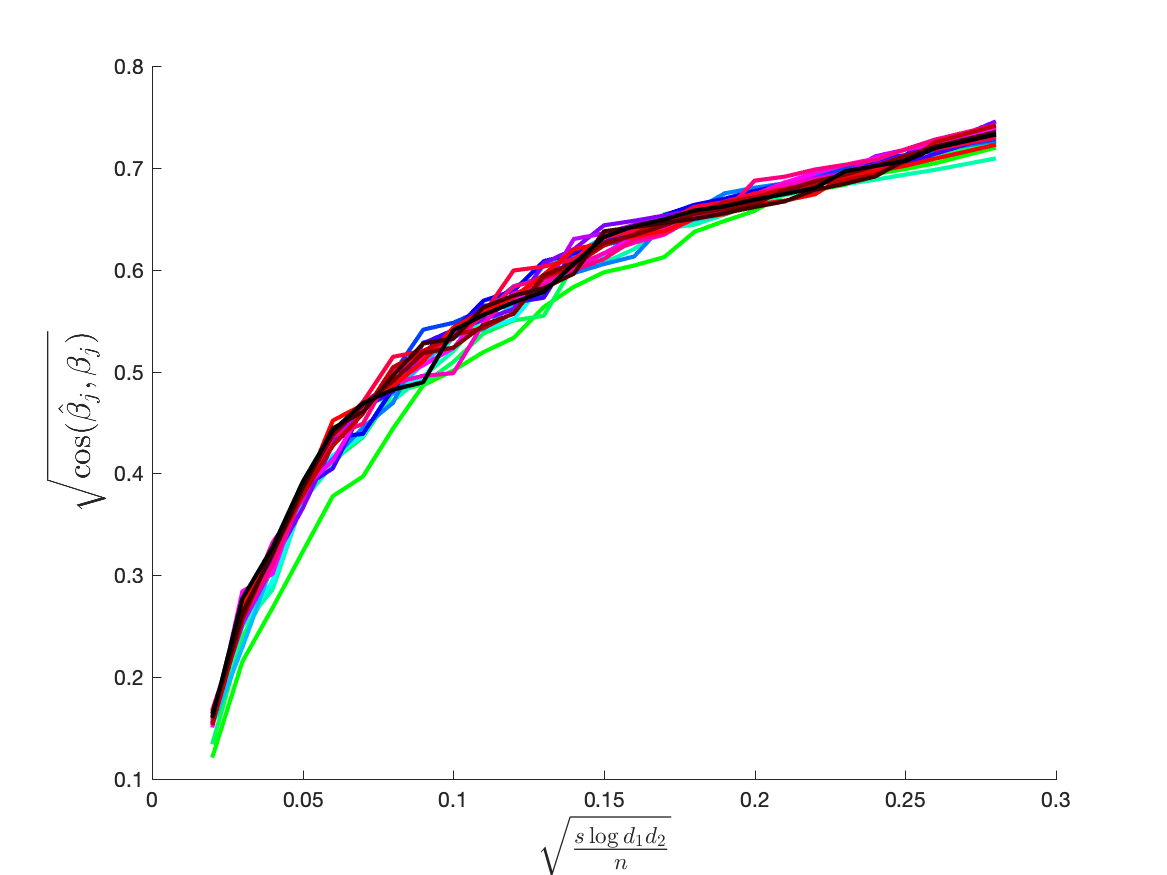}}
	\subfigure[Beta for $f^{(4)}$]{\label{fbeta4}\includegraphics[width=49.9mm]{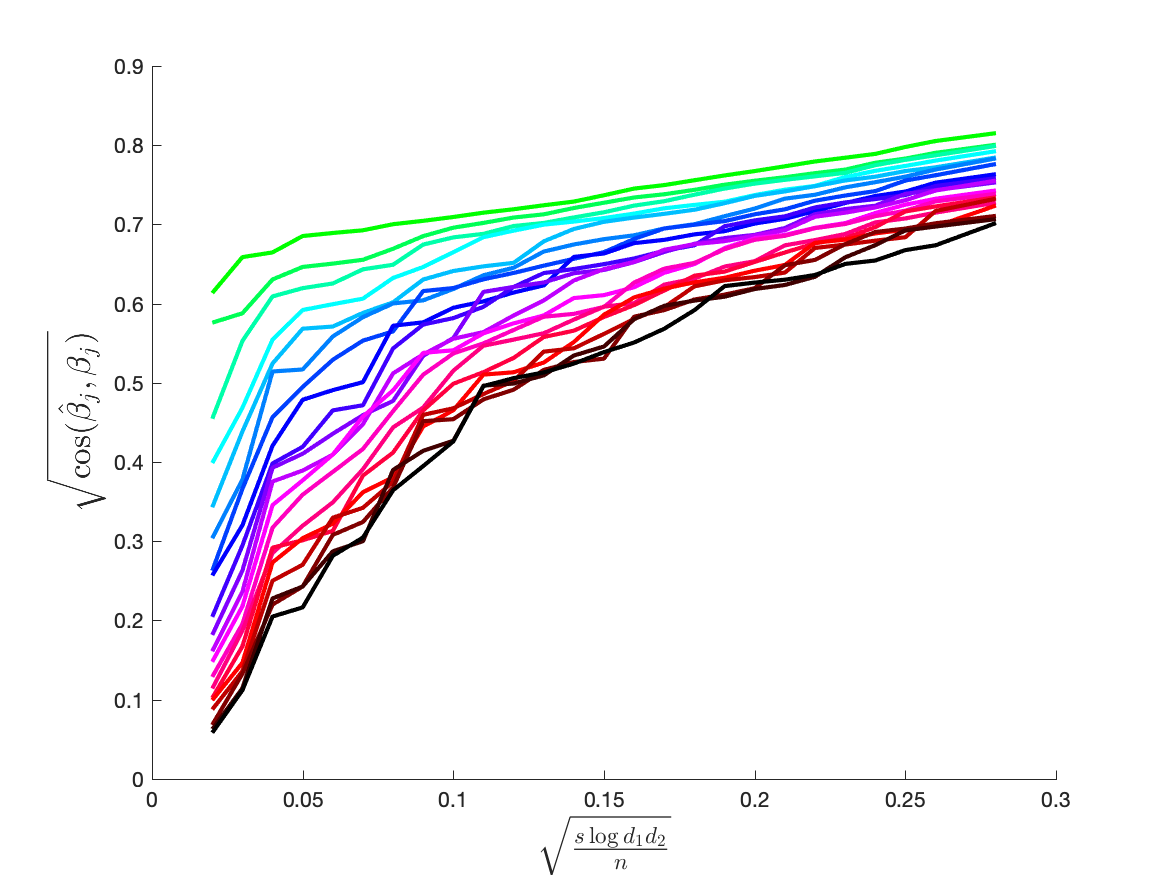}}
	\subfigure[Beta for $f^{(5)}$]{\label{fbeta5}\includegraphics[width=49.9mm]{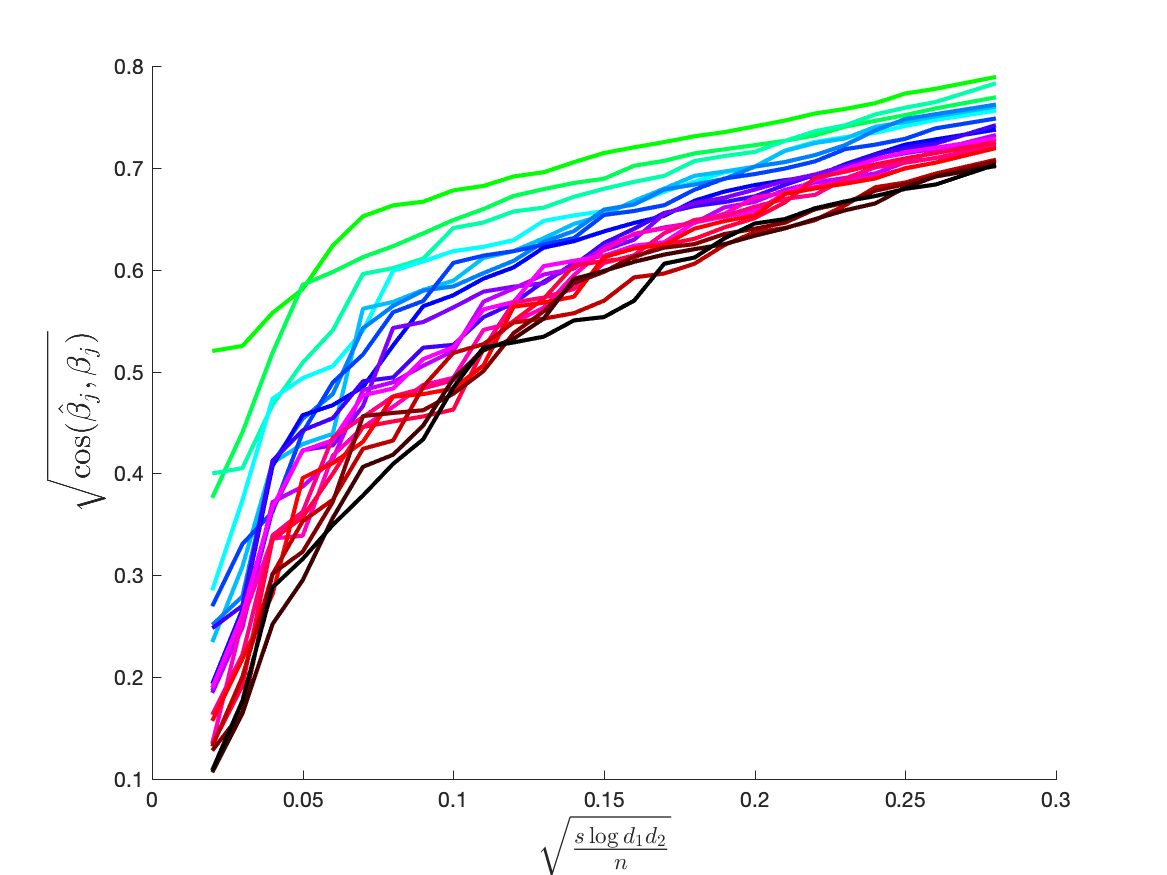}}
	\subfigure[Beta for $f^{(6)}$]{\label{fbeta6}\includegraphics[width=49.9mm]{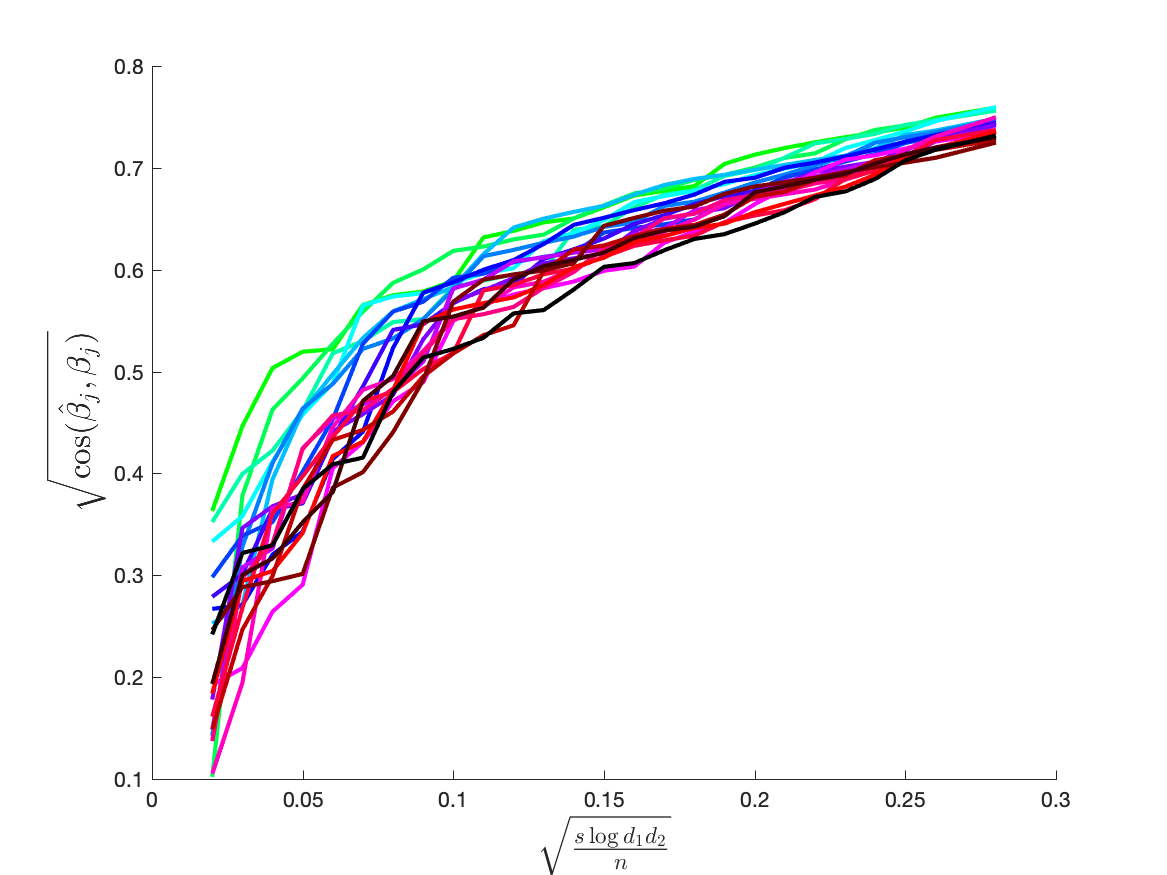}}
	\includegraphics[width=3.5cm,height=0.34cm]{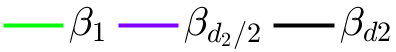}
	\caption{{Sparse vector estimation for Gaussian and Beta design. This figure shows cosine distance trend for error of estimating single sparse parameter in model (\ref{mod:1}). The color varies from green to black when $k$ varies from $1$ to $d_2$. The first and third row correspond to the linear link function, while the second and fourth row correspond to the quadratic link function.}}\label{fig:4}
\end{figure}

\begin{figure}[p]
	\centering     
	\subfigure[Gamma for $f^{(1)}$]{\label{fgamma1}\includegraphics[width=49.9mm]{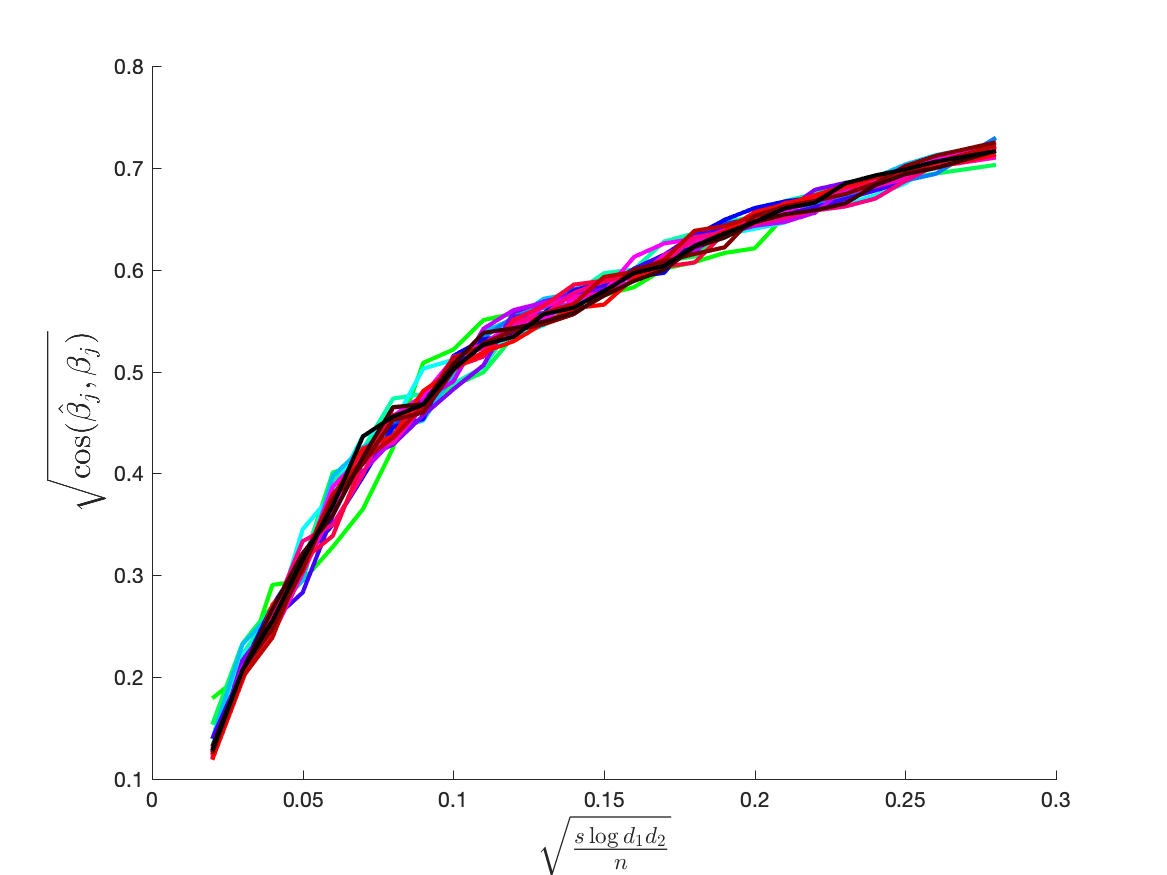}}
	\subfigure[Gamma for $f^{(2)}$]{\label{fgamma2}\includegraphics[width=49.9mm]{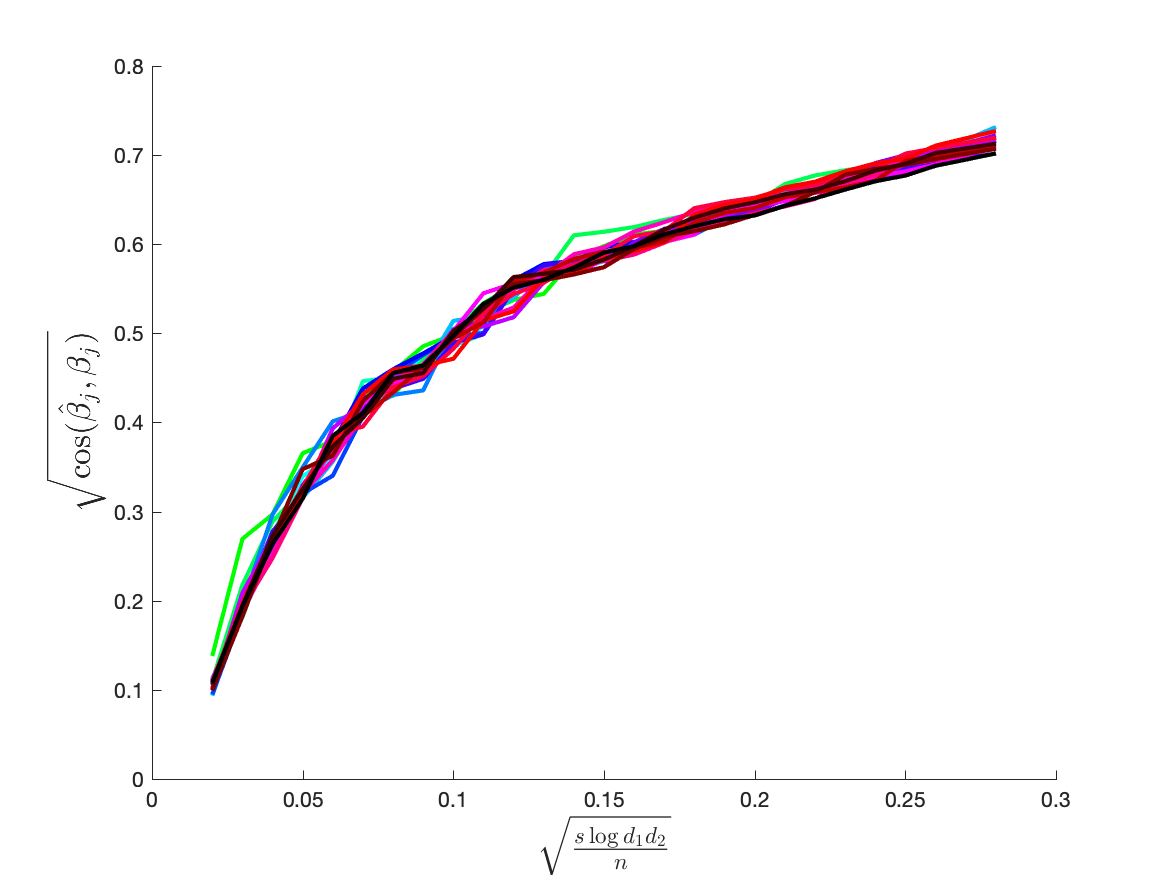}}
	\subfigure[Gamma for $f^{(3)}$]{\label{fgamma3}\includegraphics[width=49.9mm]{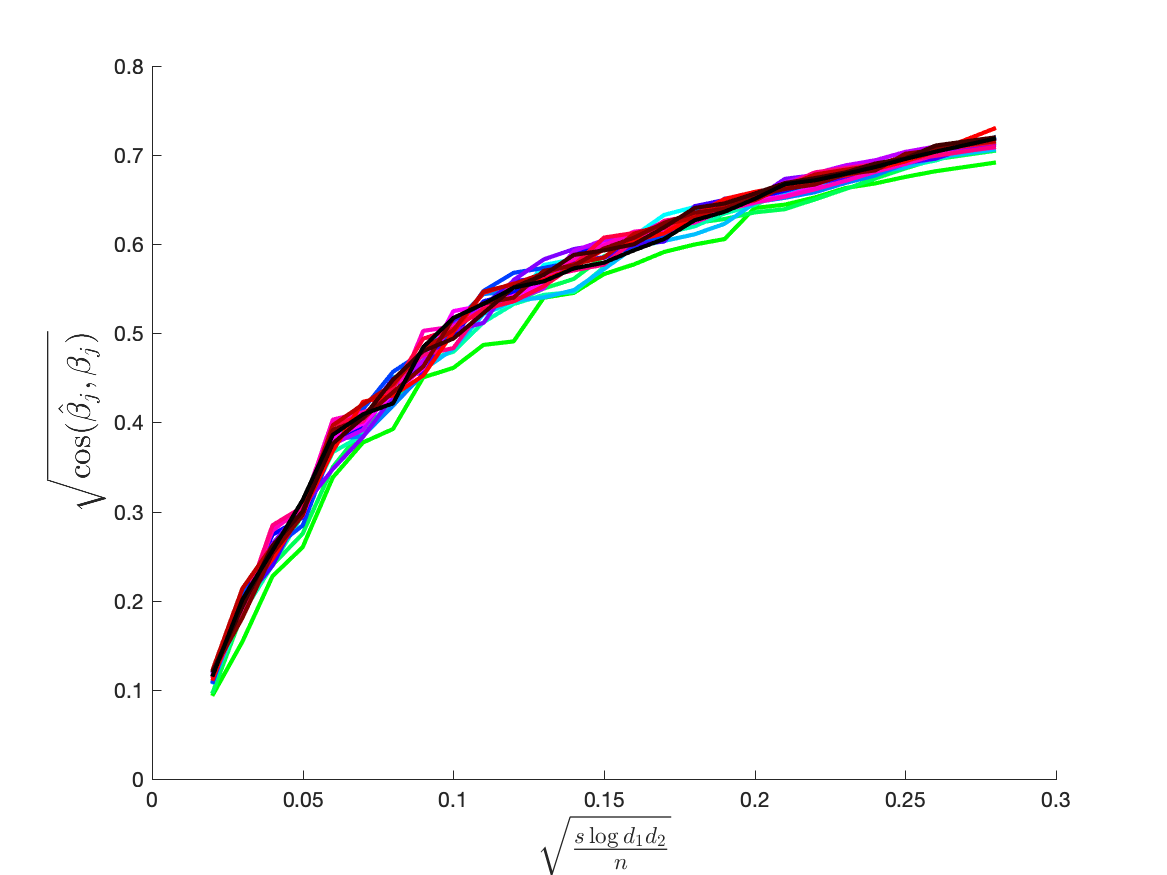}}
	\subfigure[Gamma for $f^{(4)}$]{\label{fgamma4}\includegraphics[width=49.9mm]{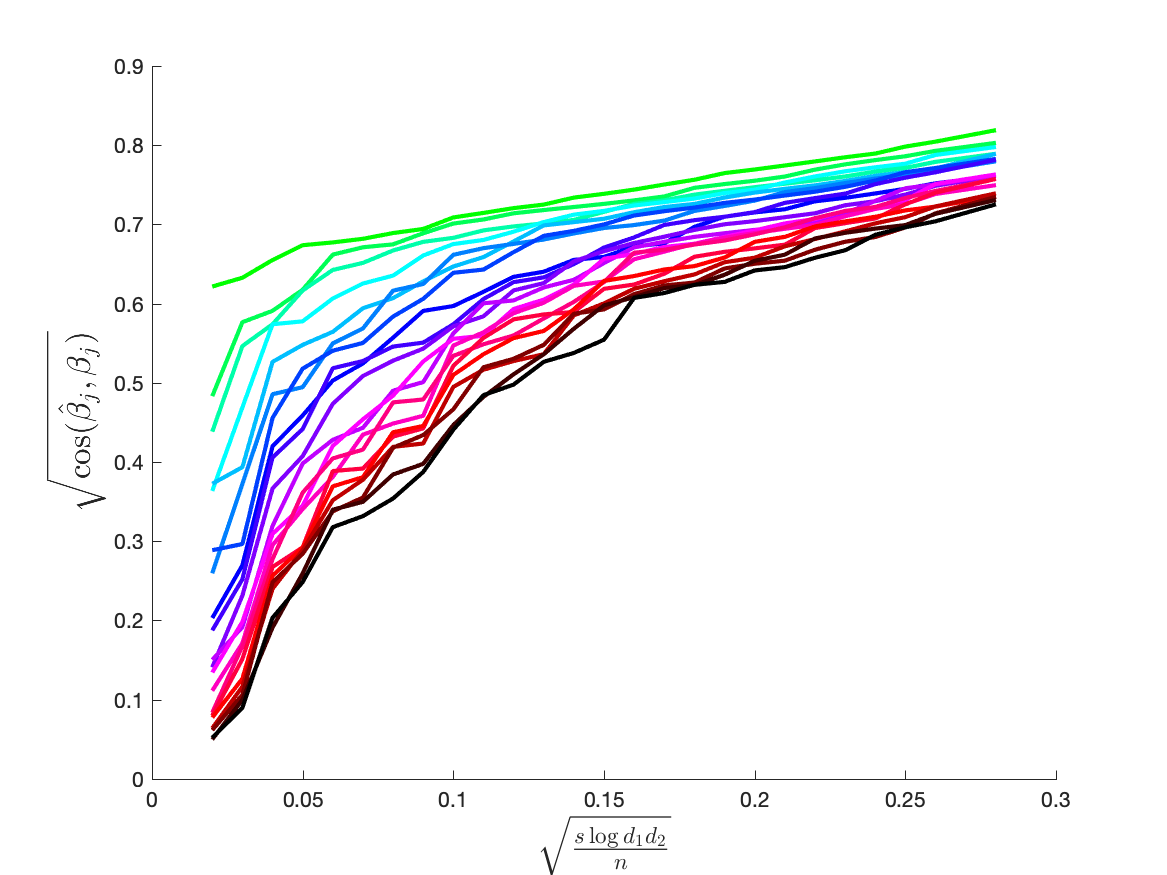}}
	\subfigure[Gamma for $f^{(5)}$]{\label{fgamma5}\includegraphics[width=49.9mm]{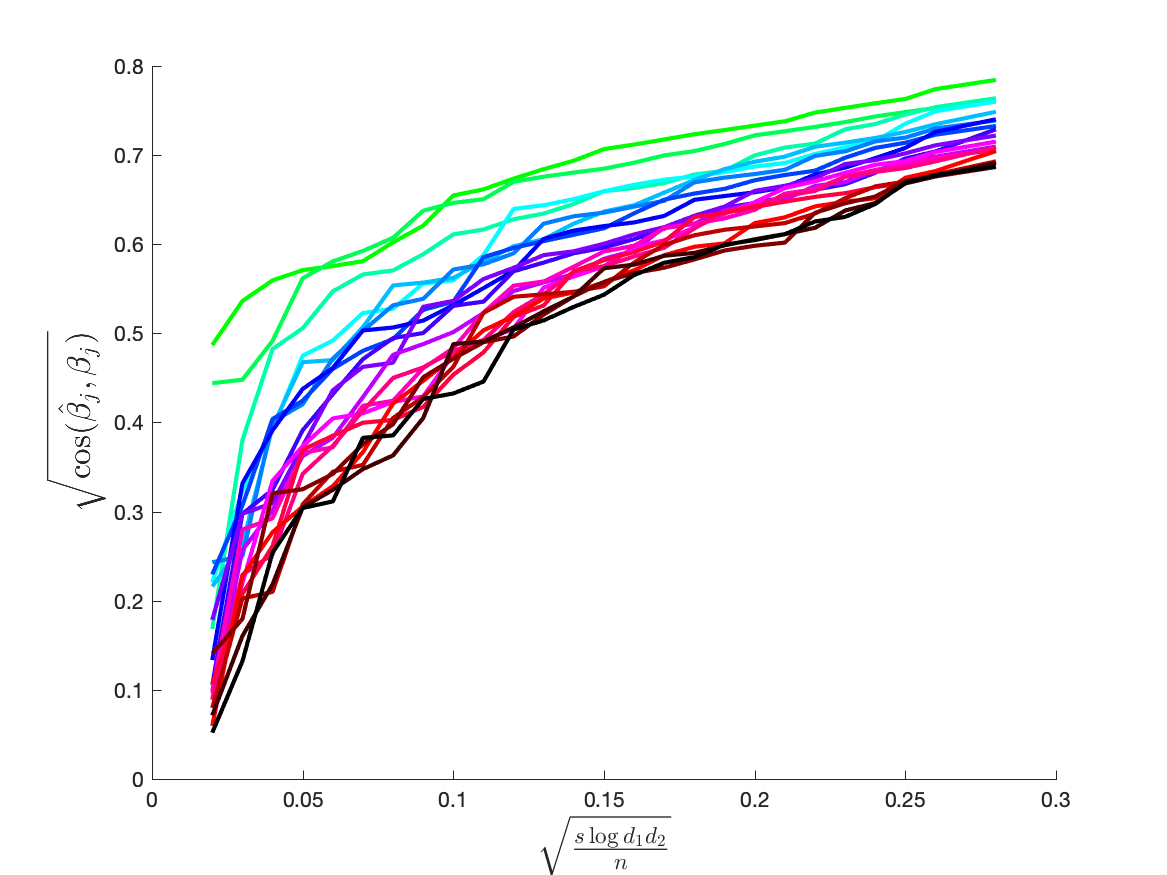}}
	\subfigure[Gamma for $f^{(6)}$]{\label{fgamma6}\includegraphics[width=49.9mm]{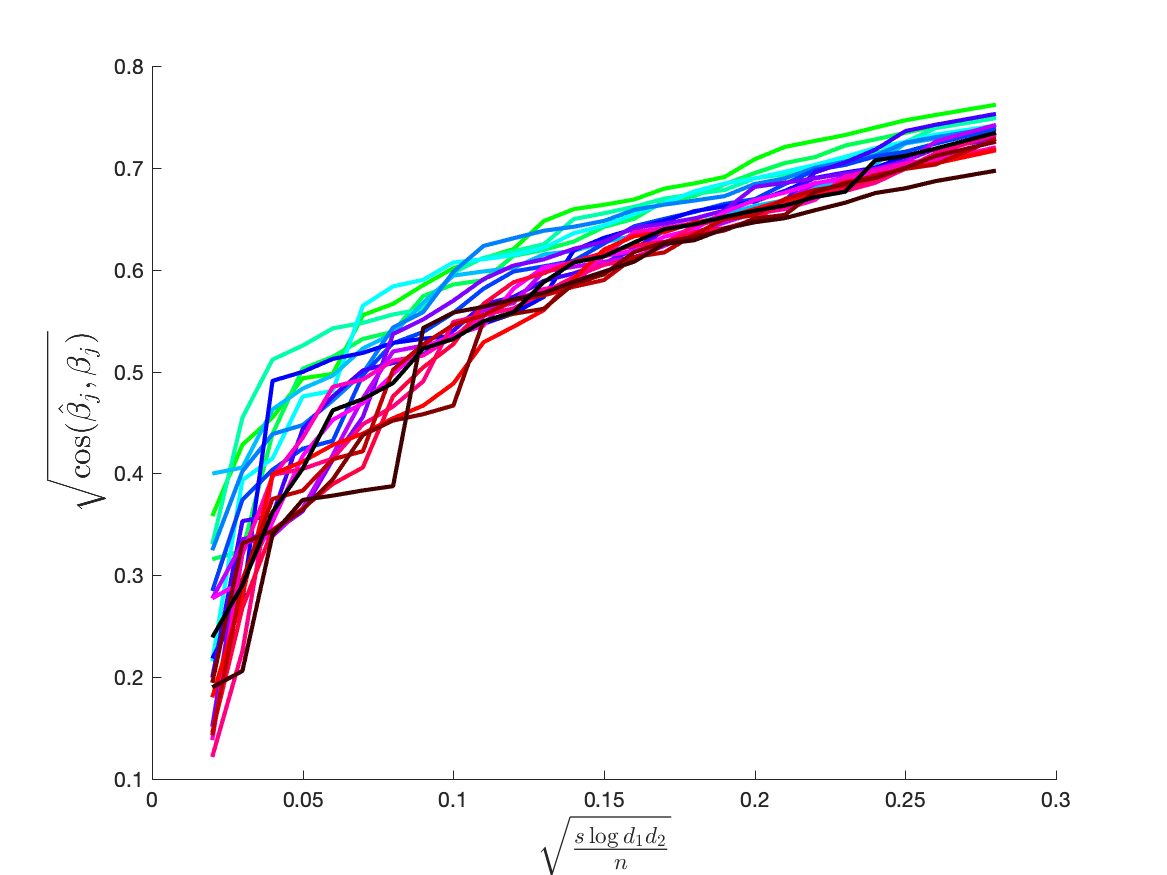}}
	\subfigure[$t_{13}$ for $f^{(1)}$]{\label{ft131}\includegraphics[width=49.9mm]{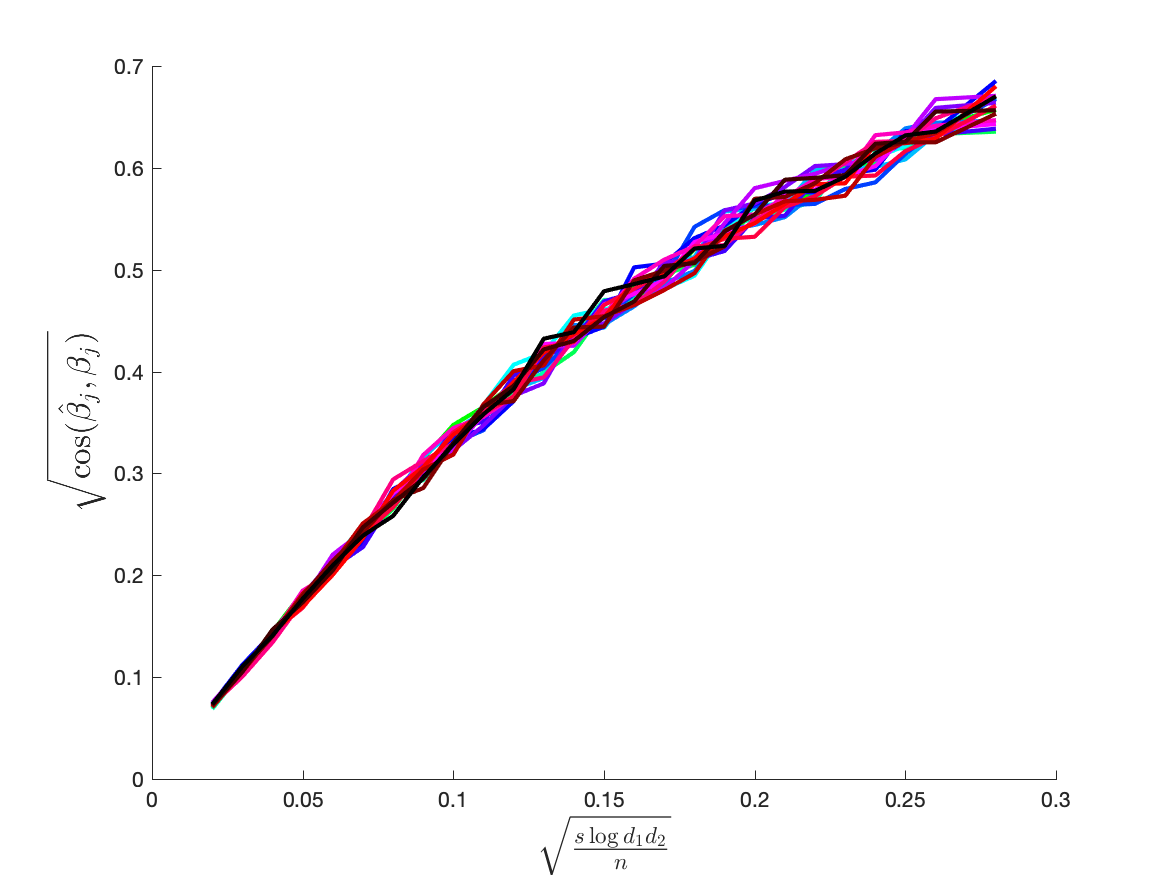}}
	\subfigure[$t_{13}$ for $f^{(2)}$]{\label{ft132}\includegraphics[width=49.9mm]{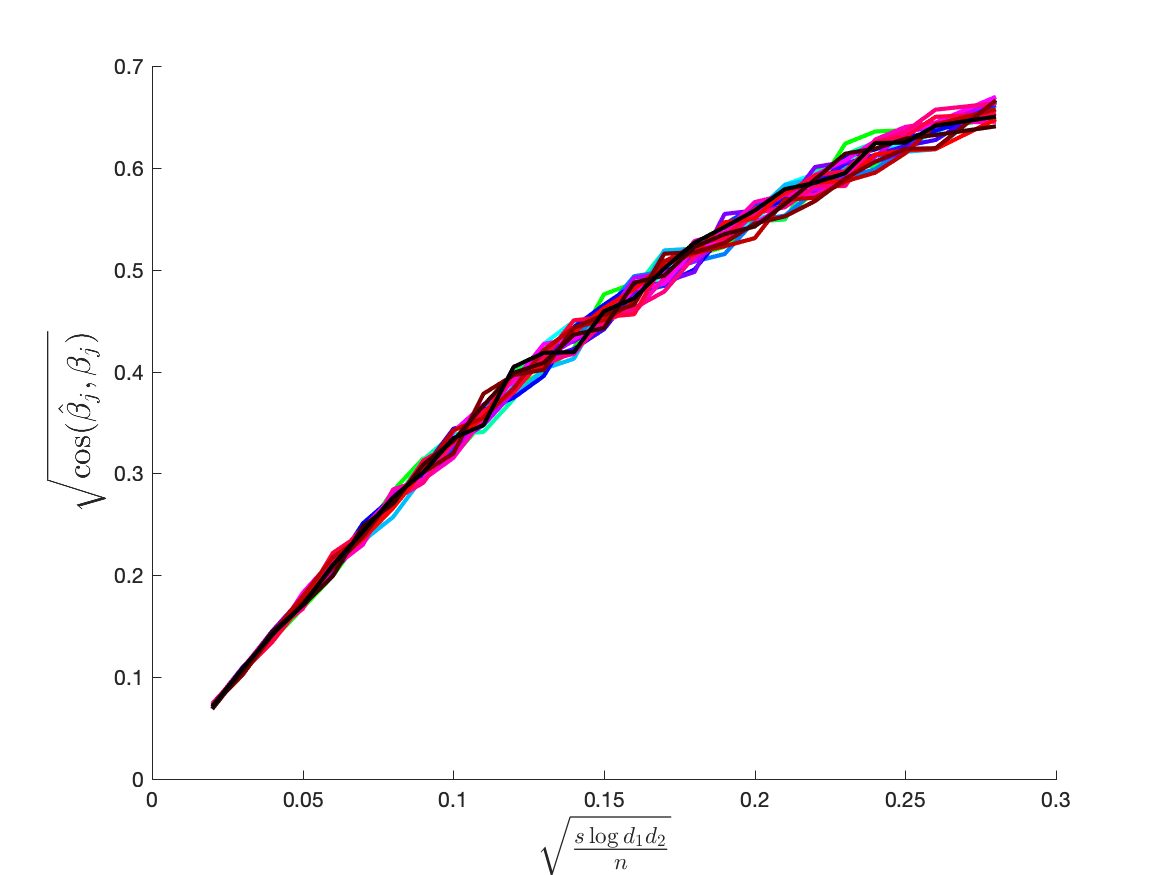}}
	\subfigure[$t_{13}$ for $f^{(3)}$]{\label{ft133}\includegraphics[width=49.9mm]{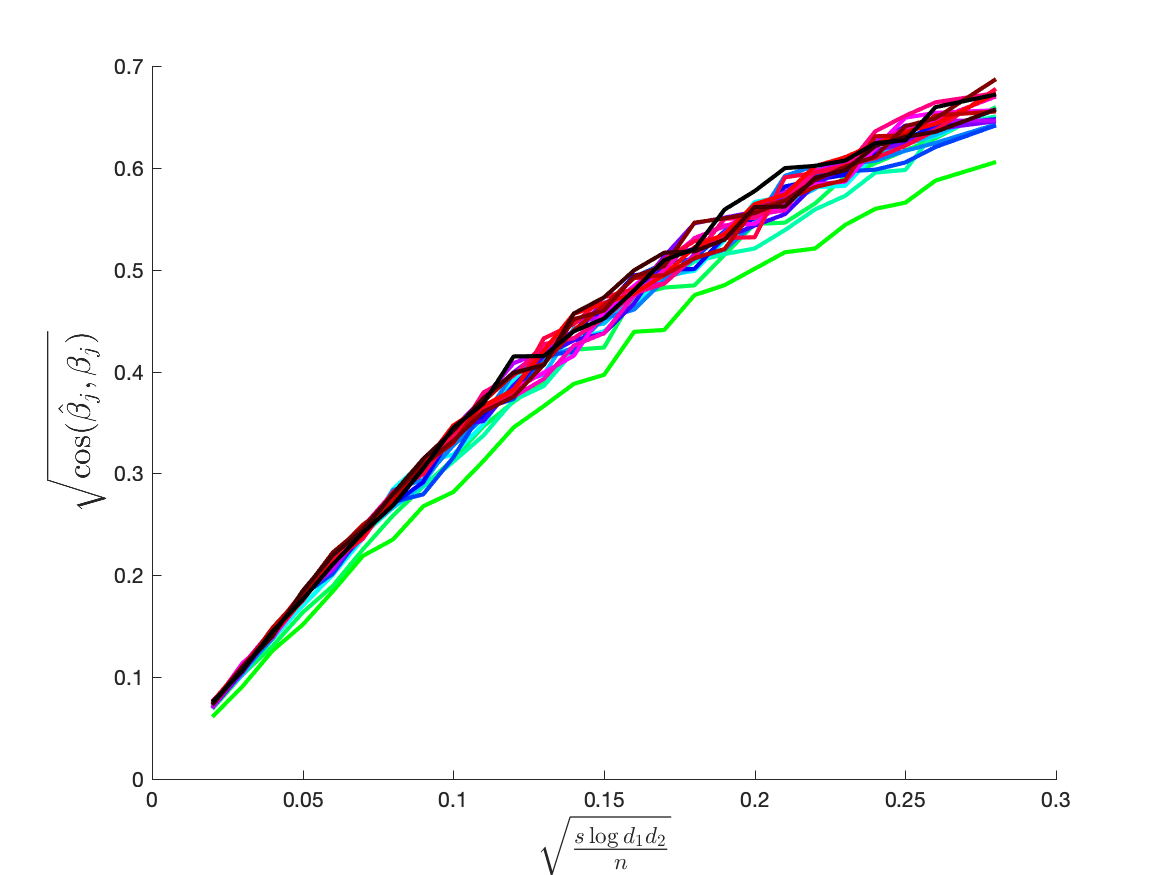}}
	\subfigure[$t_{13}$ for $f^{(4)}$]{\label{ft134}\includegraphics[width=49.9mm]{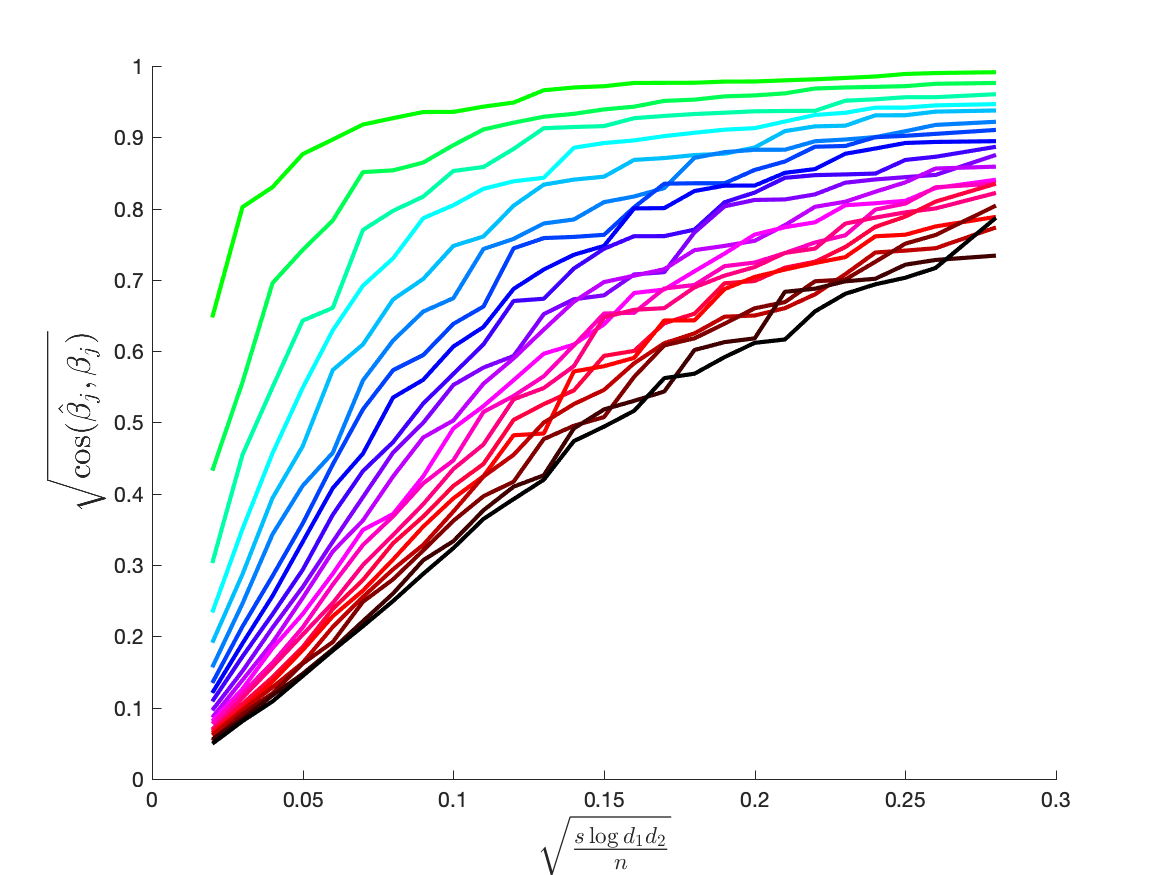}}
	\subfigure[$t_{13}$ for $f^{(5)}$]{\label{ft135}\includegraphics[width=49.9mm]{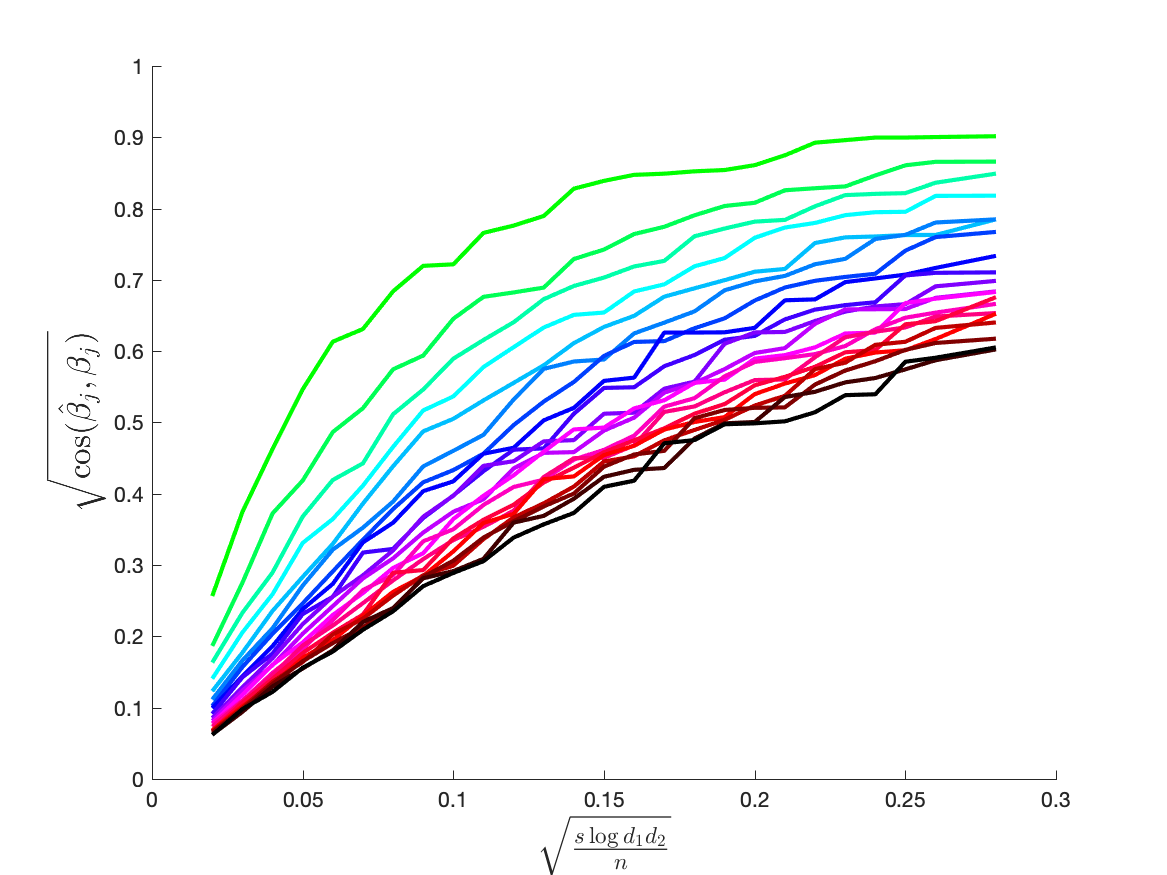}}
	\subfigure[$t_{13}$ for $f^{(6)}$]{\label{ft136}\includegraphics[width=49.9mm]{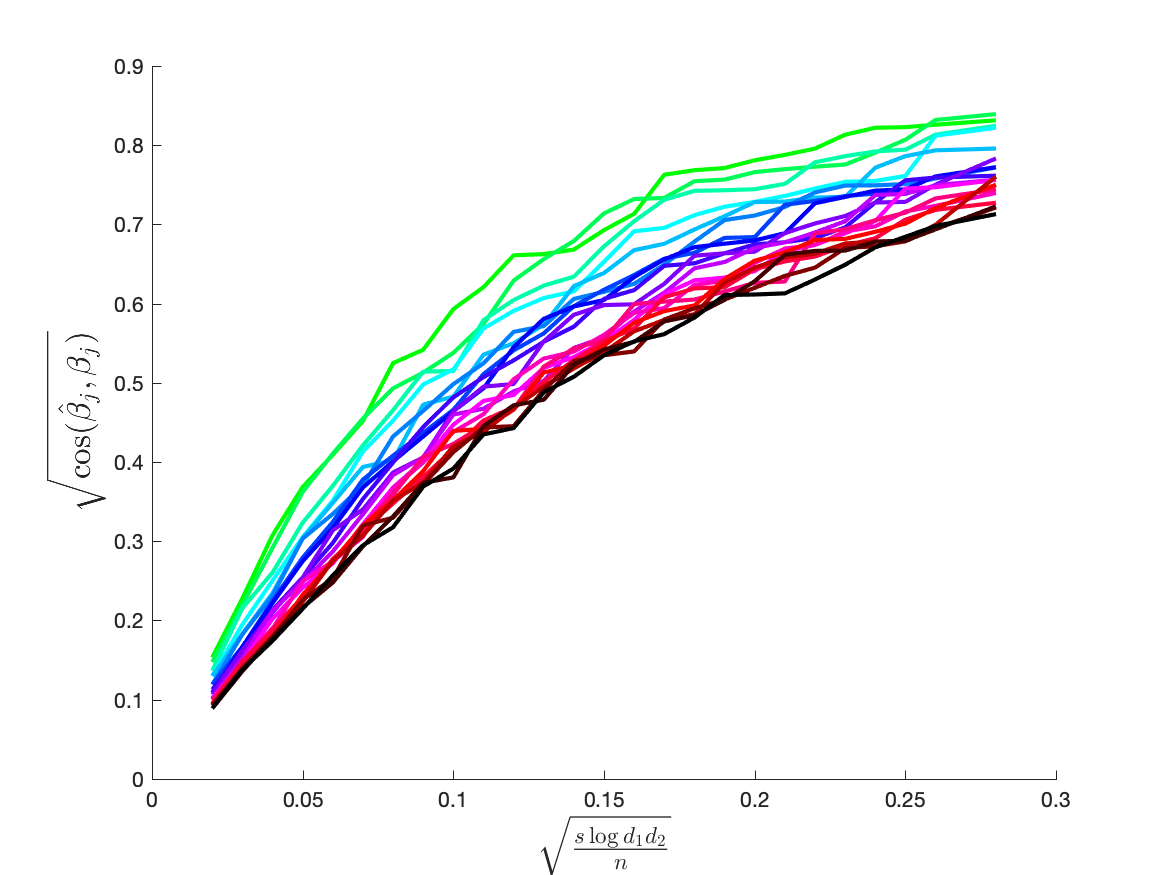}}
	\includegraphics[width=3.5cm,height=0.34cm]{Figure/betaleg.png}
	\caption{{Sparse vector estimation for Gamma and $t$ design. This figure shows cosine distance trend for error of estimating single sparse parameter in model (\ref{mod:1}). The color varies from green to black when $k$ varies from $1$ to $d_2$. The first and third row correspond to the linear link function, while the second and fourth row correspond to the quadratic link function.}}\label{fig:5}
\end{figure}

\begin{figure}[!htp]
	\centering     
	\subfigure[Rayleigh for $f^{(1)}$]{\label{fray1}\includegraphics[width=49.9mm]{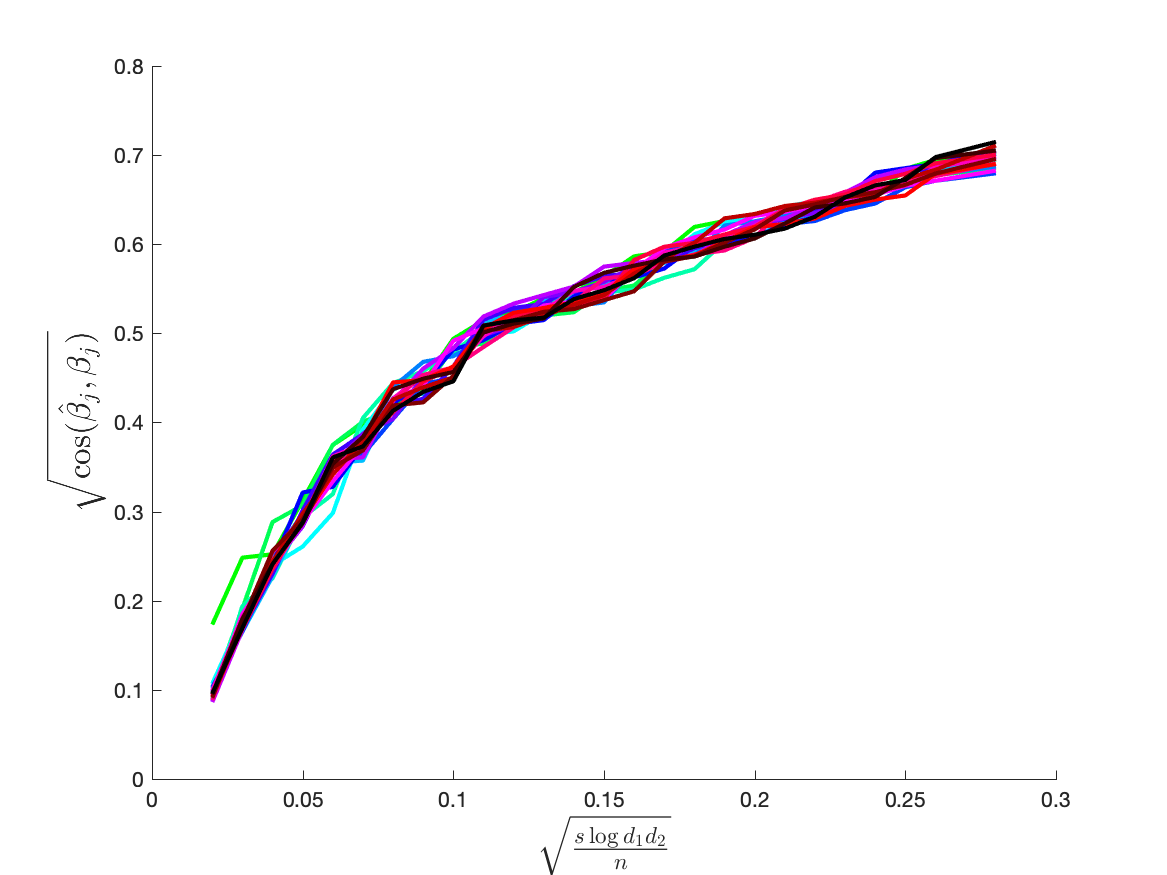}}
	\subfigure[Rayleigh for $f^{(2)}$]{\label{fray2}\includegraphics[width=49.9mm]{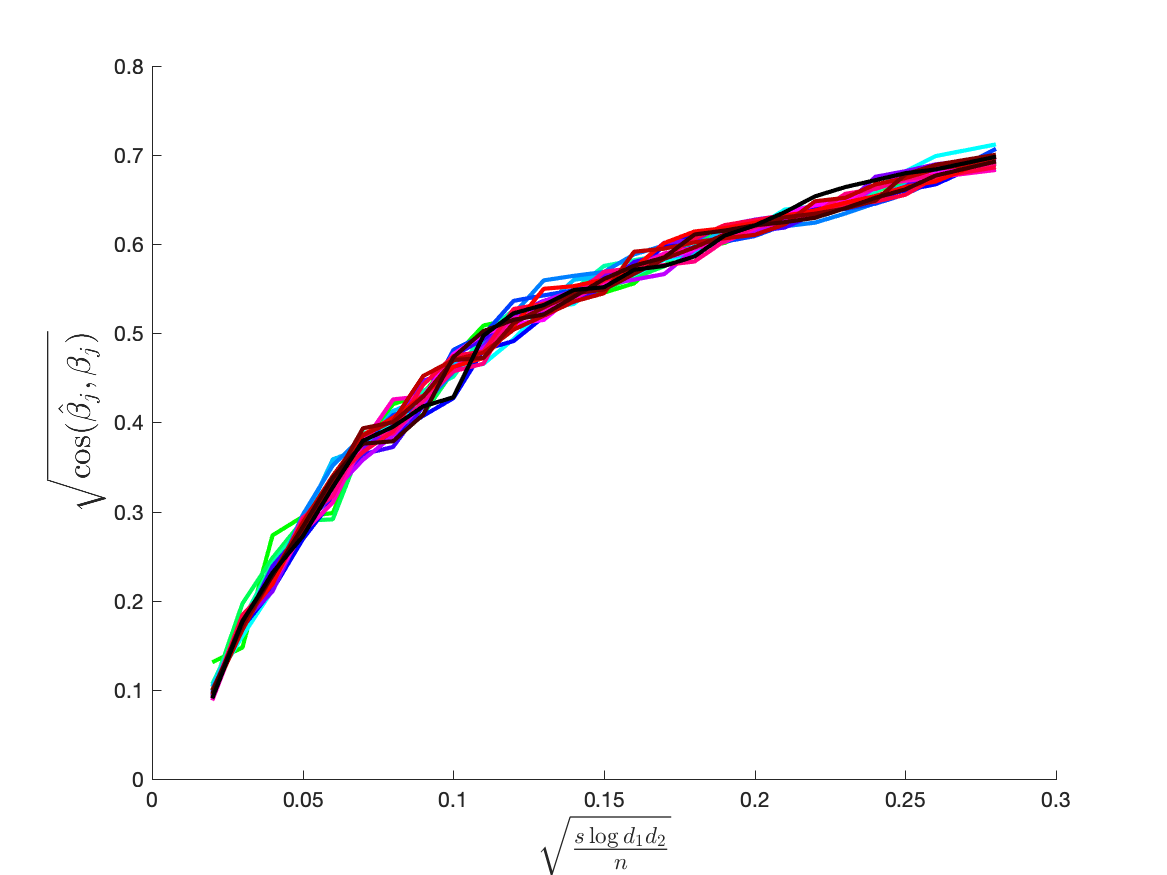}}
	\subfigure[Rayleigh for $f^{(3)}$]{\label{fray3}\includegraphics[width=49.9mm]{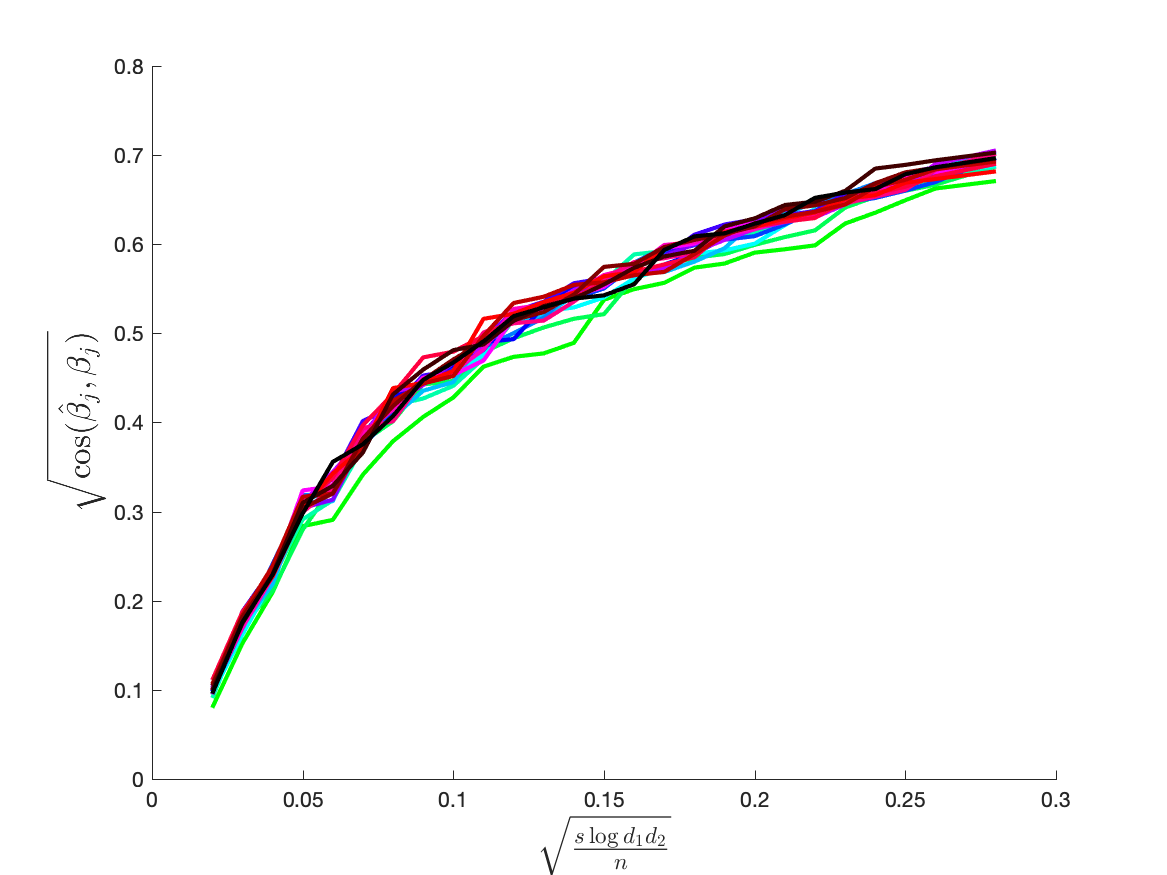}}
	\subfigure[Rayleigh for $f^{(4)}$]{\label{fray4}\includegraphics[width=49.9mm]{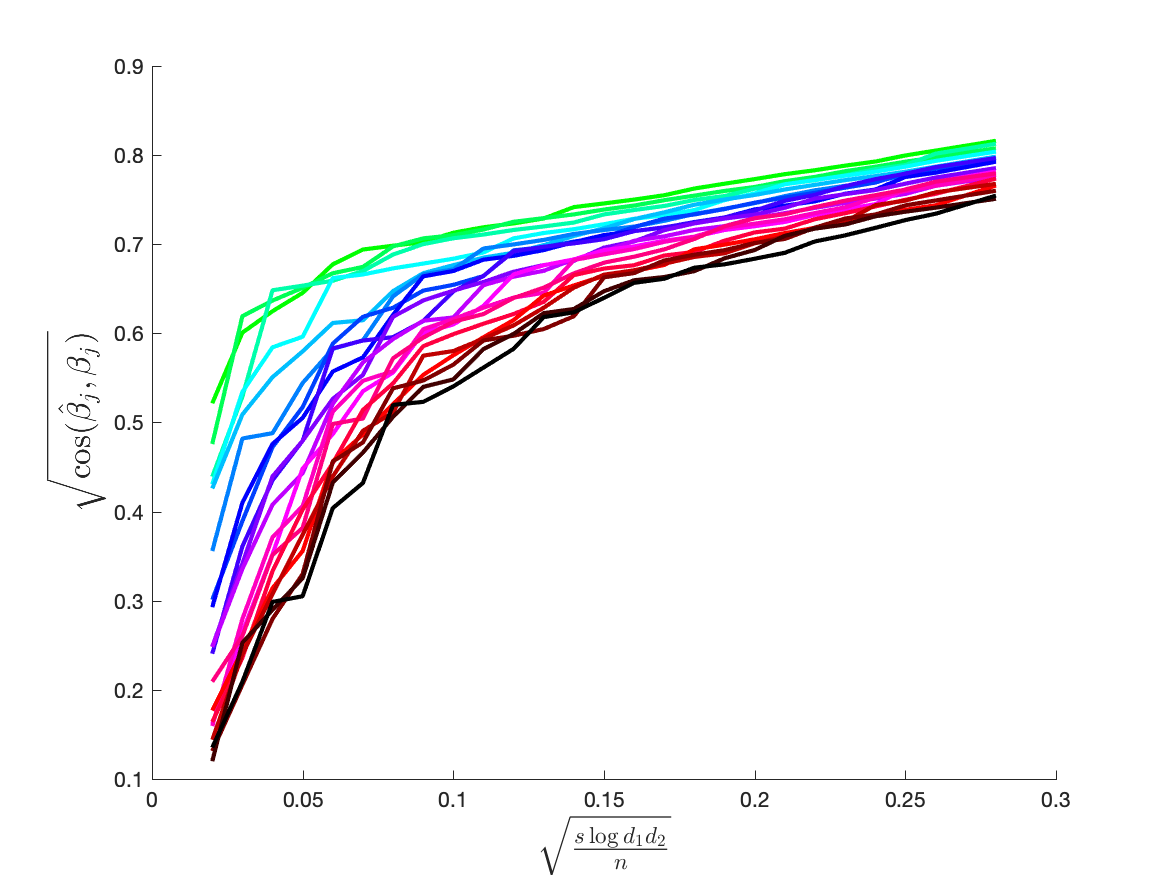}}
	\subfigure[Rayleigh for $f^{(5)}$]{\label{fray5}\includegraphics[width=49.9mm]{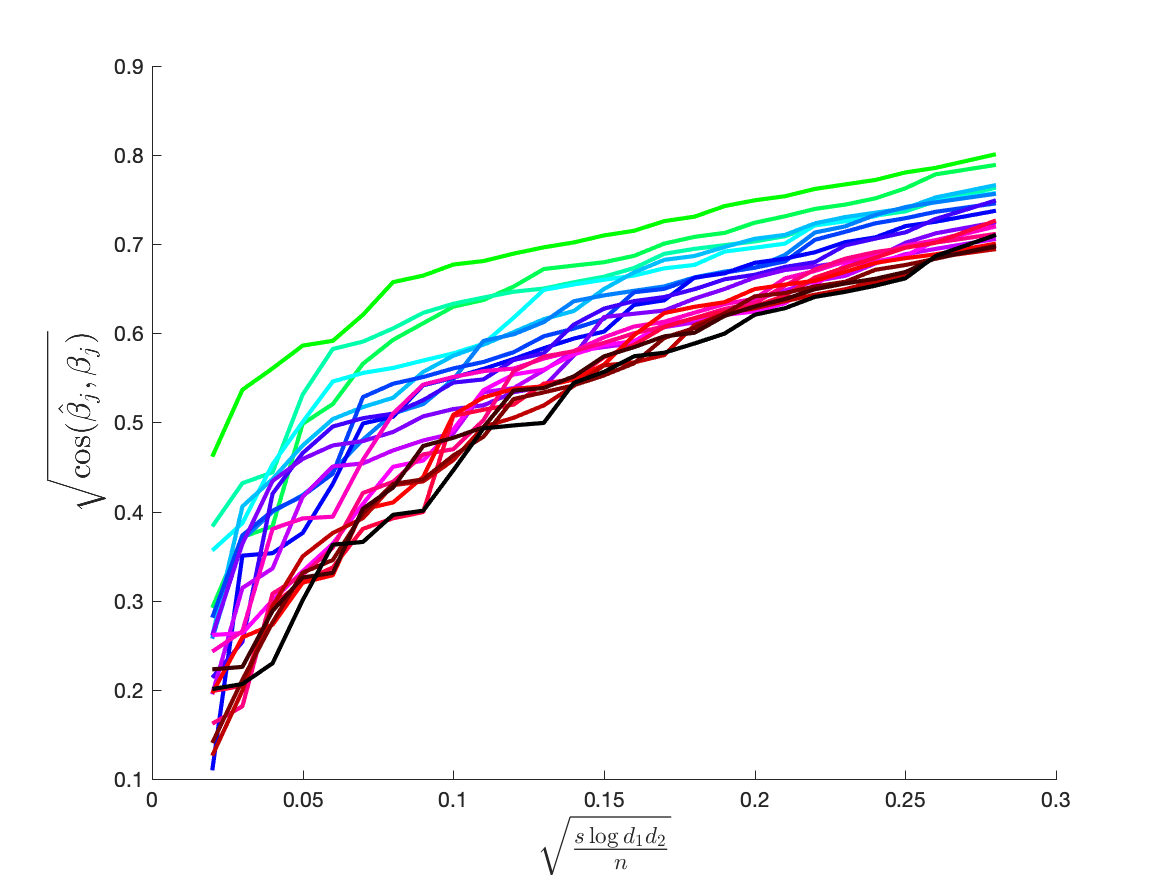}}
	\subfigure[Rayleigh for $f^{(6)}$]{\label{fray6}\includegraphics[width=49.9mm]{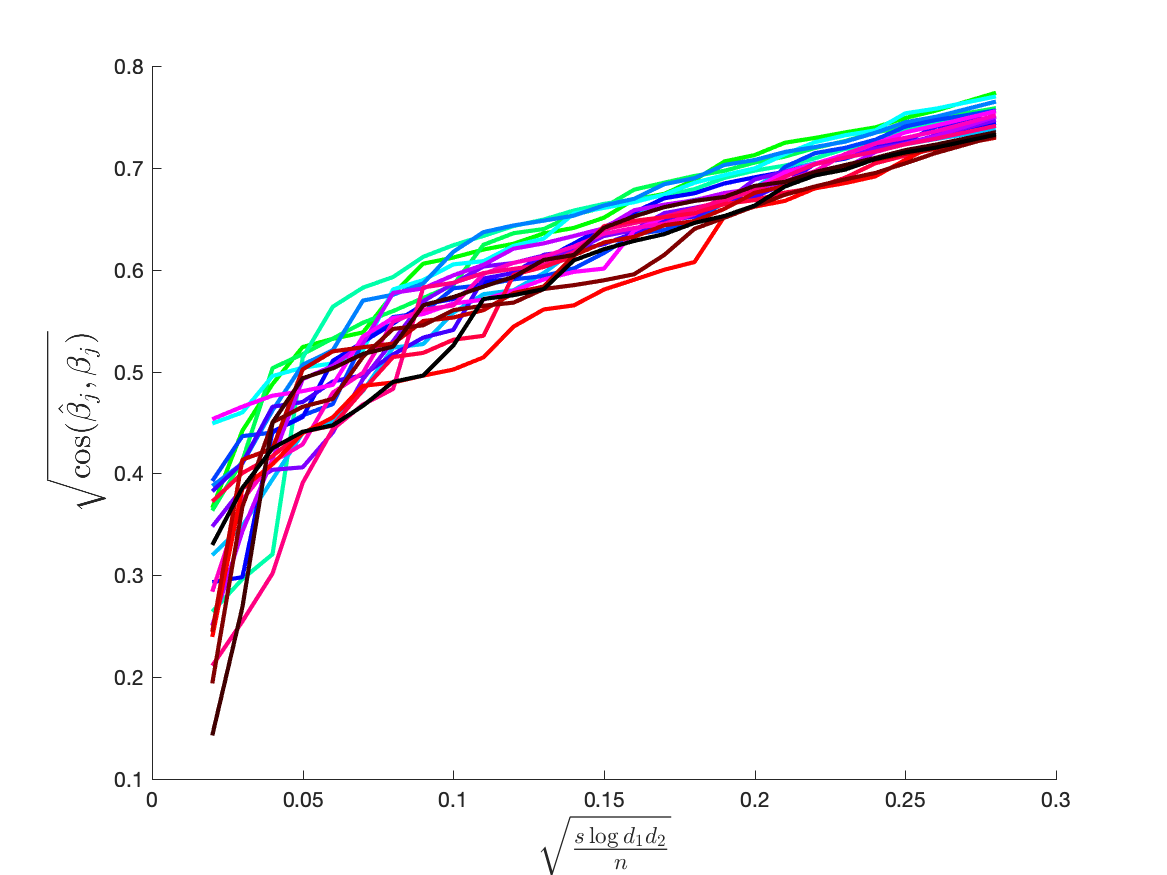}}
	\subfigure[Weibull for $f^{(1)}$]{\label{fwei1}\includegraphics[width=49.9mm]{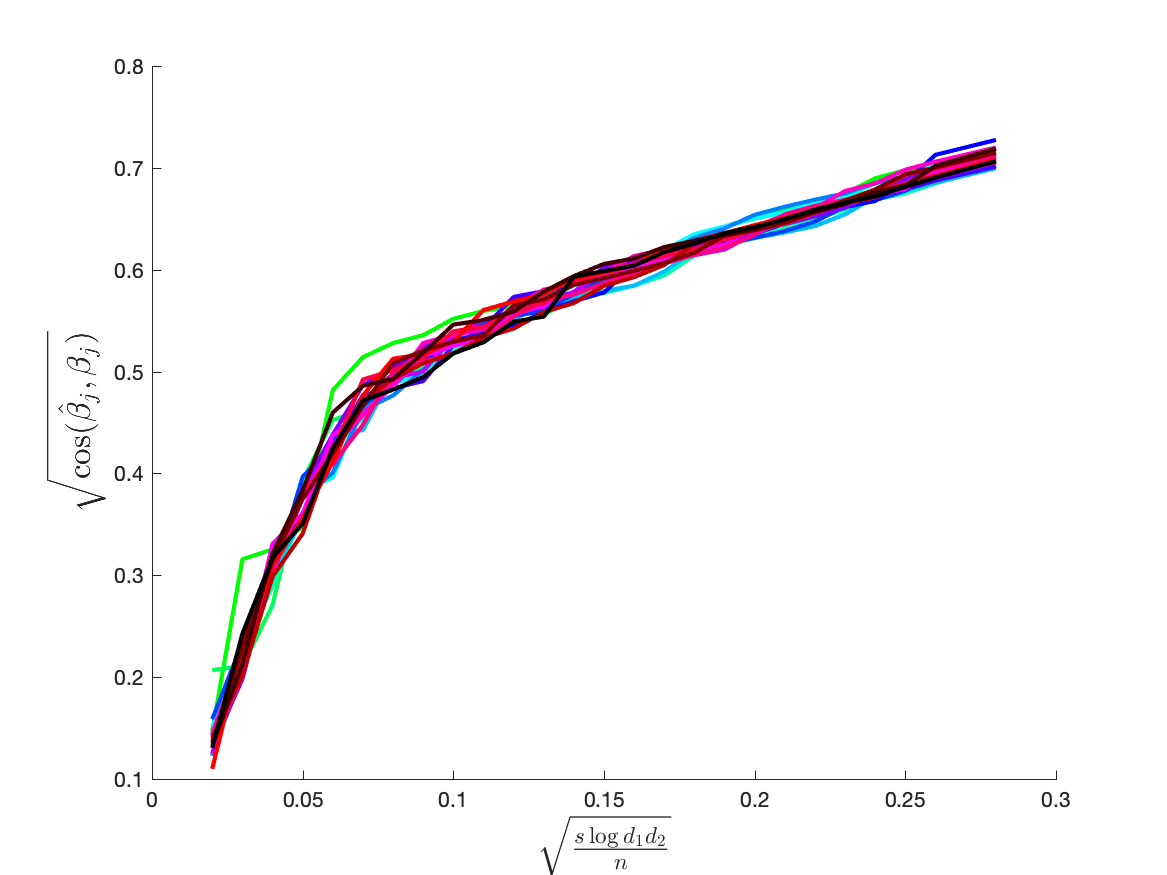}}
	\subfigure[Weibull for $f^{(2)}$]{\label{fwei2}\includegraphics[width=49.9mm]{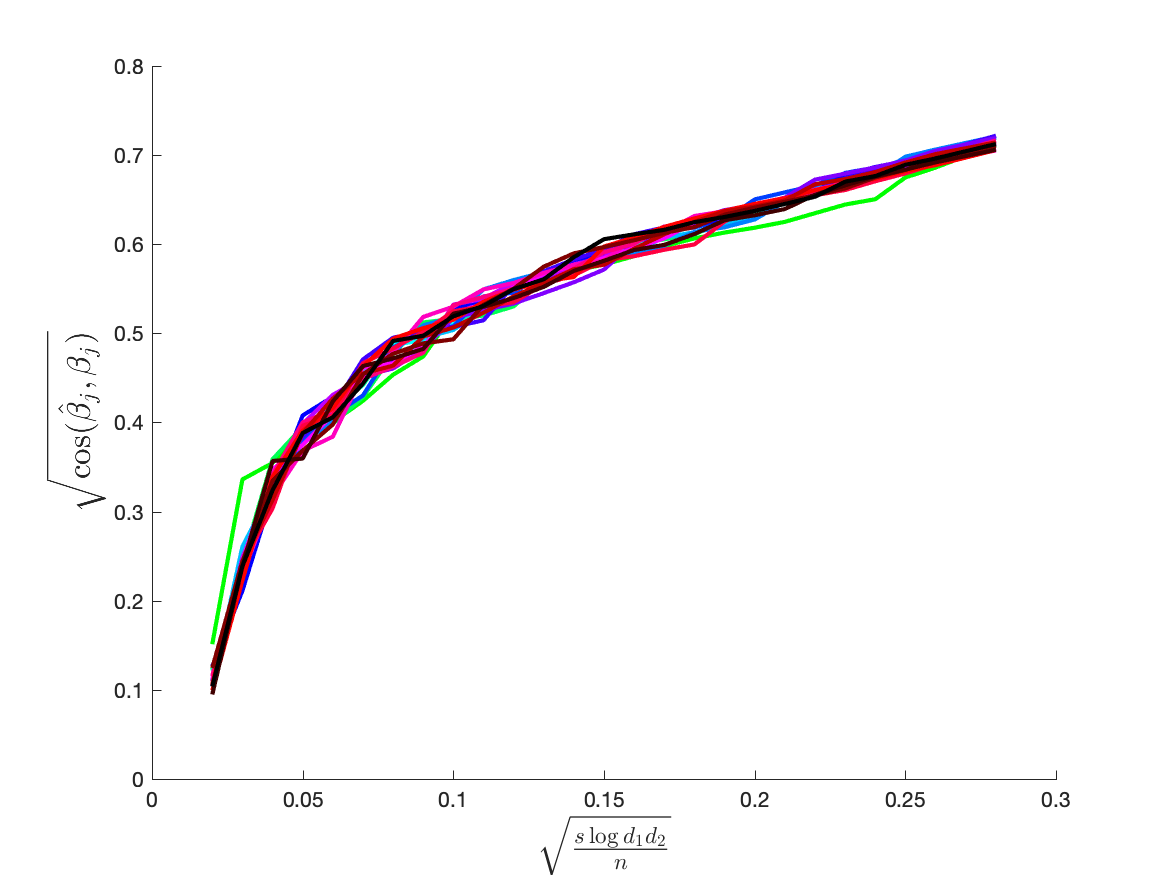}}
	\subfigure[Weibull for $f^{(3)}$]{\label{fwei3}\includegraphics[width=49.9mm]{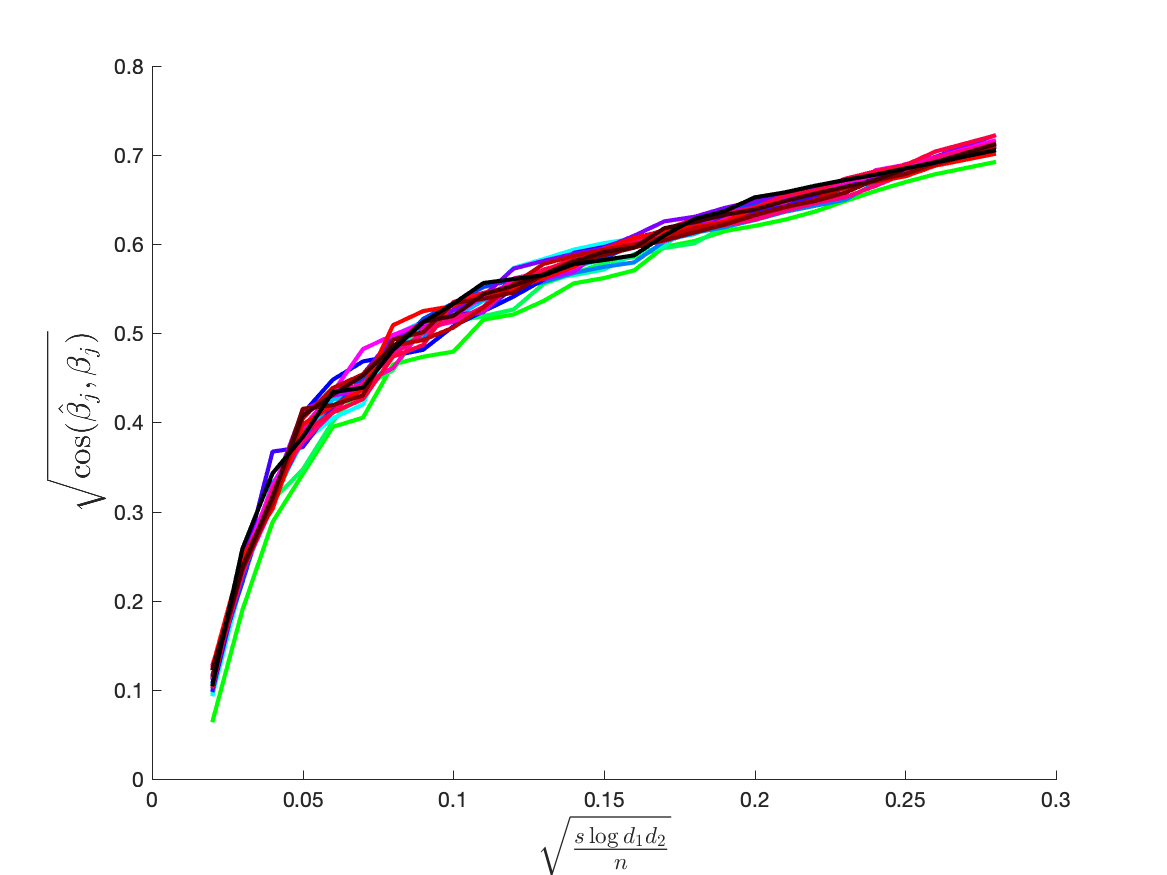}}
	\subfigure[Weibull for $f^{(4)}$]{\label{fwei4}\includegraphics[width=49.9mm]{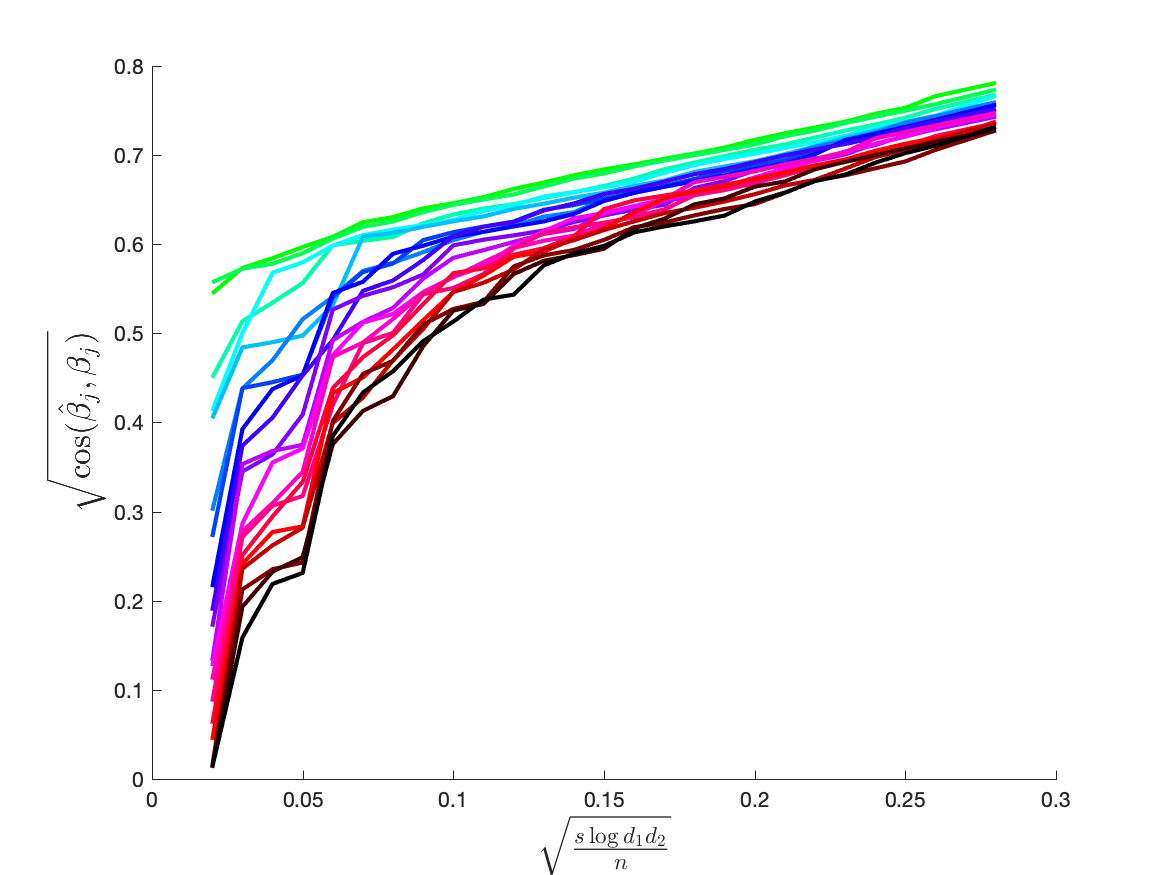}}
	\subfigure[Weibull for $f^{(5)}$]{\label{fwei5}\includegraphics[width=49.9mm]{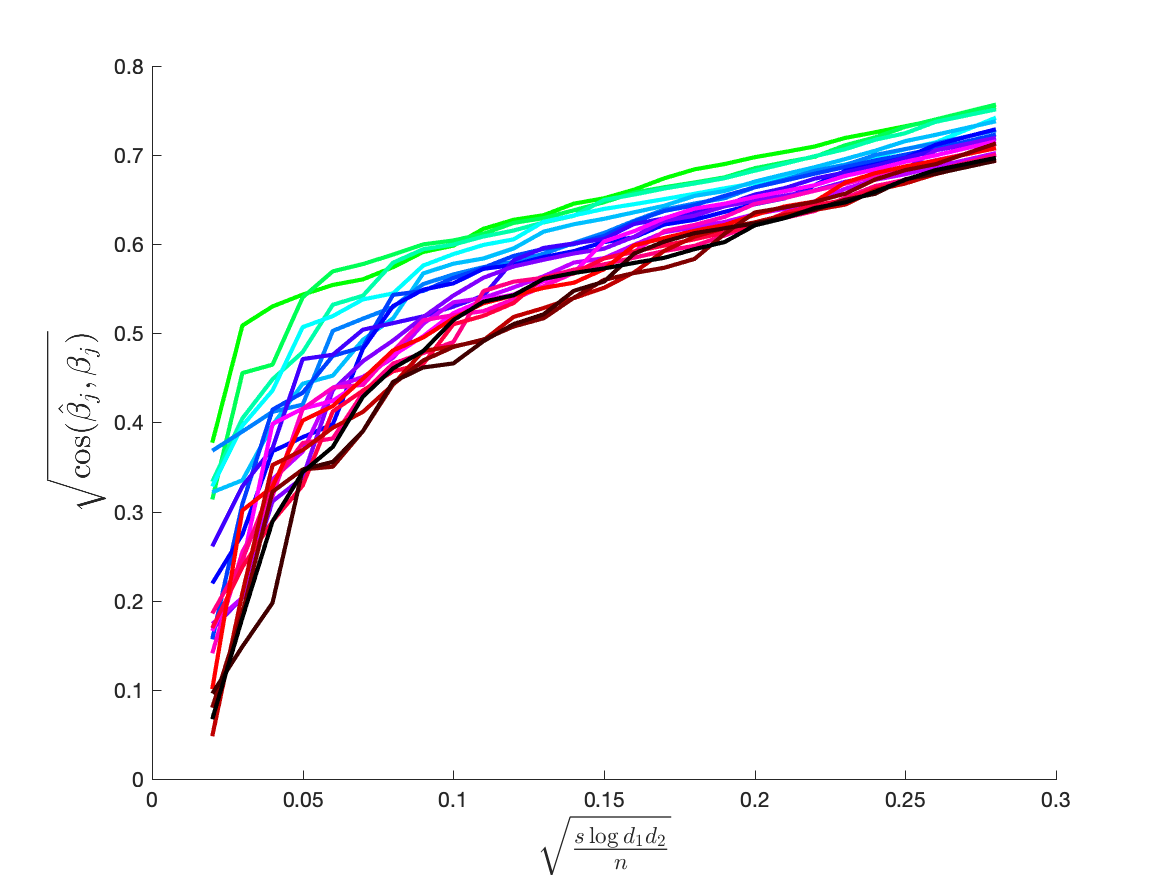}}
	\subfigure[Weibull for $f^{(6)}$]{\label{fwei6}\includegraphics[width=49.9mm]{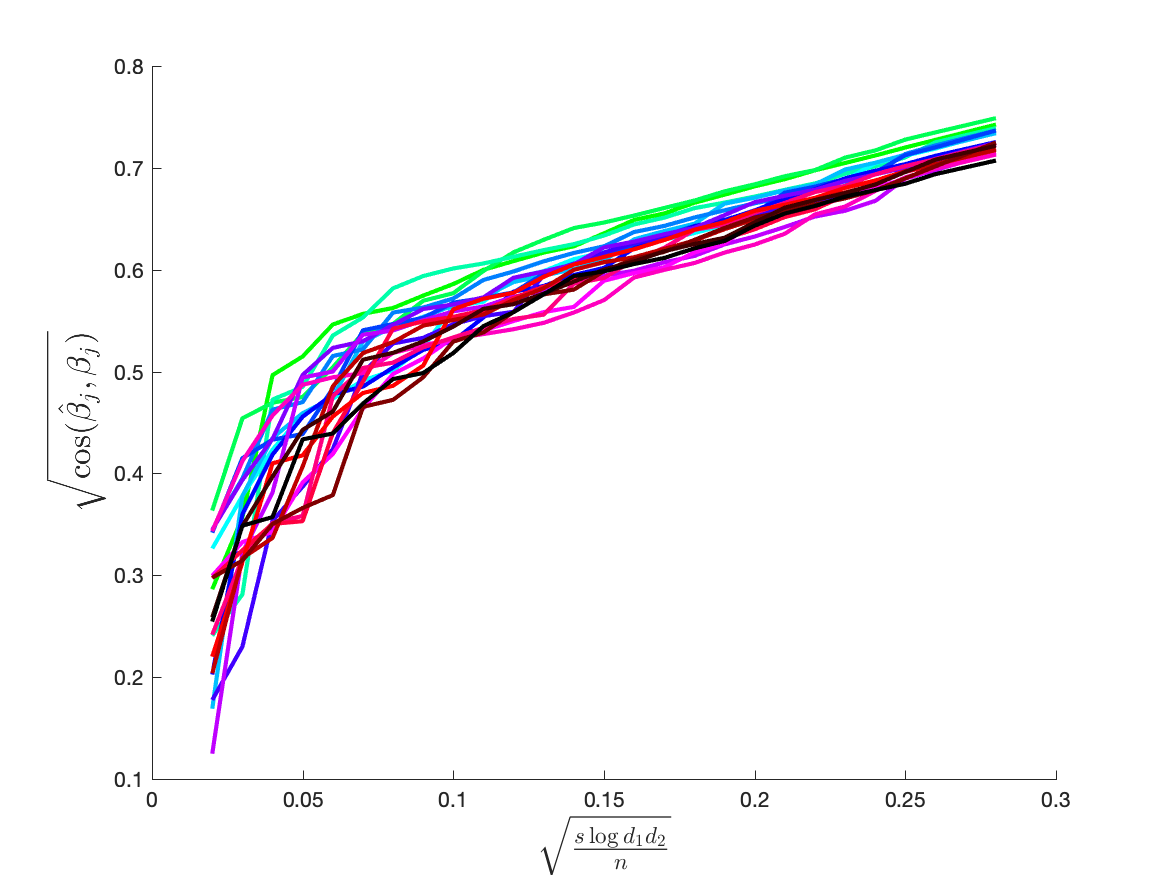}}
	\includegraphics[width=3.5cm,height=0.34cm]{Figure/betaleg.png}
	\caption{{Sparse vector estimation for Rayleigh and Weibull design. This figure shows cosine distance trend for error of estimating single sparse parameter in model (\ref{mod:1}). The color varies from green to black when $k$ varies from $1$ to $d_2$. The first and third row correspond to the linear link function, while the second and fourth row correspond to the quadratic link function.}}\label{fig:6}
\end{figure}

\begin{figure}[t]
	\centering     
	\subfigure[Production]{\label{tra1:prod}\includegraphics[width=49.9mm]{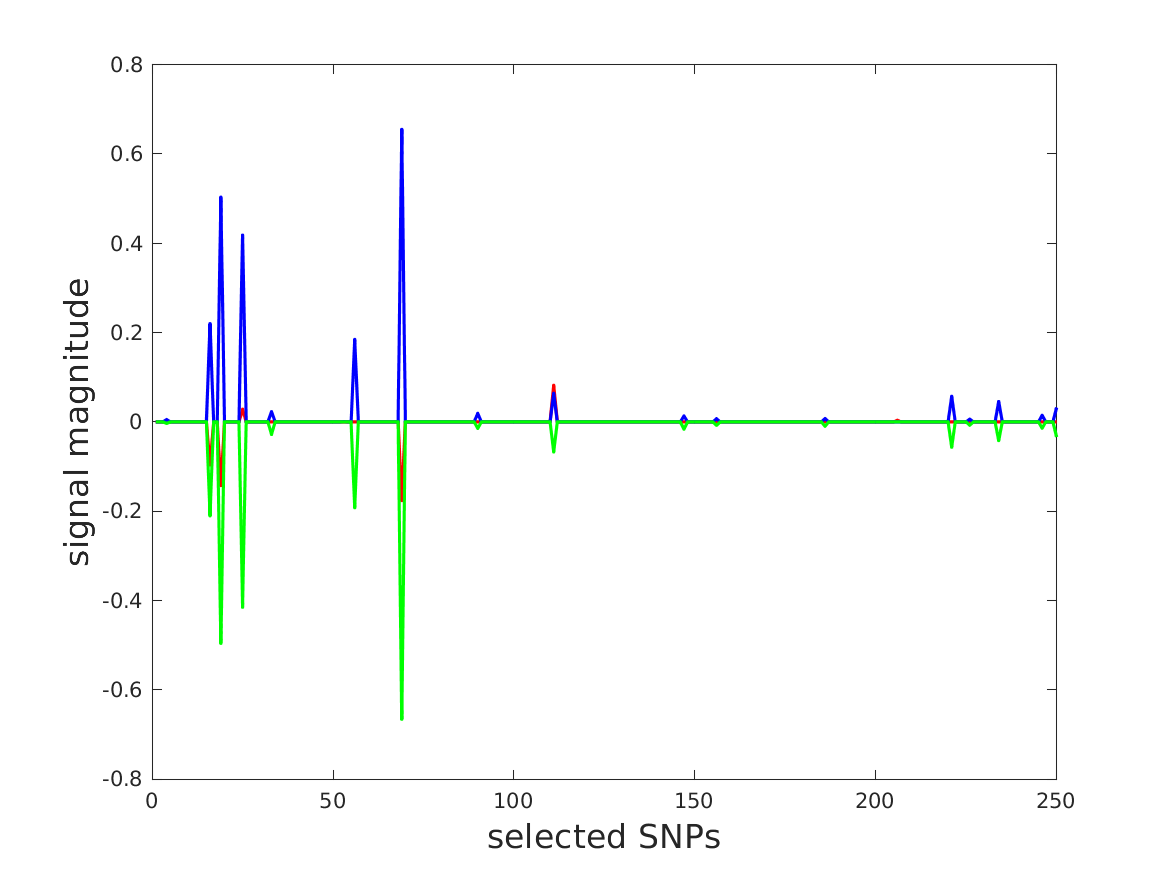}}
	\subfigure[Leaf rust]{\label{tra1:rust}\includegraphics[width=49.9mm]{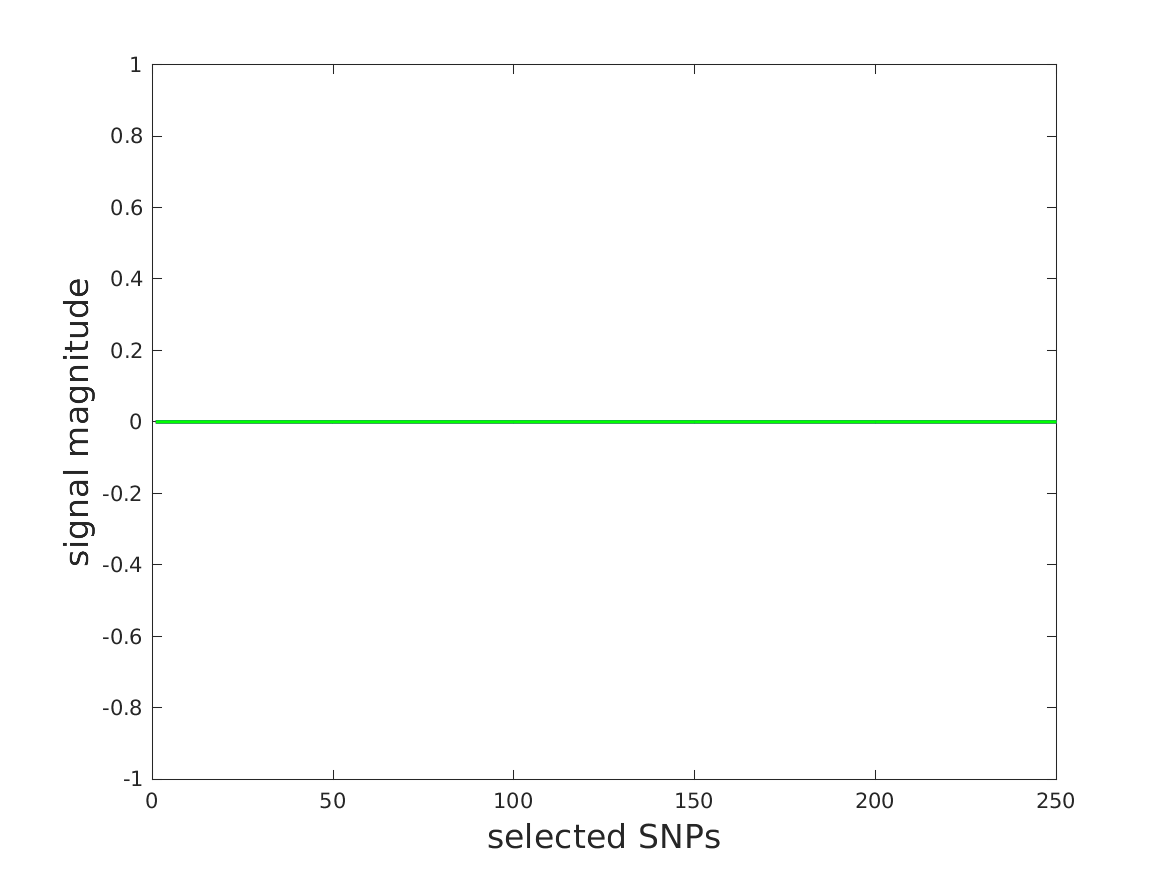}}
	\subfigure[Green beans]{\label{tra1:green}\includegraphics[width=49.9mm]{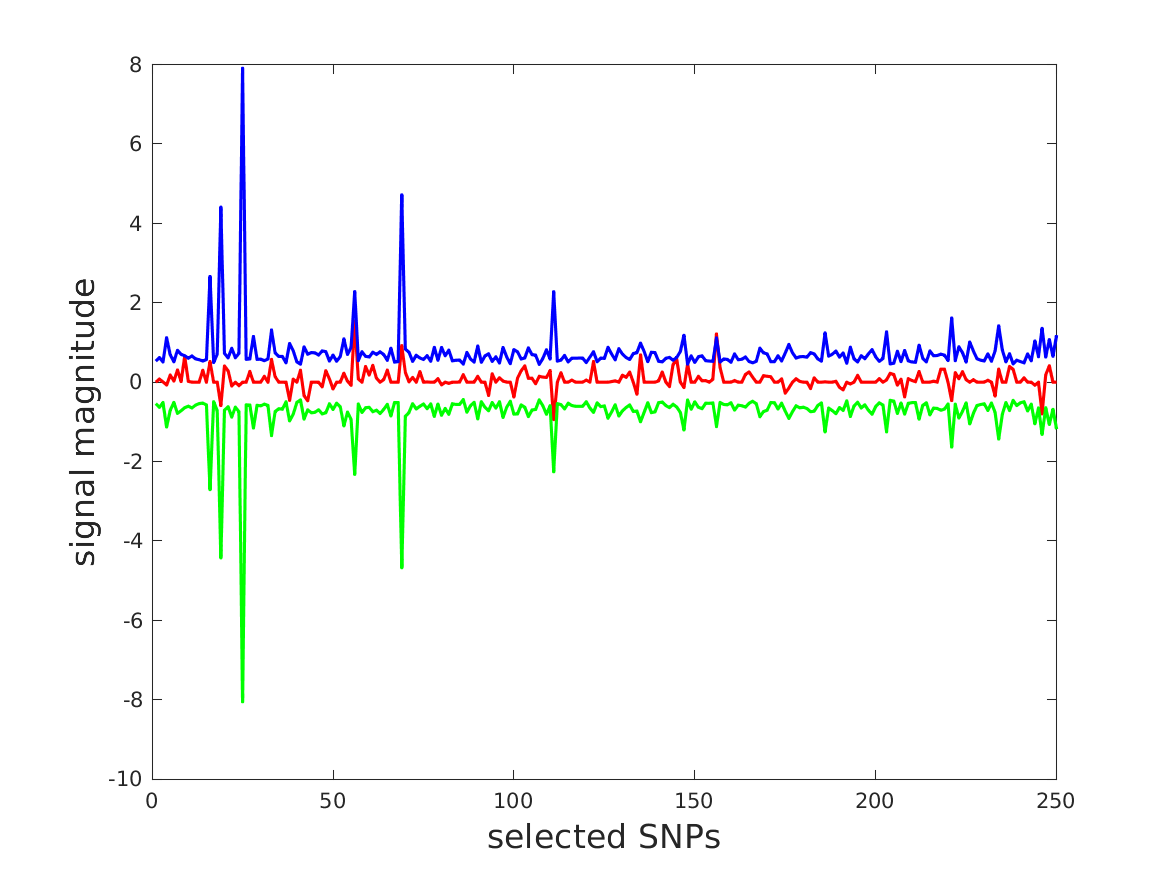}}
	\subfigure[Production]{\label{tra2:prod}\includegraphics[width=49.9mm]{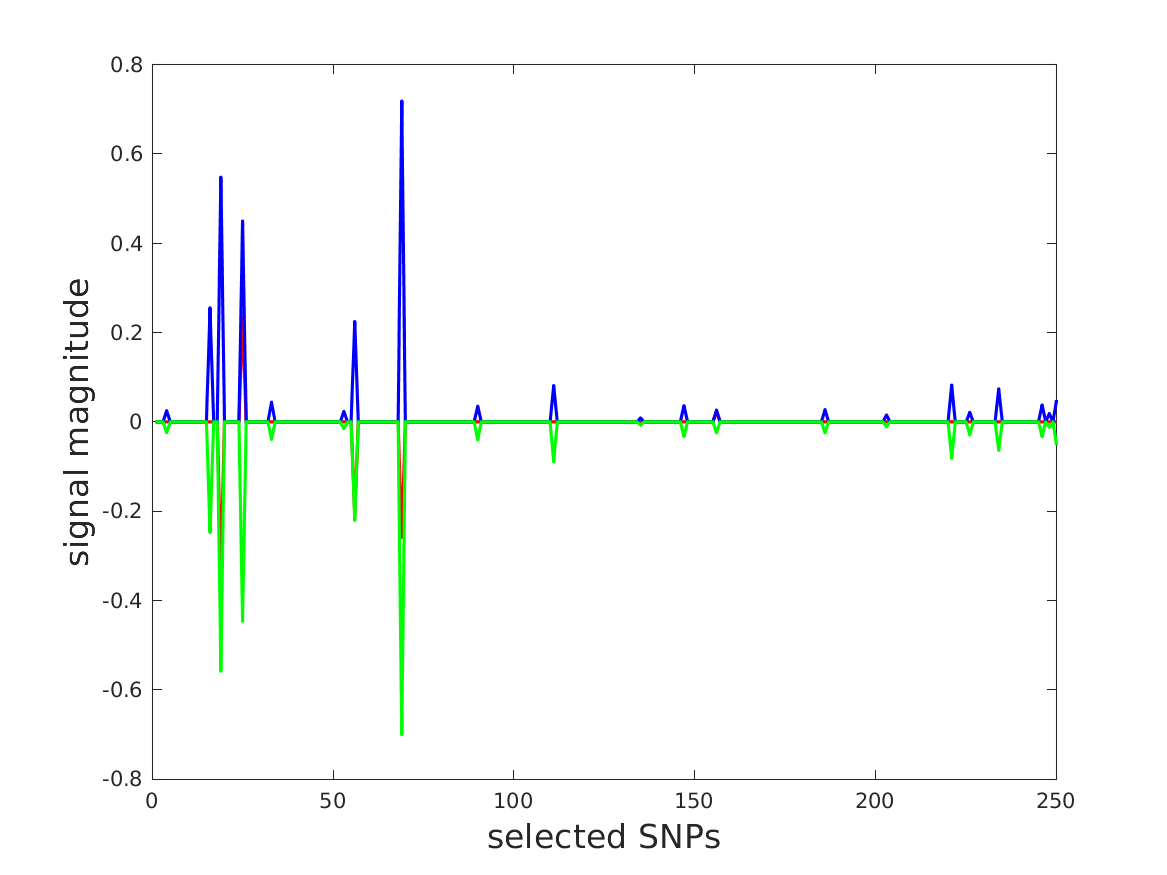}}
	\subfigure[Leaf rust]{\label{tra2:rust}\includegraphics[width=49.9mm]{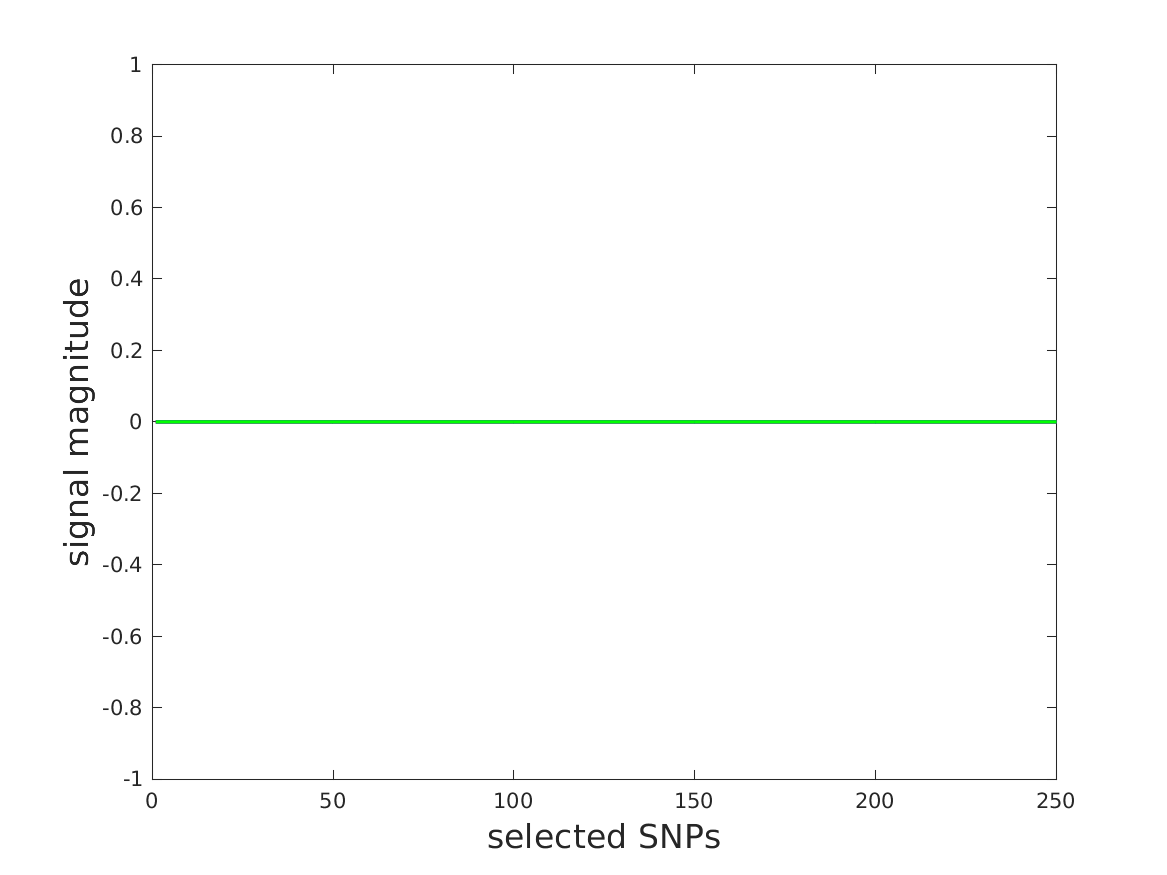}}
	\subfigure[Green beans]{\label{tra2:green}\includegraphics[width=49.9mm]{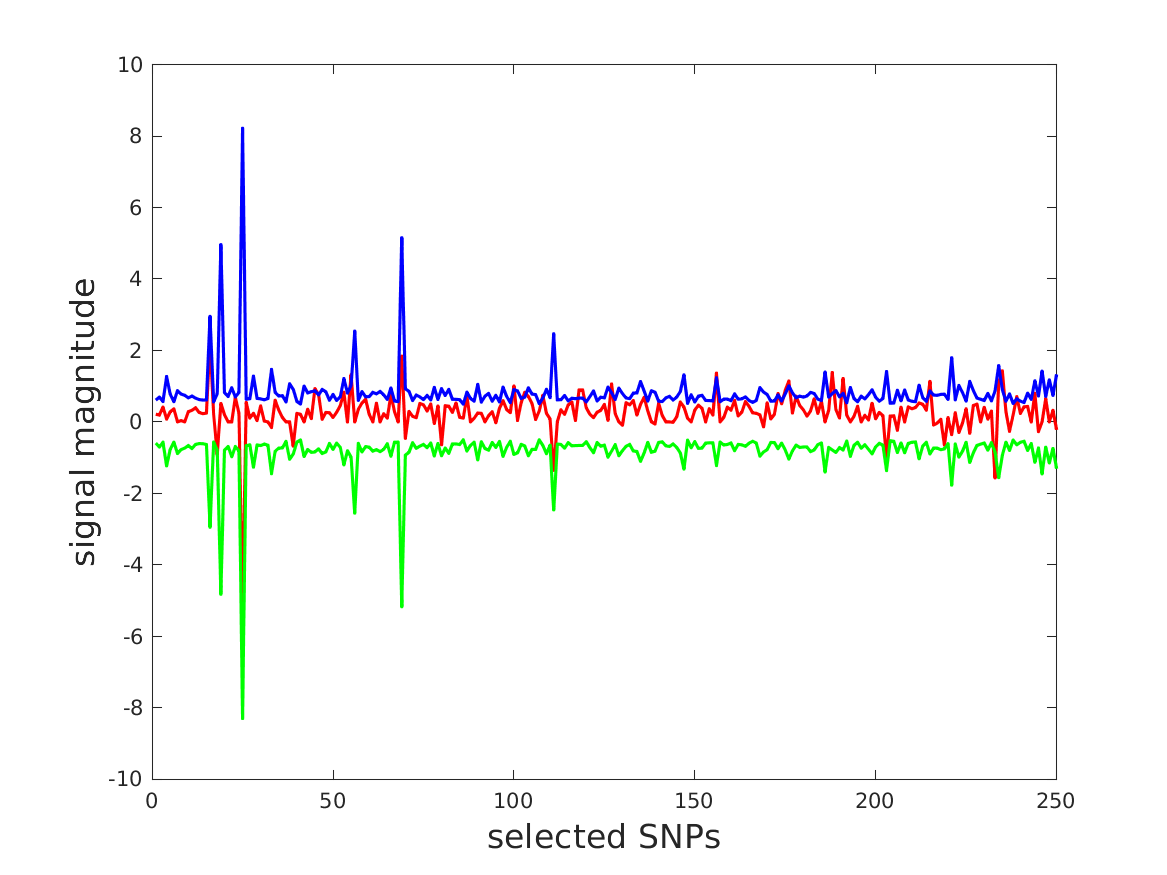}}
	\caption{{Signal trajectories for $95\%$ confidence interval (parametric bootstrapping). The $(l, i)$ plot shows the confidence interval trajectory of $(\bbeta^{l, i})^\star$ where $l = 1, 2$ indexes the confounder and $i = 1, 2, 3$ indexes the response. The blue line indicates the upper bound, the red line indicates the estimator, and the green line indicates the lower bound.}}\label{fig:12}
\end{figure}

We present additional simulations results for the setting discussed in
Section~\ref{sec:single-sparse-vector} and \ref{realdata}. Figure \ref{fig:3} illustrates the error trend for estimating $\tbeta_1$, $\tbeta_{d_2/2}$ and $\tbeta_{d_2}$ under different link functions and $\bx$ following one of the following distributions: $t_{13}$, Rayleigh or Weibull. Figure~\ref{fig:4}--\ref{fig:6} plot the error trend for all $d_2$ parameters. From the plots we observe that the linear trend will be more and more obvious when $k$ varies from $1$ to $d_2$, especially
for quadratic link functions. Note also that the number of samples
needed to get comparable results decrease dramatically as $k$
increases. Figure \ref{fig:12} shows $95\%$ confidence interval for the signal trajectory of the real dataset we analyzed in Section \ref{realdata}, which is obtained by doing parametric bootstrapping.

\bibliography{paper}

\end{document}